\documentclass{article}
\PassOptionsToPackage{numbers, compress, sort}{natbib}

\usepackage[final]{neurips_2022}

\usepackage[utf8]{inputenc} 
\usepackage[T1]{fontenc}    
\usepackage[pagebackref]{hyperref}       
 \hypersetup{
	colorlinks=true,
	linkcolor=magenta,
	filecolor=magenta,
	citecolor=magenta,      
	urlcolor=magenta,
}
\renewcommand*{\backref}[1]{}
\renewcommand*{\backrefalt}[4]{%
	\ifcase #1 Not cited.%
	\or        Cited on page~#2.%
	\else      Cited on pages~#2.%
	\fi}

\usepackage{url}            
\usepackage{booktabs}       
\usepackage{amsfonts}       
\usepackage{nicefrac}       
\usepackage{microtype}      
\usepackage{xcolor}         
\definecolor{dodgerblue}{HTML}{1e90ff}

\usepackage{amsmath}
\usepackage{amssymb}
\usepackage{mathtools}
\DeclarePairedDelimiter{\floor}{\lfloor}{\rfloor}

\usepackage{enumitem}

\usepackage[linesnumbered,ruled]{algorithm2e}

\SetCommentSty{mycommfont}

\usepackage{cleveref}
\crefname{figure}{Figure}{Figures}
\crefname{section}{Section}{Sections}
\crefname{lemma}{Lemma}{Lemmas}
\crefname{proposition}{Proposition}{Propositions}
\crefname{algorithm}{Algorithm}{Algorithms}
\crefname{appendix}{Appendix}{Appendices}
\crefname{table}{Table}{Tables}
\usepackage{graphicx}
\graphicspath{ {./figures/} }
\usepackage{caption}
\usepackage{subcaption}

\usepackage{amsthm}
\makeatletter
\newtheorem*{rep@theorem}{\rep@title}
\newcommand{\newreptheorem}[2]{%
	\newenvironment{rep#1}[1]{%
		\def\rep@title{#2 \ref{##1}}%
		\begin{rep@theorem}}%
		{\end{rep@theorem}}}
\makeatother

\newtheorem{proposition}{Proposition}
\newtheorem{lemma}{Lemma}

\newreptheorem{proposition}{Proposition}
\newreptheorem{lemma}{Lemma}

\DeclareMathOperator*{\argmax}{arg\,max}

\usepackage{placeins}
\allowdisplaybreaks
\title{Joint Entropy Search for Multi-Objective Bayesian Optimization}

\author{%
	Ben Tu\textsuperscript{\textdagger} \quad Axel Gandy\textsuperscript{\textdagger} \quad Nikolas Kantas\textsuperscript{\textdagger} \quad Behrang Shafei\textsuperscript{$\ddagger$}\\
	\textsuperscript{\textdagger}Imperial College London\\
	\textsuperscript{$\ddagger$}BASF SE\\
	\texttt{ben.tu16@imperial.ac.uk}
}

\begin{document}

\maketitle
\begin{abstract}
Many real-world problems can be phrased as a multi-objective optimization problem, where the goal is to identify the best set of compromises between the competing objectives. Multi-objective Bayesian optimization (BO) is a sample efficient strategy that can be deployed to solve these vector-valued optimization problems where access is limited to a number of noisy objective function evaluations. In this paper, we propose a novel information-theoretic acquisition function for BO called Joint Entropy Search (JES), which considers the joint information gain for the optimal set of inputs and outputs. We present several analytical approximations to the JES acquisition function and also introduce an extension to the batch setting. We showcase the effectiveness of this new approach on a range of synthetic and real-world problems in terms of the hypervolume and its weighted variants.
\end{abstract}

\section{Introduction}
Bayesian optimization (BO) has demonstrated a lot of success in solving black-box optimization problems in various domains such as machine learning \cite{snoek2012anips, snoek2015icml, wu2019joesata}, chemistry \cite{felton2021c, gomez-bombarelli2018acs}, robotics \cite{calandra2016amai, berkenkamp2021ml} and clinical trials \cite{nour2020asc, sui2018icml}. The procedure works by maintaining a probabilistic model of the observed data in order to guide the optimization procedure into regions of interest. Specifically, at each iteration the black-box function is evaluated at one or more input locations that maximizes an acquisition function on the model. Implicitly, this function strikes a balance between exploring new areas and exploiting areas that have been shown to be promising. In this work, we consider the more general problem, where the black-box function of interest is vector-valued. This increases the difficulty of the problem because there are now many directions in which the objectives can be improved, in contrast to the single-objective setting where there is only one. Informally, the end goal of multi-objective optimization is to identify a collection of points that describe the best trade-offs between the different objectives.
\\ \\
There are several ways to define an acquisition function for multi-objective BO. A popular strategy is random scalarization \cite{knowles2006itec,paria2020uai}, which works by transforming the multi-objective problem into a distribution of single-objective problems. These approaches are appealing because they enable the use of standard single-objective acquisition functions. 
A weakness of this approach is that it relies on random sampling to encourage exploration and therefore the performance of this method might suffer early on when the scale of the objectives is unknown or when either the input space or the objective space is high-dimensional \cite{daulton2022uai, paria2020uai}. Another popular class of multi-objective acquisition functions are improvement-based. These strategies focus on improving a performance metric over sets, for example the hypervolume indicator \cite{daulton2020anips, daulton2021anips, yang2019jgo, emmerich2006itec} or the R2 indicator \cite{deutz2019emo}. The main drawback of these approaches is that the performance of these methods can be biased towards a single performance metric, which can be inadequate to assess the multi-objective aspects of the problem \cite{zitzler2003itec}. There are also many other multi-objective acquisition functions discussed in the literature, which mainly differ by how they navigate the exploration-exploitation trade-off \cite{konakoviclukovic2020anips, picheny2015sc, picheny2019jgo, binois2020jmlr, binois2021a}.
\\ \\
Instead of relying on scalarizations or an improvement-based criterion, this paper considers the perspective where the goal of interest is to improve the posterior distribution over the optimal points. We propose a novel information-theoretic acquisition function called the Joint Entropy Search (JES), which assesses how informative an observation will be in learning more about the joint distribution of the optimal inputs and outputs. This acquisition function combines the advantages of existing information-theoretic methods, which focus solely on improving the posterior of either the optimal inputs \cite{hernandez-lobato2016icml, garrido-merchan2019n, garrido-merchan2021a} or the optimal outputs \cite{belakaria2019anips, belakaria2021j, suzuki2020icml}. We connect JES with the existing information-theoretic acquisition functions by showing that it is an upper bound to these utilities. 
\\ \\
After acceptance of this work, we learnt of a parallel line of inquiry by Hvarfner et al. \cite{hvarfner2022aa}, who independently came up with the same JES acquisition function \eqref{eqn:jes}. Their work focussed on the single-objective setting and the approximation scheme they devised is subtly different to the ones we present. We see our work as being complementary to theirs because we both demonstrate the effectiveness of this new acquisition function in different settings. 

\paragraph{Contributions and organization.} In \cref{sec:prelim}, we set up the problem and introduce the novel JES acquisition function. In \cref{sec:approx_jes}, we present a catalogue of conditional entropy estimates to approximate this utility and present a simple extension to the batch setting. These approximations are analytically tractable and differentiable, which means that we can take advantage of gradient-based optimization. The main results that we developed here can be viewed as direct extensions to the recent work in the Bayesian optimization literature by Suzuki et al. \cite{suzuki2020icml} and Moss et al. \cite{moss2021jmlr}. In \cref{sec:performance_criteria}, we present a discussion on the hypervolume indicator and explain how it can be a misleading performance criterion because it is sensitive to the scale of the objectives. We show that information-theoretic approaches are naturally invariant to reparameterization of the objectives, which make them well-suited for multi-objective black-box optimization. For a more complete picture of performance, we propose a novel weighted hypervolume strategy (\cref{app:hypervolume_indicator}), which allows us to assess the performance of a multi-objective algorithm over different parts of the objective space. In \cref{sec:experiments}, we demonstrate the effectiveness of JES on some synthetic and real-life multi-objective problems. Finally in \cref{sec:conclusion}, we provide some concluding remarks. Additional results and proofs are presented in the Appendix.

\section{Preliminaries}
\label{sec:prelim}

We consider the problem of maximizing a vector-valued function $f: \mathbb{X} \rightarrow \mathbb{R}^M$ over a bounded space of inputs $\mathbb{X} \subset \mathbb{R}^D$. To define the maximum $\max_{\mathbf{x} \in \mathbb{X}} f(\mathbf{x})$, we appeal to the Pareto partial ordering in $\mathbb{R}^M$. For the rest of this paper, we will denote vectors by $\mathbf{y} = (y^{(1)},\dots, y^{(M)}) \in \mathbb{R}^M$, the non-negative real numbers by $\mathbb{R}_{\geq 0}$ and diagonal matrices by $\text{diag}(\cdot)$.

\paragraph{Pareto domination.} 
We say a vector $\mathbf{y} \in \mathbb{R}^M$ weakly Pareto dominates another vector $\mathbf{y}' \in \mathbb{R}^M$ if it performs just as well in all objectives if not better: $\mathbf{y} \succeq \mathbf{y'} \iff \mathbf{y} - \mathbf{y'} \in \mathbb{R}_{\geq 0}^M$. Additionally, if the vectors are not equivalent, $\mathbf{y} \neq \mathbf{y}'$, then we say strict Pareto domination holds: $\mathbf{y} \succ \mathbf{y'} \iff \mathbf{y} - \mathbf{y'} \in \mathbb{R}_{\geq 0}^M \setminus \{\mathbf{0}_M\}$, where $\mathbf{0}_M$ is the $M$-dimensional zero vector. This binary relation can be further extended to define domination among sets. Let $A, B \subset \mathbb{R}^M$ be sets, if the set $B$ lies in the weakly dominated region of $A$, namely $B \subseteq \mathbb{D}_{\preceq}(A) = \cup_{\mathbf{a} \in A} \{\mathbf{y} \in \mathbb{R}^M: \mathbf{y} \preceq \mathbf{a}\}$, then we say $A$ weakly dominates $B$, denoted by $A \succeq B$. In addition, if it also holds that the dominated regions are not equal, $\mathbb{D}_{\preceq}(A) \neq \mathbb{D}_{\preceq}(B)$, we say strict Pareto domination holds, denoted by $A \succ B$.


\paragraph{Multi-objective optimization.}
The goal of multi-objective optimization is to identify the Pareto optimal set of inputs $\mathbb{X}^* = \argmax_{\mathbf{x} \in \mathbb{X}} f(\mathbf{x}) \subseteq \mathbb{X}$. The Pareto set is defined as the set of inputs whose objective vectors are not strictly Pareto dominated by another: $\mathbf{x}^* \in \mathbb{X}^* \iff \mathbf{x}^* \in \mathbb{X} \text{ and } \nexists \mathbf{x} \in \mathbb{X} \text{ such that } f(\mathbf{x}) \succ f(\mathbf{x}^*)$. The image of the Pareto set in the objective space $\mathbb{Y}^* = f(\mathbb{X}^*) = \max_{\mathbf{x} \in \mathbb{X}} f(\mathbf{x})$ is called the Pareto front. For convenience of notation, we will denote the set of Pareto optimal input-output pairs by $(\mathbb{X}^*, \mathbb{Y}^*)$.

\paragraph{Bayesian Optimization}

\begin{figure}
	\begin{subfigure}{0.24\linewidth}
		\includegraphics[width=1\linewidth]{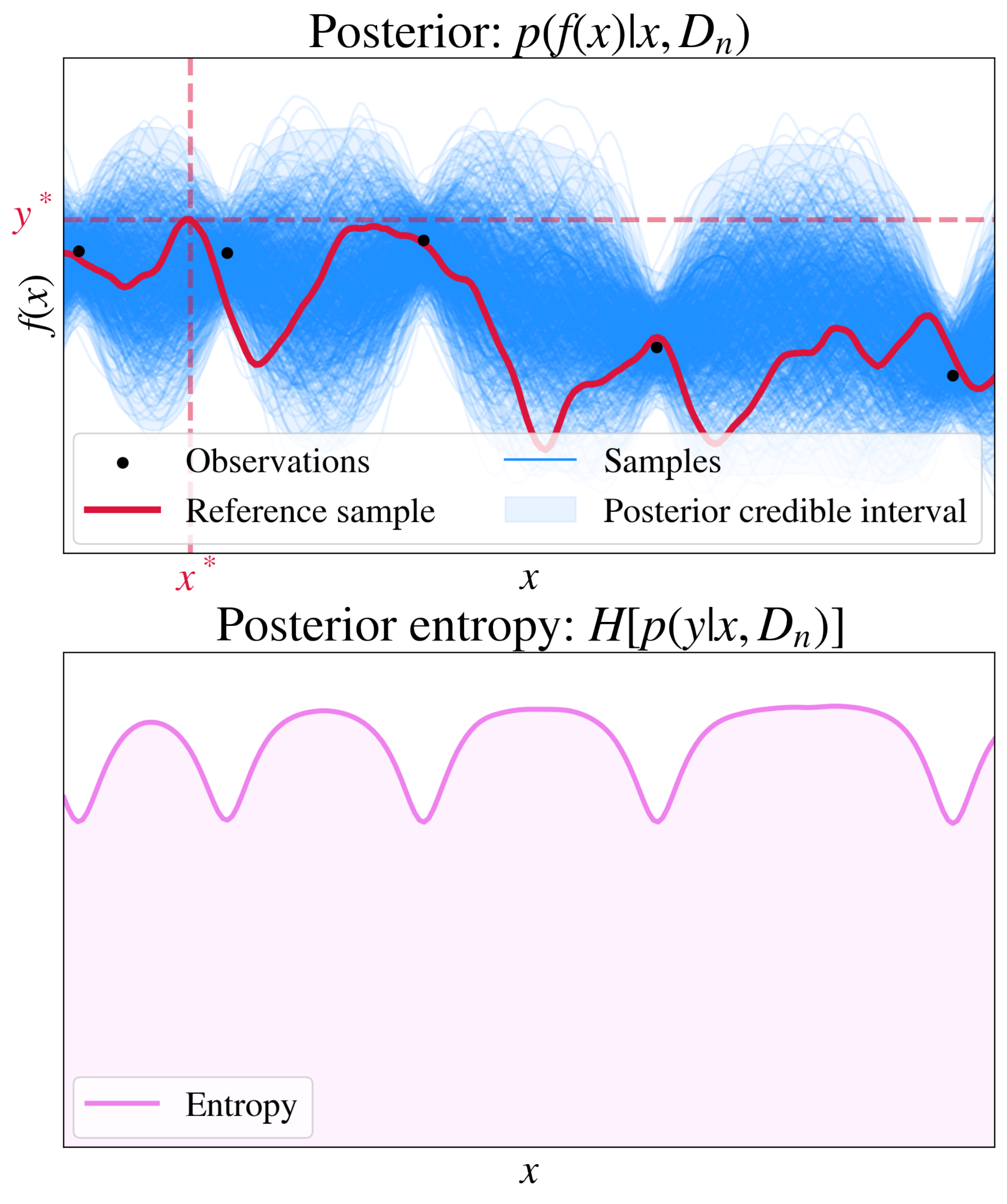}
		\centering
	\end{subfigure}
	\begin{subfigure}{0.24\linewidth}
		\includegraphics[width=1\linewidth]{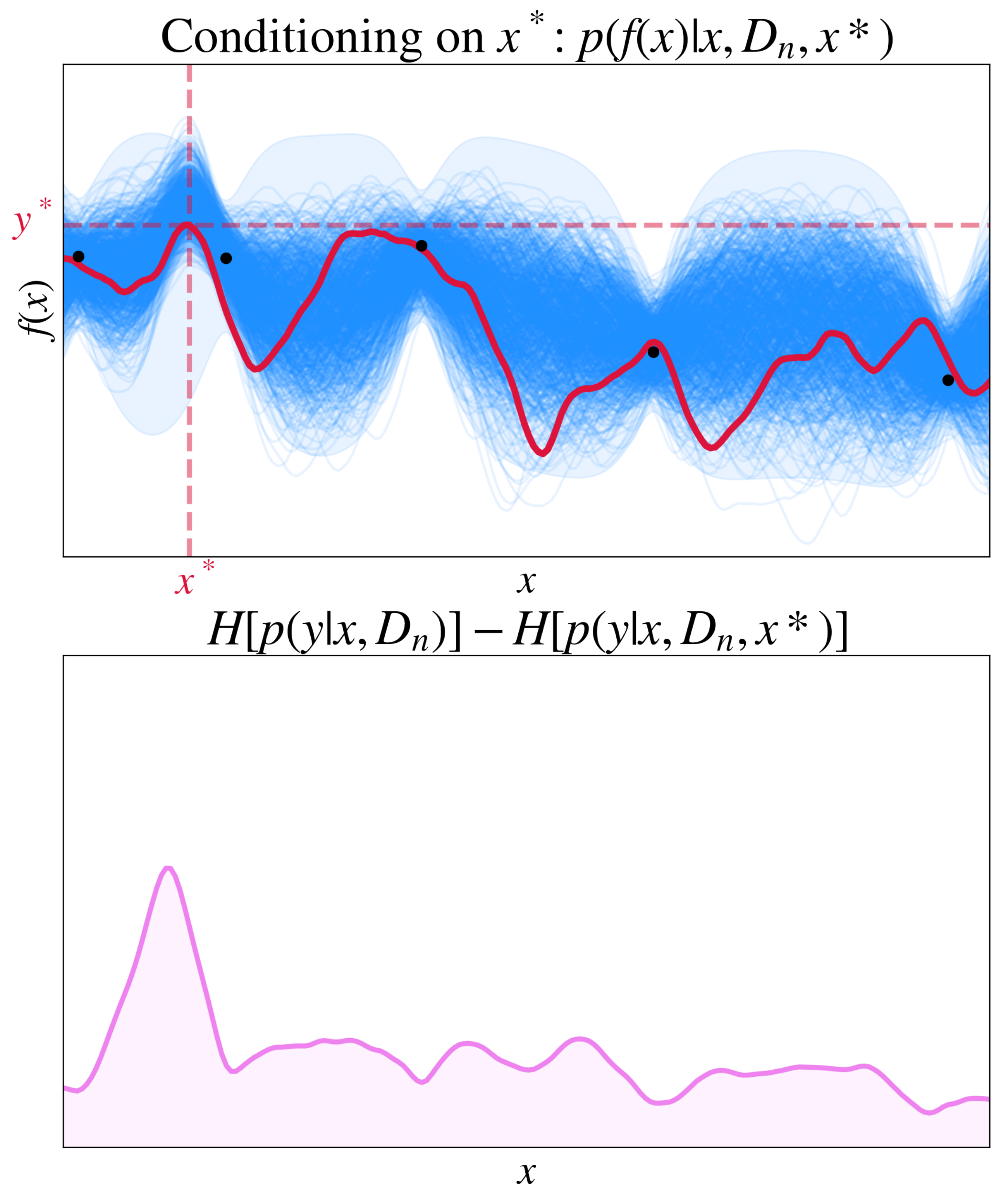}
		\centering
	\end{subfigure}
	\begin{subfigure}{0.24\linewidth}
		\includegraphics[width=1\linewidth]{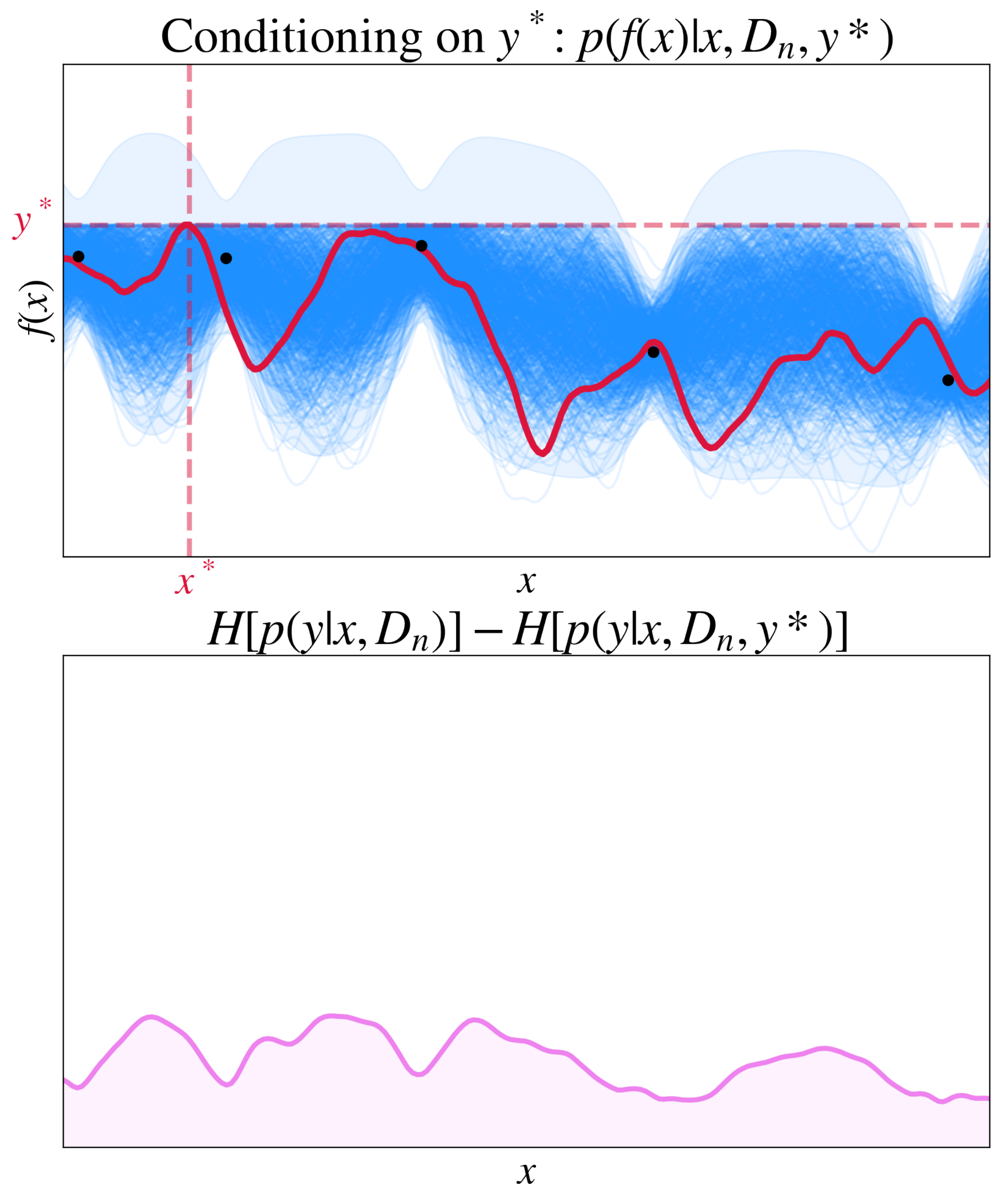}
		\centering
	\end{subfigure}
	\begin{subfigure}{0.24\linewidth}
		\includegraphics[width=1\linewidth]{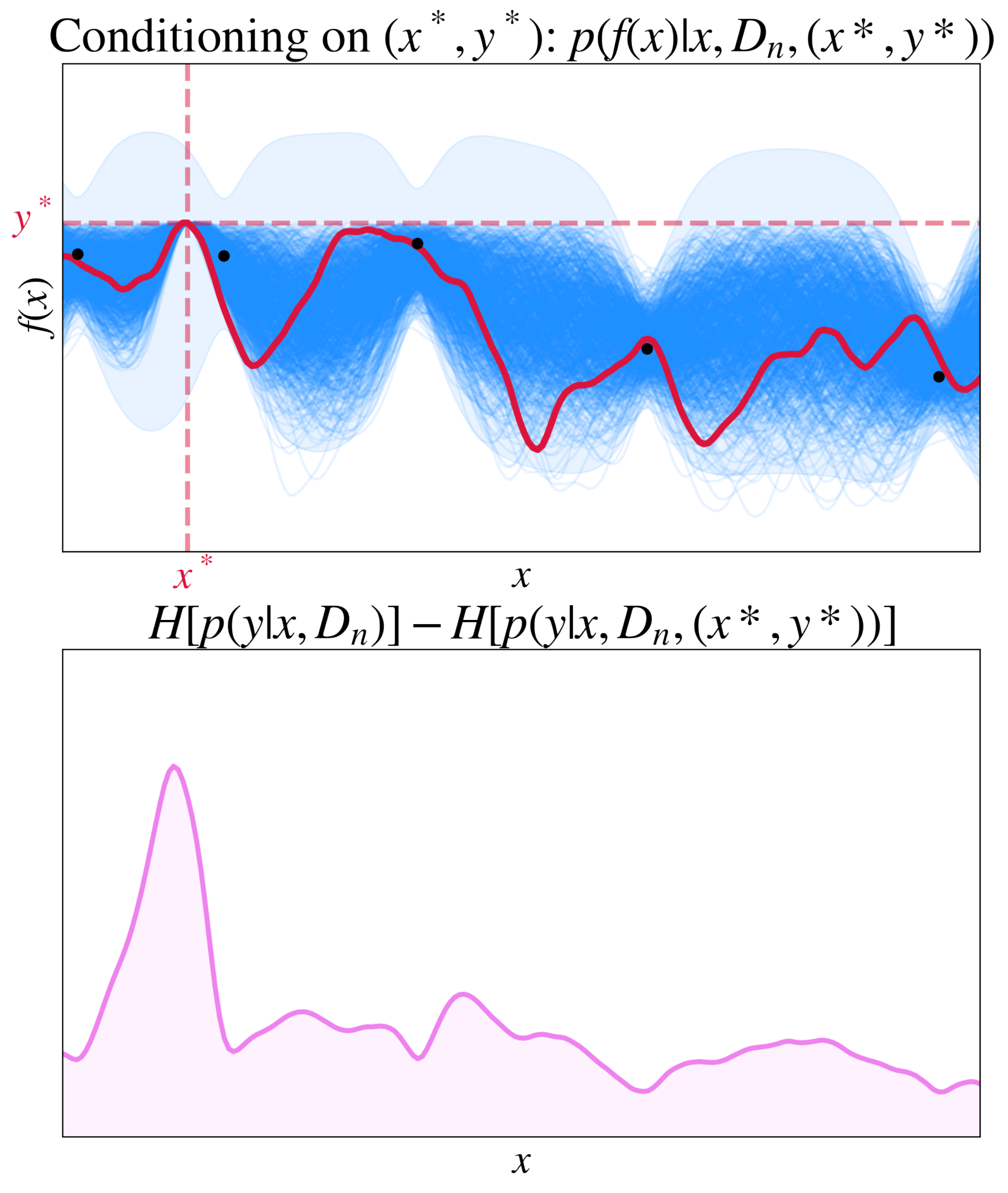}
		\centering
	\end{subfigure}
	\centering
	\caption{Comparison of the samples (top) and change in entropy (bottom) for the posterior and conditional distributions. The red line in the posterior plots denotes the reference sample that is used to obtain the maximizer $x^*$ and maximum $y^*$, whilst the shaded blue region is the $95\%$ credible interval of the posterior $p(f(x)| x, D_n)$. Conditioning on $x^*$ reduces the entropy for all inputs according to how correlated it is with $x^*$. Conditioning on $y^*$ reduces the entropy for all inputs according to the posterior probability that the objective surpasses $y^*$. Conditioning on $(x^*, y^*)$ leads to a drop in entropy based on both the input correlation with $x^*$ and the posterior probability of exceeding $y^*$.}
	\label{fig:posterior}
\end{figure}

is a sample efficient global optimization strategy, which relies on a probabilistic model in order to decide which points to query. In \cref{app:bo}, we present the pseudo-code for the standard BO procedure---for more details see \cite{shahriari2016pi, brochu2010a, frazier2018a}. 
In this work, we will use independent Gaussian process priors \cite{rasmussen2006} on each objective, $f^{(m)} \sim \text{GP}(\mu_0^{(m)}, \Sigma_0^{(m)})$, where $\mu^{(m)}: \mathbb{X} \rightarrow \mathbb{R}$ is the mean function and $\Sigma^{(m)}: \mathbb{X} \times \mathbb{X} \rightarrow \mathbb{R}$ is the covariance function for objective $m$. The observations at location $\mathbf{x} \in \mathbb{X}$ will be assumed to be corrupted with additive Gaussian noise, $\mathbf{y} = f(\mathbf{x}) + \boldsymbol{\epsilon}$, where $\boldsymbol{\epsilon} \sim \mathcal{N}(0, \text{diag}(\boldsymbol{\sigma}(\mathbf{x})))$ denotes the observation noise with variance $\boldsymbol{\sigma}(\mathbf{x}) \in \mathbb{R}_{\geq 0}^M$. After $n$ evaluations, we have a data set $D_n = \{(\mathbf{x}_t, \mathbf{y}_t)\}_{t=1,\dots,n}$. The posterior model $p(f|D_n)$ is a collection of independent Gaussian processes $f^{(m)}|D_n \sim \text{GP}(\mu_n^{(m)}, \Sigma_n^{(m)})$. The explicit expressions for the mean and covariance are presented in \cref{app:gp}. The main focus of this work is on designing the acquisition function, $\alpha: \mathbb{X} \rightarrow \mathbb{R}$, which is used to select the inputs: $\mathbf{x}_{n+1} = \argmax_{\mathbf{x} \in \mathbb{X}} \alpha(\mathbf{x}| D_n)$.

\paragraph{Information-theoretic acquisition functions} focus on maximizing the gain in information from the next observation and a function of the probabilistic model. Initial work in BO focussed on picking points to learn more about the distribution of the maximizer $p(\mathbb{X}^*| D_n)$. Specifically, the goal of interest was to maximize the mutual information between the observation $\mathbf{y}$ and the Pareto set $\mathbb{X}^*$ conditional on the current data set $D_n$:
\begin{equation}
	\alpha^{\text{PES}}(\mathbf{x}| D_n) 
	= \text{MI}(\mathbf{y}; \mathbb{X}^*| \mathbf{x}, D_n)
	= H[p(\mathbf{y}| \mathbf{x}, D_n)]
	- \mathbb{E}_{p(\mathbb{X}^*| D_n)}[H[p(\mathbf{y}| \mathbf{x}, D_n, \mathbb{X}^*)]]
	\label{eqn:pes}
\end{equation}
where $H[p(\mathbf{x})] = - \int p(\mathbf{x}) \log p(\mathbf{x}) d\mathbf{x}$ represents the differential entropy. This acquisition function is commonly referred to as predictive entropy search (PES) \cite{hernandez-lobato2014anips, hernandez-lobato2016icml, shah2015anips}, but it was formerly\footnote{The difference in the naming convention stems solely from the approximation strategy used to estimate the mutual information. At a high level, ES applies expectation propagation \cite{minka2001uai} to estimate $p(\mathbb{X}^*| D_n \cup \{\mathbf{x}, \mathbf{y}\})$,
whilst PES applies expectation propagation to estimate $p(\mathbf{y}| \mathbf{x}, D_n, \mathbb{X}^*)$.} known as entropy search (ES) \cite{villemonteix2008jgo, hennig2012jmlr}. Despite the importance of obtaining more information about the maximizer, the PES acquisition function is heavily dependent on the approximation of $p(\mathbf{y}| \mathbf{x}, D_n, \mathbb{X}^*)$, which is both computationally difficult to implement and optimize. This motivated researchers to consider a simpler scheme that focusses on learning more about the distribution of the maximum $p(\mathbb{Y}^*| D_n)$. The resulting acquisition function is known as the max-value entropy search (MES) \cite{wang2017icml, belakaria2019anips, hoffman2015nwbo, suzuki2020icml}:
\begin{equation}
	\alpha^{\text{MES}}(\mathbf{x}| D_n) 
	= \text{MI}(\mathbf{y}; \mathbb{Y}^*| \mathbf{x}, D_n)
	= H[p(\mathbf{y}| \mathbf{x}, D_n)]
	- \mathbb{E}_{p(\mathbb{Y}^*| D_n)}[H[p(\mathbf{y}| \mathbf{x}, D_n, \mathbb{Y}^*)]].
\end{equation}
Unlike PES, the conditional probability $p(\mathbf{y}| \mathbf{x}, D_n, \mathbb{Y}^*)$ arising in MES can be approximated and optimized more easily because some approximations lead to closed-form expressions. Despite the favourable properties of MES, the primary goal of interest is to identify the location of the maximizer $\mathbb{X}^*$ and not necessarily the value of the maximum $\mathbb{Y}^*$. To combine the advantages of both of these approaches, we propose the joint entropy search acquisition function, which focusses on learning more about the joint distribution of the optimal points $p((\mathbb{X}^*, \mathbb{Y}^*)| D_n)$:
\begin{align}
	\begin{split}
	\alpha^{\text{JES}}(\mathbf{x}| D_n) 
	&= \text{MI}(\mathbf{y}; (\mathbb{X}^*, \mathbb{Y}^*)| \mathbf{x}, D_n)
	\\
	&= H[p(\mathbf{y}| \mathbf{x}, D_n)]
	- \mathbb{E}_{p((\mathbb{X}^*, \mathbb{Y}^*)| D_n)}[H[p(\mathbf{y}| \mathbf{x}, D_n, (\mathbb{X}^*, \mathbb{Y}^*))]].
	\end{split}
	\label{eqn:jes}
\end{align}
The JES acquisition function inherits the advantages of the PES and MES acquisition functions because it considers the knowledge learnt about the optimal points and is also simple to implement---more details in the next section. The following proposition shows that we can also interpret JES as an upper bound to both the PES and MES acquisition function.
\begin{proposition}
	The JES is an upper bound to any convex combination of the PES and MES acquisition functions: \normalfont{ $\alpha^{\text{JES}}(\mathbf{x}| D_n) \geq 
	\beta \alpha^{\text{PES}}(\mathbf{x}| D_n) + (1- \beta) \alpha^{\text{MES}}(\mathbf{x}| D_n)$, for any $\beta \in [0, 1]$}.
	\label{prop:upper_bound}
\end{proposition}
In \cref{fig:posterior}, we illustrate the subtle differences between the different information-theoretic acquisition functions. More specifically, we visualise the difference between the conditional distributions arising in each acquisition function for a single-objective problem using one sample of the optimal points. 


\paragraph{Remark.} In the BO literature it is common to distinguish between single-objective and multi-objective acquisition functions by appending `MO' to the end of the acronym. For notational simplicity, we opt against this convention in this paper. In \cref{app:single_objective}, we emphasize the main differences that arise when computing the information-theoretic algorithms in both settings.
\section{Approximating JES}
\label{sec:approx_jes}
\begin{figure}[!htb]
	\includegraphics[width=0.225\linewidth]{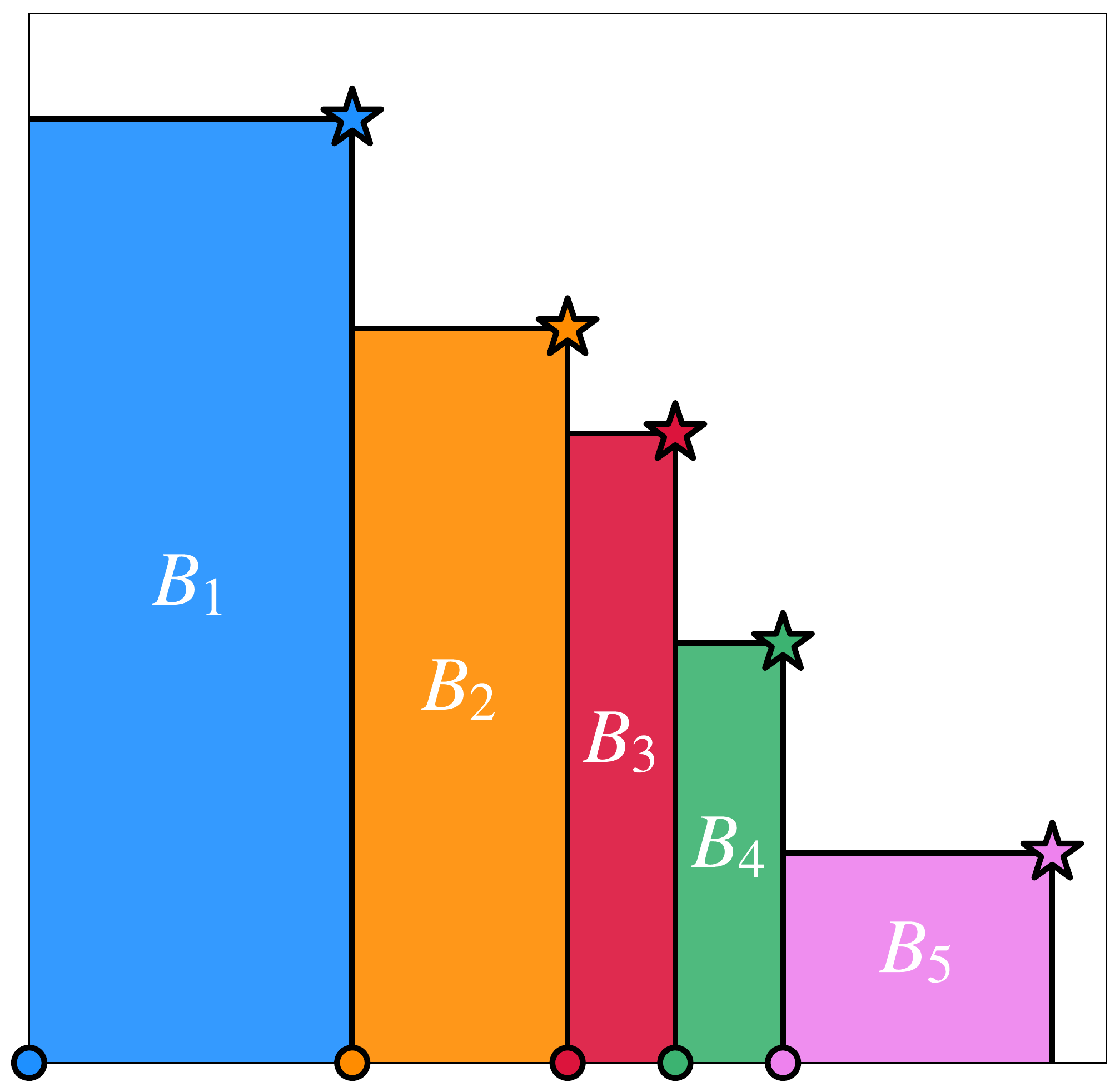}
	\hspace{0.02\linewidth}
	\includegraphics[width=0.26\linewidth]{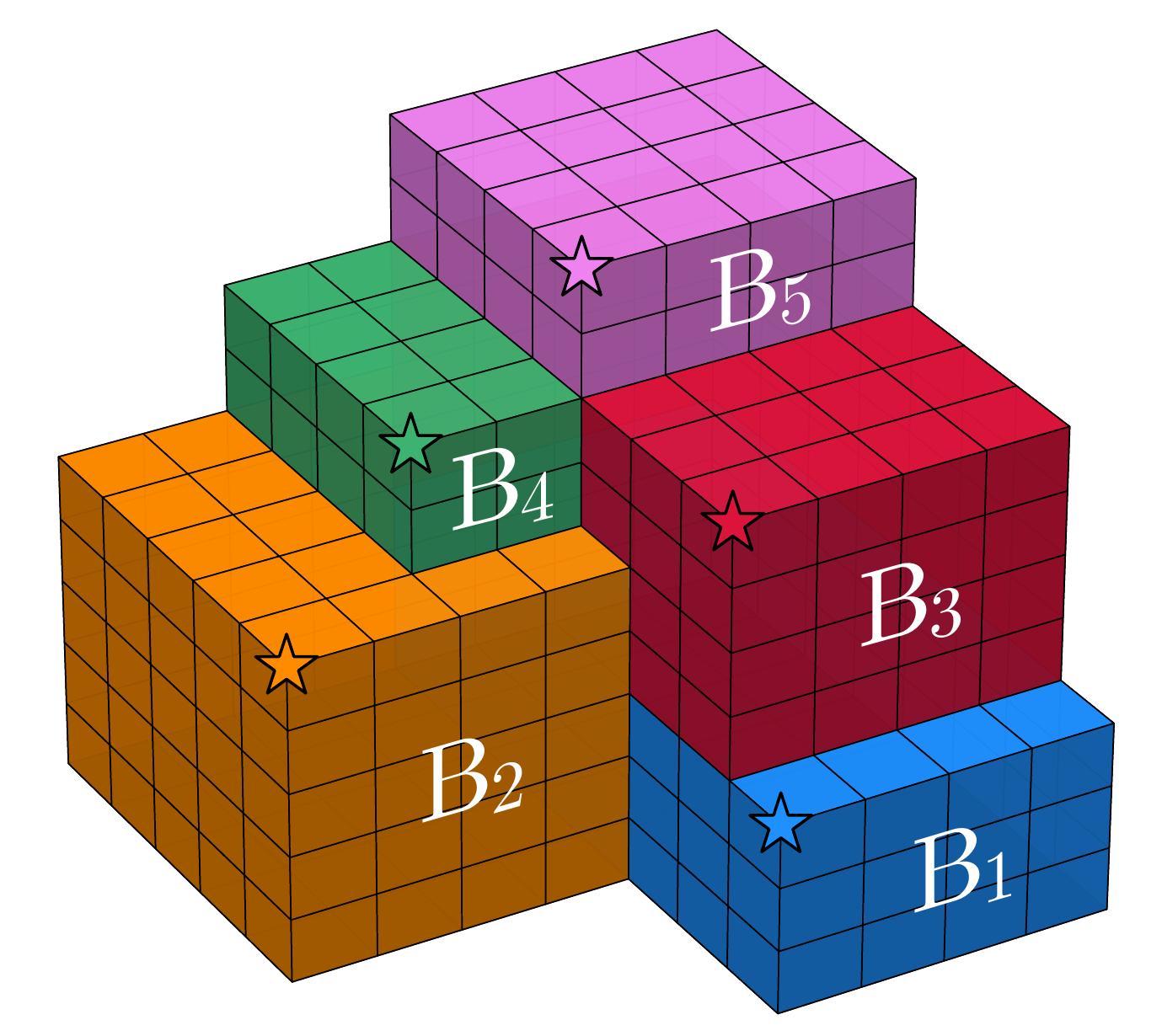}
	\centering
	\caption{Box decompositions for a two-dimensional and three-dimensional Pareto front.}
	\label{fig:box_decompositions}
\end{figure}

In this section, we present several approximations to the JES acquisition function \eqref{eqn:jes} and a simple extension to the batch setting. The first term in the JES criterion \eqref{eqn:jes} is the entropy of a multivariate normal distribution:
\begin{equation}
	H[p(\mathbf{y}| \mathbf{x}, D_n)] = \frac{M}{2}\log(2\pi e) + \frac{1}{2} \sum_{m=1}^M \log (\Sigma_n^{(m)}(\mathbf{x}, \mathbf{x}) + \sigma^{(m)}(\mathbf{x})).
	\label{eqn:prior_entropy}
\end{equation}
The second term is an intractable expectation which is approximated by drawing Monte Carlo samples from $p((\mathbb{X}^*, \mathbb{Y}^*)| D_n)$. The conditional entropy $H[p(\mathbf{y}| \mathbf{x}, D_n, (\mathbb{X}^*, \mathbb{Y}^*))]$ is also an intractable quantity which has to be estimated. 
The overall approximation of \eqref{eqn:jes} will take the form
\begin{equation}
	\begin{split}
		\hat{\alpha}^{\text{JES}}(\mathbf{x}| D_n) = 
		H[p(\mathbf{y}| \mathbf{x}, D_n)]
		- \frac{1}{S} \sum_{s=1}^S h((\mathbb{X}^*_s, \mathbb{Y}^*_s); \mathbf{x}, D_n),
	\end{split}
	\label{eqn:jes_base}
\end{equation}
where $h$ denotes the conditional entropy estimate and $(\mathbb{X}^*_s, \mathbb{Y}^*_s) \sim p((\mathbb{X}^*, \mathbb{Y}^*)| D_n)$ are the Monte Carlo samples. The distribution $p(\mathbf{y}| \mathbf{x}, D_n, (\mathbb{X}^*, \mathbb{Y}^*))$ is very challenging to work with because it enforces the global optimality condition that the function lies below the Pareto front $f(\mathbb{X}) \preceq \mathbb{Y}^*$. Instead of enforcing global optimality, we make the common simplifying assumption as in \cite{wang2017icml, suzuki2020icml, moss2021jmlr} and only enforce the optimality condition at the considered location: $f(\mathbf{x}) \preceq \mathbb{Y}^*$. 
By applying Bayes' theorem, the resulting density of interest becomes
\begin{align}
	p(\mathbf{y}| \mathbf{x}, D_{n*}, f(\mathbf{x}) \preceq \mathbb{Y}^*)
	&=\frac{p(f(\mathbf{x}) \preceq \mathbb{Y}^*| \mathbf{x}, D_{n+})}
	{p(f(\mathbf{x}) \preceq \mathbb{Y}^*| \mathbf{x}, D_{n*})}
	p(\mathbf{y}| \mathbf{x}, D_{n*}),
	\label{eqn:density}
\end{align}
where we have denoted the augmented data sets by $D_{n*} = D_{n} \cup (\mathbb{X}^*, \mathbb{Y}^*)$ and $D_{n+} = D_{n*} \cup \{(\mathbf{x}, \mathbf{y})\}$. We will refer to the quantity $p(f(\mathbf{x}) \preceq \mathbb{Y}^*)$ as the cumulative distribution function (CDF). The following lemma shows that this CDF can be computed analytically when the set $\mathbb{Y}^*\subset \mathbb{R}^M$ is discrete. This is a standard result \cite{keane2012aj, couckuyt2014jgo, picheny2015sc, suzuki2020icml} which can be derived by first partitioning the region of integration, $\mathbb{D}_{\preceq}(\mathbb{Y}^*) = \cup_{\mathbf{y}^* \in \mathbb{Y}^*} \{\mathbf{z} \in \mathbb{R}^M: \mathbf{z} \preceq \mathbf{y}^*\}$, into a collection of hyperrectangle subsets and then summing up the individual contributions---see \cref{fig:box_decompositions} for a visual. This partition can be computed using an incremental approach (Algorithm 1 of \cite{lacour2017c&or}), which has a cost of $O(|\mathbb{Y}^*|^{\floor{M/2} + 1})$. In the single-objective setting, the maximum is a single point $y^* \in \mathbb{R}$ and the box-decomposition is simply the interval $\mathbb{D}_{\preceq}(\{y^*\}) = (-\infty, y^*]$.

\begin{lemma}
	Let $\mathbb{Y}^*\subset \mathbb{R}^M$ be a finite set and $\mathbf{z} \sim N(\mathbf{a}, \text{diag}(\mathbf{b}))$ be an $M$-dimensional multivariate normal with mean $\mathbf{a} \in \mathbb{R}^M$ and variances $\mathbf{b} \in \mathbb{R}^M_{\geq 0}$. Let $\mathbb{D}_{\preceq}(\mathbb{Y}^*) = \bigcup_{j=1}^J B_j = \bigcup_{j=1}^J \prod_{m=1}^M (l^{(m)}_j, u^{(m)}_j]$ be the box decomposition of the dominated space, then
	\begin{align}
		p(\mathbf{z} \preceq \mathbb{Y}^*)
		&= \sum_{j=1}^J \prod_{m=1}^M
		\left[
		\Phi
		\left(
		\frac{u_j^{(m)} - a^{(m)}}{\sqrt{b^{(m)}}} 
		\right)
		-
		\Phi
		\left(
		\frac{l_j^{(m)} - a^{(m)}}{\sqrt{b^{(m)}}} 
		\right)
		\right].
		\label{eqn:cumulative_distribution}
	\end{align}
	\label{lemma:cumul}
\end{lemma}

In \cref{alg:jes}, we present the pseudo-code for the estimation of the JES acquisition function at a single candidate input. Several variables that calculated within the algorithm are independent of the input (coloured in blue). For computational efficiency, we only compute these variables once and then save them to memory for later use. 

\begin{algorithm}
	\SetKwInOut{Input}{Input}
	\DontPrintSemicolon
	\Input{A candidate $\mathbf{x}$; the data set $D_n$.}
	\tcp*[l]{\small Cached variables are coloured in blue.}
	Compute the initial entropy $h_0 = H[p(\mathbf{y}|\mathbf{x}, D_n)]$.
	\\
	\For{$s=1,\dots,S$}{
		Sample a path \textcolor{blue}{$f_s \sim p(f| D_n)$.}
		\label{alg:jes:sample}
		\\
		Compute the Pareto optimal points
		\textcolor{blue}{$\mathbb{X}^*_s = \argmax_{\mathbf{x}' \in \mathbb{X}} f_s(\mathbf{x}')$} and
		\textcolor{blue}{$\mathbb{Y}^*_s = f_s(\mathbb{X}^*_s)$}. 
		\label{alg:jes:moo}
		\\
		Compute the box decomposition
		\textcolor{blue}{$\mathbb{D}_{\preceq}(\mathbb{Y}^*_s) = \bigcup_{j=1}^J B_j$}.
		\label{alg:jes:bd}
		\\
		Compute the conditional \textcolor{blue}{$p(f| D_n \cup (\mathbb{X}^*_s, \mathbb{Y}^*_s))$}.
		\label{alg:jes:condition}
		\\
		Compute the estimate $h_s = h((\mathbb{X}^*_s, \mathbb{Y}^*_s); \mathbf{x}, D_n)$.
		\label{alg:jes:h}
	}
	\Return $\hat{\alpha}^{\text{JES}}(\mathbf{x}| D_n) = h_0 - \frac{1}{S} \sum_{s=1}^S h_s$.
	\caption{Joint Entropy Search (JES).}
	\label{alg:jes}
\end{algorithm}





\subsection{Estimating the conditional entropy}
\label{sec:approx_h}
\begin{figure}
\begin{subfigure}[t]{0.24\textwidth}
	\includegraphics[width=1\linewidth]{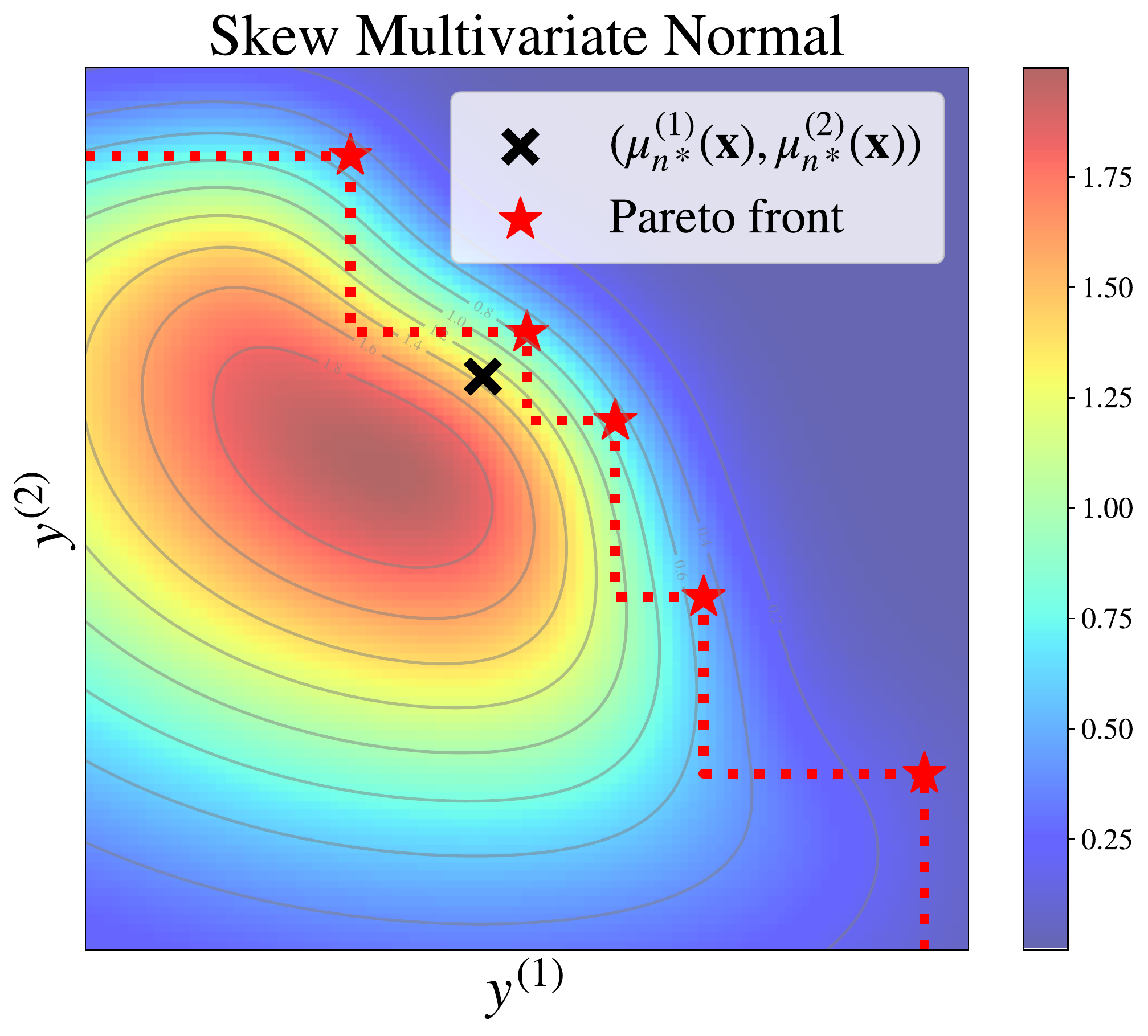}
	\caption{}
	\label{fig:skew_mvn}
\end{subfigure}
\begin{subfigure}[t]{0.24\textwidth}
	\includegraphics[width=1\linewidth]{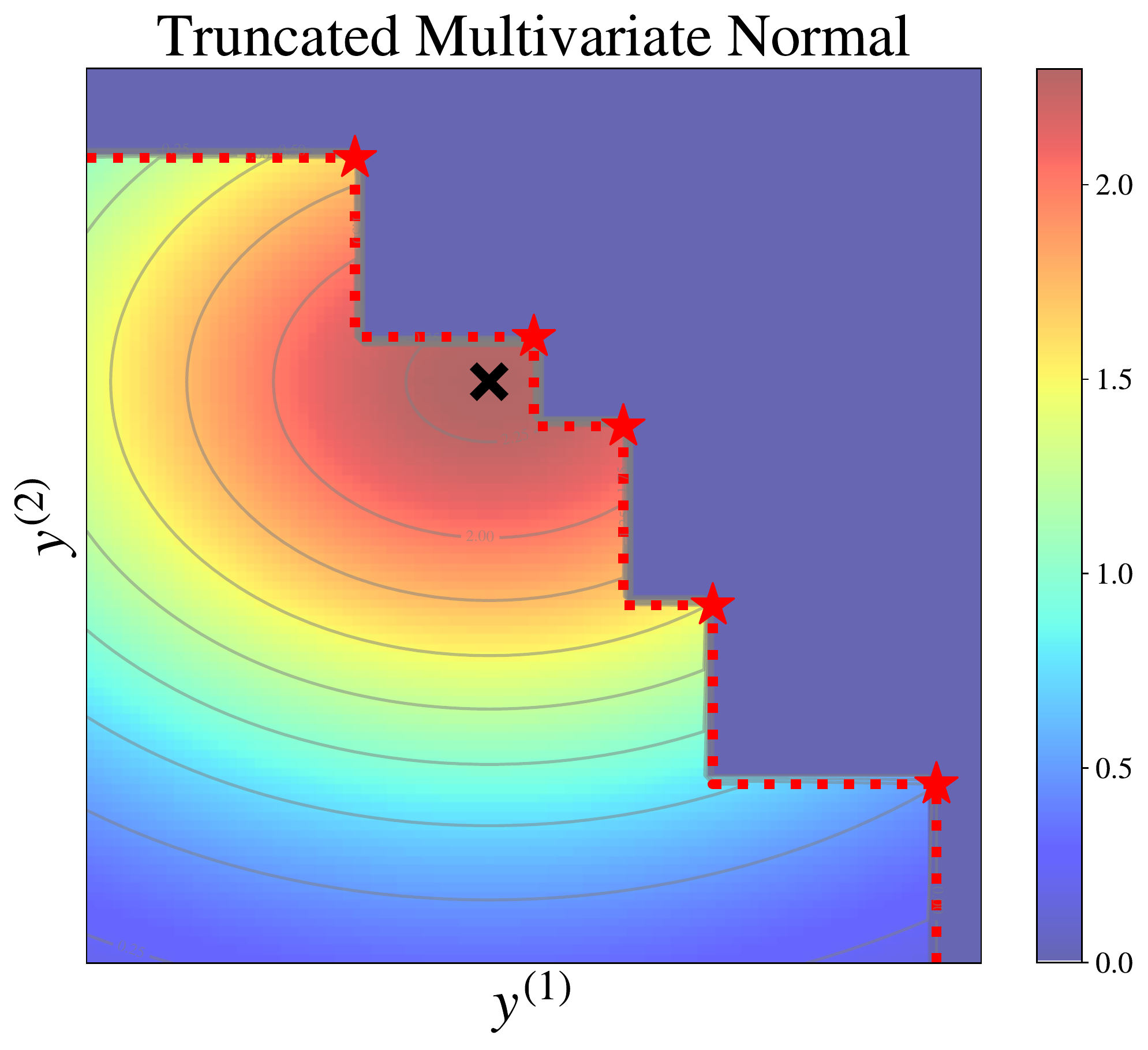}
	\caption{}
	\label{fig:trunc_mvn}
\end{subfigure}
\begin{subfigure}[t]{0.24\textwidth}
	\includegraphics[width=1\linewidth]{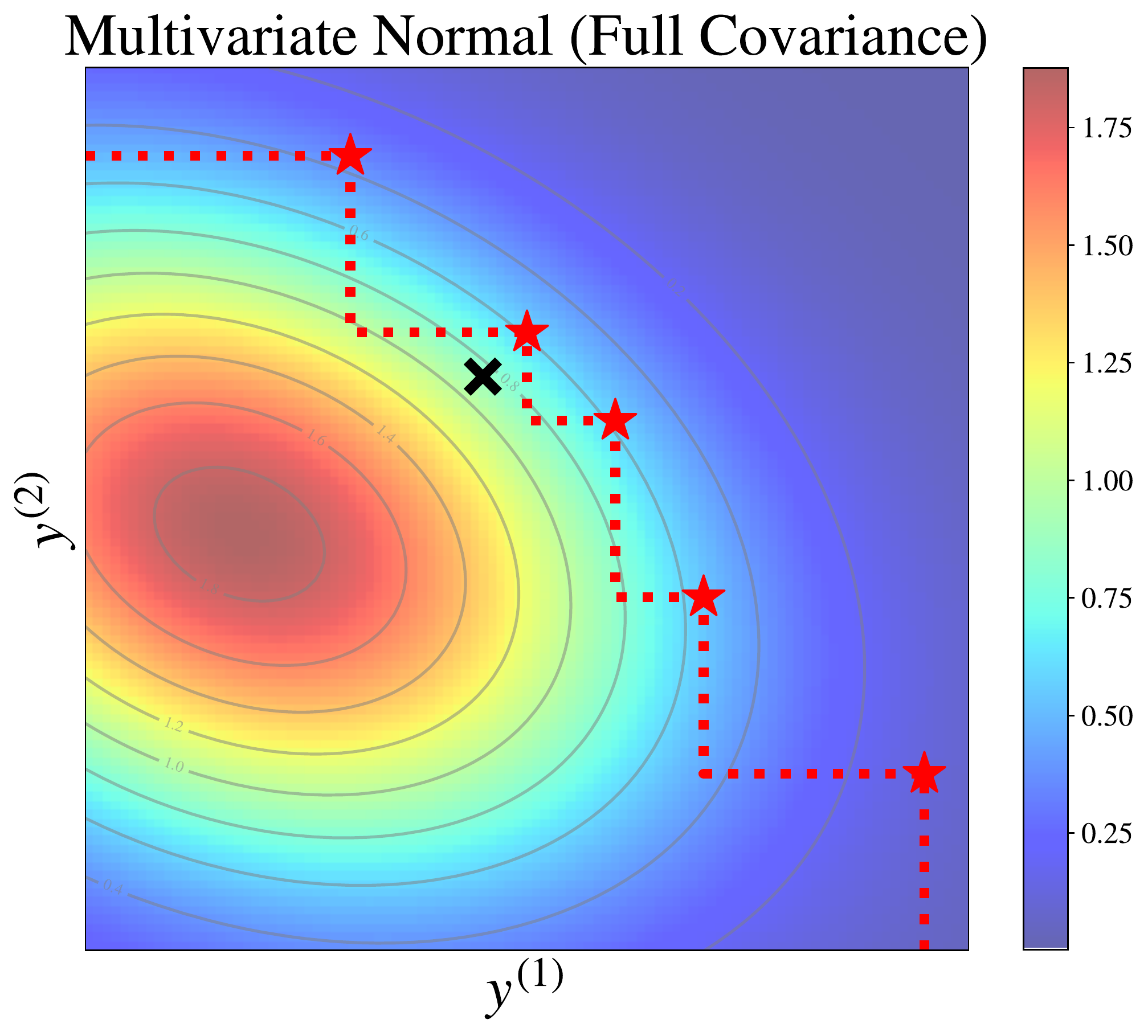}
	\caption{}
	\label{fig:mvn_full_cov}
\end{subfigure}
\begin{subfigure}[t]{0.24\textwidth}
	\includegraphics[width=1\linewidth]{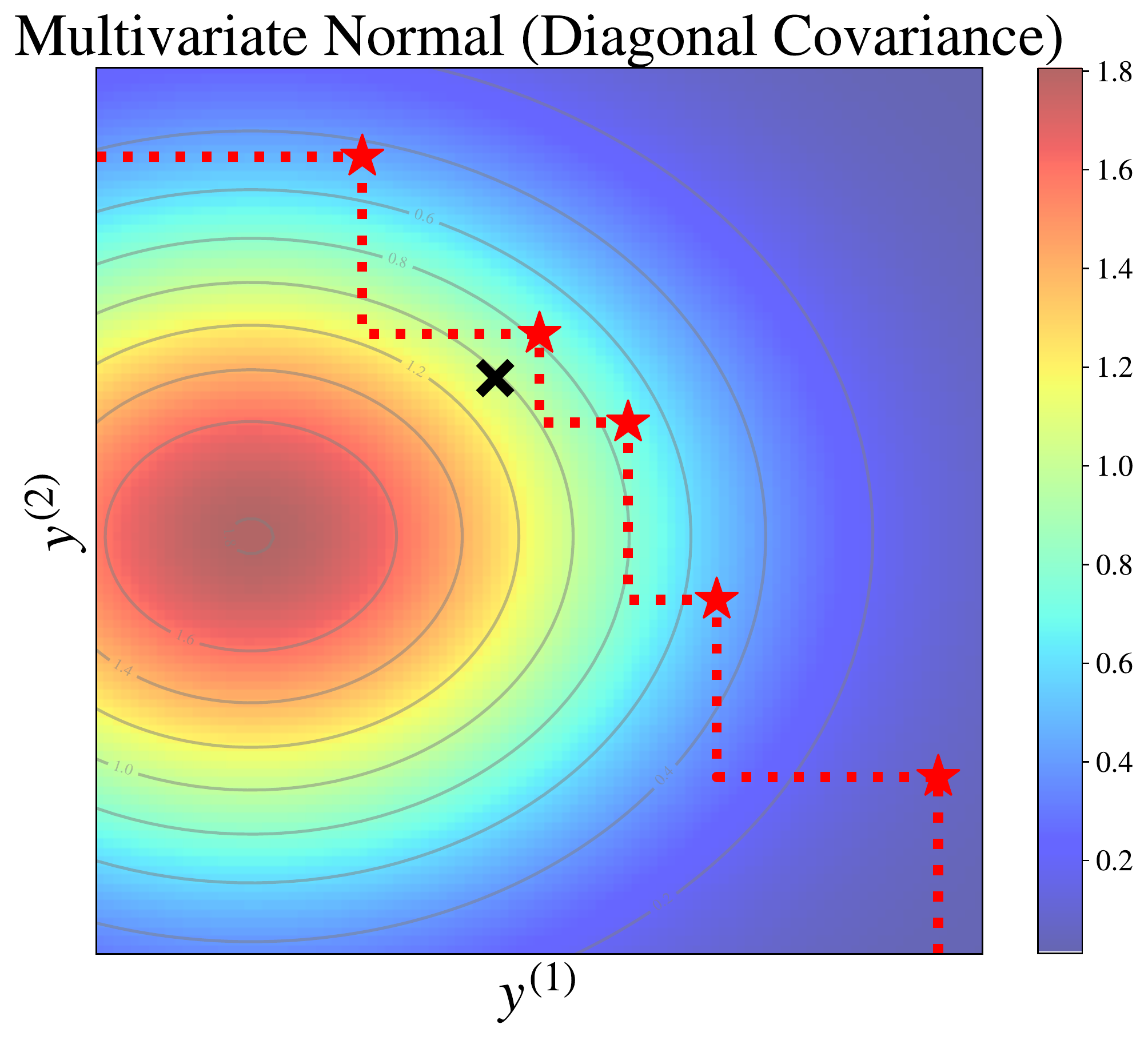}
	\caption{}
	\label{fig:mvn_diag_cov}
\end{subfigure}
\centering
\caption{Comparison of the density approximations to the skew multivariate normal distribution $p(\mathbf{y}| \mathbf{x}, D_{n*}, f(\mathbf{x}) \preceq \mathbb{Y}^*)$ shown in (a) for a single input $\mathbf{x} \in \mathbb{X}$ and sample Pareto front $\mathbb{Y}^*$. A zero noise assumption leads to the truncated multivariate normal approximation shown (b), whilst a moment matching approach leads to the multivariate normal approximations in (c) and (d).}
\label{fig:mvn_comparison}
\end{figure}

The entropy of \eqref{eqn:density} can be written as an $M$-dimensional expectation over the multivariate normal distribution $p(\mathbf{y}| \mathbf{x}, D_{n*}) = \mathcal{N}(\mathbf{y}; \boldsymbol{\mu}_{n*}(\mathbf{x}), \boldsymbol{\Sigma}_{n*}(\mathbf{x}, \mathbf{x}))$:
\begin{align}
	\begin{split}
		&H[p(\mathbf{y}| \mathbf{x}, D_{n*}, f(\mathbf{x}) \preceq \mathbb{Y}^*)]
		\\
		&\quad =
		- \mathbb{E}_{p(\mathbf{y}| \mathbf{x}, D_{n*})}
		\left[
		\frac{p(f(\mathbf{x}) \preceq \mathbb{Y}^*| \mathbf{x}, D_{n+})}
		{p(f(\mathbf{x}) \preceq \mathbb{Y}^*| \mathbf{x}, D_{n*})} 
		\log 
		\left(
		\frac{p(f(\mathbf{x}) \preceq \mathbb{Y}^*| \mathbf{x}, D_{n+})}
		{p(f(\mathbf{x}) \preceq \mathbb{Y}^*| \mathbf{x}, D_{n*})}
		p(\mathbf{y}| \mathbf{x}, D_{n*}) 
		\right)
		\right].
	\end{split}
	\label{eqn:posterior_entropy}
\end{align}
To simplify the notation, we define the $m$-th standardized value by 
\begin{equation}
	\gamma_m(z) = (z- \mu_{n*}^{(m)}(\mathbf{x})) / \sqrt{\Sigma_{n*}^{(m)}(\mathbf{x}, \mathbf{x})}
\end{equation}
for any scalar $z \in \mathbb{R}$. Using this function together with \cref{lemma:cumul}, we denote the cumulative distribution $p(f(\mathbf{x}) \preceq \mathbb{Y}^*| \mathbf{x}, D_{n*})$ by $W = \sum_{j=1}^J W_j = \sum_{j=1}^J \prod_{m=1}^M W_{j, m}$, where 
\begin{equation}
	W_{j, m} = \Phi(\gamma_m(u_j^{(m)})) - \Phi(\gamma_m(l_j^{(m)}))
\end{equation}
are the differences appearing in \eqref{eqn:cumulative_distribution}. 
Moreover, we denote the differences of the first derivative of $W_{j, m}$ and the negative of the second derivative (with respect to $\gamma_m$) by 
\begin{gather}
	G_{j, m} = \phi(\gamma_m(u_j^{(m)})) - \phi(\gamma_m(l_j^{(m)})),
	\\
	V_{j, m} = \gamma_m(u_j^{(m)}) \phi(\gamma_m(u_j^{(m)})) - \gamma_m(l_j^{(m)})\phi(\gamma_m(l_j^{(m)})),
\end{gather}
where $\phi$ is the probability density function of a standard normal distribution. In the setting where the observation noise is zero, the conditional distribution is a truncated multivariate normal, which is known to have an analytical equation formula for the entropy (Theorem 3.1. in \cite{suzuki2020icml}). In \cref{app:zero_variance}, we construct an ad hoc extension to this expression when the observation noise is non-zero. 
\\ \\
In the noisy setting, the distribution of interest is a type of multivariate skew normal distribution, which is known to not have an analytical form for the entropy \cite{arellano-valle2013sjs}. As a result, we propose two approximation strategies to estimate this entropy. The first strategy is to approximate the integral using Monte Carlo. The details of the Monte Carlo estimate $h^{\text{JES-MC}}$ is described in \cref{app:monte_carlo}. The second strategy is to directly approximate the distribution with one that exhibits an analytical entropy. We consider the most obvious choice, which is a multivariate normal distribution with the same first two moments. The same strategy was proposed in \cite{moss2021jmlr} for the single-objective multi-fidelity MES acquisition function. By a standard result (Chapter 12 of \cite{cover2006}), the entropy of this approximating distribution is actually an upper bound for the entropy of interest: $H[p(\mathbf{y}| \mathbf{x}, D_{n*}, f(\mathbf{x}) \preceq \mathbb{Y}^*)]
\leq \frac{M}{2}\log(2\pi e)
+ \frac{1}{2} \log \det \mathbb{V}\text{ar}
(\mathbf{y}| \mathbf{x}, D_{n*}, f(\mathbf{x}) \preceq \mathbb{Y}^*)$.
The following result shows that the these central moments can be computed analytically.
\begin{proposition}
	Under the modelling set-up outlined in \cref{sec:prelim}, for an input $\mathbf{x} \in \mathbb{X}$ the first and second central moment of $p(\mathbf{y}| \mathbf{x}, D_{n*}, f(\mathbf{x}) \preceq \mathbb{Y}^*)$ are
	\begin{align*}
		\mathbb{E}[y^{(m)}| \mathbf{x}, D_{n*}, f(\mathbf{x}) \preceq \mathbb{Y}^*] = \mu_{n*}^{(m)}(\mathbf{x}) 
		-  \frac{\sqrt{\Sigma_{n*}^{(m)}(\mathbf{x}, \mathbf{x})}}{W} \sum_{j=1}^J W_j \frac{G_{j, m}}{W_{j, m}}
	\end{align*}
	and
	\begin{align*}
		&\text{\normalfont Cov}\left(
		y^{(m)}, y^{(m')}
		\Big \vert \mathbf{x}, D_{n*}, f(\mathbf{x}) \preceq \mathbb{Y}^*
		\right)
		\\
		&=
		\begin{cases}
			\frac{\sqrt{\Sigma_{n*}^{(m)}(\mathbf{x}, \mathbf{x})} \sqrt{\Sigma_{n*}^{(m')}(\mathbf{x}, \mathbf{x})}}{W} 
			\sum_{j=1}^J  W_j
			\frac{G_{j, m}}{W_{j, m}}
			\left(
			\frac{G_{j, m'}}{W_{j, m'}}
			-
			\frac{1}{W} \sum_{j'=1}^J W_{j'} \frac{G_{j', m'}}{W_{j', m'}}  
			\right),
			& 
			m \neq m'; 
			\\
			\Sigma_{n*}^{(m)}(\mathbf{x}, \mathbf{x}) + \sigma^{(m)}(\mathbf{x}) - 
			\frac{ \Sigma_{n*}^{(m)}(\mathbf{x}, \mathbf{x})}{W} 
			\left(
			\sum_{j=1}^J W_j \frac{V_{j, m}}{W_{j, m}}
			+ \frac{1}{W} \left(
			\sum_{j=1}^J W_j
			\frac{G_{j, m}}{W_{j, m}}
			\right)^2
			\right),
			&
			m = m'.
		\end{cases}
	\end{align*}
	\label{prop:skew_normal_moments}
\end{proposition}
\noindent
As an upper bound on the conditional entropy leads to a lower bound on the mutual information, we will refer to the resulting conditional entropy estimate as the JES-LB estimate:
\begin{equation}
	h^{\text{JES-LB}}((\mathbb{X}^*, \mathbb{Y}^*); \mathbf{x}, D_n) =
	\frac{M}{2}\log(2\pi e) + \frac{1}{2} \log \det \mathbb{V}\text{ar}
	(\mathbf{y}| \mathbf{x}, D_{n*}, f(\mathbf{x}) \preceq \mathbb{Y}^*).
	\label{eqn:h_lb}
\end{equation}
We could obtain a further lower bound by ignoring the off-diagonal terms in the covariance matrix. We dub the resulting approximation as the JES-LB2 entropy estimate:
\begin{equation}
	h^{\text{JES-LB2}}((\mathbb{X}^*, \mathbb{Y}^*); \mathbf{x}, D_n) =
	\frac{M}{2}\log(2\pi e) + \frac{1}{2} \sum_{m=1}^M \log \mathbb{V}\text{ar}
	(y^{(m)} | \mathbf{x}, D_{n*}, f(\mathbf{x}) \preceq \mathbb{Y}^*).
	\label{eqn:h_lb2}
\end{equation}
\cref{fig:mvn_comparison} presents an illustration of the different density approximations that are used within the various conditional entropy estimates. An important remark is that all the conditional entropy estimates that we have developed here can also be applied to estimate MES. The only difference in the MES algorithm is that we no longer apply the conditioning step (\cref{alg:jes:condition} of \cref{alg:jes}) because we are interested in estimating $H[p(\mathbf{y}| \mathbf{x}, D_n, f(\mathbf{x}) \preceq \mathbb{Y}^*)]$ as opposed to \eqref{eqn:posterior_entropy}. Consequently, the MES acquisition function is cheaper to evaluate because the cost of evaluating the posterior variance at a single input is $O(n^2)$, whereas JES incurs a cost of $O((n+|\mathbb{Y}^*|)^2)$---more details are presented in the cost analysis in \cref{app:cost_analysis}.
%
%

\subsection{Batch evaluations} 

Evaluating the JES acquisition functions for a batch of points $\mathbf{x}^{[1:q]} = (\mathbf{x}^{[1]}, \dots, \mathbf{x}^{[q]}) \in \mathbb{X}^q$ is expensive because the entropy estimates now depends on the $q$-dimensional normal CDF and its derivatives. To circumvent this issue, we follow the example of \cite{moss2021jmlr} and propose a suboptimal batch approach by upper bounding the expensive joint conditional entropy term by the sum of the individual entropies: $H[p(\mathbf{y}^{[1:q]}| \mathbf{x}^{[1:q]}, D_{n*}, f(\mathbb{X}) \preceq \mathbb{Y}^*)] \leq \sum_{i=1}^q H[p(\mathbf{y}^{[i]}| \mathbf{x}^{[i]}, D_{n*}, f(\mathbb{X}) \preceq \mathbb{Y}^*)]$. The resulting $q$-batch lower bound JES estimate is given by
\begin{equation}
	\hat{\alpha}^{q\text{LB-JES}}(\mathbf{x}^{[1:q]}| D_n) 
	= H[p(\mathbf{y}^{[1:q]}| \mathbf{x}, D_n)]
	- \frac{1}{S} \sum_{s=1}^S \sum_{i=1}^q h((\mathbb{X}^*_s, \mathbb{Y}^*_s); \mathbf{x}^{[i]}, D_n),
	\label{eqn:qjes_base}
\end{equation}
where $h$ is the conditional entropy estimate and
\begin{equation}
	H[p(\mathbf{y}^{[1:q]}| \mathbf{x}, D_n)] = \frac{M}{2} \log(2 \pi e) + \frac{1}{2} \sum_{m=1}^M \log \det (\Sigma_n^{(m)}(\mathbf{x}^{[1:q]}, \mathbf{x}^{[1:q]}) + \text{diag}(\sigma^{(m)}(\mathbf{x}^{[1:q]})))
\end{equation}
is the initial entropy. This acquisition function is defined over a $qD$-dimensional space, which becomes more difficult to optimize as $q$ increases. Alternatively, we can maximize this function greedily by sequentially selecting the best input conditioned on the previously chosen points. This greedy procedure satisfies an $e^{-1}$ regret bound when the acquisition function is submodular \cite{wilson2018anips}. In \cref{app:submodularity}, we show that this batch acquisition function is indeed submodular. 

\FloatBarrier

\section{Performance criteria}
\label{sec:performance_criteria}
\begin{figure}
	\begin{subfigure}[t]{0.48\textwidth}
		\includegraphics[width=.48\linewidth]{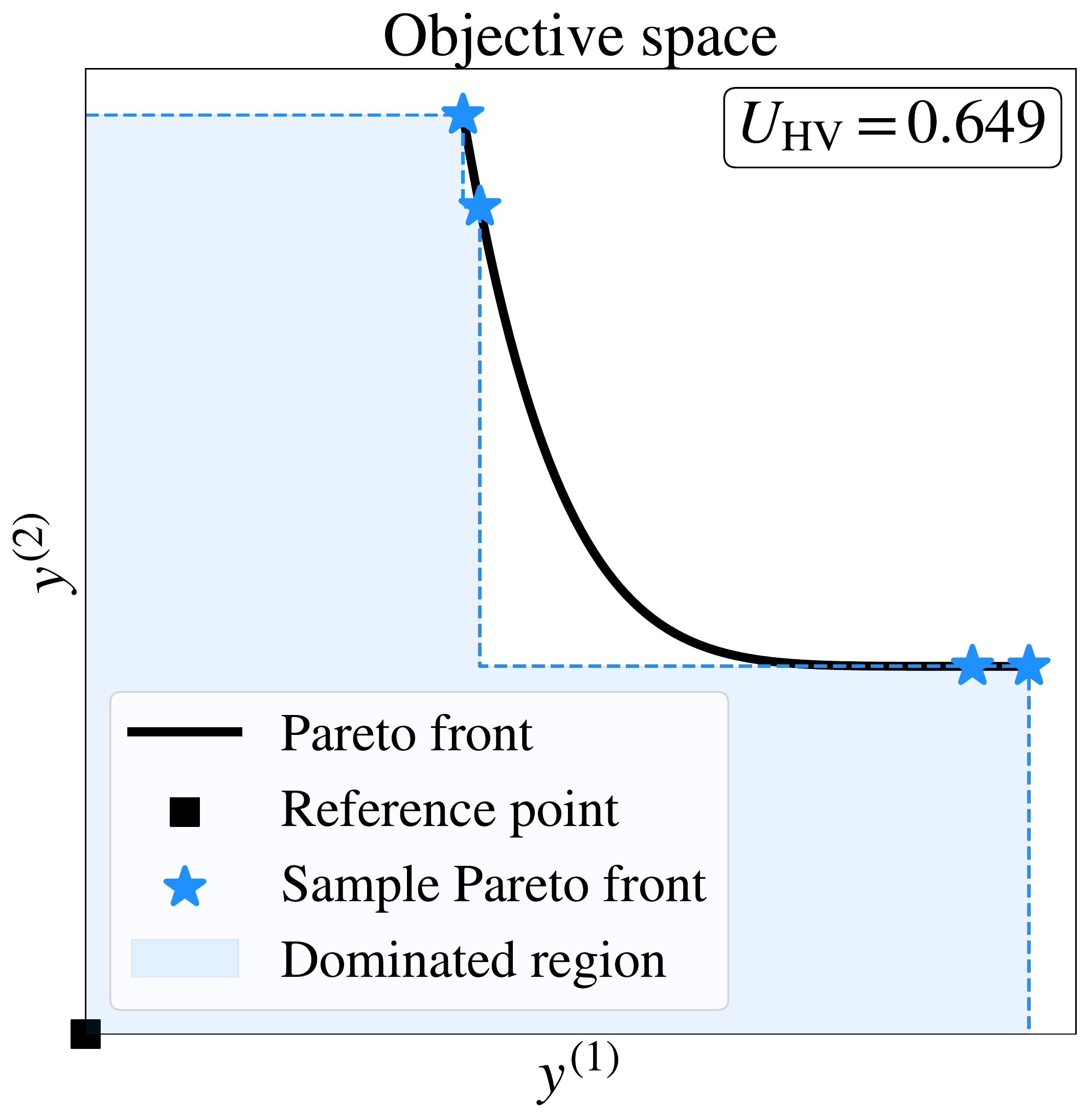}
		\includegraphics[width=.48\linewidth]{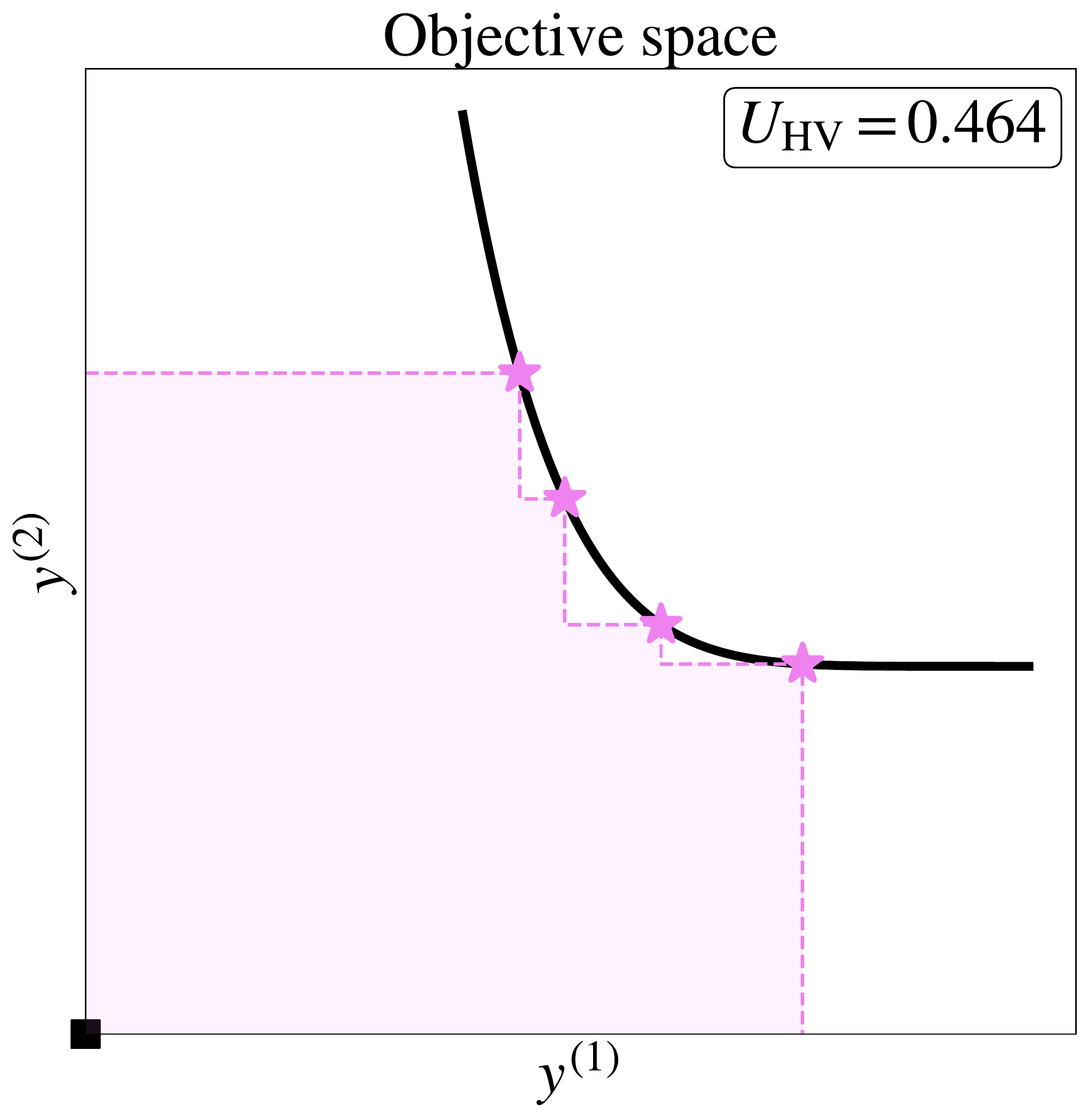}
		\caption{}
		\label{fig:thv}
	\end{subfigure}
	\begin{subfigure}[t]{0.48\textwidth}
		\includegraphics[width=.48\linewidth]{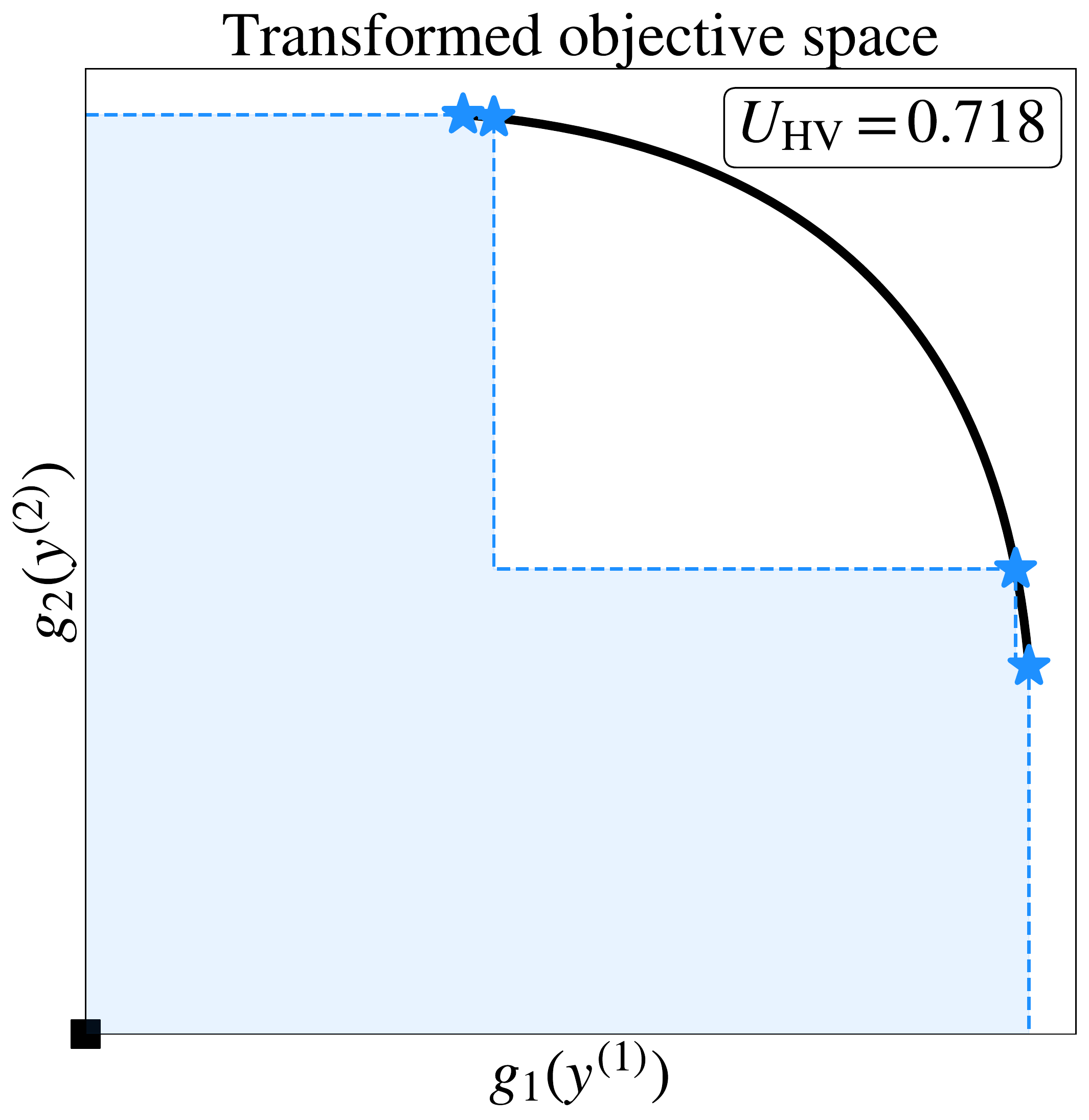}
		\includegraphics[width=.48\linewidth]{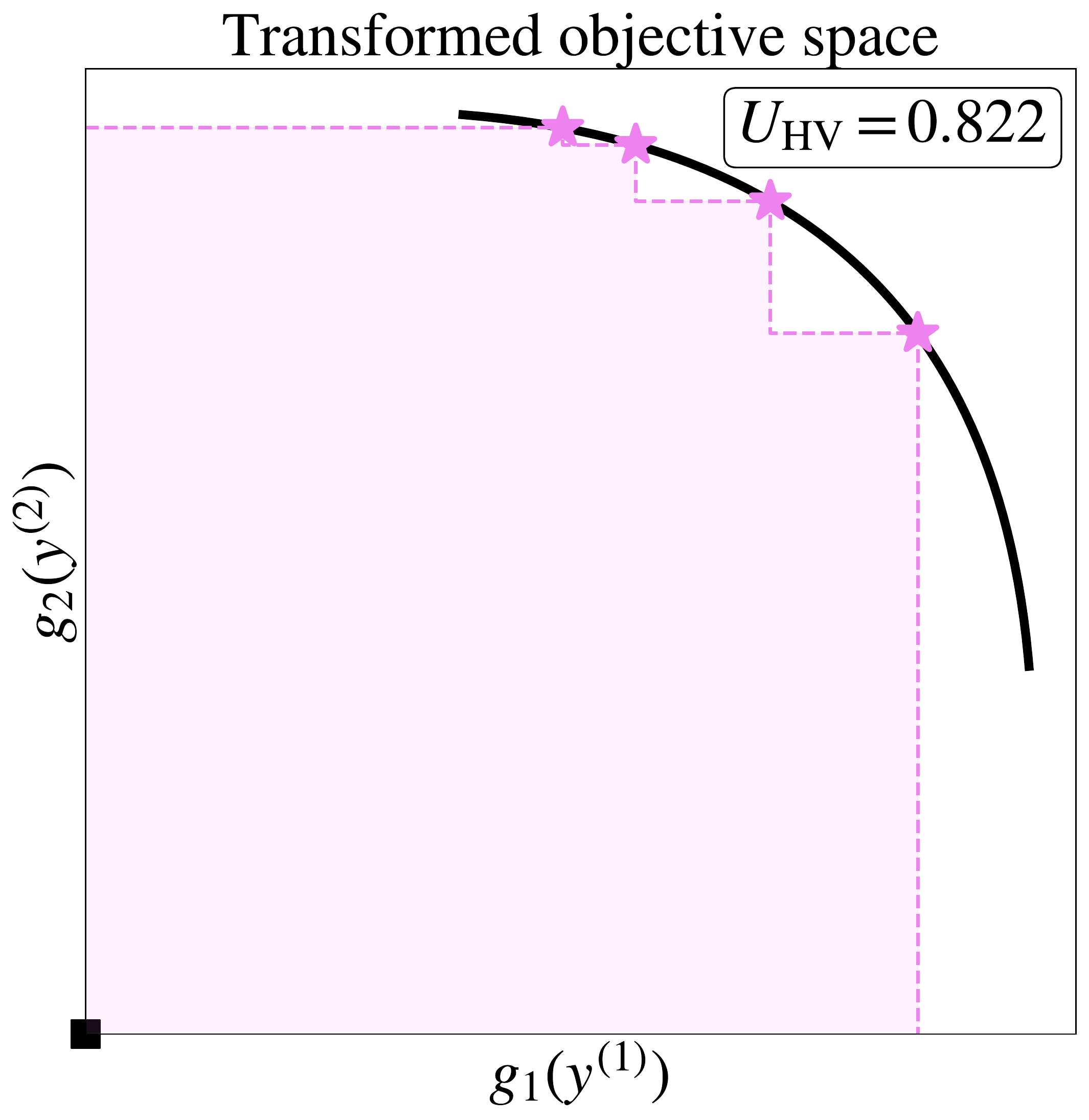}
		\caption{}
		\label{fig:hv}
	\end{subfigure}
	\centering
	\caption{Comparison of the HV of two sample Pareto fronts in (a) the standard objective space $\mathbf{y} = (y^{(1)}, y^{(2)})$ and (b) the transformed objective space $g(\mathbf{y}) = (g_1(y^{(1)}), g_2(y^{(2)}))$ described in \cref{app:hypervolume_indicator}. The HV indicator prefers a different set depending on the choice of parameterization.}
\end{figure}

In multi-objective optimization, the most common way to measure performance is by comparing the approximate Pareto set $\hat{\mathbb{X}}^*$ against the optimal Pareto set $\mathbb{X}^*$ in the objective space: $d(f(\hat{\mathbb{X}}^*), f(\mathbb{X}^*))$ where $d: 2^{\mathbb{R}^M} \times 2^{\mathbb{R}^M} \rightarrow \mathbb{R}$ is a function that measures the discrepancy between the sets of objective vectors.
Existing work in multi-objective BO mainly focusses on the hypervolume (HV) discrepancy, $d_{\text{HV}}(A, B) = |U_{\text{HV}}(A) - U_{\text{HV}}(B)|$, where the HV indicator, $U_{\text{HV}}(A) = \int_{\mathbb{R}^M} \mathbb{I}[\mathbf{r} \preceq \mathbf{z} \preceq A] d\mathbf{z}$, is defined as the volume between a reference point 
$\mathbf{r} \in \mathbb{R}^M$ and a set $A \subset \mathbb{R}^M$. The general guidance is to set reference point to be slightly worse than the nadir, which is the vector consisting of the worst possible points, $\min_{\mathbf{x} \in \mathbb{X}} f^{(m)}(\mathbf{x})$, for objectives $m=1,\dots,M$---see \cite{ishibuchi2018ec} for more details.
\\ \\
An attractive feature of the HV indicator is that it is Pareto complete (or compliant) in the sense that a better set will lead to a larger HV \cite{zitzler2003itec}: $A \succ B \implies U_{\text{HV}}(A) > U_{\text{HV}}(B)$, if we assume the sets $A$ and $B$ are finite. The reverse implication known as Pareto compatibility does not hold for the HV indicator \cite{zitzler2003itec}. In other words, the HV can be used to discriminate between sets where one dominates another, but it cannot be relied upon when the sets are incomparable. Not all incomparable sets are treated equally by the HV indicator \cite{zitzler1998ppsn}. For instance \cref{fig:thv} shows an example where the HV indicator places more emphasis on the end points of the Pareto front. On other hand, if we apply a monotonically increasing transformation $g_m:\mathbb{R} \rightarrow \mathbb{R}$ to each objective, the Pareto set will not change, whereas the HV comparison will (\cref{fig:hv}). Implicitly, the HV indicator assumes that a linear change in one objective is equivalent to a linear change in another. This assumption might not necessarily reflect the decision maker's outlook and this is something that is typically overlooked when designing and benchmarking multi-objective optimization algorithms.
The following result shows that information-theoretic acquisitions function are in fact agnostic to the choice of parameterization. 

\begin{proposition}
	The information-theoretic acquisition functions {\normalfont$\alpha^{\text{PES}}$}, {\normalfont$\alpha^{\text{MES}}$} and {\normalfont$\alpha^{\text{JES}}$} are invariant to reparameterization of the objective space that are consistent with the Pareto ordering relations. For example, {\normalfont $\alpha^{\text{JES}}(\mathbf{x}| D_n) = \text{MI}(\mathbf{y}; (\mathbb{X}^*, \mathbb{Y}^*)| D_n) = \text{MI}(g(\mathbf{y}); (\mathbb{X}^*, g(\mathbb{Y}^*))| D_n)$}, where the $g_m:\mathbb{R} \rightarrow \mathbb{R}$ is a strictly monotonically increasing function acting only on the $m$-th objective.
	\label{prop:invariance}
\end{proposition}



To benchmark the algorithms, we use both the standard HV discrepancy (\cref{sec:experiments}) and the HV discrepancy under different parameterizations (\cref{app:experiments}). To easily obtain a family of parameterizations, we devise a novel weighting approach in \cref{app:hypervolume_indicator}, which exploits the fact that the HV indicator can be written as an expectation over a uniform distribution on the $(M-1)$-dimensional hypercube \cite{zhang2020icml, deng2019itec}. We observe that it is possible to assess the performance at different locations of the objective space by using alternate distributions over the hypercube. We call the resulting metric the generalized hypervolume (GHV). In our experiments, we found that the performance of each algorithm changed with regards to the choice of parameterization, but the JES approaches tended to be one of the strongest performers throughout.
\section{Experiments}
\label{sec:experiments}
We empirically evaluate the JES acquisition function on a range of synthetic and real-world benchmark problems. We compare this approach with some popular acquisition functions in multi-objective BO: TSEMO \cite{bradford2018jgo}, ParEGO \cite{knowles2006itec}, NParEGO \cite{daulton2021anips}, EHVI \cite{daulton2020anips}, NEHVI \cite{daulton2021anips}, PES \cite{garrido-merchan2019n, garrido-merchan2021a} and MES-0 \cite{suzuki2020icml}. We have also included the MES-LB, MES-LB2 and MES-MC acquisition functions, which can be easily derived from the conditional entropy estimates that we developed here. All algorithms are based on the open source Python library BoTorch \cite{balandat2020anips}, which uses features from GPyTorch \cite{gardner2018anips} for Gaussian process regression and PyTorch \cite{paszke2019anips} for automatic differentiation. All experiments are repeated using $100$ different initial seeds and we generate the Pareto set recommendation $\hat{\mathbb{X}}^*$ of $50$ points by maximizing the posterior mean using a multi-objective solver (NSGA2 \cite{deb2002itec} from the Pymoo library \cite{blank2020ia}).  The complete details of the experiments are outlined in \cref{app:experiments}, whilst the code is available at \href{https://github.com/benmltu/JES}{https://github.com/benmltu/JES}.

\subsection{Benchmarks}

\paragraph{Synthetic benchmark.} We consider the ZDT2 \cite{deb2002itec} benchmark with $D=6$ inputs and $M=2$ objectives. We corrupt the observations with additive Gaussian noise with zero-mean and standard deviation set to approximately $10\%$ of the objective ranges.

\paragraph{Chemical reaction.}  This benchmark considers a nucleophilic aromatic substitution reaction (SnAr) between 2,4-difluoronitrobenzene and pyrrolidine in ethanol to produce a mixture of a desired product and two side-products \cite{hone2017rce}. The design space comprises of $D=4$ components relating to the initial conditions. The goal is to optimize $M=2$ objectives, namely the space time yield and the environmental impact. We apply a logarithm transform to the objectives and contaminate the observations with additive Gaussian noise with zero-mean and standard deviation set to approximately $3\%$ of the resulting objective ranges in order to emulate a potential real-world scenario.

\paragraph{Pharmaceutical manufacturing.} This problem is concerned with optimizing the Penicillin production process outlined in \cite{liang2021n2asw}. The design space is made up of $D=7$ elements that control the initial condition of the reactions. The goal is to optimize $M=3$ objectives, which relates to the yield, the amount of carbon dioxide released and the time to ferment. We include additive zero-mean Gaussian noise with a standard deviation set to approximately $1\%$ of the objective ranges. 

\paragraph{Marine design.} This problem considers optimizing a family of bulk carriers subject to the constraints imposed for ships travelling through the Panama Canal \cite{sen1998, parsons2004josr}. The design space is made up of $D=6$ variables that determine the architecture of the carriers. The goal of this problem is to maximize the annual cargo, whilst minimizing the transportation cost and the ship weight subject to some design constraints. We consider the reformulation in \cite{tanabe2020asc}, which converts the constraints into another objective. For this reformulated $M=4$ objective problem, we corrupt the observations with additive zero-mean Gaussian noise with standard deviation set to approximately $0.5 \%$ of the objective ranges. 

\subsection{Results and discussion}

\begin{figure}
	\includegraphics[width=1\linewidth]{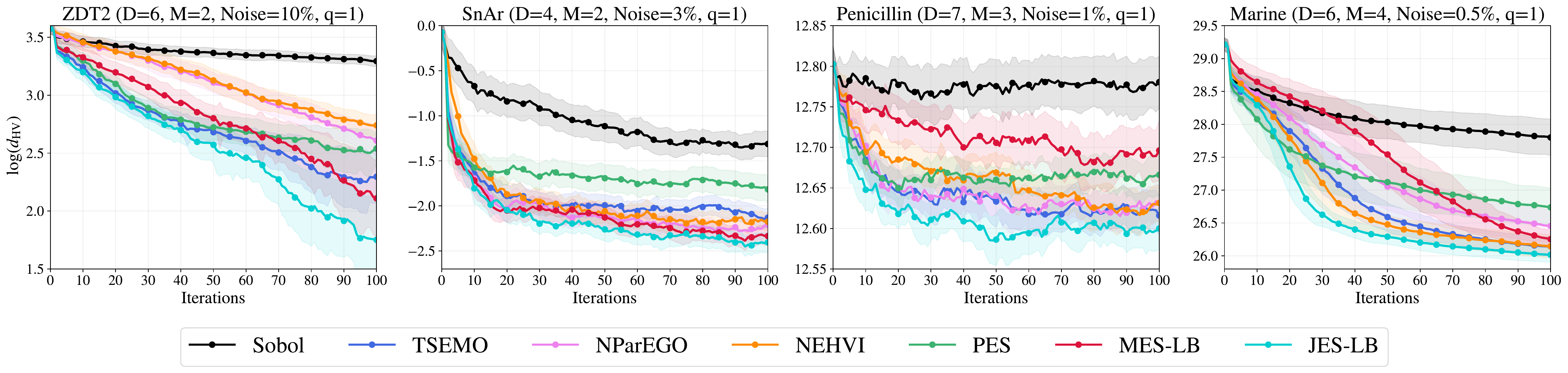}
	\includegraphics[width=1\linewidth]{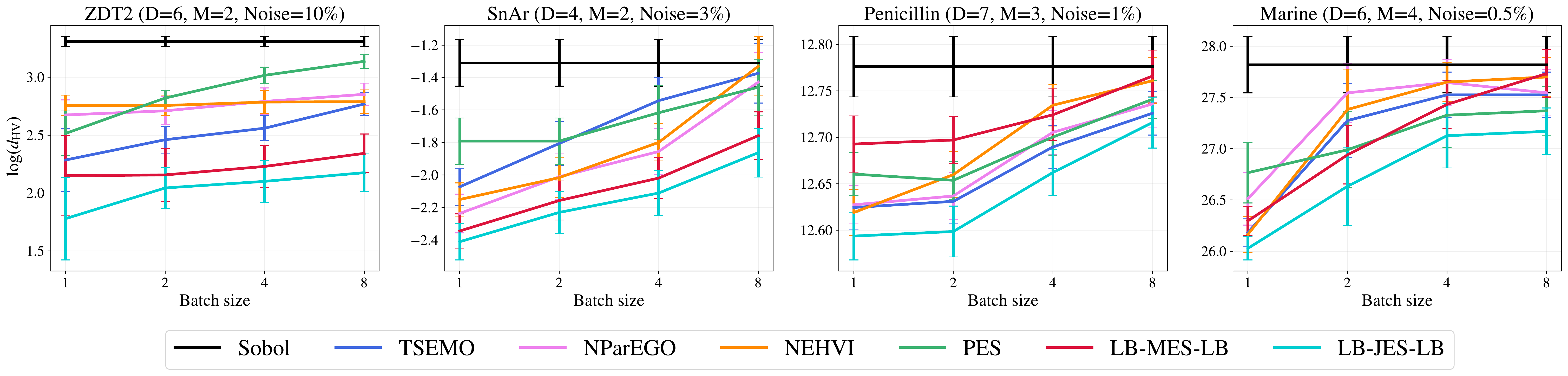}
	\centering
	\caption{A comparison of the mean logarithm HV discrepancy with two standard errors over one hundred runs for the four benchmark problems on a subset of the algorithms. We present the sequential and batch results on the top and bottom, respectively.}
	\label{fig:experiments_hv}
\end{figure}

We present the log HV discrepancy results for both the sequential and batch experiments in \cref{fig:experiments_hv}.  The JES approach is consistently one of the stronger performing algorithms for these set of experiments. A similar conclusion is reached when we consider the weighted variant of the hypervolume in \cref{app:ghv}.

\paragraph{Conditional entropy estimates.} We compared the performance of the different conditional entropy estimates for both the JES and MES acquisition function in \cref{app:conditional_entropy_estimates}. We observed that in the majority of cases all the estimates exhibit similar performance. As a result, we recommend using the cheapest approximation, which is usually the lower bound estimates, judging from the wall times presented in \cref{app:wall_times}. 

\paragraph{Acquisition wall times.} The wall times in \cref{app:wall_times} indicate that the cost of acquiring a new point with JES is comparable with NEHVI, slightly more expensive than MES, but cheaper than PES. We note that the wall times for all methods can be improved by taking advantage of parallelization. In particular, for entropy based methods we used a gradient-free optimizer to sequentially optimize the multi-objective samples (\cref{alg:jes:moo} in \cref{alg:jes}), whereas in practice we should of ideally solved these problems in parallel using a gradient-based optimizer such as \cite{liu2021anips}.

\paragraph{Querying high performing points.} In certain domains it might be useful to directly query high-performing points because the final decision will be restricted to only the sampled locations $X_N$. In \cref{app:in-sample}, we investigated the performance when such a restriction was made. In this setting, we observed that the information-theoretic approaches were occasionally outperformed by the improvement and scalarization based acquisition functions, which picked points more greedily. We observed that the entropy based approaches had a tendency to pick points that are more informative for the posterior over the optimal points as opposed to directly selecting a point that is known to perform well. To address this setting, we recommend combining information-theoretic acquisition functions with an epsilon greedy approach, where points are occasionally picked according to a greedy strategy such as maximizing a function of the posterior mean.

\begin{figure}
	\includegraphics[width=1\linewidth]{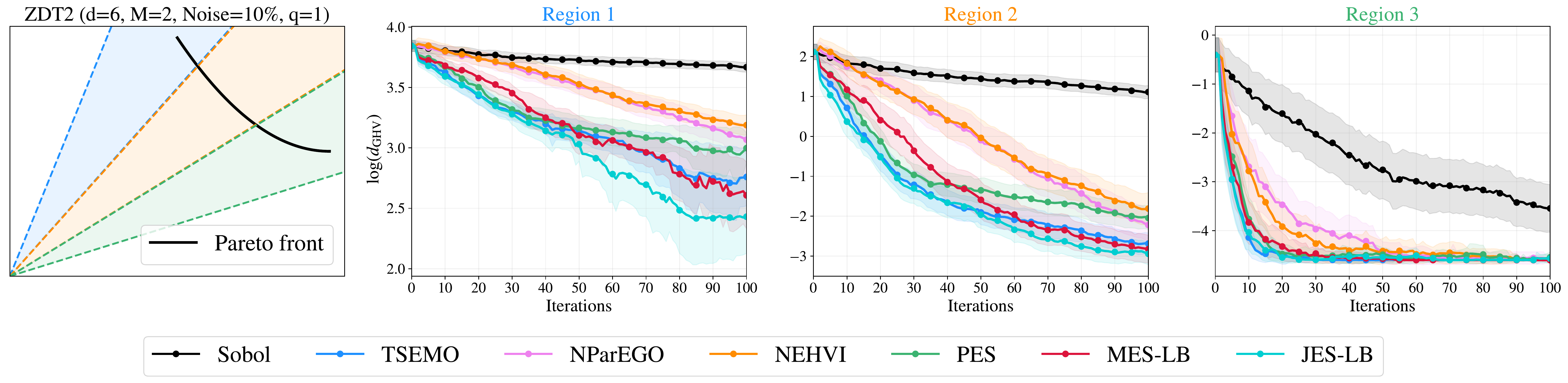}
	\includegraphics[width=1\linewidth]{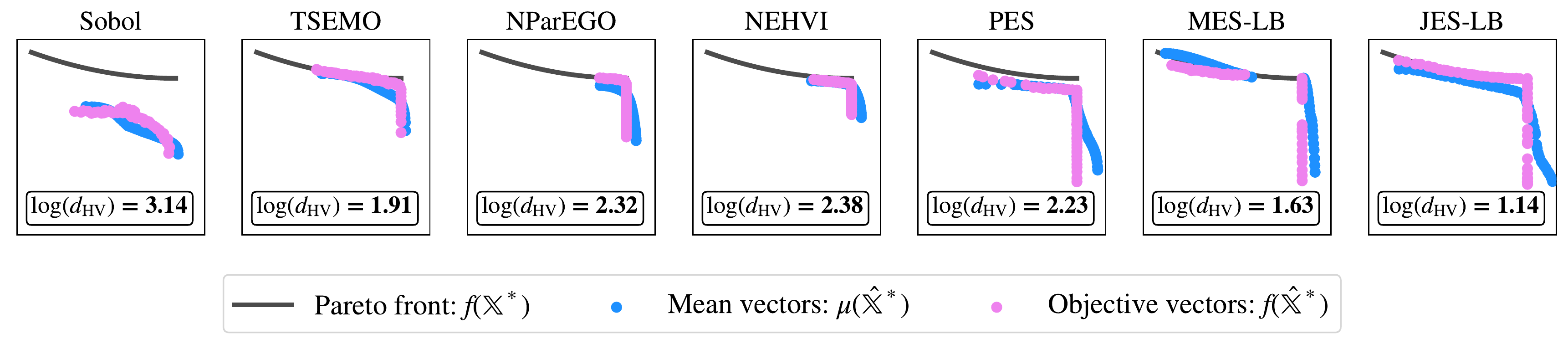}
	\centering
	\caption{An example of the generalized hypervolume on the ZDT2 benchmark. On the top plot we present the mean logarithm GHV discrepancy results for three different regions. On the bottom plot, we present the Pareto front and its approximation at the final time for the run which achieved the top 20th percentile on the standard hypervolume.}
	\label{fig:zdt2_example}
\end{figure}

\paragraph{Assessing local performance.} Using the generalized hypervolume, we can target different parts of the objective space in order to get a much better picture of performance. We demonstrate this on a simple bi-objective example in \cref{fig:zdt2_example}, where we assess that quality of the approximations at three different regions of the objective space. We observe that all of the BO algorithms were quickly able to identify the right section of the Pareto front. Evidently, the main source of difficulty for this problem arises from approximating the points in the left section of the Pareto front, which favours the second objective. This observation would not be apparent if we focussed solely on the standard HV.

\paragraph{General guidance.} The ideal acquisition functions is problem dependent and strongly depends on the decision maker's plans and goals. In a completely black-box setting, where there is no immediate preferences, \cref{prop:invariance} and the empirical results motivates the usage of information-theoretic acquisition functions, which treats all points on the Pareto front as equally desirable a priori.

\section{Conclusion}
\label{sec:conclusion}
We introduced JES, a novel information-theoretic acquisition function for multi-objective BO. To approximate this acquisition function, we presented several approximations to the conditional entropy and also a simple extension for the batch setting. Experimental results suggest that JES is very competitive with existing acquisition functions in terms of the HV discrepancy and its weighted variants. The main limitation of the JES acquisition function is that it relies on routines such as box decompositions and multi-objective optimization of function samples, which can be expensive to execute for very large problems. Future work could focus on improving the scalability of these information-theoretic methods and extending it to more general settings, which include constrained, decoupled and multi-fidelity optimization---see \cref{app:extensions} for more details.

\begin{ack}
	BT was supported by the EPSRC StatML CDT programme EP/S023151/1 and BASF SE, Ludwigshafen am Rhein. NK was partially funded by JPMorgan Chase \& Co. under J.P. Morgan A.I. Faculty Research Awards 2021.
\end{ack}

{
\small
\bibliographystyle{plainnatremove}
\bibliography{ms}
}

\newpage
\section*{Checklist}
\begin{enumerate}

	\item For all authors...
	\begin{enumerate}
		\item Do the main claims made in the abstract and introduction accurately reflect the paper's contributions and scope?
		\answerYes{}
		\item Did you describe the limitations of your work?
		\answerYes{}
		\item Did you discuss any potential negative societal impacts of your work?
		\answerNA{}
		\item Have you read the ethics review guidelines and ensured that your paper conforms to them?
		\answerYes{}
	\end{enumerate}

	\item If you are including theoretical results...
	\begin{enumerate}
		\item Did you state the full set of assumptions of all theoretical results?
		\answerYes{}
		\item Did you include complete proofs of all theoretical results?
		\answerYes{The proofs for the main results are in the appendix.}
	\end{enumerate}

	\item If you ran experiments...
	\begin{enumerate}
		\item Did you include the code, data, and instructions needed to reproduce the main experimental results (either in the supplemental material or as a URL)?
		\answerYes{The code is in the supplementary material and is available at \href{https://github.com/benmltu/JES}{https://github.com/benmltu/JES}.}
		\item Did you specify all the training details (e.g., data splits, hyperparameters, how they were chosen)?
		\answerYes{The experimental set-up is described in \cref{app:experiments}.}
		\item Did you report error bars (e.g., with respect to the random seed after running experiments multiple times)?
		\answerYes{}
		\item Did you include the total amount of compute and the type of resources used (e.g., type of GPUs, internal cluster, or cloud provider)?
		\answerYes{The computing resources are described in the caption of the wall times plots e.g. \cref{app:wall_times}.}
	\end{enumerate}

	\item If you are using existing assets (e.g., code, data, models) or curating/releasing new assets...
	\begin{enumerate}
		\item If your work uses existing assets, did you cite the creators?
		\answerYes{}
		\item Did you mention the license of the assets?
		\answerNo{We only cited the original code and corresponding manuscript in \cref{app:experiments}.}
		\item Did you include any new assets either in the supplemental material or as a URL?
		\answerYes{The code is in the supplementary material and will be made public in due time.}
		\item Did you discuss whether and how consent was obtained from people whose data you're using/curating?
		\answerNA{}
		\item Did you discuss whether the data you are using/curating contains personally identifiable information or offensive content?
		\answerNA{}
	\end{enumerate}

	\item If you used crowdsourcing or conducted research with human subjects...
	\begin{enumerate}
		\item Did you include the full text of instructions given to participants and screenshots, if applicable?
		\answerNA{}
		\item Did you describe any potential participant risks, with links to Institutional Review Board (IRB) approvals, if applicable?
		\answerNA{}
		\item Did you include the estimated hourly wage paid to participants and the total amount spent on participant compensation?
		\answerNA{}
	\end{enumerate}

\end{enumerate}

\newpage
\appendix
{\Large \textbf{Appendix to: \\[0.5cm] Joint Entropy Search for Multi-Objective Bayesian Optimization}}
\section{Basics}
\subsection{Bayesian optimization pseudocode}
\label{app:bo}
Bayesian optimization consists of two steps that are iterated until the budget of function evaluations $N$ is exhausted. The first step is modelling, this is where posterior of the probabilistic model $p(f | D_n)$ is computed. The second step is acquisition, this is where a utility function is optimized in order to determine the next location to query. The pseudo-code for this procedure is presented in \cref{alg:bo}. For a more thorough overview on Bayesian optimization consult the references \cite{brochu2010a, shahriari2016pi, frazier2018a}.
\\ \\
\begin{algorithm}[H]
	\SetKwInOut{Input}{Input}
	\DontPrintSemicolon
	\Input{A black-box function $f$.}
	Initialize the probabilistic model $p(f)$.
	\\
	\For{$n=0,\dots,N-1$}{
		Optimize for the next point
		$\mathbf{x}_{n+1} = \argmax_{\mathbf{x} \in \mathbb{X}} \alpha(\mathbf{x}| D_n)$.
		\\
		Evaluate the function
		$\mathbf{y}_{n+1} = f(\mathbf{x}_{n+1}) + \boldsymbol{\epsilon}$.
		\\
		Augment the data set
		$D_{n+1} = D_n \cup \{(\mathbf{x}_{n+1}, \mathbf{y}_{n+1})\}$.
		\\
		Compute the posterior $p(f| D_n)$.
	}
	\Return $D_N$ and $p(f| D_N)$.
	\vspace{0.2cm}
	\caption{Multi-objective Bayesian Optimization.}
	\label{alg:bo}
\end{algorithm}
\subsection{Posterior Gaussian process}
\label{app:gp}
Under the independent Gaussian observation model described in \cref{sec:prelim}, the posterior $p(f(X)|D_n)$ evaluated at a vector $X \subset \mathbb{X}$, is a collection of Gaussian processes \cite{rasmussen2006} with mean
\begin{equation}
	\begin{split}
		&\mu_{n}^{(m)}(X) 
		\\
		&= \mu_{0}^{(m)}(X) + \Sigma_{0}^{(m)}(X, X_n) \left(
		\Sigma_{0}^{(m)}(X_n, X_n) + \text{diag}(\sigma^{(m)}(X_n)) 
		\right)^{-1}
		(Y_n^{(m)} - \mu_{0}^{(m)}(X_n))
	\end{split}
	\label{eqn:post_mean}
\end{equation}
and covariance 
\begin{align}
	\begin{split}
		&\Sigma_{n}^{(m)}(X, X) 
		\\
		&= \Sigma_{0}^{(m)}(X, X) - \Sigma_{0}^{(m)}(X, X_n) \left(
		\Sigma_{0}^{(m)}(X_n, X_n) + \text{diag}(\sigma^{(m)}(X_n)) 
		\right)^{-1}
		\Sigma_{0}^{(m)}(X_n, X),
	\end{split}
	\label{eqn:post_cov}
\end{align}
where we denote the collection of sampled locations by $(X_n)_t = \mathbf{x}_t$ and observations $(Y_n^{(m)})_t = y_t^{(m)}$ for objectives $m=1,\dots,M$ and data points $t=1,\dots,n$.  
\subsection{Sampling from a Gaussian process}
\label{app:gp_sampling}
Exact sampling of a one-dimensional Gaussian process over a finite set, $X \subset \mathbb{X}$, scales cubically with the number of points $|X|$. This cubic cost arises form the inversion of the covariance matrix. As we want to sample global maximizers and maximums $(\mathbb{X}^*, \mathbb{Y}^*)$, exact sampling over the whole space $\mathbb{X}$ is not computational feasible. We could restrict the sampling to a discrete subset of $\mathbb{X}$, but this would require designing a principled strategy to select a reasonable subset, for example we could try triangulating between existing points \cite{gramacy2021a}. To avoid this difficulty, we follow the example of previous work \cite{wang2017icml, hernandez-lobato2014anips}, which considers generating approximate samples via random Fourier features \cite{rahimi2008anips, wilson2020icml, wilson2021jmlr}. The strategy centres around approximating the covariance kernel using a feature representation $\Sigma_0^{(m)}(\mathbf{x}, \mathbf{x}') \approx \varphi(\mathbf{x})^T \varphi(\mathbf{x}')$. For a stationary covariance kernel $\Sigma_0^{(m)}$, the Fourier transform of the kernel is a non-negative measure that can be normalized to obtain a probability distribution $p(\boldsymbol{\theta})$---this result follows from Bochner's Theorem \cite{bochner1959} . For the exponential and Matern kernels, this probability distribution is known in closed-form \cite{rahimi2008anips, bradford2018jgo}. Assuming $\Sigma_0^{(m)}$ is stationary, an approximate sample of the Gaussian process prior GP$(\mu_0^{(m)}, \Sigma_0^{(m)})$ can be written as a Bayesian linear model 
\begin{equation}
	f^{(m)}_0(\cdot) = \mu_{0}^{(m)}(\cdot) + \sum_{i=1}^L \omega_i \varphi_i (\cdot)
\end{equation}
where $\varphi_i(\mathbf{x}) = \sqrt{2/L} \cos(\boldsymbol{\theta}_i^T\mathbf{x} + \tau_i)$ are the Fourier features depending on the random variables $\tau_i \sim U(0, 2\pi)$ and $\boldsymbol{\theta}_i \sim p(\boldsymbol{\theta})$, whilst $\omega_i \sim \mathcal{N}(0, 1)$ are the random weights for $i=1,\dots,L$. The posterior samples of GP$(\mu_n^{(m)}, \Sigma_n^{(m)})$ can be obtained by adding an additional pathwise update via Matherons' rule \cite{wilson2020icml, wilson2021jmlr}: 
\begin{equation}
	f^{(m)}_n(\cdot) = f_0^{(m)}(\cdot) + \sum_{t=1}^n \kappa_t \Sigma_0^{(m)}(\cdot, \mathbf{x}_t),
\end{equation}
where $\kappa_t = (\Sigma_0^{(m)}(X_n, X_n) + \text{diag}(\sigma^{(m)}(X_n)))^{-1}(Y_n - f_0^{(m)}(X_n))$. For a more comprehensive overview of sampling from a Gaussian process refer to \cite{wilson2021jmlr}.
\section{Related work} 
\label{app:related_work}

\subsection{Conditional entropy estimates}
In this work we devised a number of conditional entropy estimates for the JES acquisition function. These estimates can also be used by the MES acquisition function. In this brief section, we elaborate on the similarities of our estimates with the existing work relating to the MES acquisition function. 
\\ \\
Firstly, the noiseless entropy estimate \eqref{eqn:h_0} was derived recently in \cite{suzuki2020icml} to extend the noiseless MES acquisition function to the multi-objective setting. This extension was called the Pareto frontier entropy search (PFES) and it is equivalent to the acquisition function we call MES-0. An earlier attempt to extend the MES to noiseless setting resulted in the MESMO \cite{belakaria2019anips}  acquisition function. The MESMO acquisition function is equivalent to the PFES acquisition function if we crudely approximates the dominated space with a single box $\mathbb{D}_{\preceq}(\mathbb{Y}^*) \approx B = (-\infty,  \max_{\mathbf{y} \in \mathbb{Y}^*} y^{(1)}] \times \cdots \times (-\infty,  \max_{\mathbf{y} \in \mathbb{Y}^*} y^{(M)}]$. Empirically, this approximation leads to a deterioration in performance \cite{suzuki2020icml}.
\\ \\
The closest work to the lower bound entropy estimate is the GIBBON \cite{moss2021jmlr} acquisition function. The GIBBON acquisition function was derived as a lower bound to the multi-fidelity MES acquisition function for a single-objective problem. The MES-LB and MES-LB2 can be interpreted as the multi-objective extensions of GIBBON for the single fidelity setting. 
\\ \\
The Monte Carlo entropy estimate has been used before for the multi-fidelity MES acquisition in the single-objective setting \cite{takeno2020icml, moss2021mlkdd}. The MES-MC estimate can be interpreted as the multi-objective extension of these approaches for the single fidelity setting.

\subsection{BAX}
The JES acquisition function bares some similarity with a special case of the Bayesian Algorithm Execution (BAX) algorithm proposed in \cite{neiswanger2021icml}. Specifically, if we set $\mathcal{O}_{\mathcal{A}}(f) = (\mathbb{X}^*, \mathbb{Y}^*)$ in \cite{neiswanger2021icml}, then the corresponding BAX acquisition function would reduce to
\begin{equation}
	\begin{split}
		\alpha^{\text{BAX}}(\mathbf{x}| D_n) 
		&= H[p(\mathbf{y}| \mathbf{x}, D_n)]
		- \mathbb{E}_{p((\mathbb{X}^*, \mathbb{Y}^*)| D_n)}[H[p(\mathbf{y}| \mathbf{x}, D_n \cup (\mathbb{X}^*, \mathbb{Y}^*))]].
	\end{split}
	\label{eqn:bax}
\end{equation}
In the JES acquisition function, we condition on the augmented data set $D_{n*} = D_n \cup (\mathbb{X}^*, \mathbb{Y}^*)$ and the optimality condition $f(\mathbb{X}) \preceq \mathbb{Y}^*$ for the density arising in the conditional entropy term. Whereas in BAX, we only condition on the augmented data set. The BAX acquisition function is a lower bound to JES because conditioning never increases the entropy (Theorem 2.6.5 of \cite{cover2006}).
\section{Single-objective setting}
\label{app:single_objective}
In this section, we discuss the main differences between the single-objective and multi-objective information-theoretic acquisition functions. At a high level, the only difference between the two settings is whether we use the total ordering over real numbers or the Pareto partial ordering over vectors. Note that in the single-objective setting, $M=1$, the Pareto partial ordering coincides with the standard total ordering order real numbers, that is to say the definition of the binary relations $\preceq$ and $\prec$ coincides with the standard inequality signs $\leq$ and $<$, respectively. As a result, it is generally possible for multi-objective acquisition functions to be used in the single-objective setting, which is definitely the case for the estimates of our JES acquisition function. We will now comment individually on each information-theoretic acquisition functions on the more subtle differences between the single and multi-objective algorithms.

\paragraph{PES.} There are two main difference between the single-objective and multi-objective PES algorithm. Firstly, in the single-objective setting, we sample a single maximizer, $\mathbf{x}^* \sim p(\mathbf{x}^*|D_n)$, based on the standard total ordering, whereas in the multi-objective setting we sample a discrete set of maximizers, $\mathbb{X}^* \sim p(\mathbb{X}^*|D_n)$, based on the Pareto partial ordering. Secondly, the equations governing the expectation propagation updates can be different depending on how the target density is factorised and the modelling assumptions that are made. Therefore, setting $M=1$ for the multi-objective PES algorithm (also known as PESMO) described in \cite{garrido-merchan2019n} might not exactly recover the same result for a different single-objective implementation of the PES algorithm. In our code we follow the equations proposed in \cite{garrido-merchan2019n}, which accounts for single, multi-objective, batch and/or constrained setting.

\paragraph{MES.} There are two main difference between the single-objective and multi-objective MES algorithm. Firstly, in the single-objective setting, we sample a single maximum, $y^* \sim p(y^*|D_n)$, based on the standard total ordering, whereas in the multi-objective setting we sample a discrete set of maximums, $\mathbb{Y}^* \sim p(\mathbb{Y}^*|D_n)$, based on the Pareto partial ordering. Secondly, in the single-objective setting the box decomposition is readily available without any effort, $\mathbb{D}_{\preceq}(\{y^*\}) = (-\infty, y^*]$, whereas in the multi-objective setting we have to compute it. The MESMO algorithm \cite{belakaria2019anips} and the PFES algorithm \cite{suzuki2020icml} both reduce to the single-objective MES algorithm \cite{wang2017icml} when we set $M=1$. As mentioned before in \cref{app:related_work}, the MESMO algorithm is a special case of the PFES algorithm when using a crude approximation to the box decomposition of $\mathbb{D}_{\preceq}(\mathbb{Y}^*)$, which turns out to be exact when there is only one objective.

\paragraph{JES.} There are two main difference between the single-objective and multi-objective JES algorithm. Firstly, in the single-objective setting, we sample a single optimal point, $(\mathbf{x}^*, y^*) \sim p((\mathbf{x}^*, y^*)|D_n)$, based on the standard total ordering, whereas in the multi-objective setting we sample a discrete set of optimal points, $(\mathbb{X}^*, \mathbb{Y}^*) \sim p((\mathbb{X}^*, \mathbb{Y}^*)|D_n)$, based on the Pareto partial ordering. Secondly, in the single-objective setting the box decomposition is readily available without any effort, $\mathbb{D}_{\preceq}(\{y^*\}) = (-\infty, y^*]$, whereas in the multi-objective setting we have to compute it. As mentioned before in \cref{sec:approx_jes}, if we do not perform the conditioning step in \cref{alg:jes}, we obtain the MES algorithm.

\section{Proof of results}
\label{app:proofs}
Before we begin the proofs, we will restate some important notation. Let $p(\mathbf{y}| \mathbf{x}, D_{n*}) = \mathcal{N}(\mathbf{y}; \boldsymbol{\mu}_{n*}(\mathbf{x}), \boldsymbol{\Sigma}_{n*}(\mathbf{x}, \mathbf{x}))$ denote the probability density at a point $\mathbf{x} \in \mathbb{X}$ conditional on the data set $D_{n*} = D_n \cup (\mathbb{X}^*, \mathbb{Y}^*)$. Then we denote the $m$-th standardized function by
\begin{equation}
	\gamma_m(z) = \frac{z- \mu_{n*}^{(m)}(\mathbf{x})}{\sqrt{\Sigma_{n*}^{(m)}(\mathbf{x}, \mathbf{x})}}
	\label{eqn:gamma}
\end{equation}
for $m=1,\dots,M$. For the discrete set of points $\mathbb{Y}^* \subset \mathbb{R}^M$, we decompose the dominated region into $J$ boxes: 
\begin{equation}
	\mathbb{D}_{\preceq}(\mathbb{Y}^*) = \bigcup_{j=1}^J B_j = \bigcup_{j=1}^J \prod_{m=1}^M (l^{(m)}_j, u^{(m)}_j],
\end{equation}
where $\mathbf{l}_j = (l^{(1)}_j, \dots, l^{(M)}_j)$ are lower bounds and $\mathbf{u}_j = (u^{(1)}_1, \dots, u^{(M)}_J)$ are the upper bounds for boxes $j=1,\dots,J$. Using the box decomposition, we define
\begin{equation}
	W_{j, m} = \Phi(\gamma_m(u_j^{(m)})) - \Phi(\gamma_m(l_j^{(m)}))
	\label{eqn:wjm}
\end{equation}
for boxes $j=1,\dots,J$ and objectives $m=1,\dots,M$, where $\Phi$ is the cumulative distribution function of a standard normal distribution. The first derivative of $W_{j, m}$ (with respect to $\gamma_m$) is denoted by
\begin{equation}
	G_{j, m} = \phi(\gamma_m(u_j^{(m)})) - \phi(\gamma_m(l_j^{(m)}))
	\label{eqn:gjm}
\end{equation}
and the negative of the second derivative by
\begin{equation}
	V_{j, m} = \gamma_m(u_j^{(m)}) \phi(\gamma_m(u_j^{(m)})) - \gamma_m(l_j^{(m)})\phi(\gamma_m(l_j^{(m)})),
	\label{eqn:vjm}
\end{equation}
for boxes $j=1,\dots,J$ and objectives $m=1,\dots,M$, where $\phi$ is the probability density function of a standard normal distribution.
\subsection{Proof of Proposition \ref{prop:upper_bound}} 
\begin{repproposition}{prop:upper_bound}
	The JES is an upper bound to any convex combination of the PES and MES acquisition functions: \normalfont{ $\alpha^{\text{JES}}(\mathbf{x}| D_n) \geq 
		\beta \alpha^{\text{PES}}(\mathbf{x}| D_n) + (1- \beta) \alpha^{\text{MES}}(\mathbf{x}| D_n)$, for any $\beta \in [0, 1]$}.
\end{repproposition}

\paragraph{Proof.} The upper bound property follows from the standard result that conditioning on more variables will never increase the entropy (Theorem 2.6.5 of \cite{cover2006}): $H(A|B) \leq H(A)$ for random variables $A$ and $B$. Using this result, $H[p(\mathbf{y}| \mathbf{x}, D_n, (\mathbb{X}^*, \mathbb{Y}^*))] \leq \max(H[p(\mathbf{y}| \mathbf{x}, D_n, \mathbb{X}^*)], H[p(\mathbf{y}| \mathbf{x}, D_n, \mathbb{Y}^*)])$. Plugging this inequality into \eqref{eqn:jes}, we obtain the $\alpha^{\text{JES}}(\mathbf{x}| D_n) \geq \max(\alpha^{\text{PES}}(\mathbf{x}| D_n), \alpha^{\text{MES}}(\mathbf{x}| D_n))$, which implies the result.
\\
\null \hfill $\blacksquare$
\subsection{Proof of Lemma \ref{lemma:cumul}}
\label{lemma:cumul:proof}
\begin{replemma}{lemma:cumul}
	Let $\mathbb{Y}^*\subset \mathbb{R}^M$ be a finite set and $\mathbf{z} \sim N(\mathbf{a}, \text{diag}(\mathbf{b}))$ be an $M$-dimensional multivariate normal with mean $\mathbf{a} \in \mathbb{R}^M$ and variances $\mathbf{b} \in \mathbb{R}^M_{\geq 0}$. Let $\mathbb{D}_{\preceq}(\mathbb{Y}^*) = \bigcup_{j=1}^J B_j = \bigcup_{j=1}^J \prod_{m=1}^M (l^{(m)}_j, u^{(m)}_j]$ be the box decomposition of the dominated space, then
	\begin{align}
		p(\mathbf{z} \preceq \mathbb{Y}^*)
		&= \sum_{j=1}^J \prod_{m=1}^M
		\left[
		\Phi
		\left(
		\frac{u_j^{(m)} - a^{(m)}}{\sqrt{b^{(m)}}} 
		\right)
		-
		\Phi
		\left(
		\frac{l_j^{(m)} - a^{(m)}}{\sqrt{b^{(m)}}} 
		\right)
		\right].
	\end{align}
\end{replemma}

\paragraph{Proof.} Under the assumptions of \cref{lemma:cumul}, the following series of equations holds:
\begin{align*}
	\begin{split}
		p(\mathbf{z} \preceq \mathbb{Y}^*)
		&= \int_{\mathbb{D}_{\preceq}(\mathbb{Y}^*)} p(\mathbf{z}) d\mathbf{z}
		= \sum_{j=1}^J \int_{B_j} 
		\left(
		\prod_{m=1}^M p(z^{(m)})
		\right) 
		d\mathbf{z}
		\overset{(1)}{=} \sum_{j=1}^J \prod_{m=1}^M
		\int_{l_j^{(m)}}^{u_j^{(m)}}
		p(z^{(m)})
		dz^{(m)}
		\\
		&\overset{(2)}{=} \sum_{j=1}^J \prod_{m=1}^M
		\left[
		\Phi\left(\frac{u_j^{(m)} - a^{(m)}}{\sqrt{b^{(m)}}} \right)
		-
		\Phi\left(\frac{l_j^{(m)} - a^{(m)}}{\sqrt{b^{(m)}}} \right)
		\right].
	\end{split}
\end{align*}
(1) This step follows from the box decomposition of $\mathbb{D}_{\preceq}(\mathbb{Y}^*)$ and the independence between the components of $\mathbf{z}$. (2) This step follows from the definition of the CDF.
\\
\null \hfill $\blacksquare$
\subsection{Proof of Proposition \ref{prop:skew_normal_moments}}
\label{prop:skew_normal_moments:proof}

\begin{repproposition}{prop:skew_normal_moments}
	Under the modelling set-up outlined in \cref{sec:prelim}, for an input $\mathbf{x} \in \mathbb{X}$ the first and second central moment of $p(\mathbf{y}| \mathbf{x}, D_{n*}, f(\mathbf{x}) \preceq \mathbb{Y}^*)$ are
	\begin{align*}
		\mathbb{E}[y^{(m)}| \mathbf{x}, D_{n*}, f(\mathbf{x}) \preceq \mathbb{Y}^*] = \mu_{n*}^{(m)}(\mathbf{x}) 
		-  \frac{\sqrt{\Sigma_{n*}^{(m)}(\mathbf{x}, \mathbf{x})}}{W} \sum_{j=1}^J W_j \frac{G_{j, m}}{W_{j, m}}
	\end{align*}
	and
	\begin{align*}
		&\text{\normalfont Cov}\left(
		y^{(m)}, y^{(m')}
		\Big \vert \mathbf{x}, D_{n*}, f(\mathbf{x}) \preceq \mathbb{Y}^*
		\right)
		\\
		&=
		\begin{cases}
			\frac{\sqrt{\Sigma_{n*}^{(m)}(\mathbf{x}, \mathbf{x})} \sqrt{\Sigma_{n*}^{(m')}(\mathbf{x}, \mathbf{x})}}{W} 
			\sum_{j=1}^J  W_j
			\frac{G_{j, m}}{W_{j, m}}
			\left(
			\frac{G_{j, m'}}{W_{j, m'}}
			-
			\frac{1}{W} \sum_{j'=1}^J W_{j'} \frac{G_{j', m'}}{W_{j', m'}}  
			\right),
			& 
			m \neq m'; 
			\\
			\Sigma_{n*}^{(m)}(\mathbf{x}, \mathbf{x}) + \sigma^{(m)}(\mathbf{x}) - 
			\frac{ \Sigma_{n*}^{(m)}(\mathbf{x}, \mathbf{x})}{W} 
			\left(
			\sum_{j=1}^J W_j \frac{V_{j, m}}{W_{j, m}}
			+ \frac{1}{W} \left(
			\sum_{j=1}^J W_j
			\frac{G_{j, m}}{W_{j, m}}
			\right)^2
			\right),
			&
			m = m'.
		\end{cases}
	\end{align*}
\end{repproposition}

\paragraph{Proof.} In the noisy setting, the density of interest is
\begin{align*}
	p(\mathbf{y}| \mathbf{x}, D_{n*}, f(\mathbf{x}) \preceq \mathbb{Y}^*)
	=
	\frac{p(f(\mathbf{x}) \preceq \mathbb{Y}^*| \mathbf{x}, D_{n+})}
	{p(f(\mathbf{x}) \preceq \mathbb{Y}^*| \mathbf{x}, D_{n*})} 
	p(\mathbf{y}| \mathbf{x}, D_{n*}).
\end{align*}
From \cref{lemma:cumul},
\begin{align}
	\begin{split}
			p(f(\mathbf{x}) \preceq \mathbb{Y}^*| \mathbf{x}, D_{n*})
			&= \sum_{j=1}^J \prod_{m=1}^M W_{j, m}
			= \sum_{j=1}^J W_j
			= W.
		\end{split}
	\label{eqn:W}
\end{align}
To obtain a tractable expression for the cumulative distribution $p(f(\mathbf{x}) \preceq \mathbb{Y}^*| \mathbf{x}, D_{n+})$, we first compute the conditional distribution $p(\mathbf{y}| \mathbf{x}, D_{n+})$. 
By standard Gaussian conditioning $p(\mathbf{y}| \mathbf{x}, D_{n+}) = \mathcal{N}(\mathbf{y}; \boldsymbol{\mu}_{n+}(\mathbf{x}), \Sigma_{n+}^{(m)} (\mathbf{x}, \mathbf{x}))$ where
\begin{align*}
	\mu_{n+}^{(m)}(\mathbf{x})
	&= \mu_{n*}^{(m)}(\mathbf{x}) + 
	\Sigma_{n*}^{(m)} (\mathbf{x}, \mathbf{x})
	(\Sigma_{n*}^{(m)} (\mathbf{x}, \mathbf{x}) + \sigma^{(m)}(\mathbf{x}))^{-1}
	(y^{(m)} - \mu_{n*}^{(m)}(\mathbf{x}))
	\\
	&= \mu_{n*}^{(m)}(\mathbf{x}) + \rho_m (\mathbf{x})
	\frac{\sqrt{\Sigma_{n*}^{(m)} (\mathbf{x}, \mathbf{x})}}
	{\sqrt{\Sigma_{n*}^{(m)} (\mathbf{x}, \mathbf{x}) + \sigma^{(m)}(\mathbf{x})}}
	(y^{(m)} - \mu_{n*}^{(m)}(\mathbf{x}))
\end{align*}
and
\begin{align*}
	\Sigma_{n+}^{(m)} (\mathbf{x}, \mathbf{x}) 
	&= \Sigma_{n*}^{(m)}(\mathbf{x}, \mathbf{x})  - \Sigma_{n*}^{(m)} (\mathbf{x}, \mathbf{x}) 
	\left({\Sigma_{n*}^{(m)} (\mathbf{x}, \mathbf{x}) + \sigma^{(m)}(\mathbf{x})} \right)^{-1} 
	\Sigma_{n*}^{(m)}(\mathbf{x}, \mathbf{x})  (\mathbf{x}, \mathbf{x})
	\\
	&= \Sigma_{n*}^{(m)} (\mathbf{x}, \mathbf{x}) (1 - \rho_m^2(\mathbf{x}))
\end{align*}
with $\rho_m := \rho_m(\mathbf{x}) = \sqrt{\Sigma_{n*}^{(m)} (\mathbf{x}, \mathbf{x})} / \sqrt{\Sigma_{n*}^{(m)} (\mathbf{x}, \mathbf{x}) + \sigma^{(m)}(\mathbf{x})}$ denoting correlation between the observation $y^{(m)}$ and the objective value $f^{(m)}(\mathbf{x})$, for objectives $m=1,\dots,M$. Using the box decomposition, the posterior CDF is equal to
\begin{align}
	&p(f(\mathbf{x}) \preceq \mathbb{Y}^*| \mathbf{x}, D_{n+})
	= \sum_{j=1}^J \prod_{m=1}^M 
	\left[
	\Phi\left(
	\frac{u_j^{(m)} - \mu_{n+}^{(m)}(\mathbf{x})}{\sqrt{\Sigma_{n+}^{(m)}(\mathbf{x}, \mathbf{x})}} 
	\right)
	-
	\Phi\left(
	\frac{l_j^{(m)} - \mu_{n+}^{(m)}(\mathbf{x})}{\sqrt{\Sigma_{n+}^{(m)}(\mathbf{x}, \mathbf{x})}} 
	\right)
	\right].
	\label{eqn:W+}
\end{align}
To simplify the notation we perform a change of variable:
\begin{align*}
	\gamma_m^{+}(z): = \frac{z - \mu_{n+}^{(m)}(\mathbf{x})}{\sqrt{\Sigma_{n+}^{(m)}(\mathbf{x}, \mathbf{x})}}
	&= \frac{z^{(m)} - \mu_{n*}^{(m)}(\mathbf{x}) - 
		\rho_m 
		\frac{\sqrt{\Sigma_{n*}^{(m)} (\mathbf{x}, \mathbf{x})}}
		{\sqrt{\Sigma_{n*}^{(m)} (\mathbf{x}, \mathbf{x}) + \sigma^{(m)}(\mathbf{x})}}
		(y^{(m)} - \mu_{n*}^{(m)}(\mathbf{x}))}
	{\sqrt{\Sigma_{n*}^{(m)}(\mathbf{x}, \mathbf{x})(1 - \rho_m^2)}}
	\\
	&= \frac{
		\frac{z^{(m)} - \mu_{n*}^{(m)}(\mathbf{x})}{\sqrt{\Sigma_{n*}^{(m)}(\mathbf{x}, \mathbf{x}})} - 
		\rho_m
		\frac{(y^{(m)} - \mu_{n*}^{(m)}(\mathbf{x}))}
		{\sqrt{\Sigma_{n*}^{(m)} (\mathbf{x}, \mathbf{x}) + \sigma^{(m)}(\mathbf{x})}}
	}
	{\sqrt{1 - \rho_m^2}}
	= \frac{\gamma_m(z) - \rho_m \bar{y}^{(m)}}{\sqrt{1 - \rho_m^2}}
\end{align*}
where $\bar{y}^{(m)} = (y^{(m)} - \mu_{n*}^{(m)}(\mathbf{x})) / \sqrt{\Sigma_{n*}^{(m)}(\mathbf{x}, \mathbf{x}) + \sigma^{(m)}(\mathbf{x})}$ is the standardized observation, which is distributed according to a standard normal random variable for objectives $m=1,\dots,M$. To derive the moments of $p(\mathbf{y}| \mathbf{x}, D_{n*}, f(\mathbf{x}) \preceq \mathbb{Y}^*)$, we first obtain the moment generating function of the standardized observation $\mathbf{\bar{y}}$.
\begin{align*}
	&M_{\mathbf{\bar{y}}}(\mathbf{t})
	= \frac{1}{W}
	\int_{\mathbb{R}^M} 
	e^{\mathbf{t}^T \mathbf{\bar{y}}}
	p(f(\mathbf{x}) \preceq \mathbb{Y}^*| \mathbf{x}, D_{n+})
	p(\mathbf{\bar{y}}| \mathbf{x}, D_{n*})
	d\mathbf{\bar{y}}
	\\
	&= \frac{1}{W}
	\sum_{j=1}^J \prod_{m=1}^M 
	\int_{\mathbb{R}}
	e^{t^{(m)} \bar{y}^{(m)}}
	\left[\Phi(\gamma_m^{+}(u_j^{(m)})) - \Phi(\gamma_m^{+}(l_j^{(m)})) \right]
	\phi(\bar{y}^{(m)})
	d\bar{y}^{(m)}
	\\
	&= \frac{1}{W}
	\sum_{j=1}^J \prod_{m=1}^M 
	e^{\frac{(t^{(m)})^2}{2}}
	\int_{\mathbb{R}}
	\left[\Phi(\gamma_m^{+}(u_j^{(m)})) - \Phi(\gamma_m^{+}(l_j^{(m)})) \right]
	\phi(\bar{y}^{(m)} - t^{(m)})
	d\bar{y}^{(m)}.
\end{align*}
By Lemma 2 in \cite{azzalini1985sjs}, the expectation of the normal CDF is given by $\int_{\mathbb{R}} \Phi(a z + b) \phi(z) dz = \Phi ( b / \sqrt{1 + a^2} )$ for any constants $a, b \in \mathbb{R}$. Using this result,
\begin{align*}
	&\int_{\mathbb{R}}
	\left[\Phi(\gamma_m^{+}(u_j^{(m)})) - \Phi(\gamma_m^{+}(l_j^{(m)})) \right]
	\phi(\bar{y}^{(m)} - t^{(m)})
	d\bar{y}^{(m)}
	\\
	&= \int_{\mathbb{R}}
	\left[
	\Phi\left(
	\frac{\gamma_m(u_j^{(m)}) - \rho_m \bar{y}^{(m)}}{\sqrt{1 - \rho_m^2}}
	\right) 
	- \Phi\left(
	\frac{\gamma_m(l_j^{(m)}) - \rho_m \bar{y}^{(m)}}{\sqrt{1 - \rho_m^2}}
	\right) 
	\right]
	\phi(\bar{y}^{(m)} - t^{(m)})
	d\bar{y}^{(m)}
	\\
	&= \int_{\mathbb{R}}
	\left[
	\Phi\left(
	\frac{\gamma_m(u_j^{(m)}) - \rho_m (\bar{y}^{(m)} + t^{(m)})}{\sqrt{1 - \rho_m^2}}
	\right) 
	- \Phi\left(
	\frac{\gamma_m(l_j^{(m)}) - \rho_m (\bar{y}^{(m)} + t^{(m)})}{\sqrt{1 - \rho_m^2}}
	\right) 
	\right]
	\phi(\bar{y}^{(m)})
	d\bar{y}^{(m)}
	\\
	&= \Phi(\gamma_m(u_j^{(m)}) - \rho_m t^{(m)}) 
	- \Phi(\gamma_m(l_j^{(m)}) - \rho_m t^{(m)}).
\end{align*}
The first moment $p(\bar{y}^{(m)}| \mathbf{x}, D_{n*}, f(\mathbf{x}) \preceq \mathbb{Y}^*)$ can be obtained by evaluating the first derivative of $M_{\mathbf{\bar{y}}}(\mathbf{t})$ with respect to $t^{(m)}$ at $\mathbf{t}=\mathbf{0}_M$.
\begin{align*}
	&\mathbb{E}[\bar{y}^{(m)}| \mathbf{x}, D_{n*}, f(\mathbf{x}) \preceq \mathbb{Y}^*]
	= \frac{\partial}{\partial t^{(m)}} M_{\mathbf{\bar{y}}}(\mathbf{t}) \Big \vert_{\mathbf{t}=\mathbf{0}_M}
	\\
	&= \frac{1}{W}
	\sum_{j=1}^J \prod_{m' \neq m}
	\left[\Phi(\gamma_{m'}(u_j^{(m')})) - \Phi(\gamma_{m'}(l_j^{(m')})) \right] 
	\left(- \rho_m \left(\phi(\gamma_m(u_j^{(m)})) - \phi(\gamma_m(l_j^{(m)})) \right) \right)
	\\
	&= - \frac{\rho_m}{W} \sum_{j=1}^J \frac{W_j}{W_{j, m}} G_{j, m}.
\end{align*}
Differentiating a second time, we can obtain the second moments. For $m \neq m'$,
\begin{align*}
	\mathbb{E}[\bar{y}^{(m)} \bar{y}^{(m')}| \mathbf{x}, D_{n*}, f(\mathbf{x}) \preceq \mathbb{Y}^*]
	&= \frac{\partial}{\partial t^{(m)} \partial t^{(m')}} M_{\mathbf{\bar{y}}}(\mathbf{t}) \Big \vert_{\mathbf{t}=\mathbf{0}_M}
	\\
	&
	= \frac{\rho_m \rho_{m'}}{W} \sum_{j=1}^J \frac{W_j}{W_{j, m} W_{j, m'}} G_{j, m} G_{j, m'}
\end{align*}
and for $m = m'$,
\begin{align*}
	&\mathbb{E}[(\bar{y}^{(m)})^2| \mathbf{x}, D_{n*}, f(\mathbf{x}) \preceq \mathbb{Y}^*]
	= \frac{\partial^2}{(\partial t^{(m)})^2} M_{\mathbf{\bar{y}}}(\mathbf{t}) \Big \vert_{\mathbf{t}=\mathbf{0}_M}
	\\
	&= 1 + \frac{1}{W}
	\sum_{j=1}^J \prod_{m' \neq m}
	\left[\Phi(\gamma_{m'}(u_j^{(m')})) - \Phi(\gamma_{m'}(l_j^{(m')})) \right] 
	\\
	&\quad \quad
	\times \left(
	- \rho_m^2 \left(\gamma_m(u_j^{(m)})\phi(\gamma_m(u_j^{(m)})) - \gamma_m(l_j^{(m)}) \phi(\gamma_m(l_j^{(m)})) \right)
	\right)
	\\
	&= 1 - \frac{\rho_m^2}{W}
	\sum_{j=1}^J \frac{W_j}{W_{j, m}} V_{j, m}.
\end{align*}
The moments of $y^{(m)}$ can be now derived by reversing the initial linear transformation: $y^{(m)} = \bar{y}^{(m)} \sqrt{\Sigma_{n*}^{(m)}(\mathbf{x}, \mathbf{x}) + \sigma^{(m)}(\mathbf{x})}  + \mu_{n*}^{(m)}(\mathbf{x}) $. The first moment is
\begin{equation*}
	\mathbb{E}[y^{(m)}| \mathbf{x}, D_{n*}, f(\mathbf{x}) \preceq \mathbb{Y}^*] = \mu_{n*}^{(m)}(\mathbf{x}) 
	-  \frac{\sqrt{\Sigma_{n*}^{(m)}(\mathbf{x}, \mathbf{x})}}{W} \sum_{j=1}^J W_j \frac{W_{j, m}}{G_{j, m}}.
\end{equation*}
The second moment:
\begin{align*}
	\begin{split}
		&\mathbb{E}[\bar{y}^{(m)} \bar{y}^{(m')}| \mathbf{x}, D_{n*}, f(\mathbf{x}) \preceq \mathbb{Y}^*]
		\\
		&= \mathbb{E}\left[
		\frac{y^{(m)} - \mu_{n*}^{(m)}(\mathbf{x})}{\sqrt{\Sigma_{n*}^{(m)}(\mathbf{x}, \mathbf{x}) + \sigma^{(m)}(\mathbf{x})}}
		\frac{y^{(m')} - \mu_{n*}^{(m')}(\mathbf{x})}{\sqrt{\Sigma_{n*}^{(m')}(\mathbf{x}, \mathbf{x}) + \sigma^{(m')}(\mathbf{x})}}
		\Big \vert \mathbf{x}, D_{n*}, f(\mathbf{x}) \preceq \mathbb{Y}^*
		\right]  
		\\
		&= 
		\mathbb{E}\left[
		\frac{y^{(m)}y^{(m')} - y^{(m)}\mu_{n*}^{(m')}(\mathbf{x}) - y^{(m')}\mu_{n*}^{(m)}(\mathbf{x})
			+ \mu_{n*}^{(m)}(\mathbf{x}) \mu_{n*}^{(m')}(\mathbf{x})}
		{\sqrt{\Sigma_{n*}^{(m)}(\mathbf{x}, \mathbf{x}) + \sigma^{(m)}(\mathbf{x})} \sqrt{\Sigma_{n*}^{(m')}(\mathbf{x}, \mathbf{x}) + \sigma^{(m')}(\mathbf{x})}}
		\Big \vert \mathbf{x}, D_{n*}, f(\mathbf{x}) \preceq \mathbb{Y}^*
		\right].
	\end{split}
\end{align*}
This implies
\begin{align*}
	\begin{split}
		&\mathbb{E}\left[
		y^{(m)}y^{(m')}
		\Big \vert \mathbf{x}, D_{n*}, f(\mathbf{x}) \preceq \mathbb{Y}^*
		\right] 
		\\
		&=
		\sqrt{\Sigma_{n*}^{(m)}(\mathbf{x}, \mathbf{x}) + \sigma^{(m)}(\mathbf{x})}
		\sqrt{\Sigma_{n*}^{(m')}(\mathbf{x}, \mathbf{x}) + \sigma^{(m')}(\mathbf{x})}	
		\mathbb{E}[\bar{y}^{(m)} \bar{y}^{(m')}| \mathbf{x}, D_{n*}, f(\mathbf{x}) \preceq \mathbb{Y}^*]
		\\
		&\quad + \mathbb{E}[\bar{y}^{(m)}| \mathbf{x}, D_{n*}, f(\mathbf{x}) \preceq \mathbb{Y}^*] \mu_{n*}^{(m')}(\mathbf{x})
		+ \mathbb{E}[\bar{y}^{(m')}| \mathbf{x}, D_{n*}, f(\mathbf{x}) \preceq \mathbb{Y}^*] \mu_{n*}^{(m)}(\mathbf{x}) 
		\\
		&\quad - \mu_{n*}^{(m)}(\mathbf{x}) \mu_{n*}^{(m')}(\mathbf{x}).
	\end{split}
\end{align*}
Substituting in the expressions before, we have that for $m \neq m'$,
\begin{align*}
	\begin{split}
		&\mathbb{E}\left[
		y^{(m)}y^{(m')}
		\Big \vert \mathbf{x}, D_{n*}, f(\mathbf{x}) \preceq \mathbb{Y}^*
		\right]
		= \mu_{n*}^{(m)}(\mathbf{x}) \mu_{n*}^{(m')}(\mathbf{x}) + 
		\frac{\sqrt{\Sigma_{n*}^{(m)}(\mathbf{x}, \mathbf{x})} \sqrt{\Sigma_{n*}^{(m')}(\mathbf{x}, \mathbf{x})}}{W}
		\\
		&\quad \times
		\sum_{j=1}^J  W_j \left(
		\frac{G_{j, m}}{W_{j, m}}
		\frac{G_{j, m'}}{W_{j, m'}}
		-
		\frac{\mu_{n*}^{(m')}(\mathbf{x})}{\sqrt{\Sigma_{n*}^{(m')}(\mathbf{x}, \mathbf{x})}} \frac{G_{j, m}}{W_{j, m}}
		-
		\frac{\mu_{n*}^{(m)}(\mathbf{x})}{\sqrt{\Sigma_{n*}^{(m)}(\mathbf{x}, \mathbf{x})}} \frac{G_{j, m'}}{W_{j, m'}}
		\right)
	\end{split}
\end{align*}
and for $m = m'$
\begin{align*}
	\begin{split}
		&\mathbb{E}\left[
		(y^{(m)})^2
		\Big \vert \mathbf{x}, D_{n*}, f(\mathbf{x}) \preceq \mathbb{Y}^*
		\right]
		= \mu_{n*}^{(m)}(\mathbf{x})^2 + \Sigma_{n*}^{(m)}(\mathbf{x}, \mathbf{x}) + \sigma^{(m)}(\mathbf{x})
		\\
		&\quad -
		\frac{ \Sigma_{n*}^{(m)}(\mathbf{x}, \mathbf{x})}{W} \sum_{j=1}^J W_j
		\left(
		\frac{V_{j, m}}{W_{j, m}}
		+ 2 \frac{\mu_{n*}^{(m)}(\mathbf{x})}{\sqrt{\Sigma_{n*}^{(m)}(\mathbf{x}, \mathbf{x})}} \frac{G_{j, m}}{W_{j, m}}
		\right).
	\end{split}
\end{align*}
By some additional algebraic manipulation, the covariance for $m \neq m'$ is given
\begin{align*}
	\begin{split}
		&\text{Cov}\left(
		y^{(m)}, y^{(m')}
		\Big \vert \mathbf{x}, D_{n*}, f(\mathbf{x}) \preceq \mathbb{Y}^*
		\right)
		=
		\frac{\sqrt{\Sigma_{n*}^{(m)}(\mathbf{x}, \mathbf{x})} \sqrt{\Sigma_{n*}^{(m')}(\mathbf{x}, \mathbf{x})}}{W} 
		\\
		&\quad \times
		\left(
		\sum_{j=1}^J  W_j
		\frac{G_{j, m}}{W_{j, m}}
		\frac{G_{j, m'}}{W_{j, m'}}
		-
		\frac{1}{W} \left(\sum_{j=1}^J W_j \frac{G_{j, m}}{W_{j, m}} \right)
		\left(\sum_{j'=1}^J W_{j'} \frac{G_{j', m'}}{W_{j', m'}}  \right)
		\right)
	\end{split}
\end{align*}
and the variance is
\begin{align*}
	\begin{split}
		&\mathbb{V}\text{ar}\left(
		y^{(m)}
		\Big \vert \mathbf{x}, D_{n*}, f(\mathbf{x}) \preceq \mathbb{Y}^*
		\right)
		\\
		&= \Sigma_{n*}^{(m)}(\mathbf{x}, \mathbf{x}) + \sigma^{(m)}(\mathbf{x}) - 
		\frac{ \Sigma_{n*}^{(m)}(\mathbf{x}, \mathbf{x})}{W} 
		\left(
		\sum_{j=1}^J W_j \frac{V_{j, m}}{W_{j, m}}
		+ \frac{1}{W} \left(
		\sum_{j=1}^J W_j
		\frac{G_{j, m}}{W_{j, m}}
		\right)^2
		\right).
	\end{split}
\end{align*}
\null \hfill $\blacksquare$

\subsection{Proof of Proposition \ref{prop:invariance}}
\label{prop:invariance:proof}
\begin{repproposition}{prop:invariance}
	The information-theoretic acquisition functions {\normalfont$\alpha^{\text{PES}}$}, {\normalfont$\alpha^{\text{MES}}$} and {\normalfont$\alpha^{\text{JES}}$} are invariant to reparameterization of the objective space that are consistent with the Pareto ordering relations. For example, {\normalfont $\alpha^{\text{JES}}(\mathbf{x}| D_n) = \text{MI}(\mathbf{y}; (\mathbb{X}^*, \mathbb{Y}^*)| D_n) = \text{MI}(g(\mathbf{y}); (\mathbb{X}^*, g(\mathbb{Y}^*))| D_n)$}, where the $g_m:\mathbb{R} \rightarrow \mathbb{R}$ is a strictly monotonically increasing function acting only on the $m$-th objective.
\end{repproposition}

\paragraph{Proof.} We will prove the general statement that the mutual information is invariant under smooth bijective transformations. Consider two random variables $X$ and $Y$ then following equations hold:
\begin{align*}
	I(X; Y) 
	= \mathbb{E}_{p(X, Y)}\left[\log \frac{p(X, Y)}{p(X) p(Y)} \right]
	= \mathbb{E}_{p(X', Y')}\left[\log \frac{p(X', Y') |J_x| |J_y| }{p(X') |J_x| p(Y') |J_y|} \right]
	= I(X'; Y'),
\end{align*}
where $X' = g_x(X)$ and $Y' = g_y(Y)$ represents the transformed variables under some suitably defined smooth invertible functions $g_x$ and $g_y$. The expressions $J_x$ and $J_y$ correspond to the Jacobian of the transformation, which are non-zero because the transformation are assumed to be invertible. The result follows by restricting the class of bijective functions to ones where the Pareto set remains unchanged. For example, the class of monotonic increasing functions in each objective ensures this property holds: $g:\mathbb{R}^M \rightarrow \mathbb{R}^M$ such that $g(\mathbf{y}) = (g_1(y^{(1)}), \dots g_M(y^{(M)}))$ with $g_m$ being monotonically increasing.
\\
\null \hfill $\blacksquare$
\section{Noiseless entropy estimate}
\label{app:zero_variance}
In this section, we present the conditional entropy estimate for the zero observation variance setting and an ad hoc extension for noisy setting. Firstly, if we assume the observation variance is zero, the conditional distribution of interest is a truncated multivariate normal:
\begin{align}
	p(\mathbf{y}| \mathbf{x}, D_{n*}, f(\mathbf{x}) \preceq \mathbb{Y}^*)
	= \frac{p(\mathbf{y}| \mathbf{x}, D_{n*})}
	{p(f(\mathbf{x}) \preceq \mathbb{Y}^*| \mathbf{x}, D_{n*})} 
	\mathbb{I}[\mathbf{y} \preceq \mathbb{Y}^*],
	\label{eqn:noiseless_density}
\end{align}
which is known to have the following analytical equation for the entropy:
\begin{proposition}
	(Theorem 3.1. in \cite{suzuki2020icml}) Under the modelling set-up outlined in \cref{sec:prelim}, if $\mathbf{x} \in \mathbb{X}$ is an input with zero observation variance, $\boldsymbol{\sigma}(\mathbf{x}) = \mathbf{0}_M$, then the entropy of the truncated multivariate normal distribution $p(\mathbf{y}| \mathbf{x}, D_{n*}, f(\mathbf{x}) \preceq \mathbb{Y}^*)$ is given by
	\begin{align*}
			\begin{split}
					&H[p(\mathbf{y}| \mathbf{x}, D_{n*}, f(\mathbf{x}) \preceq \mathbb{Y}^*)]
					\\
					&= \frac{M}{2} \log(2\pi e) + \frac{1}{2} \sum_{m=1}^M \log(\Sigma_{n*}^{(m)}(\mathbf{x}, \mathbf{x}))
					+ \log{W} - \frac{1}{2W} \sum_{j=1}^J W_j\sum_{m=1}^M \frac{V_{j, m}}{W_{j, m}}.
				\end{split}
		\end{align*}
	\label{prop:noiseless_entropy}
\end{proposition}
\paragraph{Proof.} See \cite{suzuki2020icml} for the proof of this result.
\\
\null \hfill $\blacksquare$
\\
Assuming $\boldsymbol{\sigma}(\mathbf{x}) = \mathbf{0}_M$, we have
\begin{align*}
	\alpha^{\text{JES-}0}(\mathbf{x}| D_n)
	&= 
	H[p(\mathbf{y}| \mathbf{x}, D_n)]
	- \mathbb{E}_{p((\mathbb{X}^*, \mathbb{Y}^*)| D_n)}[H[p(\mathbf{y}| \mathbf{x}, D_{n*}, f(\mathbf{x}) \preceq \mathbb{Y}^*)]]
	\\
	\begin{split}
		&=
		- \mathbb{E}_{p((\mathbb{X}^*, \mathbb{Y}^*)| D_n)}[\log(W)]
		\\
		&+
		\frac{1}{2} \sum_{m=1}^M (\log (\Sigma_n^{(m)}(\mathbf{x}, \mathbf{x})) - \mathbb{E}_{p((\mathbb{X}^*, \mathbb{Y}^*)| D_n)}[\log (\Sigma_{n*}^{(m)}(\mathbf{x}, \mathbf{x}))])
		\\
		&\quad
		+ \mathbb{E}_{p((\mathbb{X}^*, \mathbb{Y}^*)| D_n)}\left[\frac{1}{2W}  \sum_{j=1}^J W_j\sum_{m=1}^M \frac{V_{j, m}}{W_{j, m}} \right].
	\end{split}
\end{align*}
The first term is equal to expectation of the negative log-probability of the noiseless observation lying below Pareto front:
\begin{equation*}
	- \mathbb{E}_{p((\mathbb{X}^*, \mathbb{Y}^*)| D_n)}[\log(W)]
	= - \mathbb{E}_{p((\mathbb{X}^*, \mathbb{Y}^*)| D_n)}[\log(p(\mathbf{y} \preceq \mathbb{Y}^*| \mathbf{x}, D_{n*}))].
\end{equation*}
This term accounts for the exploitation because it puts more emphasis on the points that are likely to lie above the Pareto front. The remaining terms
account for the exploration by placing more emphasis on reducing the uncertainty at input location. To still take advantage of the result in \cref{prop:noiseless_entropy} for the noisy observation setting, we propose an ad hoc extension that adjusts the exploration term to include the effects of the observation noise. Specifically, we propose a modification which replaces the difference in log variances in the exploration term,
\begin{equation*}
	\log (\Sigma_n^{(m)}(\mathbf{x}, \mathbf{x})) - \mathbb{E}_{p((\mathbb{X}^*, \mathbb{Y}^*)| D_n)}[\log (\Sigma_{n*}^{(m)}(\mathbf{x}, \mathbf{x}))]
\end{equation*}
with the differences in log variances plus observation noise,
\begin{equation*}
	\log (\Sigma_n^{(m)}(\mathbf{x}, \mathbf{x}) + \sigma^{(m)}(\mathbf{x})) - \mathbb{E}_{p((\mathbb{X}^*, \mathbb{Y}^*)| D_n)}[\log (\Sigma_{n*}^{(m)}(\mathbf{x}, \mathbf{x}) + \sigma^{(m)}(\mathbf{x})))],
\end{equation*}
where $\sigma^{(m)}(\mathbf{x}))(\mathbf{x})$ is the observation variance for objectives $m=1,\dots,M$ at $\mathbf{x} \in \mathbb{X}$. Using this adjustment, we define the resulting conditional entropy estimate by
\begin{align}
	\begin{split}
			&h^{\text{JES-}0}((\mathbb{X}^*, \mathbb{Y}^*); \mathbf{x}, D_n)	
			\\		
			&= \frac{M}{2}\log(2\pi e) + \frac{1}{2} \sum_{m=1}^M
			\log(\Sigma_{n*}^{(m)}(\mathbf{x}, \mathbf{x}) + \sigma^{(m)}(\mathbf{x})) 
			+ \log{W} 
			- \frac{1}{2W} \sum_{j=1}^J W_j\sum_{m=1}^M \frac{V_{j, m}}{W_{j, m}}.
		\end{split}
	\label{eqn:h_0}
\end{align}
Empirically, we observe that the performance of this conditional entropy estimate is in-line with the other conditional entropy approximations. 
\section{Monte Carlo entropy estimate}
\label{app:monte_carlo}
In this section, we consider estimating the entropy of $p(\mathbf{y}| \mathbf{x}, D_{n*}, f(\mathbf{x}) \preceq \mathbb{Y}^*)$ via Monte Carlo. As a reminder, the entropy of interest can be written as an expectation over $p(\mathbf{y}| \mathbf{x}, D_{n*})$:
\begin{align*}
	\begin{split}
		&H[p(\mathbf{y}| \mathbf{x}, D_{n*}, f(\mathbf{x}) \preceq \mathbb{Y}^*)]
		\\
		&=
		- \int_{\mathbb{R}^M} p(\mathbf{y}| \mathbf{x}, D_{n*}, f(\mathbf{x}) \preceq \mathbb{Y}^*)
		\log
		\left(
		p(\mathbf{y}| \mathbf{x}, D_{n*}, f(\mathbf{x}) \preceq \mathbb{Y}^*)
		\right)
		d\mathbf{y}
		\\
		&=
		- \int_{\mathbb{R}^M} \frac{p(f(\mathbf{x}) \preceq \mathbb{Y}^*| \mathbf{x}, D_{n+})} 
		{p(f(\mathbf{x}) \preceq \mathbb{Y}^*| \mathbf{x}, D_{n*})} p(\mathbf{y}| \mathbf{x}, D_{n*}) 
		\log
		\left(
		\frac{p(f(\mathbf{x}) \preceq \mathbb{Y}^*| \mathbf{x}, D_{n+})} 
		{p(f(\mathbf{x}) \preceq \mathbb{Y}^*| \mathbf{x}, D_{n*})} p(\mathbf{y}| \mathbf{x}, D_{n*}) 
		\right)
		d\mathbf{y}
		\\
		&=
		- \mathbb{E}_{p(\mathbf{y}| \mathbf{x}, D_{n*})}
		\left[
		\frac{p(f(\mathbf{x}) \preceq \mathbb{Y}^*| \mathbf{x}, D_{n+})} 
		{p(f(\mathbf{x}) \preceq \mathbb{Y}^*| \mathbf{x}, D_{n*})}
		\log 
		\left(
		\frac{p(f(\mathbf{x}) \preceq \mathbb{Y}^*| \mathbf{x}, D_{n+})} 
		{p(f(\mathbf{x}) \preceq \mathbb{Y}^*| \mathbf{x}, D_{n*})} 
		p(\mathbf{y}| \mathbf{x}, D_{n*}) 
		\right)
		\right].
	\end{split}
\end{align*}
The CDF in the denominator is $p(f(\mathbf{x}) \preceq \mathbb{Y}^*| \mathbf{x}, D_{n*}) = W$ from \eqref{eqn:W}, whilst the CDF in the numerator is
\begin{align*}
	\begin{split}
		p(f(\mathbf{x}) \preceq \mathbb{Y}^*| \mathbf{x}, D_{n+})
		&= \sum_{j=1}^J \prod_{m=1}^M 
		\left[
		\Phi\left(\gamma_m^+(u_j^{(m)})\right)
		-
		\Phi\left(\gamma_m^+(l_j^{(m)})\right)
		\right]
		\\
		&= \sum_{j=1}^J \prod_{m=1}^M W^+_{j, m}(\mathbf{y})
		= \sum_{j=1}^J W^+_{j}(\mathbf{y}) 
		= W^+(\mathbf{y})
	\end{split}
\end{align*}
from \eqref{eqn:W+}. By sampling $\mathbf{y}_i \sim p(\mathbf{y}| \mathbf{x}, D_{n*})$ for $i=1,\dots,I$, we can approximate the entropy with the following Monte Carlo average:
\begin{align*}
	\begin{split}
		H[p(\mathbf{y}| \mathbf{x}, D_{n*}, f(\mathbf{x}) \preceq \mathbb{Y}^*)]
		&=
		- \frac{1}{W}
		\mathbb{E}_{p(\mathbf{y}| \mathbf{x}, D_{n*})}
		\left[
		W^+(\mathbf{y})
		\log( W^+(\mathbf{y}) p(\mathbf{y}| \mathbf{x}, D_{n*}))
		\right]
		+ \log(W)
		\\
		&\approx h^{\text{JES-MC}}((\mathbb{X}^*, \mathbb{Y}^*); \mathbf{x}, D_n),
	\end{split}
\end{align*}
where the Monte Carlo entropy estimate is given by
\begin{align}
	h^{\text{JES-MC}}((\mathbb{X}^*, \mathbb{Y}^*); \mathbf{x}, D_n)
	&= - \frac{1}{W I} \sum_{i=1}^I W^+(\mathbf{y}_i) \log( W^+(\mathbf{y}_i) p(\mathbf{y}_i | \mathbf{x}, D_{n*}))
	+ \log(W).
	\label{eqn:h_mc}
\end{align}
Instead of generating new samples for each call of the acquisition function, we follow the general wisdom in BO \cite{wilson2018anips} and apply the reparameterization trick on the sampling distribution: $\mathbf{y}_i = \boldsymbol{\mu}_{n*}(\mathbf{x}) + \mathbf{C}_{n*}(\mathbf{x}) \mathbf{z}_i$, where $\mathbf{C}_{n*}(\mathbf{x}) \in \mathbb{R}^{M \times M}$ is the Cholesky factor of $\boldsymbol{\Sigma}_{n*}(\mathbf{x}, \mathbf{x})$ and $\mathbf{z}_i \sim \mathcal{N}(0, I_M)$ are the base samples that only need to be initialized once. The variance of this estimate could potentially be reduced by including control variates. For example,
\begin{align*}
	\begin{split}
		h^{\text{JES-MC-CV}}&((\mathbb{X}^*, \mathbb{Y}^*); \mathbf{x}, D_n)
		= h^{\text{JES-MC}}((\mathbb{X}^*, \mathbb{Y}^*); \mathbf{x}, D_n)
		\\
		&\quad + \beta_1 \left(\frac{1}{I} \sum_{i=1}^I W^+(\mathbf{y}_i) - \mathbb{E}_{p(\mathbf{y}| \mathbf{x}, D_{n*})}[W^+(\mathbf{y})] \right)
		\\
		&\quad +
		\beta_2 \left( - \frac{1}{I} \sum_{i=1}^I \log(p(\mathbf{y}_i| \mathbf{x}, D_{n*})) + \mathbb{E}_{p(\mathbf{y}| \mathbf{x}, D_{n*})}[\log(p(\mathbf{y}| \mathbf{x}, D_{n*})]
		\right),
	\end{split}
\end{align*}
where the expectations are known,
\begin{gather}
	\mathbb{E}_{p(\mathbf{y}| \mathbf{x}, D_{n*})}[W^+(\mathbf{y})] = W,
	\\
	- \mathbb{E}_{p(\mathbf{y}| \mathbf{x}, D_{n*})}[\log(p(\mathbf{y}| \mathbf{x}, D_{n*})] 
	= \frac{M}{2} \log(2 \pi e) + \frac{1}{2} \sum_{m=1}^M \log(\Sigma_{n*}^{(m)}(\mathbf{x}, \mathbf{x}) + \sigma^{(m)}(\mathbf{x}))
\end{gather}
and $\beta_1, \beta_2 \in \mathbb{R}$ are the regression coefficients. We did not assess the effect adding control variates to our Monte Carlo estimate because the standard quasi-Monte Carlo scheme with the reparameterization trick worked reasonably well out of the box. One word of caution \cite{hickernell2005ss}: combining both control variates and quasi-Monte Carlo could lead to an increase in variance if the coefficients are naively estimated.
\section{Submodularity}
\label{app:submodularity}
A set function $g: 2^{V} \rightarrow \mathbb{R}$ is submodular if for every $A, B \subset V$, $g(A \cap B) + g(A \cup B) \leq g(A) + g(B)$, where $V$ is the set of interest---for further details on submodularity refer to \cite{krause2013t}. The following proposition states that the lower bound batch acquisition function and its approximation for the JES and MES are submodular functions defined over subsets of the input space.
\begin{proposition}
	The lower bound batch acquisition function {\normalfont $\alpha^{\text{qLB-JES}}(X| D_n)$ } and its approximations {\normalfont $\hat{\alpha}^{\text{qLB-JES}}(X| D_n)$ } are submodular functions defined over subsets of $\mathbb{X}$.
\end{proposition}
\paragraph{Proof.}
Let $X = \{\mathbf{x}_i\}_{i=1,\dots,q} \subset \mathbb{X}$ denote a set of inputs. If $K_X \in \mathbb{R}^{q\times q}$ is a positive semi-definite matrix such that $K_{i, j}$ depends only on the inputs $\mathbf{x}_i \in \mathbb{X}$ and $\mathbf{x}_j \in \mathbb{X}$, then the log-determinant function $\log\det K_X$ defined over sets $X$ is submodular \cite{kulesza2012fiml}.  Similar to \cite{moss2021jmlr}, we will show that this lower bound batch acquisition function can be written as a sum of these log-determinants.
\begin{align*}
	&\alpha^{q\text{LB-JES}}(\mathbf{x}^{[1:q]}| D_n)
	\\
	&= \frac{1}{2} \sum_{m=1}^M \log \det (\Sigma_n^{(m)}(\mathbf{x}^{[1:q]}, \mathbf{x}^{[1:q]}) + \text{diag}(\sigma^{(m)}(\mathbf{x}^{[1:q]})))
	\\
	&\quad + \frac{M}{2} \log(2 \pi e)
	- \sum_{i=1}^q \mathbb{E}_{p((\mathbb{X}^*, \mathbb{Y}^*)| D_n)} \left[ 
	H[p(\mathbf{y}^{[i]}| \mathbf{x}^{[i]}, D_{n*}, f(\mathbb{X}^*) \preceq \mathbb{Y}^*)]
	\right]
	\\
	&= \frac{1}{2} \sum_{m=1}^M \log \det (\Sigma_n^{(m)}(\mathbf{x}^{[1:q]}, \mathbf{x}^{[1:q]}) + \text{diag}(\sigma^{(m)}(\mathbf{x}^{[1:q]}))) 
	+ \sum_{i=1}^q \log\left(
	e^{\zeta(\mathbf{x}^{[i]})}
	\right)
	\\
	&= \frac{1}{2} \sum_{m=1}^M \left(
	\log \det (\Sigma_n^{(m)}(\mathbf{x}^{[1:q]}, \mathbf{x}^{[1:q]}) + \text{diag}(\sigma^{(m)}(\mathbf{x}^{[1:q]}))) 
	+ \sum_{i=1}^q \log\left(
	e^{2 \zeta(\mathbf{x}^{[i]}) / M}
	\right)
	\right)
	\\
	&= \frac{1}{2} \sum_{m=1}^M \left( \log \det (\Sigma_n^{(m)}(\mathbf{x}^{[1:q]}, \mathbf{x}^{[1:q]}) + \text{diag}(\sigma^{(m)}(\mathbf{x}^{[1:q]}))) 
	+ \log \det \left( \text{diag} \left(
	e^{2\zeta(\mathbf{x}^{[1:q]})/ M}
	\right) \right)
	\right)
\end{align*}
where
\begin{align}
	\zeta(\mathbf{x}^{[i]})
	= \frac{M}{2q} \log(2 \pi e) - \mathbb{E}_{p((\mathbb{X}^*, \mathbb{Y}^*)| D_n)} 
	\left[ 
	H[p(\mathbf{y}^{[i]}| \mathbf{x}^{[i]}, D_{n*}, f(\mathbb{X}^*) \preceq \mathbb{Y}^*)]
	\right].
	\label{eqn:zeta}
\end{align}
For notational convenience, let $A^{(m)} = \Sigma_n^{(m)}(\mathbf{x}^{[1:q]}, \mathbf{x}^{[1:q]}) + \text{diag}(\sigma^{(m)}(\mathbf{x}^{[1:q]}))$ for $m=1,\dots,M$ and $B = \text{diag} \left(e^{\zeta(\mathbf{x}^{[1:q]}) / M} \right)$. Combining the terms, we obtain
\begin{align*}
	\alpha^{q\text{LB-JES}}(\mathbf{x}^{[1:q]}| D_n) 
	&= \frac{1}{2} \sum_{m=1}^M \left(\log \det (A^{(m)}) + \log \det (B^2) \right)
	= \frac{1}{2} \sum_{m=1}^M \log \det K^{(m)},
\end{align*}
where $K^{(m)} = B A^{(m)} B$ with $K^{(m)}_{i, j} = e^{\zeta(\mathbf{x}^{[i]}) / M}e^{\zeta(\mathbf{x}^{[j]}) / M} A^{(m)}_{i, j}$ for $i, j = 1,\dots, q$ and $m=1,\dots,M$. Each summand, $\log \det K^{(m)}$, is a submodular function \cite{kulesza2012fiml}. As a sum of submodular functions is submodular, we conclude using that the lower bound batch acquisition function is submodular. For the approximate batch acquisition function, the expectation in \eqref{eqn:zeta} is replaced by the corresponding Monte Carlo estimate. The submodularity derivation above continues to hold when using the conditional entropy approximations.
\\
\null \hfill $\blacksquare$
\section{Cost analysis}
\label{app:cost_analysis}
In this section we consider the costs involved in evaluating the JES acquisition function according to \cref{alg:jes}. We also include some discussion of the cost of the other information-theoretic criterion, namely MES and PES.

\paragraph{Sampling cost.} The cost of approximate sampling from a single Gaussian processes $p(f^{(m)}|D_n)$ using the random Fourier features (described briefly in \cref{app:gp_sampling}) is $O(\min(n, L)^3)$, where $L$ is the number of Fourier features---for more details refer to \cite{wilson2020icml}. 
An evaluation of a sample at a set of inputs $X \subset \mathbb{X}$ has a linear cost depending on $|X|$ \cite{wilson2020icml}.

\paragraph{Multi-objective optimization cost.} Given a sample $f_s$, we optimize for $p$ Pareto optimal points $(\mathbb{X}^*_s, \mathbb{Y}^*_s)$ using the popular genetic algorithm known as NSGA2 \cite{deb2002itec}. The cost of this algorithm is $O(M N_{\text{pop}}^2 N_{\text{gen}} N_{\text{off}})$, where $N_{\text{pop}}$ is the size of the population, $N_{\text{gen}}$ is the number of generations and $N_{\text{off}}$ is the number of offspring \cite{deb2002itec}. At a high level, NSGA2 works by evaluating the function at $N_{\text{pop}}$ locations at time $t=1$ and then moves on to evaluate $N_{\text{off}}$ candidates for the rest of the time $t=2,\dots,N_{\text{gen}}$. The location of the offspring evaluations are determined by a random heuristic motivated by the mechanisms involved in the theory of evolution: crossover, mutation, elitism and diversity. 

\paragraph{Box decomposition.} The cost of performing a single the box decomposition based on the incremental algorithm in \cite{lacour2017c&or} is $O(p^{\floor{M/2} + 1})$, where $p$ is the number of Pareto optimal points. 

\paragraph{Conditioning cost.} Conditioning on an additional $p$ data-points requires updating the Cholesky decomposition of the input covariance matrix. The cost of a single update of this kind relies on a triangular solve, which has a quadratic complexity $O(M(n+p)^2)$. 

\paragraph{Initial entropy evaluation cost.} The initial entropy \eqref{eqn:prior_entropy} can be computed directly from the posterior covariance. The cost of instantiating the caches of the Gaussian process covariance, namely $(\Sigma_{0}^{(m)}(X_n, X_n) + \text{diag}(\sigma^{(m)}(X_n))^{-1}$ is $O(n^3)$. The cost of evaluating the posterior covariance \eqref{eqn:post_cov} at point $\mathbf{x} \in \mathbb{X}$ is $O(n^2)$. In the batch case, we also have to compute a log-determinant of the $q \times q$ covariance matrix, which has a cost of $O(q^3)$. We use the implementation in the GPyTorch \cite{gardner2018anips}, which computes the approximate log-determinant, which only uses a linear cost of $O(q)$---this approximation approach is outlined in \cite{dong2017anips}. 

\paragraph{Conditional entropy evaluation cost.} Assuming the posterior model and conditioned model are instantiated, we will consider the operations involved when evaluating the conditional entropy estimates. The cost of evaluating the probability density function and CDF of a univariate normal distribution at a point $\mathbf{x} \in \mathbb{X}$ are both assumed to be $O(1)$. As a result, the dominant cost of evaluating the $h^{\text{JES-}0}$ \eqref{eqn:h_0} and $h^{\text{JES-LB}2}$ \eqref{eqn:h_lb2} comes from the evaluation of the variance. For $h^{\text{JES-LB}}$ \eqref{eqn:h_lb}, we additionally have to populate an $M \times M$ matrix and compute its log-determinant. For $h^{\text{JES-MC}}$ \eqref{eqn:h_mc}, we need to initialize a set of $I$ base samples for the reparameterization trick \cite{wilson2018anips} described in \cref{app:monte_carlo}---this has a linear one time cost of $O(M I)$. Assuming the Monte Carlo samples are generated, the rest of the operations depend linear on $I$.

\paragraph{Expectation propagation.} To approximate the density $p(\mathbf{y}| \mathbf{x}, D_n, \mathbb{X}^*)$ in the PES acquisition function \eqref{eqn:pes}, the authors of \cite{garrido-merchan2019n} consider using expectation propagation \cite{minka2001uai}. The dominant cost of this is from matrix inversions of the covariance matrix, which scales cubically with the number of data points in consideration. In the initialization phase, we prepare the expectation propagation caches, which require inverting an $(n + p) \times (n + p)$ matrix, whilst during testing we invert a $(q + p) \times (q + p)$ matrix, where $q$ is the batch size.

\paragraph{Total cost.} The total cost is the sum of the initialization cost and the query cost. Using the variables defined in \cref{tab:init_cost} and \cref{tab:estimate_cost}, we can write down the initialization and query cost of JES, MES and PES. For the calculations, we assume that we have already queried $n$ points and we want to compute the acquisition function at a batch of $q$ points using $S$ Monte Carlo samples of the Pareto set and/or front comprised of $p$ points. 
\begin{itemize}[leftmargin=*]
	\item \textbf{JES.} The initialization cost is $C_{\text{init}} + S C_{\text{sample}}(L) + S C_{\text{moo}}(N_{\text{pop}}, N_{\text{gen}}, N_{\text{off}}) + S C_{\text{bd}}(p)+ S C_{\text{cond}}(p)$. There is an additional initial cost of $S C_{\text{base-sample}}(I)$ for the Monte Carlo conditional entropy estimate. The query cost is $C_{\text{init-h}}(q) + S q C_{\text{h}}(p)$, where $C_{\text{h}} \in \{C_{\text{h-0}}, C_{\text{h-lb}}, C_{\text{h-lb2}}, C_{\text{h-mc}}\}$ is the conditional entropy estimation strategy. Keeping the other parameters fixed, the dominant query cost is $Mq^2$ when using the approximate log-determinant and $Mq^3$ when using the standard log-determinant for large $q$.
	\item \textbf{MES.} We consider the MES algorithm obtained by excluding the conditioning step described in \cref{alg:jes}. The initialization cost of this approach is $C_{\text{init}}(q) + SC_{\text{sample}}(L) + SC_{\text{moo}}(N_{\text{pop}}, N_{\text{gen}}, N_{\text{off}}) + SC_{\text{bd}}(p)$. The query cost is $C_{\text{init-h}}(q) + S q C_{\text{h}}(0)$, where $C_{\text{h}} \in \{C_{\text{h-0}}, C_{\text{h-lb}}, C_{\text{h-lb2}}, C_{\text{h-mc}}\}$ is the conditional entropy estimation strategy. Keeping the other parameters fixed, the dominant query cost is $Mq^2$ when using the approximate log-determinant and $Mq^3$ when using the standard log-determinant for large $q$.
	\item \textbf{PES.} We consider the expectation propagation approach described in \cite{garrido-merchan2019n} to approximate the PES acquisition. The batch extension is derived in a follow-up work \cite{garrido-merchan2021a}. The initialization cost of this approach is $C_{\text{init}}(q) + S C_{\text{sample}}(L) + S C_{\text{moo}}(N_{\text{pop}}, N_{\text{gen}}, N_{\text{off}}) + S C_{\text{ep0}}(p)$. The query cost is $C_{\text{init-h}}(q) + S C_{\text{ep}}(q, p)$.  Keeping the other parameters fixed, the dominant query cost is $SMq^3$ for large $q$.
\end{itemize}
Naturally, the initialization phase can be executed in parallel because we are running $S$ independent operations. For our experiments, we did not take advantage of this advantageous property when performing the sampling, multi-objective optimization and box decomposition.

\begin{table}[!htb]
	\centering
	\begin{tabular}{lll}
		\toprule
		Operation     & Reference & Cost \\
		\midrule
		Initializing the posterior $p(f|D_n)$                           & $C_{\text{init}}$        & $M n^3$    \\
		Log-determinant of a $K \times K$ matrix                        & $C_{\text{logdet}}(K)$   & $K^3$    \\
		Approximate log-determinant of a $K \times K$ matrix            & $C_{\text{alogdet}}(K)$  & $K$    \\
		Generating $I$ base samples	from $\mathcal{N} (0, \text{diag}(\mathbf{1}_M))$ & $C_{\text{base-sample}}(I)$ & $M I$ \\
		Approximate sampling of $p(f|D_n)$                              & $C_{\text{sample}}(L)$   & $M \min(n, L)^3$  \\
		Multi-objective optimization of $f_s$   & $C_{\text{moo}}(N_{\text{pop}}, N_{\text{gen}}, N_{\text{off}})$ & $M N_{\text{pop}}^2 N_{\text{gen}} N_{\text{off}}$ \\
		Box decomposition of $\mathbb{Y}^*_s$ with $p$ points           & $C_{\text{bd}}(p)$       & $p^{\floor{M/2} + 1}$  \\
		Conditioning on the Pareto optimal point $p(f| D_{n*})$         & $C_{\text{cond}}(p)$     & $M (n + p)^2$ \\
		Initialization of expectation propagation caches                & $C_{\text{ep0}}(p)$      & $M (n + p)^3$ \\
		\bottomrule
	\end{tabular}
	\bigskip
	\caption{The initialization and operation costs. The cost only includes the highest order terms and we have ignored the constant factors.}
	\label{tab:init_cost}
\end{table}

\begin{table}[!htb]
	\centering
	\begin{tabular}{lll}
		\toprule
		Operation & Reference & Cost \\
		\midrule
		Posterior covariance $\mathbf{\Sigma}_n(X, X)$ & $C_{\text{cov}}(|X|, n)$ & $M (|X|n^2 + |X|^2)$    \\
		Initial entropy $H[p(\mathbf{y}^{[1:q]} | \mathbf{x}^{[1:q]}, D_n)]$ & $C_{\text{init-h}}(q)$ & $C_{\text{cov}}(q, n) + M C_{\text{logdet}}(q)$ \\
		Conditional entropy estimate $h^{\text{JES-}0}$  & $C_{\text{h-0}}(p)$ & $C_{\text{cov}}(1, n + p)$    \\
		Conditional entropy  estimate $h^{\text{JES-LB}}$ & $C_{\text{h-lb}}(p)$ & $C_{\text{cov}}(1, n + p) + M^2 + C_{\text{logdet}}(M)$  \\
		Conditional entropy estimate $h^{\text{JES-LB}2}$ & $C_{\text{h-lb2}}(p)$ & $ C_{\text{cov}}(1, n + p)$    \\
		Conditional entropy estimate $h^{\text{JES-MC}}$  & $C_{\text{h-mc}}(p)$ & $ C_{\text{cov}}(1, n + p) + M I$    \\
		Expectation propagation & $C_{\text{ep}}(q, p)$ & $M (q + p)^3$ \\
		\bottomrule
	\end{tabular}
	\bigskip
	\caption{The cost of involved with querying after the initialization. The cost only includes the highest order terms and we have ignored the constant factors. In our experiments, we use the approximate log-determinant, which cost $C_{\text{alogdet}}$ instead of the more expensive cost of $C_{\text{logdet}}$. }
	\label{tab:estimate_cost}
\end{table}
\FloatBarrier
\section{Estimation error}
\label{app:estimation_error}
Overall, there are five sources of estimation error when approximating the batch JES acquisition function. In this section, we briefly enumerate and discuss these errors below.

\begin{enumerate}[leftmargin=*]
	\item The first source of error arises from replacing the global optimality condition with the local optimality condition. This turns out to be a lower bound approximation:
	\begin{align*}
		\alpha^{\text{JES}}(\mathbf{x}| D_n) 
		&= H[p(\mathbf{y}| \mathbf{x}, D_n)]
		- \mathbb{E}_{p((\mathbb{X}^*, \mathbb{Y}^*)| D_n)}[H[p(\mathbf{y}| \mathbf{x}, D_n, (\mathbb{X}^*, \mathbb{Y}^*))]]
		\\
		&= H[p(\mathbf{y}| \mathbf{x}, D_n)]
		- \mathbb{E}_{p((\mathbb{X}^*, \mathbb{Y}^*)| D_n)}[H[p(\mathbf{y}| \mathbf{x}, D_n \cup (\mathbb{X}^*, \mathbb{Y}^*), f(\mathbb{X}) \preceq \mathbb{Y}^*)]]
		\\
		&\geq H[p(\mathbf{y}| \mathbf{x}, D_n)]
		- \mathbb{E}_{p((\mathbb{X}^*, \mathbb{Y}^*)| D_n)}[H[p(\mathbf{y}| \mathbf{x}, D_n \cup (\mathbb{X}^*, \mathbb{Y}^*), f(\mathbf{x}) \preceq \mathbb{Y}^*)]],
	\end{align*}
	because conditioning on more variables will never increase the entropy. The error from this lower bound approximation appears also in the majority of the work on entropy based acquisition functions. Empirically, we observed that this approximation leads to a very minor change to the selected point for the sequential acquisition function. The effects of this error on the batch acquisition function is unclear because computing an unbiased estimate of the exact batch acquisition function over the whole input space is too expensive to do in practice.
	\item The second and third source of error arises from the Monte Carlo approximation of the intractable expectation and the discrete approximation of the Pareto optimal points, respectively. The error of this step could be reduced by increasing the number of samples or optimal points at the expense of more computation. In the experiments we conducted, we set the number of Monte Carlo samples as $S=10$ and the number of optimal points as $p=10$. This combination seemed to work well under our collection of problems and the sensitivity analysis we conducted in \cref{app:sensitivity_analysis} seems to indicate that there is a diminishing gain in performance when we increase the number of samples or optimal points further. 
	\item The fourth source of error comes from estimating the conditional entropy. This can be computed exactly in the noiseless setting, but has to be estimated in the other settings. The Monte Carlo estimate gives an unbiased estimate of this term, which can be made more accurate by increasing the number of samples. In the experiments presented in \cref{app:experiments}, we observed very little difference in the points that are selected when using the different conditional entropy estimates. As a result, we recommend using the cheapest estimates which in most cases is the one based on the moment-matching approach.
	\item The fifth source of error comes from estimating the batch acquisition function using the lower bound. This error is only present when the batch size is greater than one: $q > 1$. Quantifying this error is hard to do empirically because the exact batch acquisition function is too expensive to estimate unbiasedly. Nevertheless, the lower bound batch acquisition function and its approximation are still principled acquisition functions because they can be written as determinantal point processes (DPPs) \cite{kulesza2012fiml}. This property was derived and used in the proof of submodularity in \cref{app:submodularity}. 
	\\ \\
	As discussed in the BO literature \cite{kathuria2016anips, contal2013mlkdd, wang2017icmla}, batch acquisition functions based on DPPs are powerful because they can promote diverse batches in high-quality regions. The trade-off between diversity and quality (Section 3.1 in \cite{kulesza2012fiml}) is evident in the form of the DPP kernel, $K^{(m)}_{i,j} = q_i q_j A_{i, j}^{(m)}$, where $q_i = \exp(\zeta_i/M)$ can be interpreted as the quality of item i, whilst $A_{i,j}^{(m)}$ corresponds to a notion of similarity. In the approximation, we use a Monte Carlo estimate for the term $\zeta_i$, which only arises when computing the quality term. As a result, the batch that is selected with this approximation might differ slightly from the optimal batch but the diversity of the batch will still be high because of the matrix $A^{(m)}$. 
\end{enumerate}
%
\section{Contour plots}
\label{app:contours}
\begin{figure}
	\includegraphics[width=.245\linewidth]{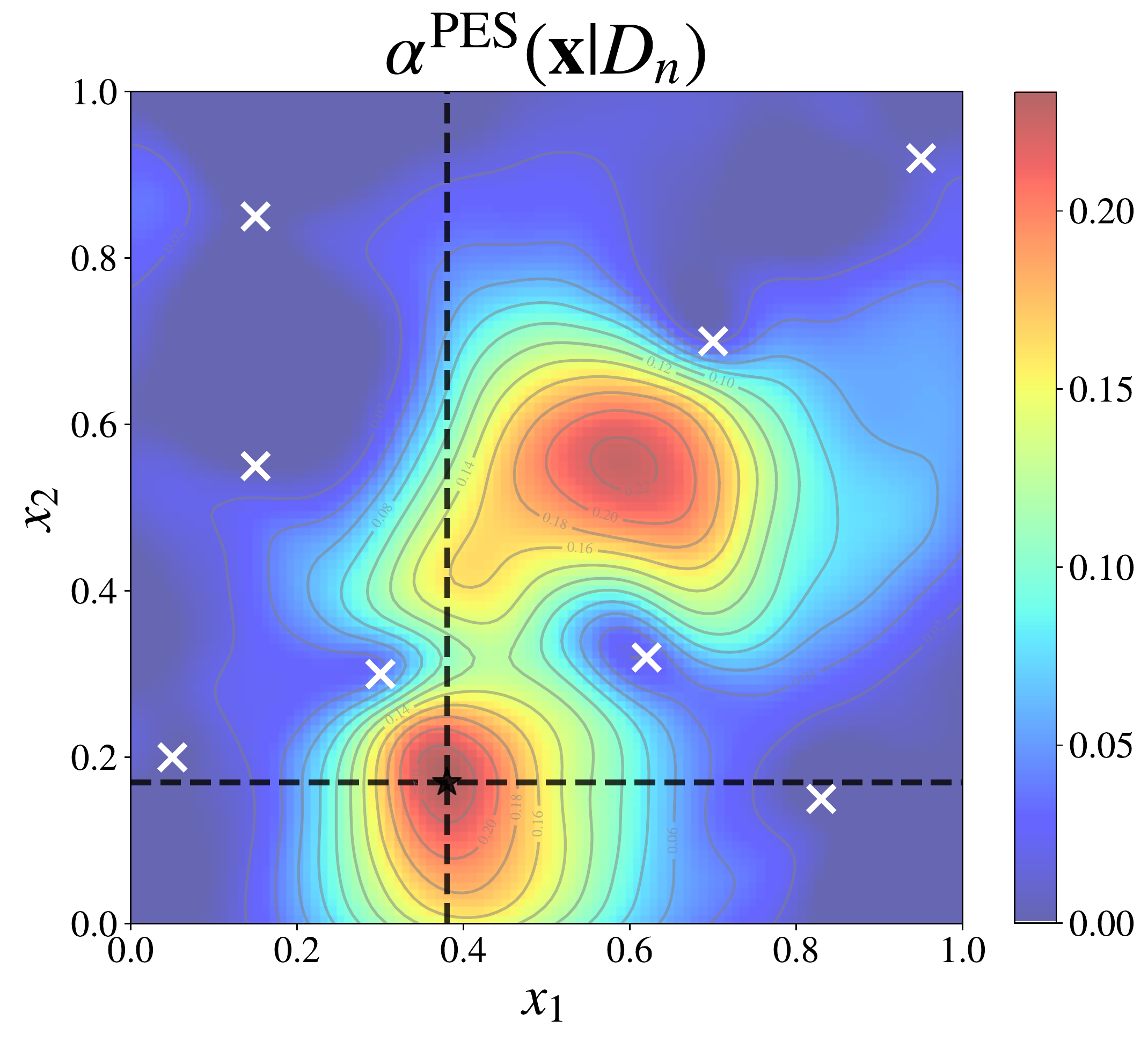}
	\includegraphics[width=.245\linewidth]{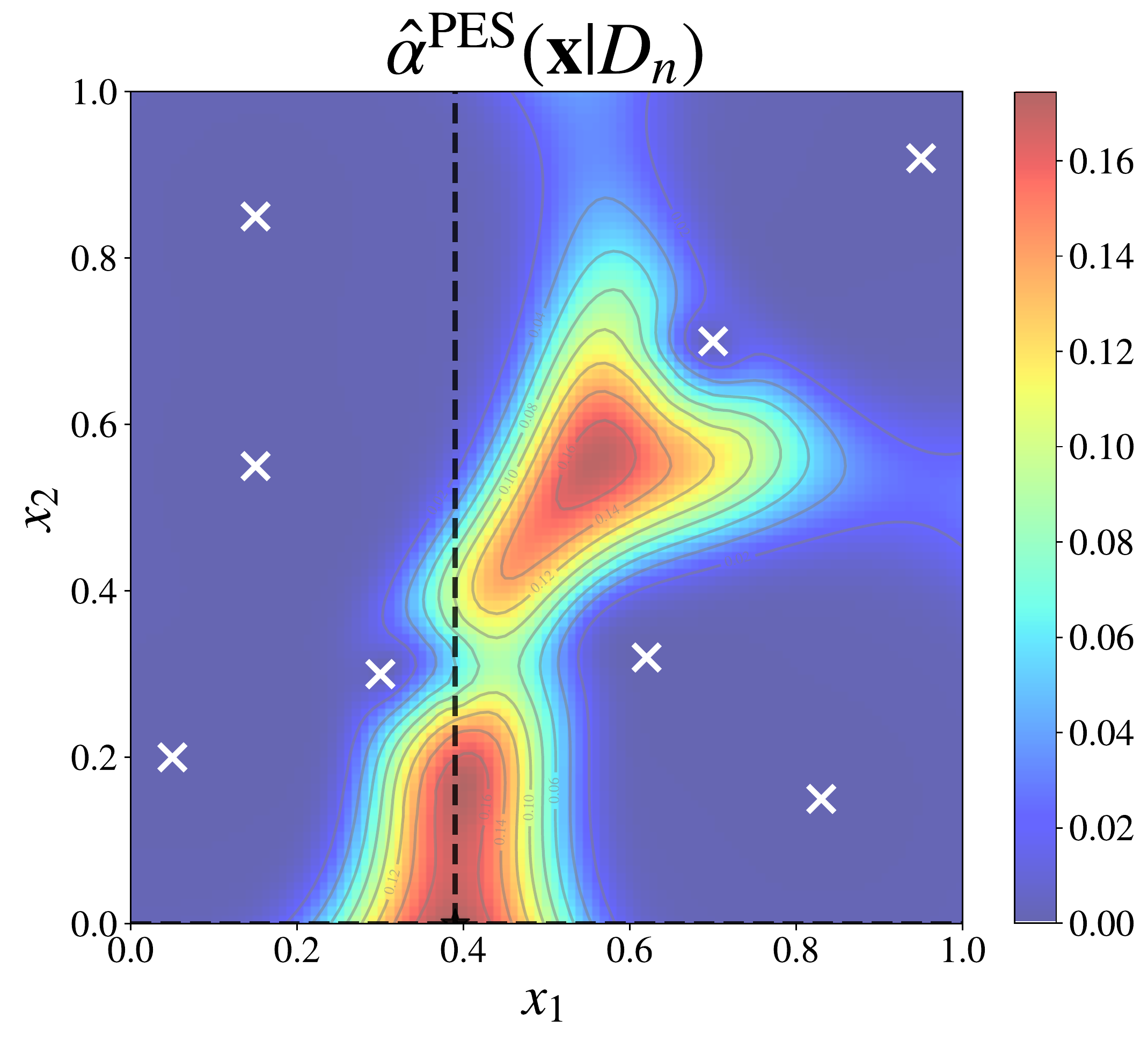}
	\includegraphics[width=.245\linewidth]{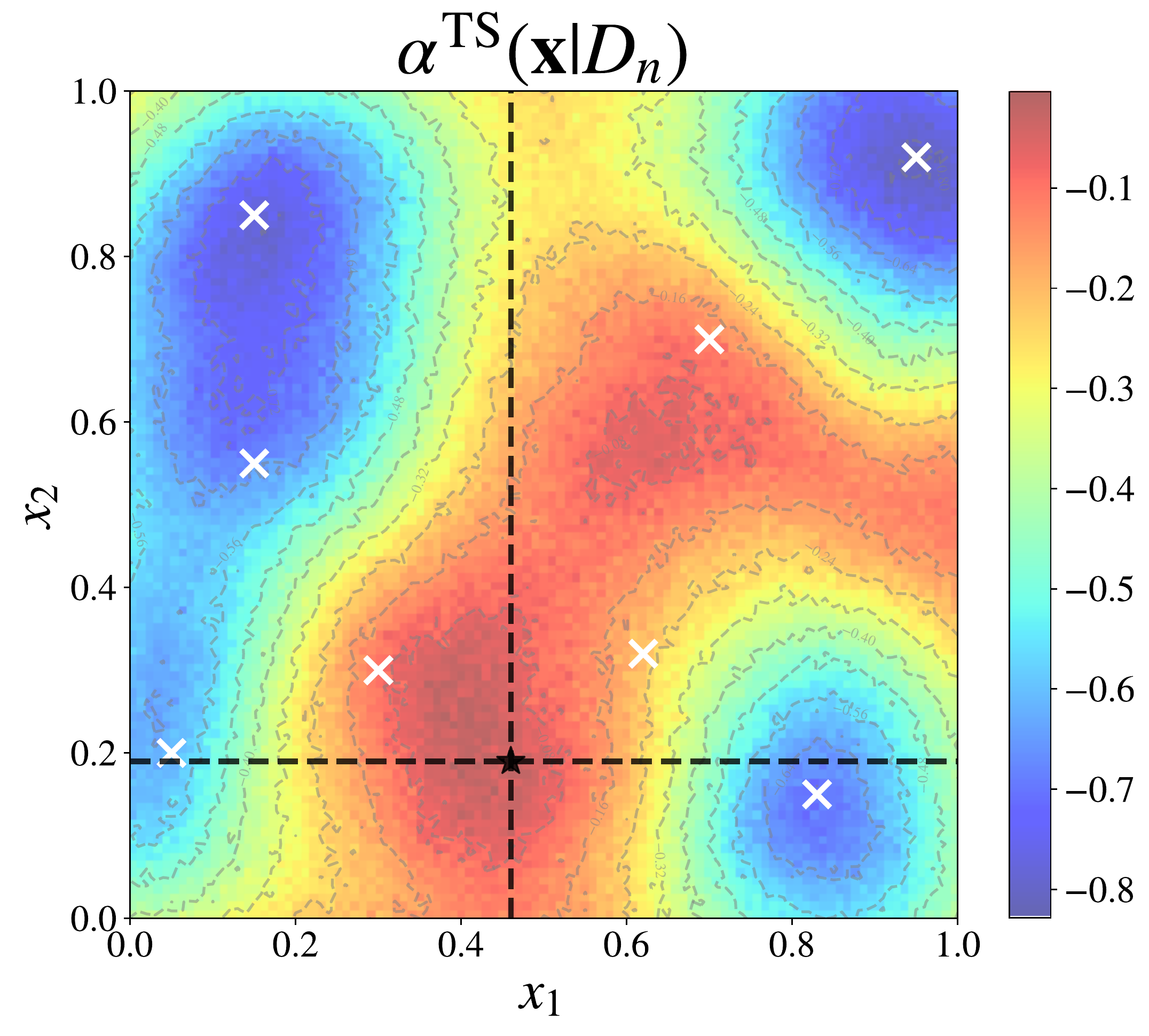}
	\centering
	\caption{The contours for the predictive entropy search acquisition and its approximation obtained via expectation propagation. For reference we also include a random Thompson sample. The location of the maximizer is highlighted using a pair of dotted black lines.}
	\label{fig:rastrigin_pes}
\end{figure}

\begin{figure}
	\includegraphics[width=.245\linewidth]{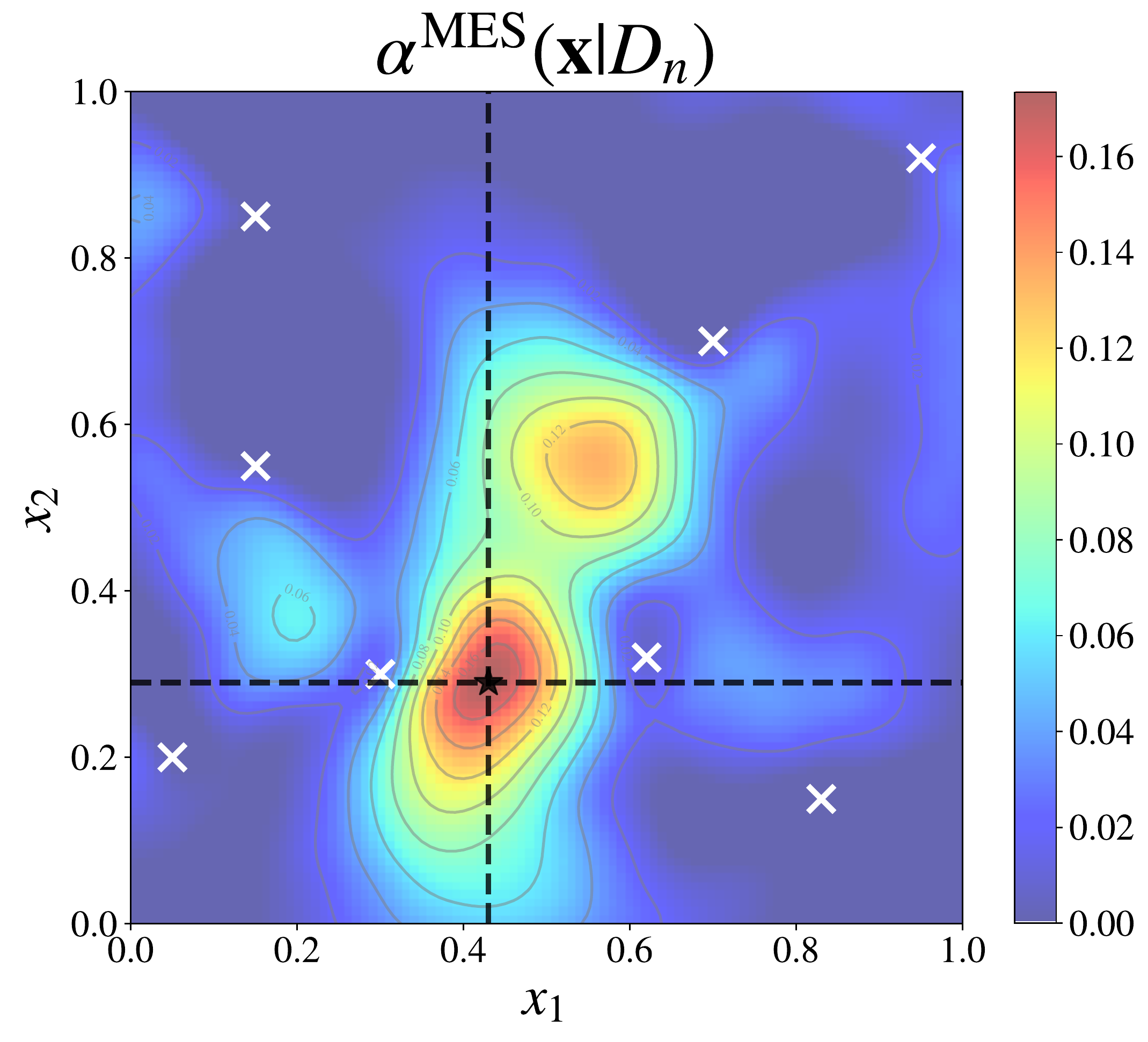}
	\includegraphics[width=.245\linewidth]{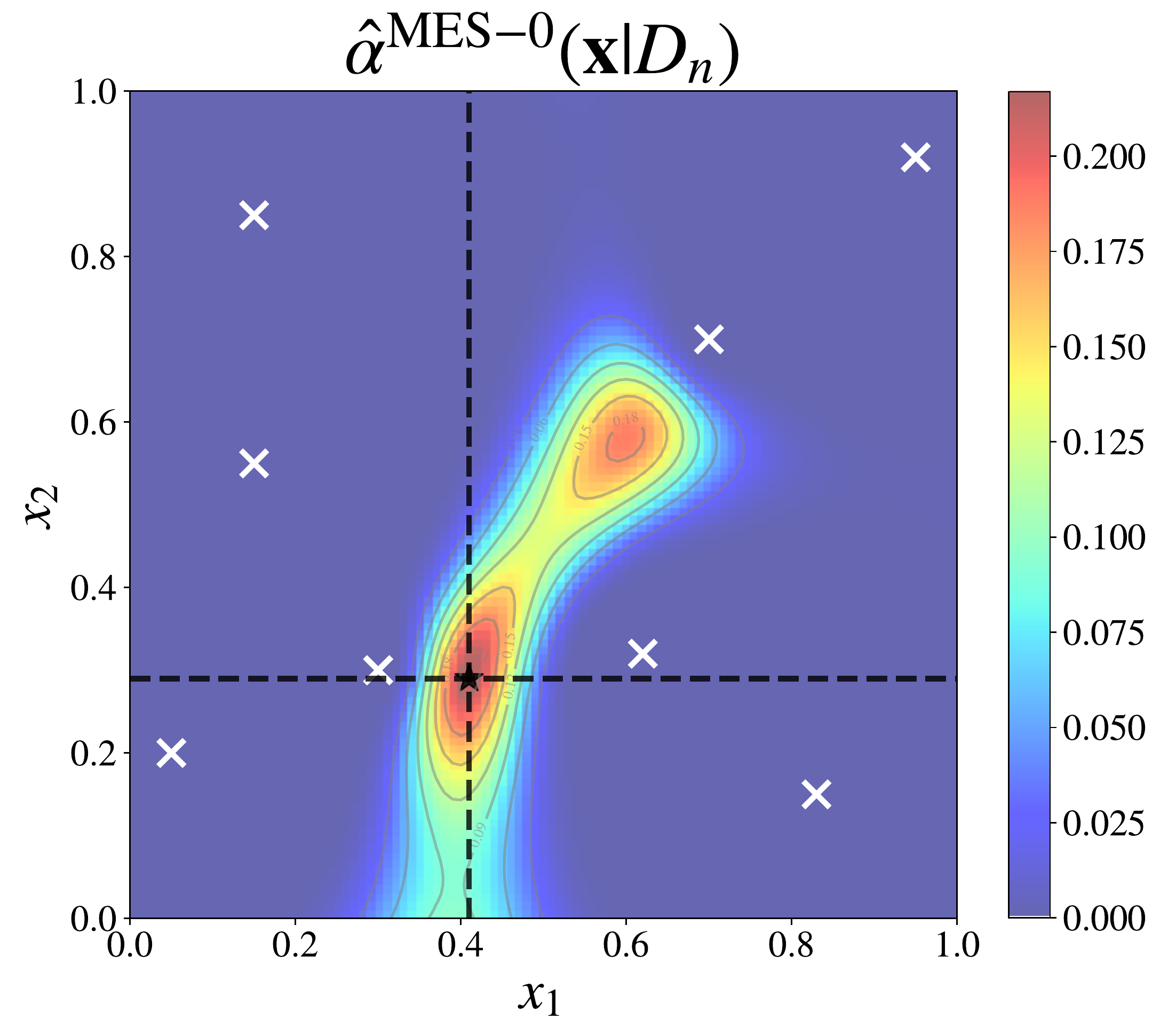}
	\includegraphics[width=.245\linewidth]{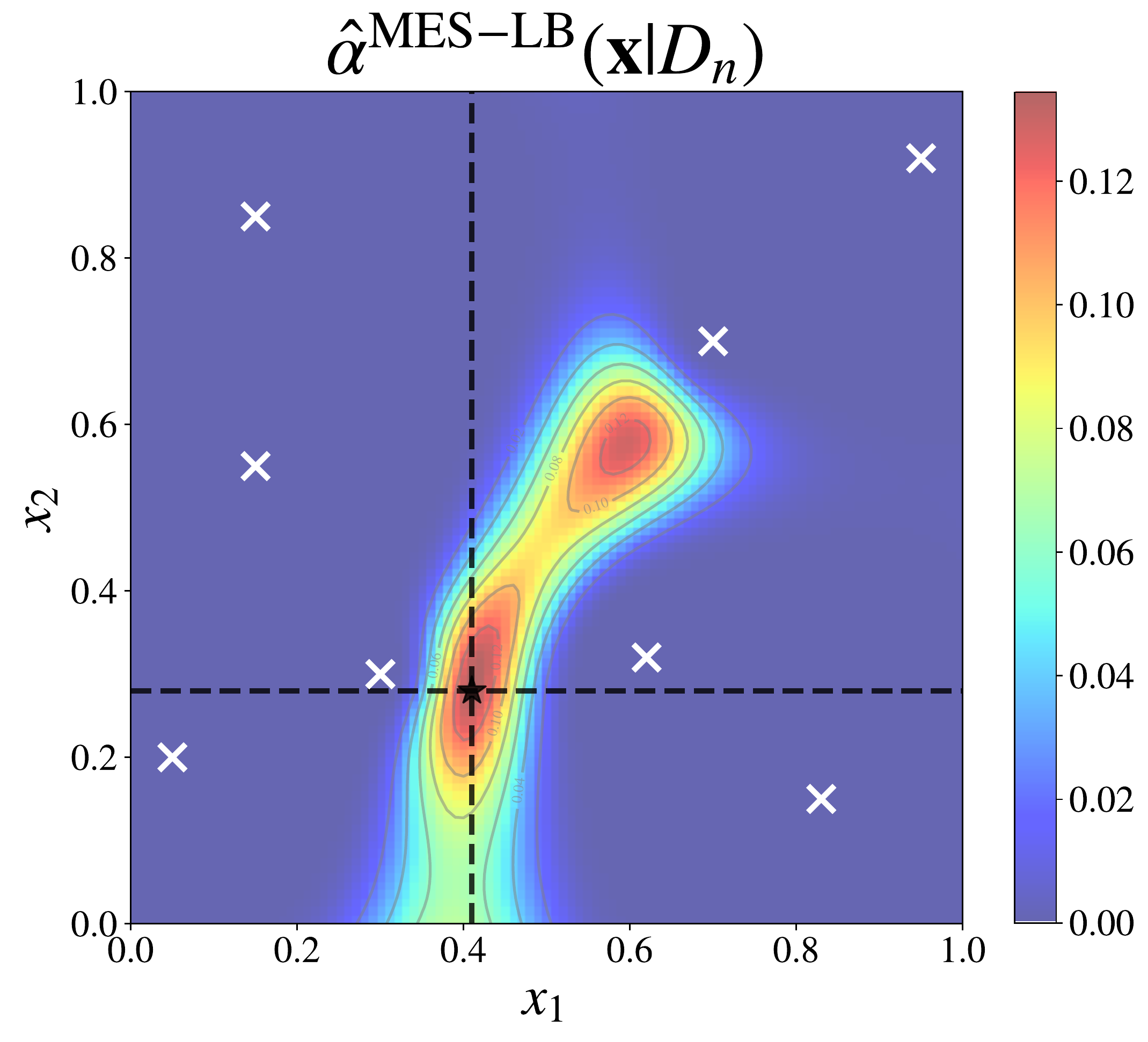}
	\includegraphics[width=.245\linewidth]{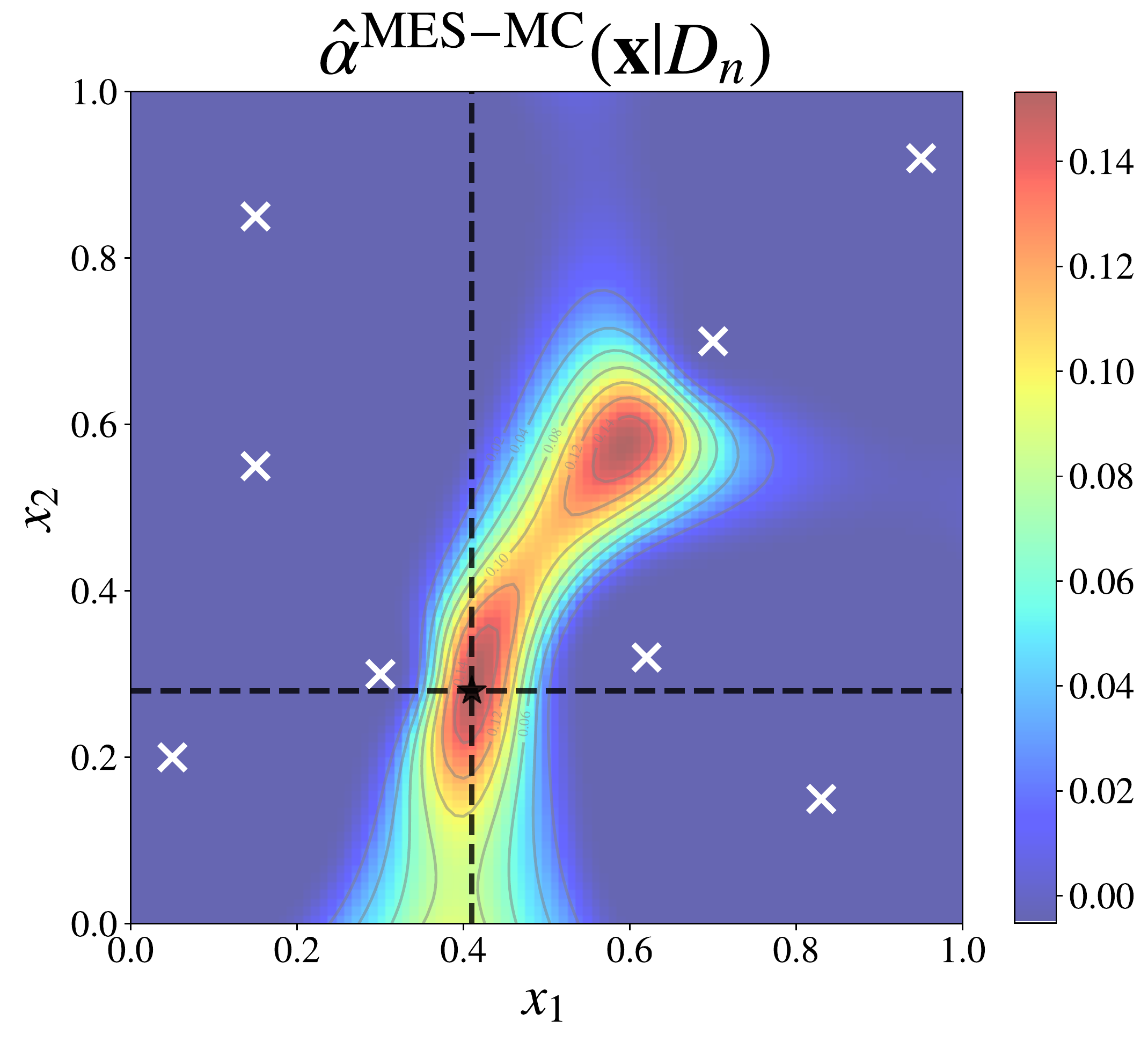}
	\centering
	\caption{The contours for the maximum value entropy search acquisition function and its approximation obtained via the zero noise approximation, moment matching and Monte Carlo. The location of the maximizer is highlighted using a pair of dotted black lines.}
	\label{fig:rastrigin_mes}
\end{figure}

\begin{figure}
	\includegraphics[width=.245\linewidth]{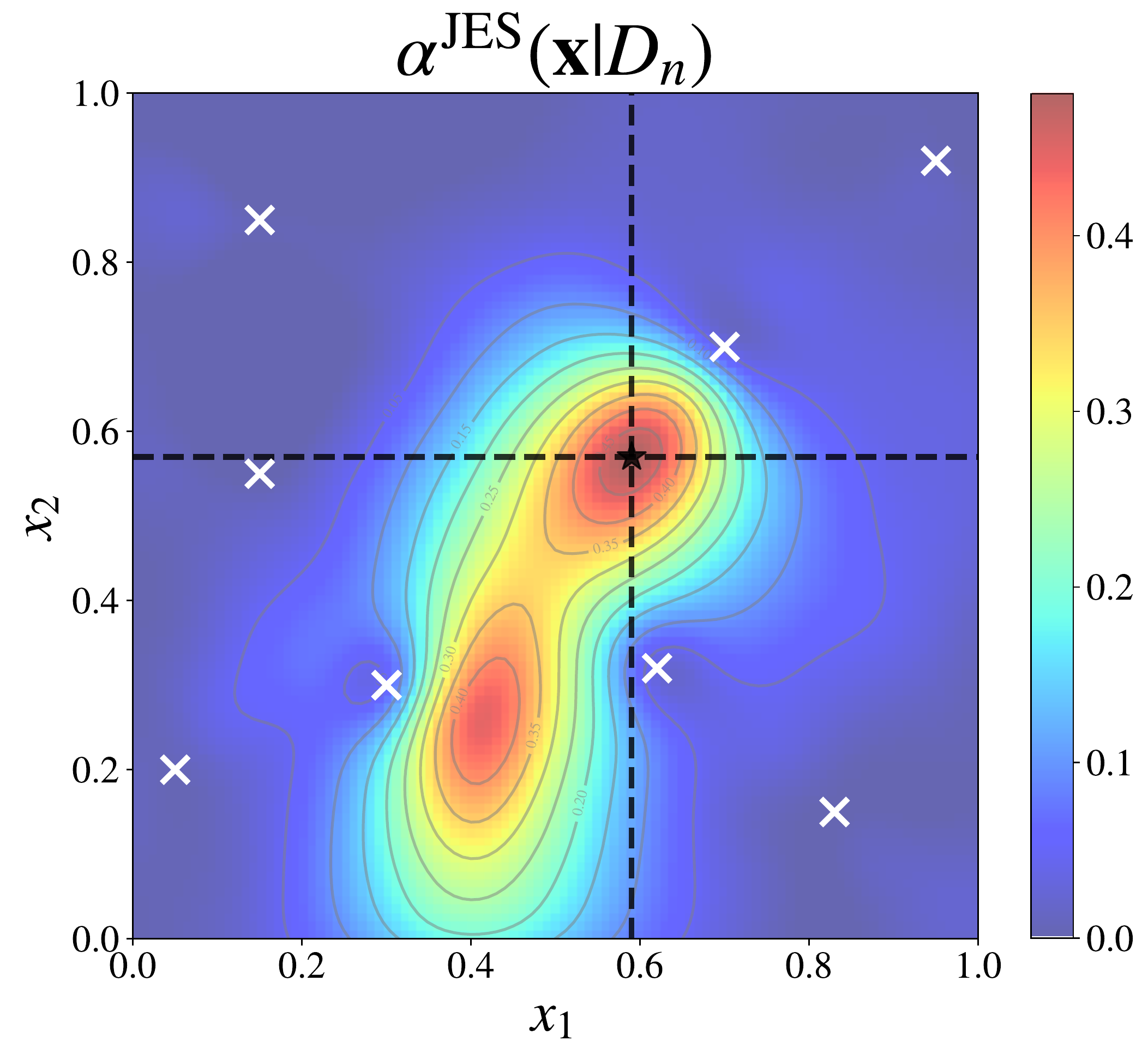}
	\includegraphics[width=.245\linewidth]{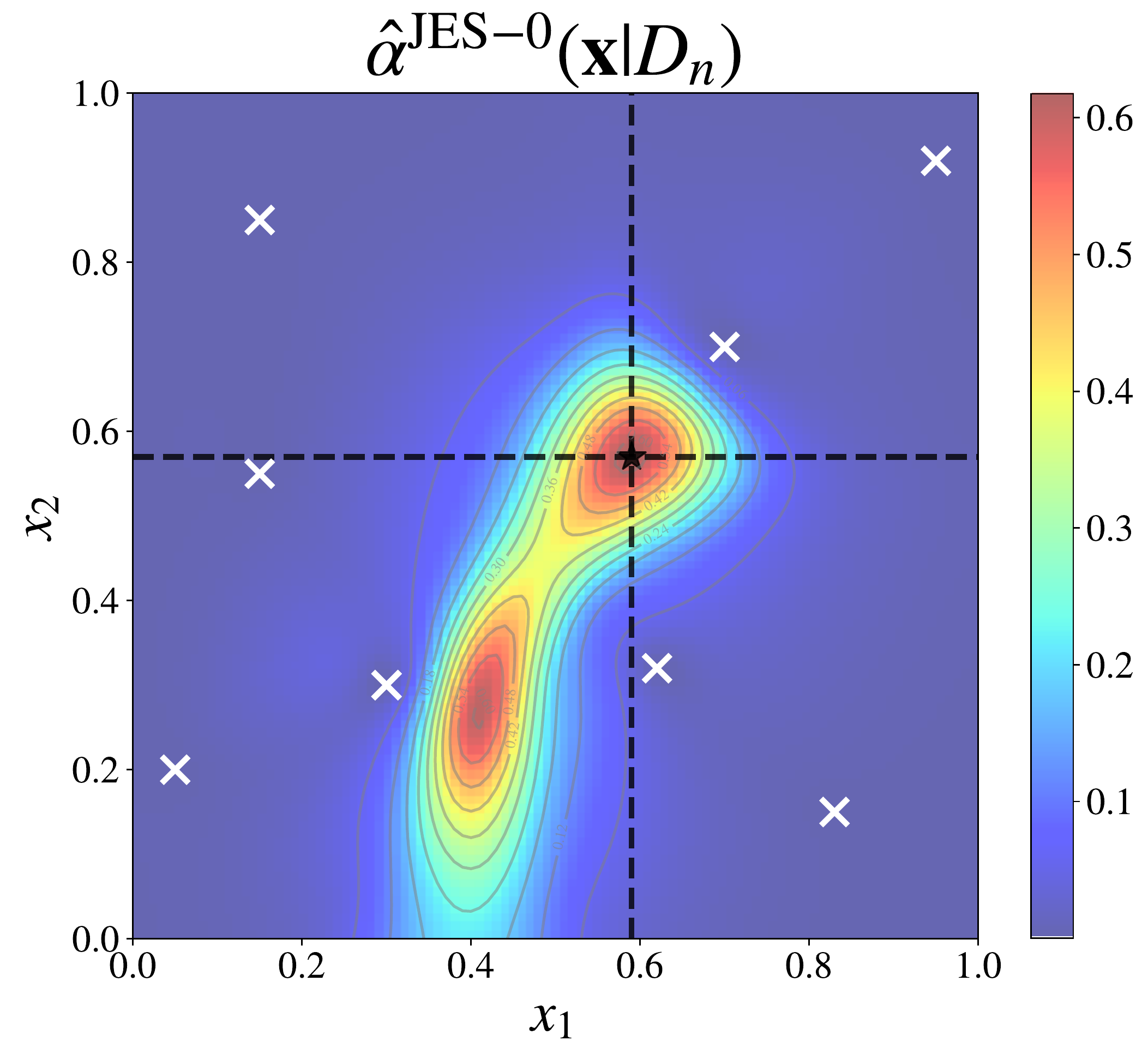}
	\includegraphics[width=.245\linewidth]{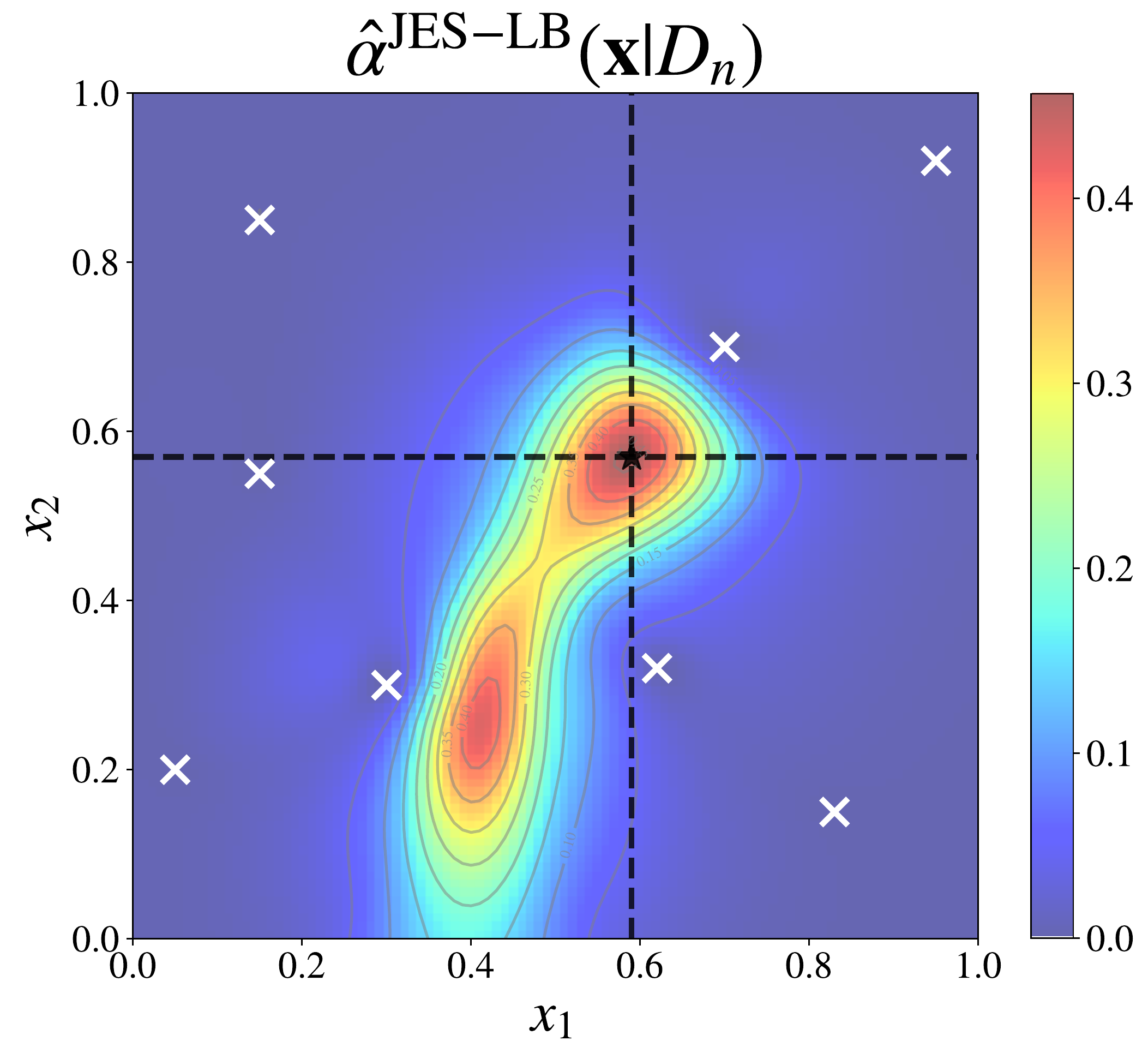}
	\includegraphics[width=.245\linewidth]{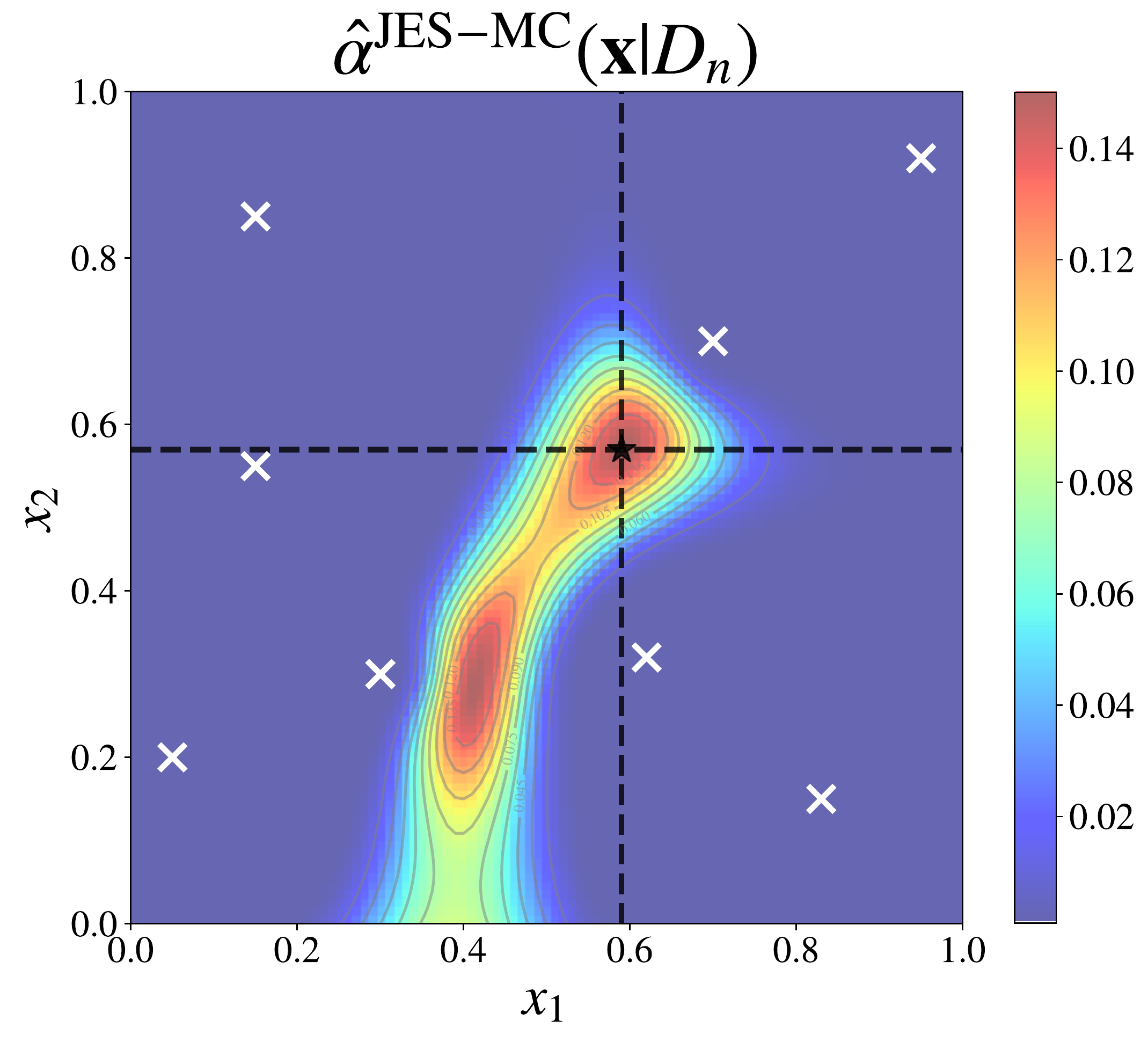}
	\centering
	\caption{The contours for the joint entropy search acquisition function and its approximation obtained via the zero noise approximation, moment matching and Monte Carlo. The location of the maximizer is highlighted using a pair of dotted black lines.}
	\label{fig:rastrigin_jes}
\end{figure}

\FloatBarrier
In this section we present some contour plots to illustrate visually the differences that emerges between the different approximation strategies for the information-theoretic acquisition functions on a single-objective problem. The approximation for PES, MES and JES are presented in \cref{fig:rastrigin_pes},  \cref{fig:rastrigin_mes} and \cref{fig:rastrigin_jes}, respectively. To obtain the ground truth for the acquisition function we run the rejection sampling scheme discussed in \cite{hernandez-lobato2014anips}. As a comparison we also include the contours of some popular single-objective acquisition functions. Visually the estimates all appear to be reasonably effective for this set-up. Interestingly, the PES and MES prefer querying around the lower mode, whilst JES prefers the upper mode which perhaps offers a better trade-off between the gain in information about both the optimal input and output.
\begin{figure}
	\begin{subfigure}[t]{0.33\textwidth}
		\includegraphics[width=1\linewidth]{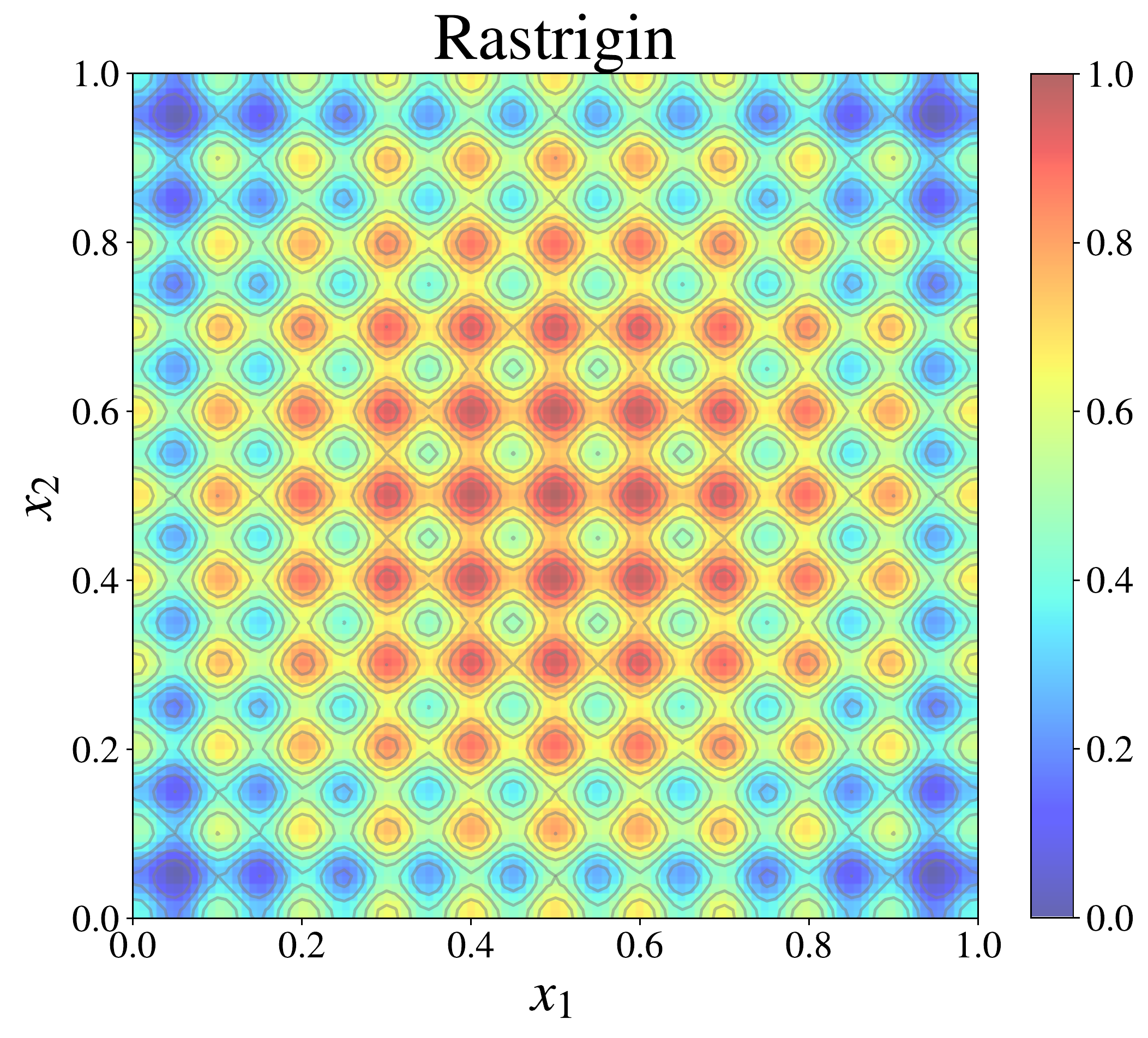}
	\end{subfigure}
	\begin{subfigure}[t]{0.33\textwidth}
		\includegraphics[width=1\linewidth]{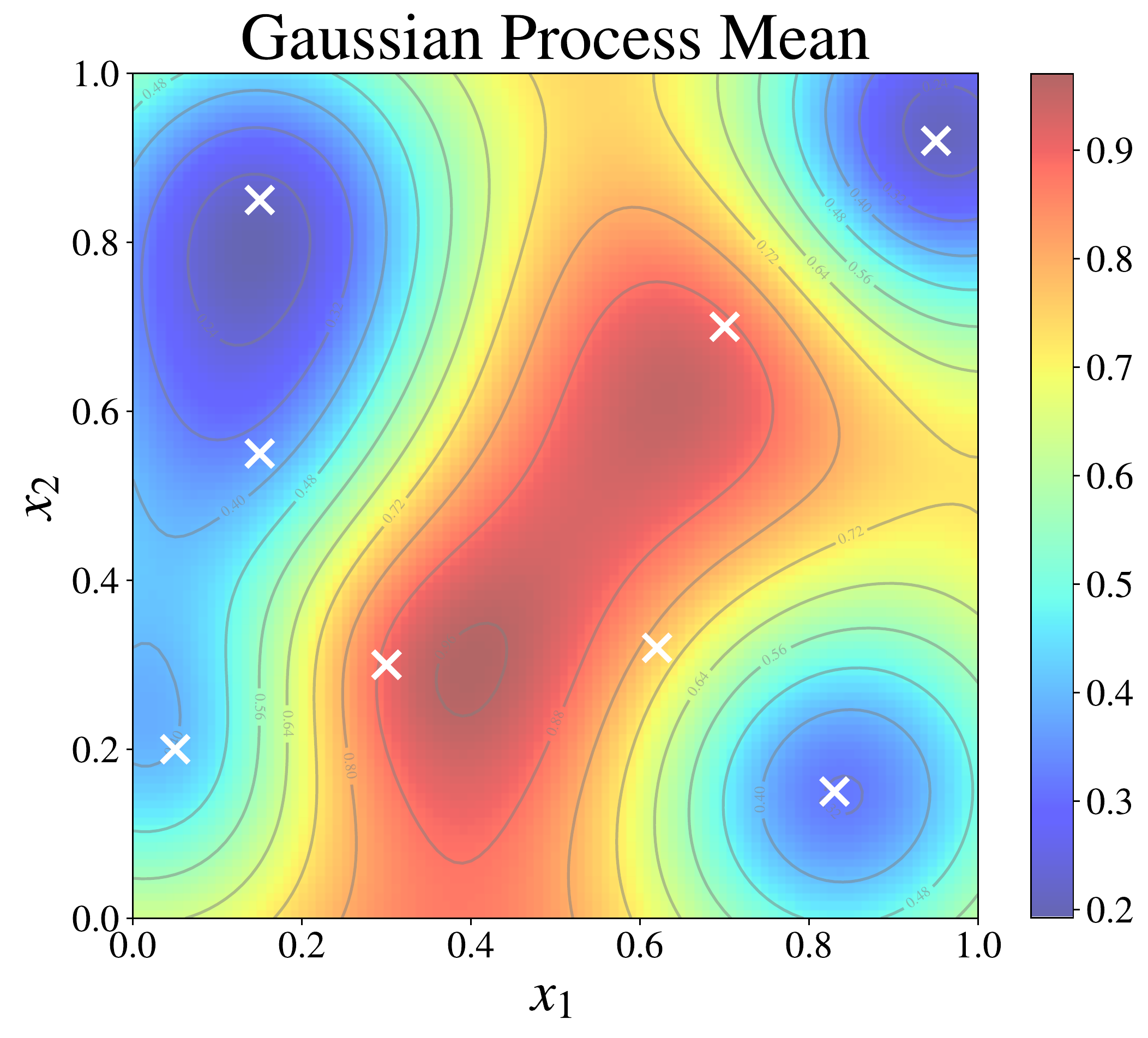}
	\end{subfigure}
	\begin{subfigure}[t]{0.33\textwidth}
		\includegraphics[width=1\linewidth]{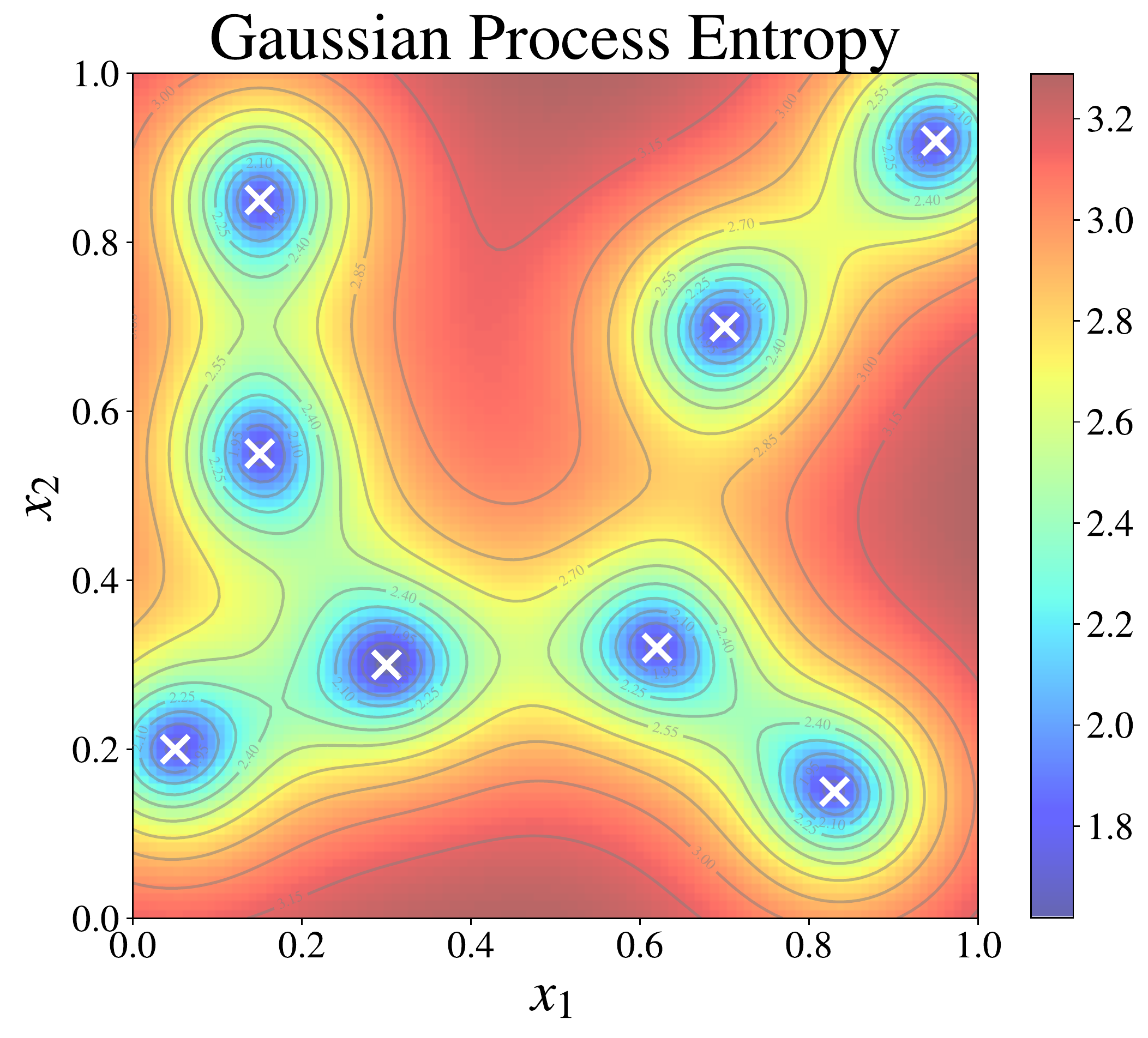}
	\end{subfigure}
	\caption{The contours for the Gaussian process posterior mean and entropy after making 8 noisy observations of the (normalized) Rastrigin objective function ($d=2, M=1$).}
\end{figure}
\begin{figure}
	\includegraphics[width=.245\linewidth]{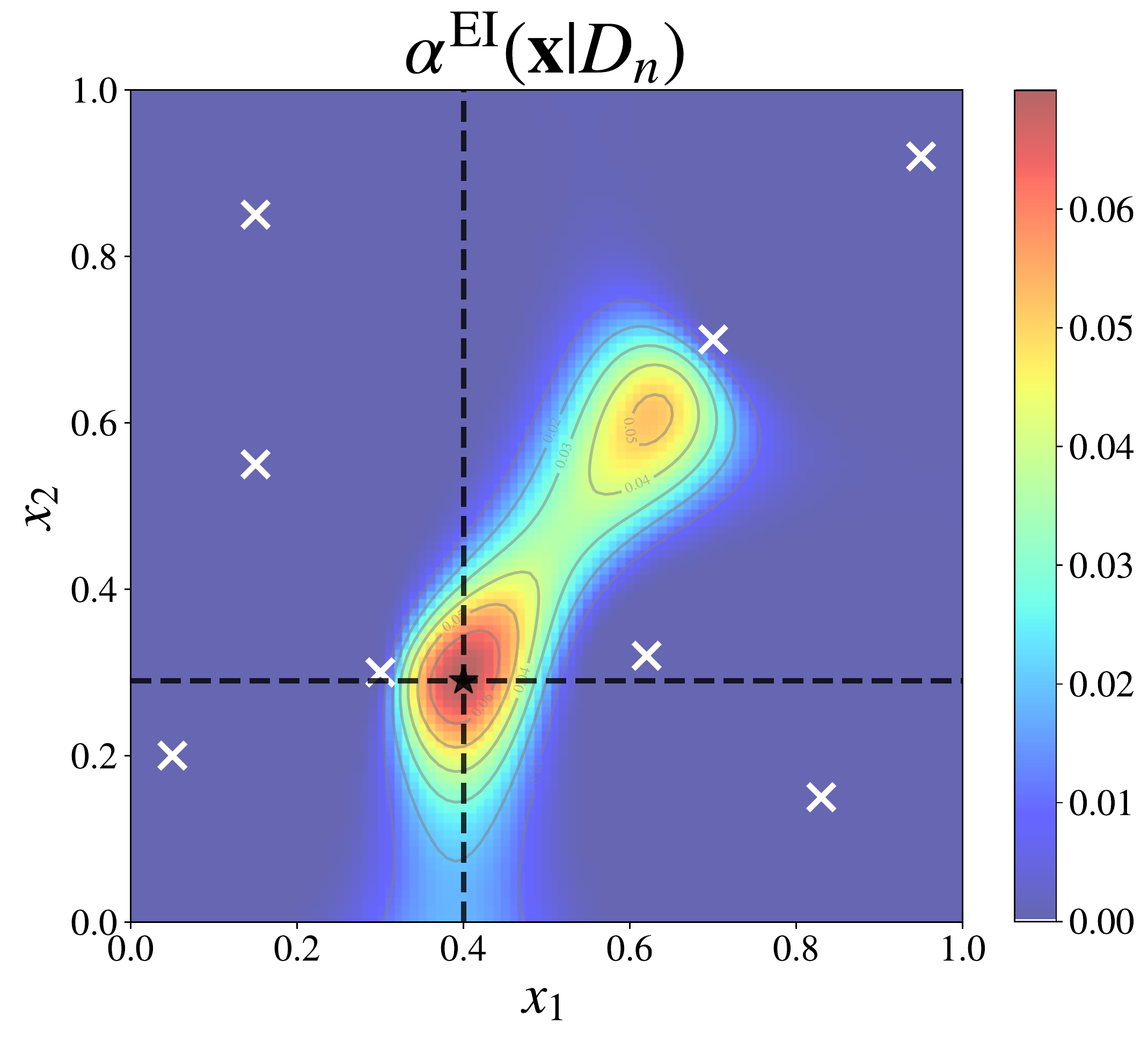}
	\includegraphics[width=.245\linewidth]{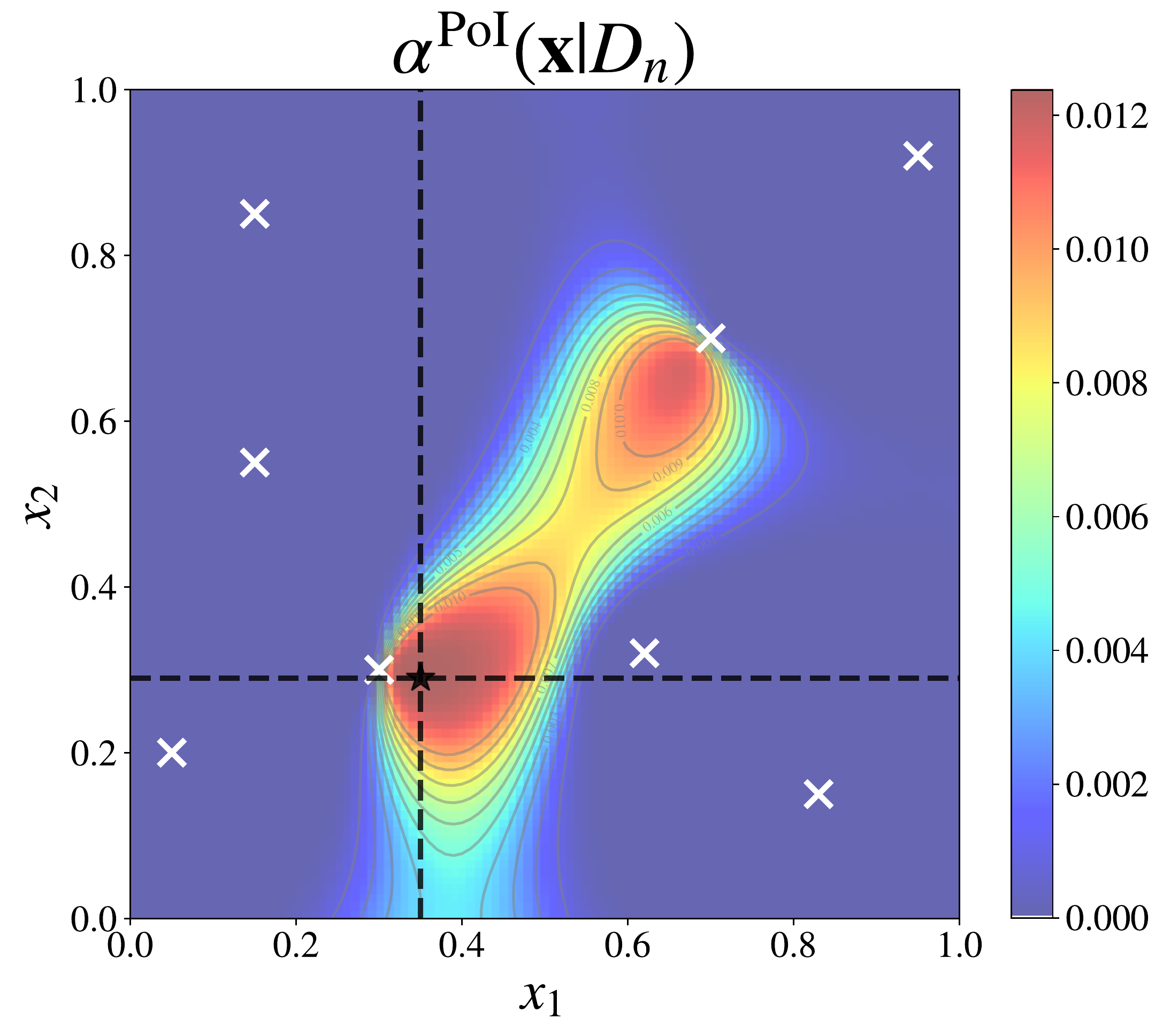}
	\includegraphics[width=.245\linewidth]{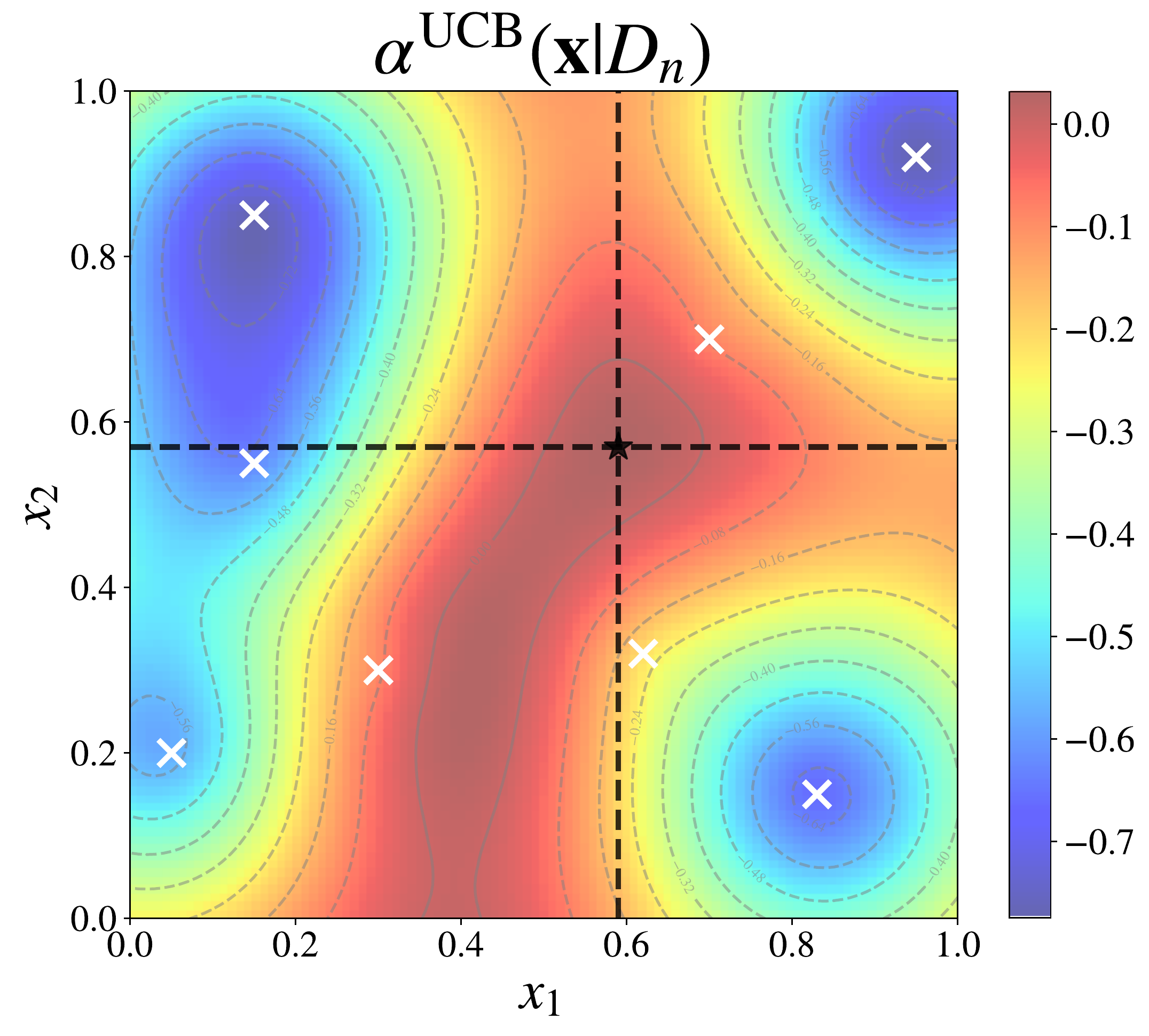}
	\includegraphics[width=.245\linewidth]{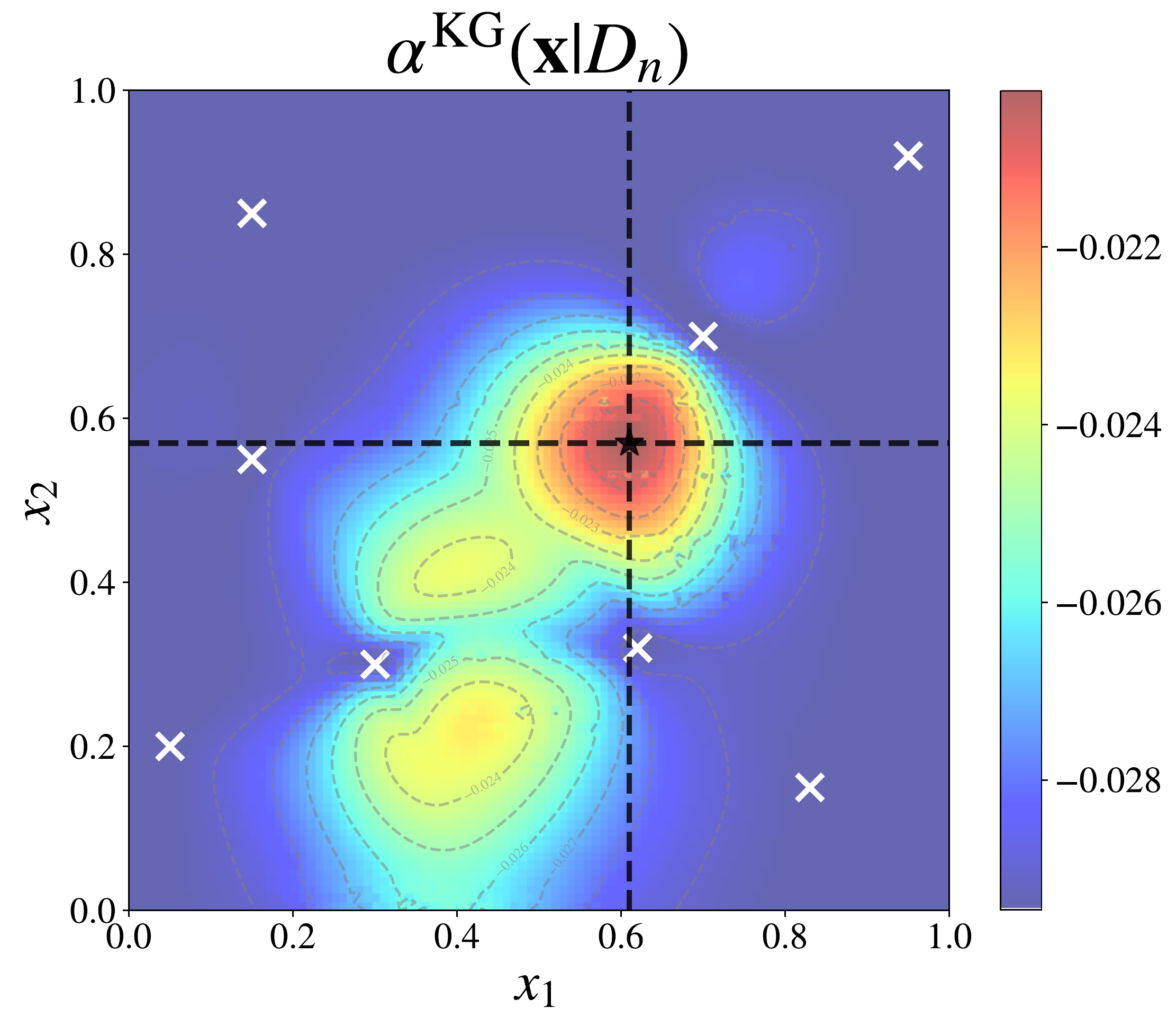}
	\centering
	\caption{The contours for the expected improvement, probability of improvement, upper confidence bound $(\beta = 3)$ and knowledge gradient acquisition function. The location of the maximizer is highlighted using a pair of dotted black lines.}
\end{figure}
\FloatBarrier
\section{Hypervolume indicator}
\label{app:hypervolume_indicator}
\begin{figure}
	\begin{subfigure}[t]{0.3\textwidth}
		\includegraphics[width=1\linewidth]{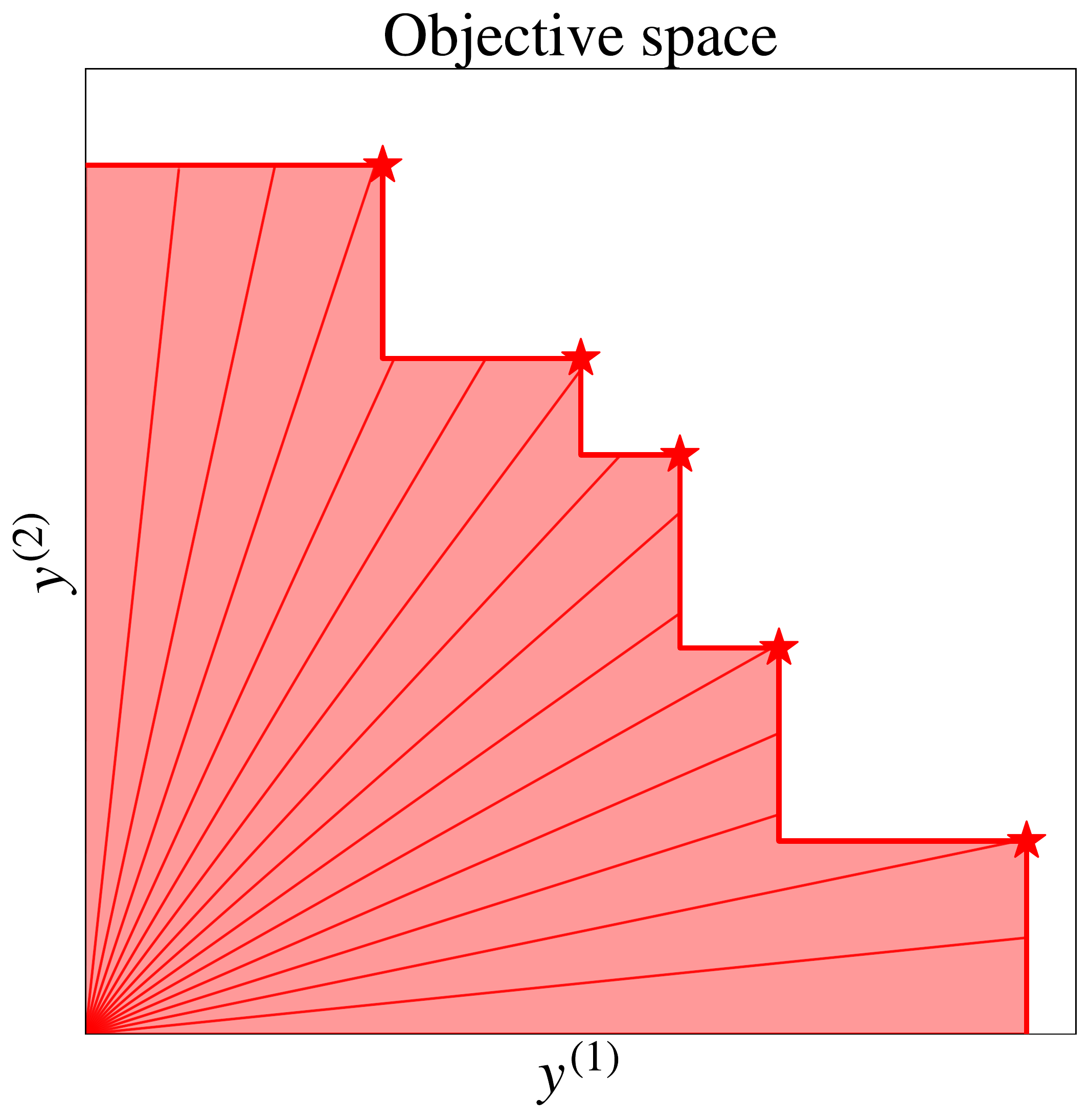}
		\caption{Beta$(1, 1)$}
		\centering
	\end{subfigure}
	\begin{subfigure}[t]{0.3\textwidth}
		\includegraphics[width=1\linewidth]{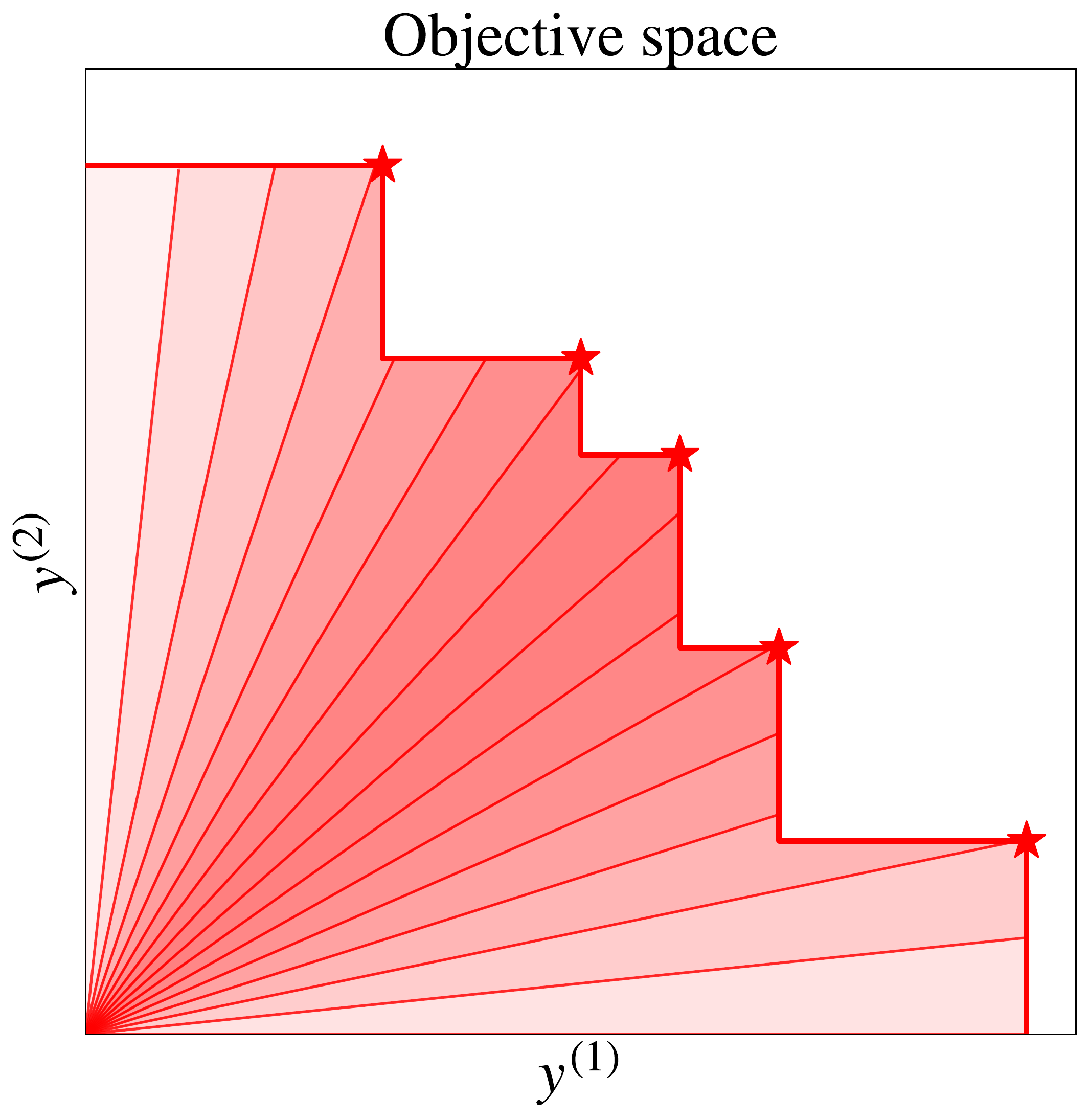}
		\caption{Beta$(2.5, 2.5)$}
		\centering
	\end{subfigure}
	\begin{subfigure}[t]{0.3\textwidth}
		\includegraphics[width=1\linewidth]{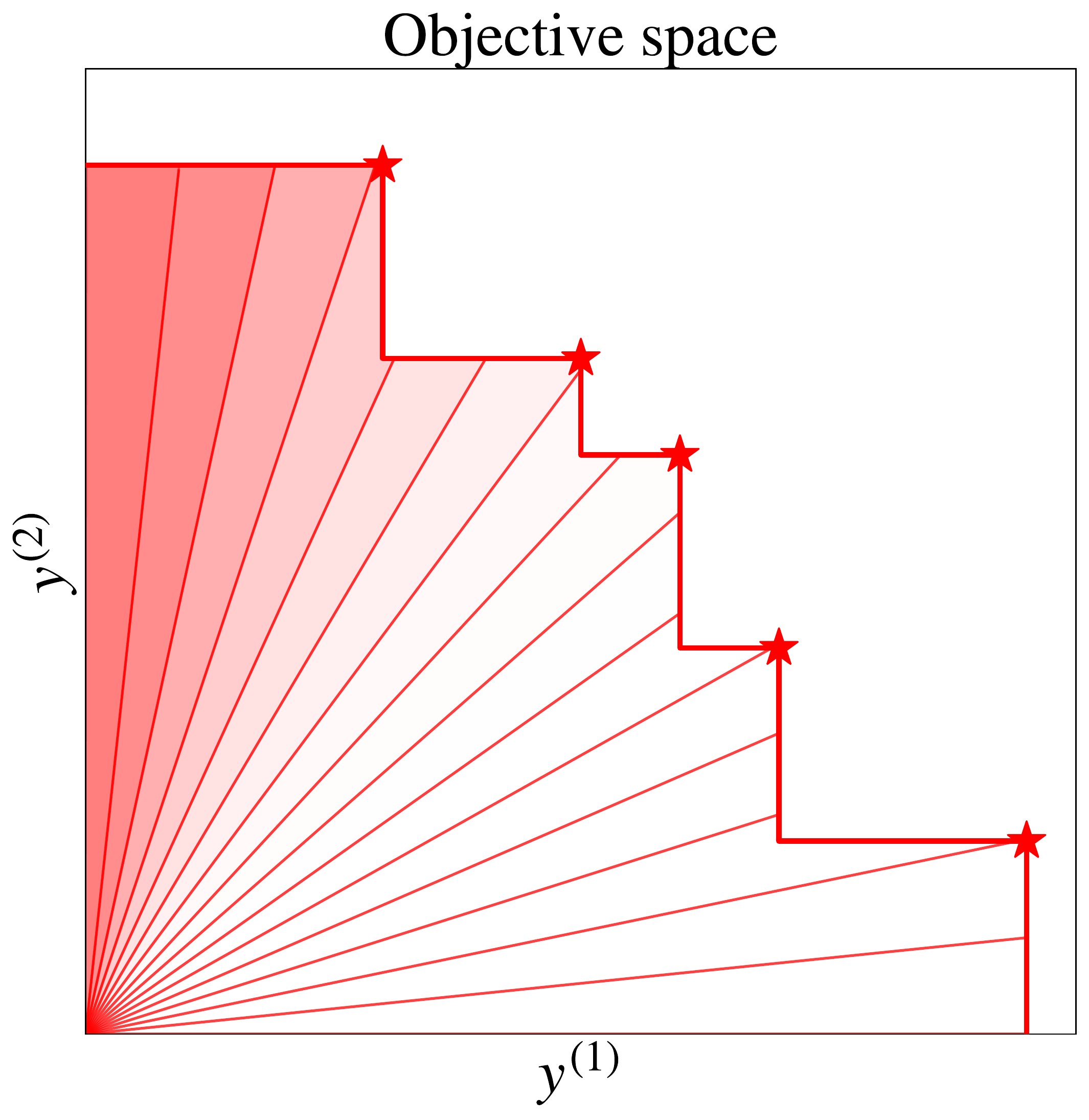}
		\caption{Beta$(5, 2)$}
		\centering
	\end{subfigure}
	\caption{The contribution of each radial segment according to the distribution $p(\mathbf{w})$.}
	\label{fig:hv_beta}
\end{figure}

\FloatBarrier
The hypervolume indicator is defined as the area of the dominated region between a reference point $\mathbf{r} \in \mathbb{R}^M$ and the set of interest $A \subset \mathbb{R}^M$: 
\begin{equation*}
	U_{\text{HV}}(A) = \int_{\mathbb{R}^M} \mathbb{I}[\mathbf{r} \preceq \mathbf{z} \preceq A] d\mathbf{z}.
\end{equation*}
As discussed in the main text (\cref{sec:performance_criteria}), the HV indicator is sensitive to the parameterizations of the objective space. In order to build a more complete picture about the performance of a multi-objective optimization algorithm, we consider tracking the HV under a number of different parameterizations. 
\begin{figure}
	\includegraphics[width=0.25\linewidth]{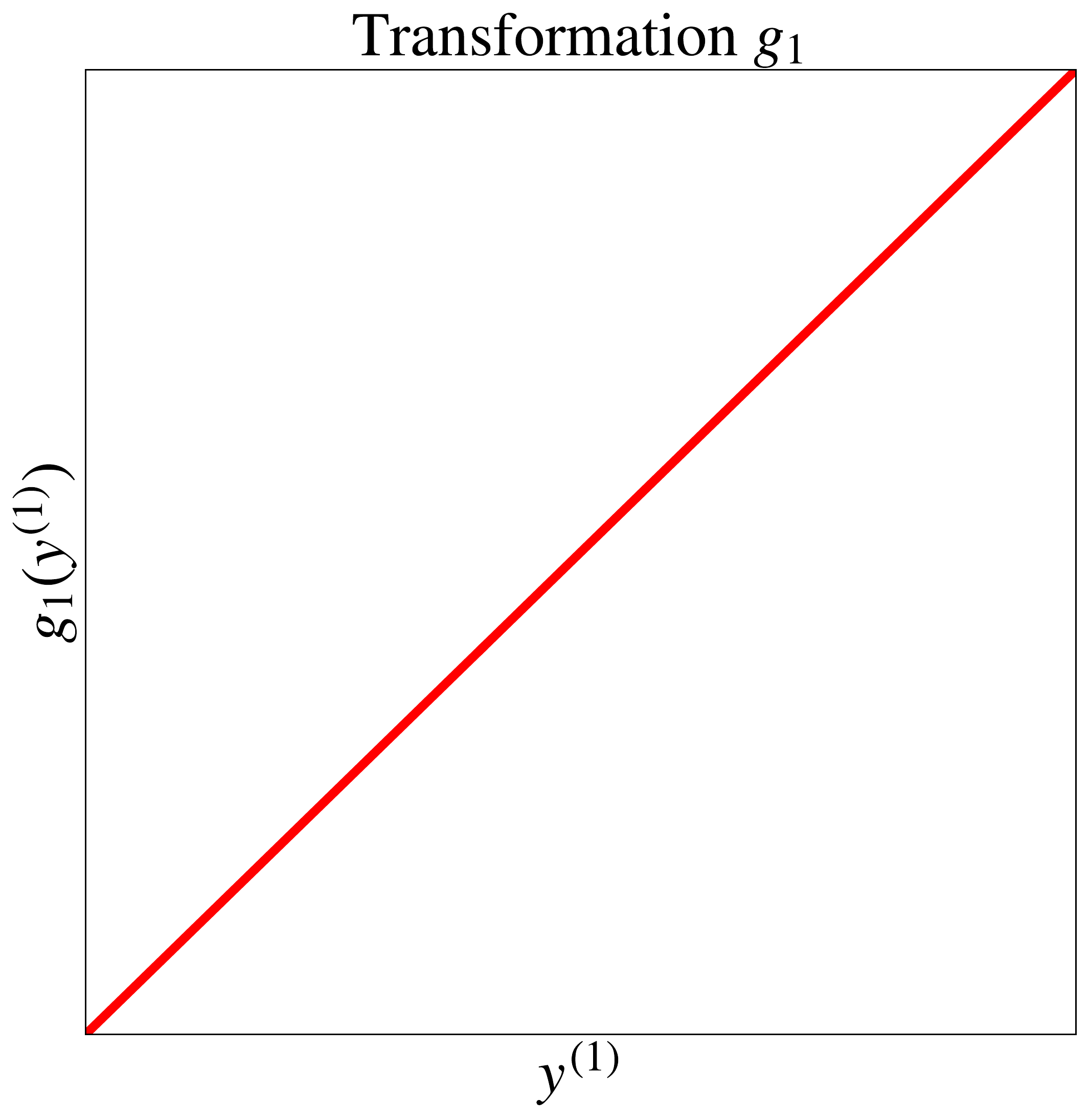}
	\includegraphics[width=0.25\linewidth]{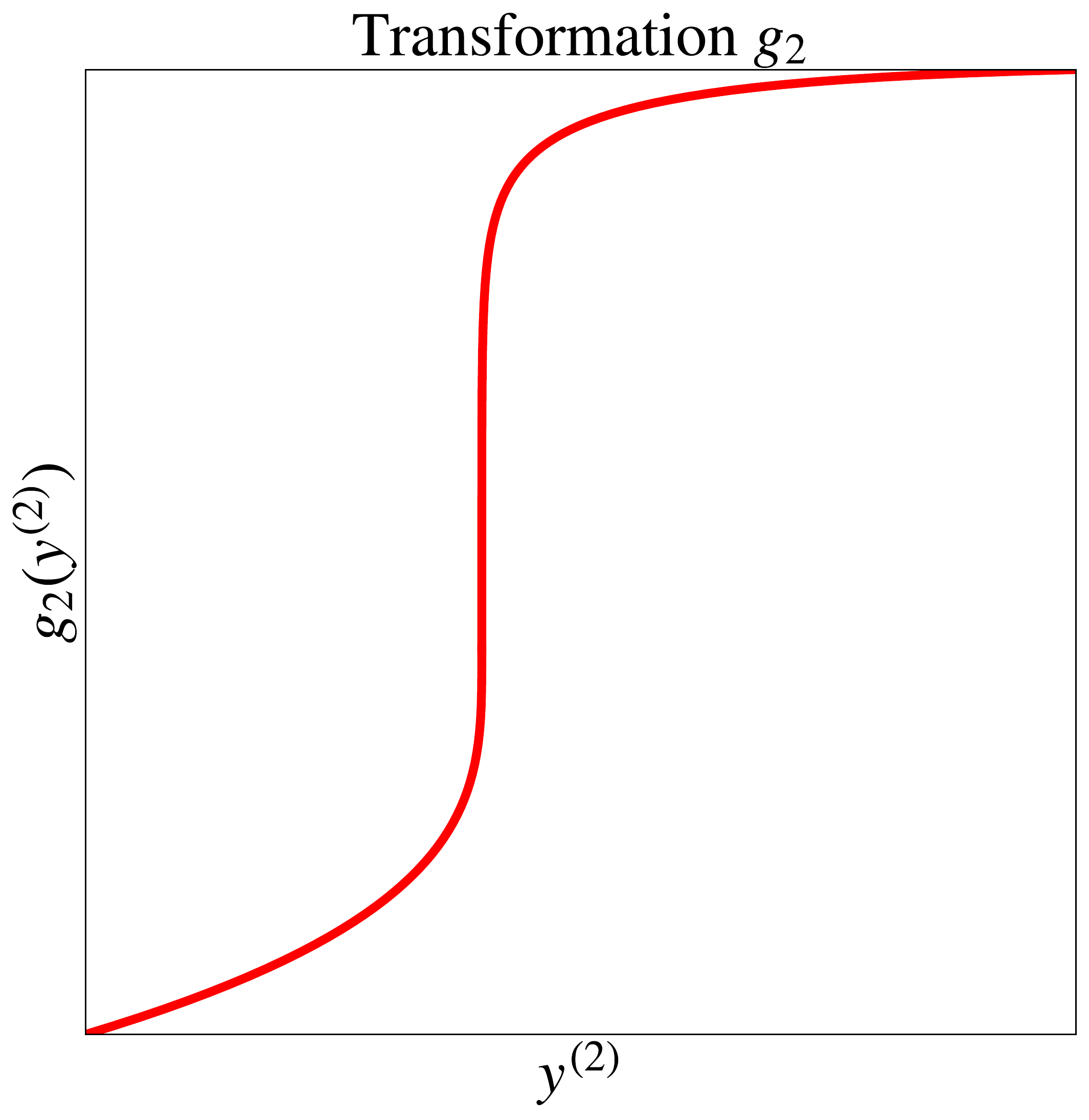}
	\centering
	\caption{The transformation functions $g_1$ and $g_2$ used in \cref{fig:hv}.}
	\label{fig:transform}
\end{figure}
Instead of deriving a family of transformation function, we consider a more general approach based on a different formulation of the HV indicator described in \cite{deng2019itec, zhang2020icml}. In particular, the HV indicator can be written as an expectation over a probability distribution by performing a change of variables into spherical polar coordinates:
\begin{equation}
	U_{\text{HV}}(A) = \frac{\pi^{M/2}}{2^M \Gamma(M/2 + 1)}
	\mathbb{E}_{p(\boldsymbol{\lambda})}\left[
	\max_{\mathbf{a} \in A} s_{\boldsymbol{\lambda}}(\mathbf{a})
	\right]
\end{equation}
where $\Gamma(\cdot)$ is the Gamma function and the scalarization function $s_{\boldsymbol{\lambda}}: \mathbb{R}^M \rightarrow \mathbb{R}$ is defined as
\begin{equation}
	s_{\boldsymbol{\lambda}}(\mathbf{a}) = \min_{m=1,\dots,M} \left(\max \left(0, (a^{(m)} - r^{(m)})/\lambda^{(m)} \right) \right)^M.
\end{equation}
For the standard HV, the inverse weight distribution $p(\boldsymbol{\lambda})$ is a uniform distribution over the surface of the $M$-dimensional unit sphere in the non-negative orthant $\mathcal{S}^{M-1}_+ = \{\boldsymbol{\lambda} \in \mathbb{R}^M_{\geq 0}: \sum_{m=1}^M (\lambda^{(m)})^2 = 1\}$. The inverse weight distribution $p(\boldsymbol{\lambda})$ controls the radial contribution for each point of $A$ towards the HV. Existing work mainly considers a uniform distribution over the weights in order to compute the standard HV. We make the novel observation that we can assess the quality of the Pareto front in different regions of the objective space by varying this weight distribution. We call the resulting utility function the generalized hypervolume (GHV) indicator, denoted by $U_{\text{GHV}}$.
\\ \\
This GHV satisfies a weaker version of the Pareto complete property: $A \succeq B \implies U_{\text{GHV}}(A) \geq U_{\text{GHV}}(B)$. This property can be proved from the fact that the scalarization function $s_{\boldsymbol{\lambda}}(\mathbf{a})$ is a monotonic increasing function in $\mathbf{a} \in A$, which implies that the Pareto order is maintained. This performance criteria satisfies the strict Pareto complete property when all of the weights have non-zero density $p(\boldsymbol{\lambda}) > 0$. Intuitively, if a set of weights had zero density, then an improvement along these direction will not always result in a larger GHV. This means that we will not always be able to distinguish between a set which strictly dominates another, which is essential for strict Pareto completeness.
\\ \\
To generate different distributions $p(\boldsymbol{\lambda})$, we exploit the fact that the set $\mathcal{S}^{M-1}_+$ is isomorphic to the $(M-1)$-dimensional unit hypercube $[0, 1]^{M-1}$. In particular, any point $\mathbf{w} \in [0, 1]^{M-1}$ can be mapped onto the sphere to a point $\boldsymbol{\lambda}(\mathbf{w}) \in \mathcal{S}^{M-1}_+$ with
\begin{align*}
	\lambda^{(1)} &= \cos \left( \frac{\pi}{2} w_1 \right)
	\\
	\lambda^{(2)} &= \sin \left( \frac{\pi}{2} w_1 \right) 
	\cos \left( \frac{\pi}{2} w_2 \right)
	\\
	\vdots
	\\
	\lambda^{(M-1)} &= \sin \left( \frac{\pi}{2} w_1 \right) \cdots \sin \left( \frac{\pi}{2} w_{M-2} \right)
	\cos \left( \frac{\pi}{2} w_{M-1} \right)
	\\
	\lambda^{(M)} &=  \sin \left( \frac{\pi}{2} w_1 \right) \cdots 
	\sin \left( \frac{\pi}{2} w_{M-2} \right) 
	\sin \left( \frac{\pi}{2} w_{M-1} \right).
\end{align*}
Consequently, we can use any distribution with a finite support to generate samples on $\mathcal{S}^{M-1}_+$. For example, we use $M-1$ independent Beta distribution Beta$(a^{(m)}, b^{(m)})$ to generate samples from $\mathbf{w} \in [0, 1]^{M-1}$. In \cref{fig:hv_beta}, we present the radial contributions to the GHV for different weight distributions $p(\mathbf{w})$.
\\ \\
To isolate performance in one particular radial region, we propose the use of a uniform distribution over some subset of $[0, 1]^{M-1}$. For example in \cref{fig:zdt2_example}, we chose three different subsets of $[0, 1]$ to isolate three different regions of the two-dimensional objective space.
\paragraph{Remark.} We are not the first to observe the importance of weighting the HV towards regions of interest. For example, an early paper \cite{zitzler2007emo} proposed another a weighted hypervolume indicator (WHV), $U_{\text{WHV}}(A) = \int_{\mathbb{R}^M} w(\mathbf{z}) \mathbb{I}[\mathbf{r} \preceq \mathbf{z} \preceq A] d\mathbf{z}$, which introduces a weight function $w: \mathbb{R}^M \rightarrow \mathbb{R}$ directly into the integral. This approach is not as flexible as the one presented here because it relied on designing hand-crafted weight functions and an involved symbolic integration procedure.
\FloatBarrier

\section{Experiments}
\label{app:experiments}
\subsection{Implementation details}
\label{app:implementation}
All of the numerical results was implemented on Python 3.8 using open-source libraries: BoTorch (0.5.1) \cite{balandat2020anips}, GPyTorch (1.6.0) \cite{gardner2018anips}, NumPy (1.21.2), Pymoo (0.5.0) \cite{blank2020ia} \cite{harris2020n}, PyTorch (1.9.0) \cite{paszke2019anips} and SciPy (1.7.3) \cite{virtanen2020nm}. All computations were performed on a computing cluster, where we restricted the computation to a single CPU core of an AMD EPYC 7742 64-Core Processor @ 2.25GHz. The code is available at \href{https://github.com/benmltu/JES}{https://github.com/benmltu/JES}.

\paragraph{Gaussian process model.} For all the problems, we normalized the inputs and standardized the observations before performing Gaussian process regression. We assume an independent model for each objective with a constant mean and Matérn 5/2 ARD kernel. Additionally, we assume a zero-mean Gaussian observation noise with a homogeneous variance $\boldsymbol{\sigma}(\mathbf{x}) = \boldsymbol{\sigma} \in \mathbb{R}^M$. For convenience, we rely on the default model in BoTorch, which additionally places Gamma priors for the kernel hyperparameters and observation variance. At each iteration of BO, we update the point estimates for the model hyperparameters and observation variance by maximizing the exact marginal log-likelihood using SciPy's default gradient optimizer. 

\paragraph{Optimizing the acquisition function.} To determine the next point, we optimize the acquisition function using multi-start L-BFGS-B \cite{virtanen2020nm}. We use the exact gradient inferred using automatic differentiation \cite{paszke2019anips} for all algorithms except for the PES. The automatic gradients inferred for the PES acquisition function occasionally failed due to the differentiability issues arising from the damping procedure within the expectation propagation update \cite{garrido-merchan2019n}. As a result, we used the default 2-point finite difference method in SciPy to approximate the gradients of the PES acquisition function. To initialize the multi-start gradient optimizer, we evaluated the acquisition function on a space-filling design of $1000D$ random points. The starting points were then chosen as the best performing points from the initial check. The number of starting points that we used differed for each algorithm based on the overall expenses. For the sequential experiments, we used $20$ starting points for PES, $5D$ starting points for the MES and JES, and $10D$ starting points for the rest.  For the batch experiments, we used $20$ starting points for PES and $5D$ starting points for the rest. For the information-theoretic acquisition functions, we could have initialized the starts near the sampled Pareto optimal points in order to promote faster convergence. We abstained from doing this because we wanted to give a fair comparison between the existing approaches. Specifically, we wanted to see how well information-based acquisition function worked out of the box without any optimization compared to the existing optimized approaches.

\paragraph{Sampling the Pareto points.} 
To sample the Pareto optimal inputs and outputs, we first sample Gaussian process paths. We achieved this by using an approximate sampling strategy based on random Fourier feaures (\cref{app:gp_sampling}). We used the implementation in BoTorch and set the number of features to $L=500$. To solve for the Pareto set and front, we used the multi-objective solver NSGA2 \cite{deb2002itec}, which was implemented in Pymoo. We set the population size to be $N_{\text{pop}} = 100$, the number of generations to be $N_{\text{gen}} = 500$ and the number of offspring to be $N_{\text{off}} = 10$. In general, the multi-objective solver outputs an approximate set of Pareto optimal points with a size less than or equal to $N_{\text{pop}}$. To truncate this set into a size $p$, we used a HV truncation strategy. In particular, we greedily selected points based on their contribution to the sample HV generated by the sample $f_s(X_n)$---the reference point is set to the current estimate of the nadir minus some error $\hat{\mathbf{r}}_n - 0.1 |\hat{\mathbf{r}}_n|$, where $\hat{r}_n^{(m)} = \min_{t=1,\dots,n} y_t^{(m)}$. This refinement strategy can be implemented quickly using the expected HV improvement strategy discussed in \cite{daulton2020anips}. We implemented this sampling and optimizing procedure in sequence, but naturally this can be executed in parallel because we are performing $S$ independent computations. As motivated by the wall times in \cref{app:wall_times}, implementing this step in parallel could be very beneficial computationally. 

\paragraph{Box decompositions.} We used the BoTorch implementation of the box decomposition strategy discussed in Algorithm 1 of \cite{lacour2017c&or}. The box decompositions were performed in sequence instead of being executed in parallel. From the wall times presented in \cref{app:wall_times}, the time required to compute the box decompositions becomes more demanding as the number of objectives increases.

\paragraph{Conditioning.} We used the fantasizing feature in BoTorch to condition the current posterior on a collection of $S$ independent Pareto samples of size $p$. We treated the Pareto samples as noisy pseudo-observations in the conditioning.

\paragraph{Acquisition functions.}
The benchmark comparison considers a range of popular acquisition functions that have all been implemented in BoTorch. We implemented our own version of the multi-objective PES, MES and JES. We now elaborate on the implementation details for each acquisition functions in the benchmark experiments.

\begin{itemize}[leftmargin=*]
	\item \textbf{TSEMO.} The Thomson sampling algorithm (TSEMO) \cite{bradford2018jgo} is a random acquisition function, which selects the point that maximizes the HV improvement according to a single sample of the Pareto front. Unlike the original paper \cite{bradford2018jgo}, we select the point that improves the HV of the sample frontier $f_s(X_n)$ as opposed to the observation frontier generated by $Y_n$. We find this adjustment to be sensible when there is observation noise. We use the same modification for the batch extension.
	
	\item \textbf{ParEGO/ NParEGO.} The random scalarization strategy (ParEGO) \cite{knowles2006itec} and its noisy counterpart (NParEGO) \cite{daulton2021anips} are one of the most popular strategies for multi-objective BO. At each iteration, a random scalarization of the objectives is obtained by randomly drawing a weight $\mathbf{w} \in \mathbb{R}^M$ from the $(M-1)$-simplex. To target all the Pareto optimal points, we use the augmented Chebyshev scalarization: $\min_{m=1,\dots,M} w^{(m)} \tilde{f}^{(m)} + 0.01 \sum_{m=1} w^{(m)} \tilde{f}^{(m)}$. Here we have denoted $\tilde{f} \propto f$ as the objective function, which has been approximately normalized to $[0, 1]^M$ using the observations $Y_n$. For the single-objective problem, a Monte Carlo estimate of the expected improvement and the noisy expected improvement for the ParEGO and NParEGO algorithm are used respectively. The number of base samples for the Monte Carlo estimates is set to $128$. For the batch setting, we sampled $q$ different weights and optimized the acquisition function sequentially.
	
	\item \textbf{EHVI/ NEHVI.} The expected hypervolume improvement (EHVI) \cite{daulton2020anips} and its noisy counterpart (NEHVI) \cite{daulton2021anips} are an improvement-based acquisition function for the HV indicator. The number of base samples for the Monte Carlo estimates is set to $128$. We set the reference point of the HV indicator to be equal to the observed nadir minus some error, $\hat{\mathbf{r}}_n - 0.1 |\hat{\mathbf{r}}_n|$, where $\hat{r}_n^{(m)} = \min_{t=1,\dots,n} y_t^{(m)}$. This dynamic reference point strategy was recommended in the supplementary material of \cite{daulton2020anips}. For the batch setting, we considered a sequentially greedy optimization strategy.
	
	\item \textbf{PES.} The predictive entropy search (PES) \cite{hernandez-lobato2014anips, hernandez-lobato2016icml, garrido-merchan2019n, garrido-merchan2021a} acquisition function is approximated using expectation propagation. We implemented this algorithm from scratch in BoTorch under the guidance of the  supplementary material in \cite{garrido-merchan2019n}. We used $S=10$ Monte Carlo samples and $p=10$ number of Pareto optimal inputs. We set the jitter for the matrix inversion to be $0.001$ and the convergence threshold for the initialization stage to be $5\%$ relative change in the mean and covariance. If the expectation propagation failed to converge, we outputted a random vector from the already sampled Pareto sets as the official recommendation. For the batch setting, we extended the approach outlined in \cite{garrido-merchan2019n}. This extension appears to be equivalent to the approach described in \cite{garrido-merchan2021a}. We optimized the resulting acquisition function using a joint optimization approach.
	
	\item \textbf{MES.} The maximum value entropy search (MES) \cite{wang2017icml, suzuki2020icml} can be approximated using all the conditional entropy estimates we devised in this paper. We used $S=10$ Monte Carlo samples and $p=10$ number of Pareto optimal outputs. For the Monte Carlo estimate MES-MC we set the number of base samples to 128. For the batch setting, we considered a greedy optimization strategy \cite{daulton2020anips, daulton2021anips}. For the batch setting, we consider the lower bound described by \eqref{eqn:qjes_base}. We optimized the resulting acquisition function using a greedy optimization approach.
		
	\item \textbf{JES.} The joint entropy search (JES) \cite{wang2017icml, suzuki2020icml} can be approximated using all the conditional entropy estimates we devised in this paper. We used $S=10$ Monte Carlo samples and $p=10$ number of Pareto optimal points. For the Monte Carlo estimate JES-MC we set the number of base samples to 128. For the batch setting, we consider the lower bound described by \eqref{eqn:qjes_base}. We optimized the resulting acquisition function using a greedy optimization approach.
\end{itemize}

\paragraph{Optimizing for the recommendation.} To obtain the recommendation set, $\hat{\mathbb{X}}^*_n$, we used the NSGA2 multi-objective solver to optimize the posterior mean $\boldsymbol{\mu}_n$. Using the Pymoo implementation, we set the population size to be $N_{\text{pop}} = 500$, the number of generations to be $N_{\text{gen}} = 500$ and the number of offspring to be $N_{\text{off}} = 10$. We select the $p=50$ points that greedily maximizes the HV generated by the mean objectives. The reference point for the HV truncation was set to the current estimate of the nadir minus some error: $\hat{\mathbf{r}}_n - 0.1 |\hat{\mathbf{r}}_n|$, where $\hat{r}_n^{(m)} = \min_{t=1,\dots,n} y_t^{(m)}$. Before applying the truncation, we first augmented the Pareto set found at this iteration with the previous recommendation set $\hat{\mathbb{X}}^*_{n-1}$ in order to guarantee that some promising solutions weren't missed due to randomness of the solver.

\paragraph{Optimizing for the Pareto set.} To obtain the baseline Pareto set, $\mathbb{X}^*$, we used the Pymoo's NSGA2 with a population size to be $N_{\text{pop}} = 1000$, the number of generations to be $N_{\text{gen}} = 5000$ and the number of offspring to be $N_{\text{off}} = 10$. 
\subsection{Benchmark problems}
\label{app:benchmarks}
We initialized each test problem with with $2(D+1)$ training points using a random space filling design. The details of the individual benchmark problems is presented below.

\paragraph{ZDT2 (D=2, M=2, Noise=10\%).}
The ZDT2 benchmark \cite{zitzler2000ec} is a bi-objective minimization problem $\min_{\mathbf{x} \in \mathbb{X}} f(\mathbf{x})$, over the $D$-dimensional hypercube $\mathbb{X} = [0, 1]^D$, where
\begin{align*}
	f^{(1)}(\mathbf{x}) &= x^{(1)}
	\\
	f^{(2)}(\mathbf{x}) &= g(\mathbf{x}) \left(1 - \left(\frac{f^{(1)}(\mathbf{x})}{g(\mathbf{x})}\right)^2 \right)
\end{align*}
with $g(\mathbf{x}) = 1 - \frac{9}{D - 1} \sum_{i=2}^D x^{(i)}$. For the experiments, we considered the maximization problem by negating the objective: $\max_{\mathbf{x} \in \mathbb{X}} (- f(\mathbf{x}))$. The standard deviation of the Gaussian observation noise is set to $\boldsymbol{\sigma}^{1/2} = (0.1, 0.8)$, which is estimated to be around $10\%$ the range of the objectives from $1000$ function evaluations. For the HV indicators we use the reference point of $\mathbf{r} = (-11, -11)$. 

\paragraph{SnAr (D=4, M=2, Noise=3\%).}
The SnAr benchmark \cite{felton2021c} is a bi-objective minimization problem $\min_{\mathbf{x} \in \mathbb{X}} f(\mathbf{x})$, which is defined over a four-dimensional space, $\mathbb{X} = [0.5, 2.0] \times [1.0, 5.0] \times [0.1, 0.5] \times [30, 120]$, where 
\begin{align*}
	f^{(1)}(\mathbf{x}) &= -\log(\texttt{STY}(\mathbf{x}))
	\\
	f^{(2)}(\mathbf{x}) &= \log(\texttt{E-Factor}(\mathbf{x}))
\end{align*}
with \texttt{STY} and \texttt{E-Factor} denoting the space-time yield and environmental factor, respectively. The functions \texttt{STY} and \texttt{E-Factor} depend on the solution of an ordinary differential equation governed by a kinetic model, where the rate constants have been estimated by empirical tests \cite{hone2017rce}. Unlike the original paper \cite{felton2021c}, we have taken the logarithm of the output to accommodate for the additive noise in the objective space. For the experiments, we considered the maximization problem by negating the objective: $\max_{\mathbf{x} \in \mathbb{X}} (- f(\mathbf{x}))$. The standard deviation of the Gaussian observation noise is set to $\boldsymbol{\sigma}^{1/2} = (0.12, 0.08)$, which is estimated to be around $3\%$ the range of the objectives from $1000$ function evaluations. For the HV indicators we use the reference point of $\mathbf{r} = (5.5, -5)$.
\FloatBarrier

\paragraph{Penicillin (D=7, M=3, Noise=1\%).}
The Penicillin benchmark \cite{liang2021n2asw} is a three objective minimization problem $\min_{\mathbf{x} \in \mathbb{X}} f(\mathbf{x})$, which is defined over a seven-dimensional space, $\mathbb{X} = [60, 120] \times [0.05, 18] \times [293, 303] \times [0.05, 18] \times [0.01, 0.5] \times [500, 700] \times [5, 6.5]$, where 
\begin{align*}
	f^{(1)}(\mathbf{x}) &= -\texttt{PenicillinConcentration}(\mathbf{x})
	\\
	f^{(2)}(\mathbf{x}) &=  \texttt{CO2Concentration}(\mathbf{x}))
	\\
	f^{(3)}(\mathbf{x}) &= \texttt{TimeToFerment}(\mathbf{x})
\end{align*}
with \texttt{PenicillinConcentration}, \texttt{CO2Concentration} and \texttt{TimeToFerment} denoting the concentration of the desirable product, the concentration of the subproduct and the time to ferment, respectively. The equations describing the pharmaceutical simulation are given in the original reference \cite{liang2021n2asw}. We used the open-source implementation available in BoTorch \cite{balandat2020anips, daulton2022icml}. For the experiments, we considered the maximization problem by negating the objective: $\max_{\mathbf{x} \in \mathbb{X}} (- f(\mathbf{x}))$. The standard deviation of the Gaussian observation noise is set to $\boldsymbol{\sigma}^{1/2} = (0.14, 0.8, 3.8)$, which is estimated to be around $1\%$ the range of the objectives from $1000$ function evaluations. For the HV indicators we use the reference point of $\mathbf{r} = (-1.85, -86.93, -514.70)$.
\FloatBarrier

\paragraph{Marine Design (D=6, M=4, Noise=0.5\%).}
This marine design benchmark \cite{sen1998, tanabe2020asc, parsons2004josr} is a four objective minimization problem $\min_{\mathbf{x} \in \mathbb{X}} f(\mathbf{x})$, which is defined over a six-dimensional space, $\mathbb{X} = [150, 274.32] \times [20, 32.31] \times [13, 25] \times [10, 11.71] \times [14, 18] \times [0.63, 0.75]$, where 
\begin{align*}
	f^{(1)}(\mathbf{x}) &= \texttt{TransportationCost}(\mathbf{x})
	\\
	f^{(2)}(\mathbf{x}) &= \texttt{Weight}(\mathbf{x})
	\\
	f^{(3)}(\mathbf{x}) &= - \texttt{AnnualCargo}(\mathbf{x})
	\\
	f^{(3)}(\mathbf{x}) &= \texttt{SumOfConstraints}(\mathbf{x})
\end{align*}
with \texttt{TransportationCost}, \texttt{Weight}, \texttt{AnnualCargo} and \texttt{SumOfConstraints} denoting the transportation cost, the ship weight, the annual cargo transport capacity and the sum of the constraint violations, respectively. The formal equations describing the functions are presented in the original references \cite{sen1998,tanabe2020asc, parsons2004josr}. For the experiments, we considered the maximization problem by negating the objective: $\max_{\mathbf{x} \in \mathbb{X}} (- f(\mathbf{x}))$. The standard deviation of the Gaussian observation noise is set to $\boldsymbol{\sigma}^{1/2} = (10, 77, 132, 0.07)$, which is estimated to be around $0.5 \%$ the range of the objectives from $1000$ function evaluations. For the HV indicators we use the reference point of $\mathbf{r} = (250, -20000, -25000, -15.0)$.
\FloatBarrier
\subsection{Sensitivity analysis}
\label{app:sensitivity_analysis}
\begin{figure}[!htb]
	\includegraphics[width=1\linewidth]{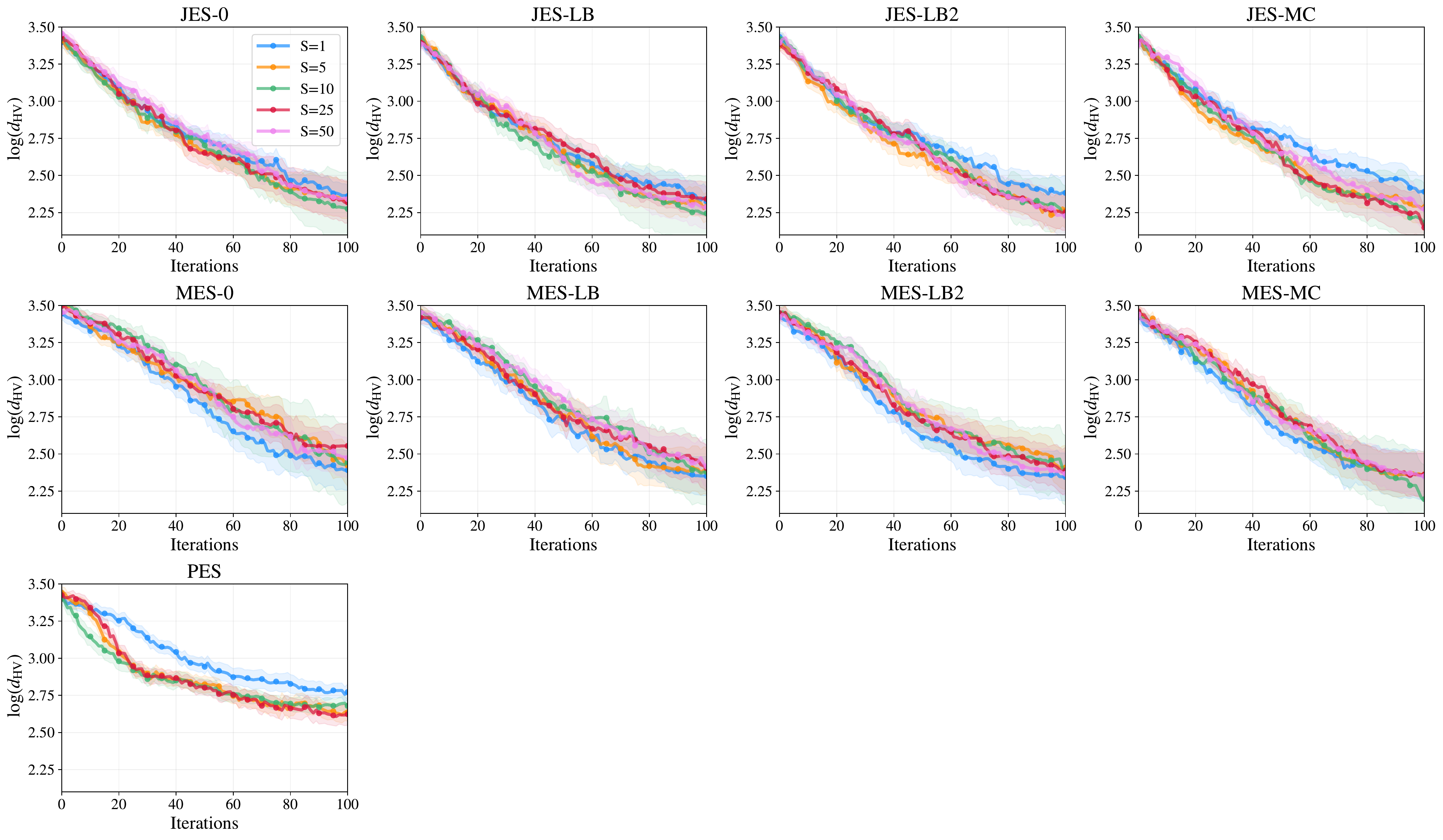}
	\centering
	\caption{A comparison of the mean logarithm HV discrepancy with two standard errors over different number of Monte Carlo samples $S$. }
	\label{fig:zdt2_sens_mc_hv}
\end{figure}

\begin{figure}[!htb]
	\includegraphics[width=1\linewidth]{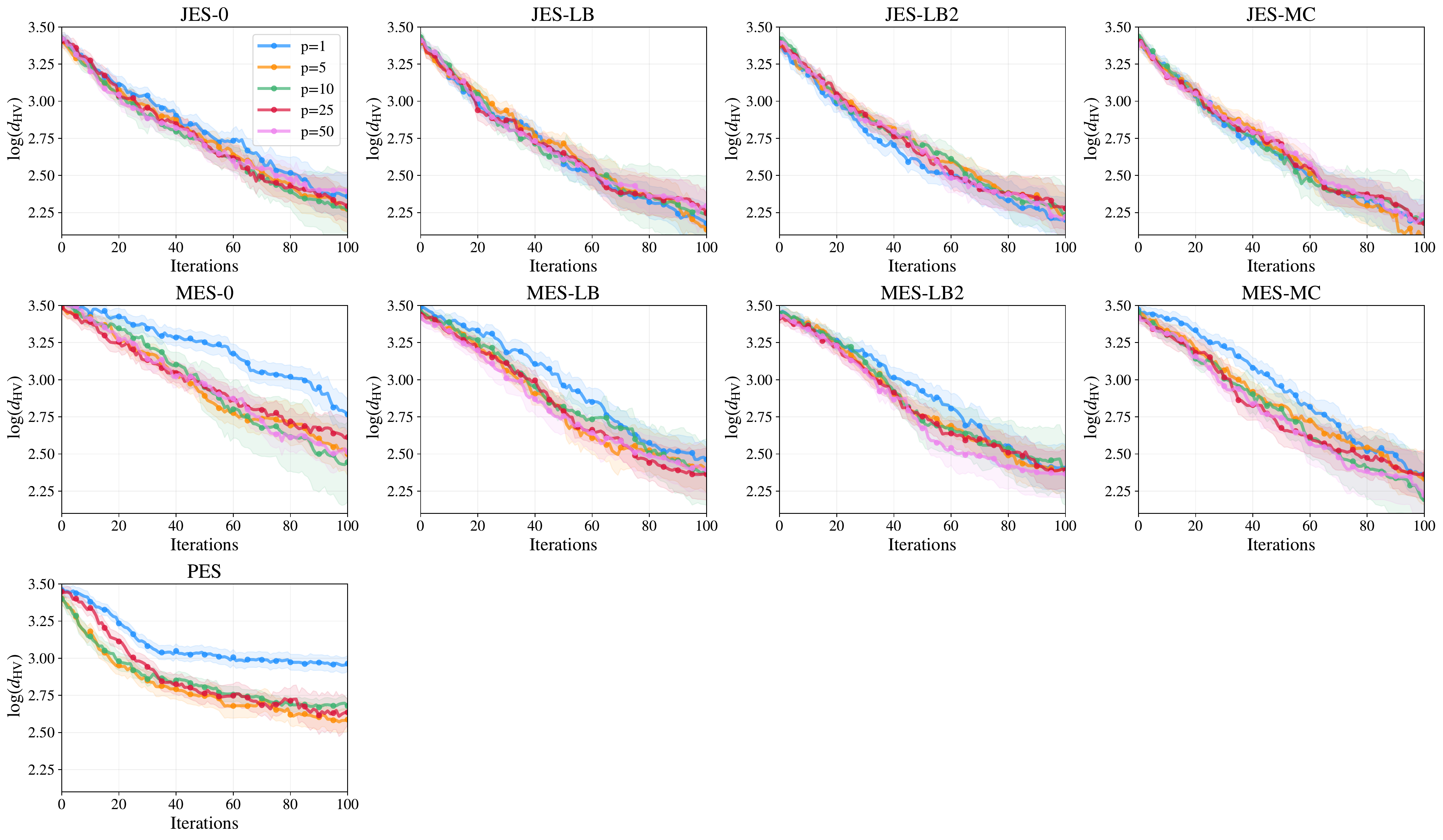}
	\centering
	\caption{A comparison of the mean logarithm HV discrepancy with two standard errors over different number of Pareto optimal samples $p$.}
	\label{fig:zdt2_sens_p_hv}
\end{figure}

\begin{figure}[!htb]
	\includegraphics[width=0.495\linewidth]{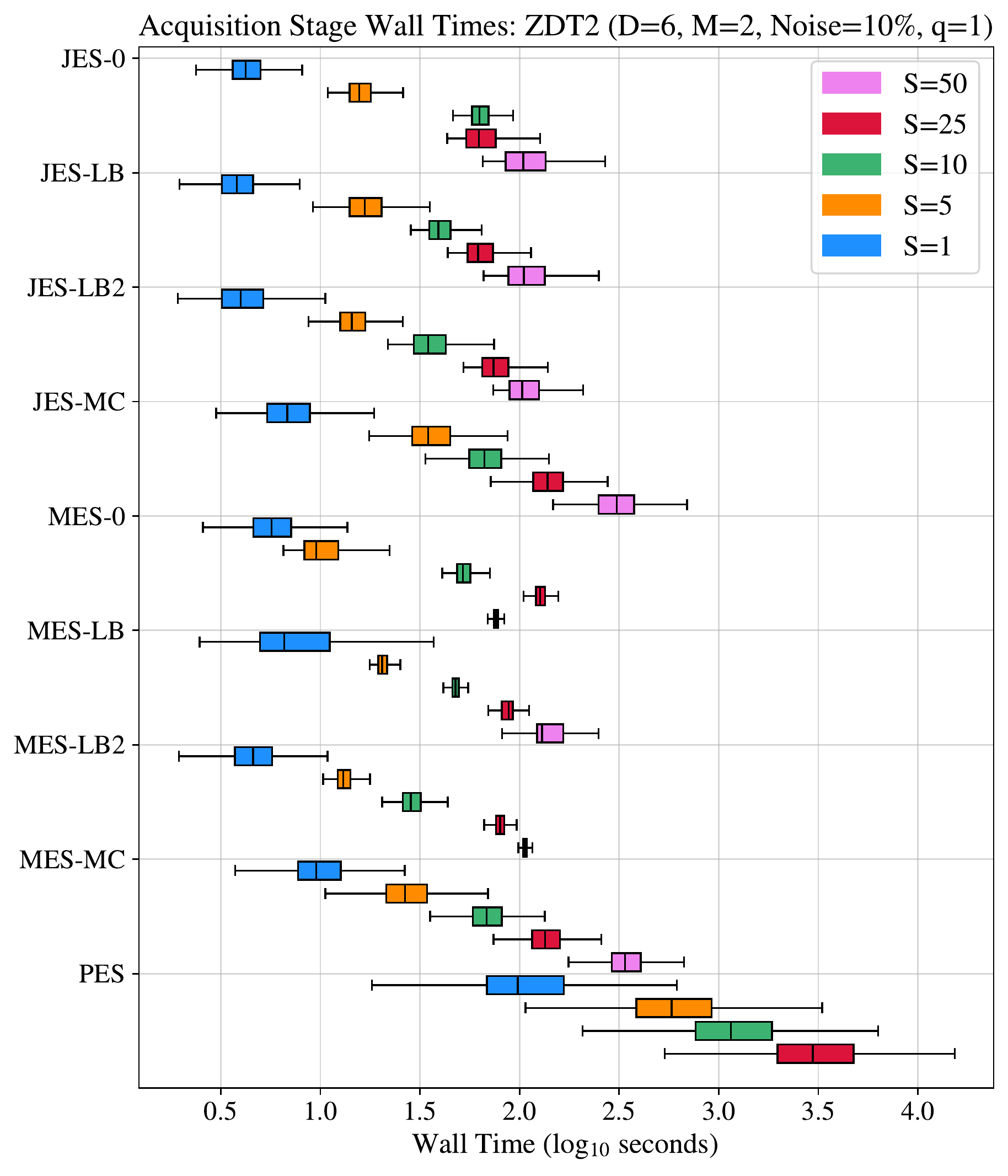}
	\includegraphics[width=0.495\linewidth]{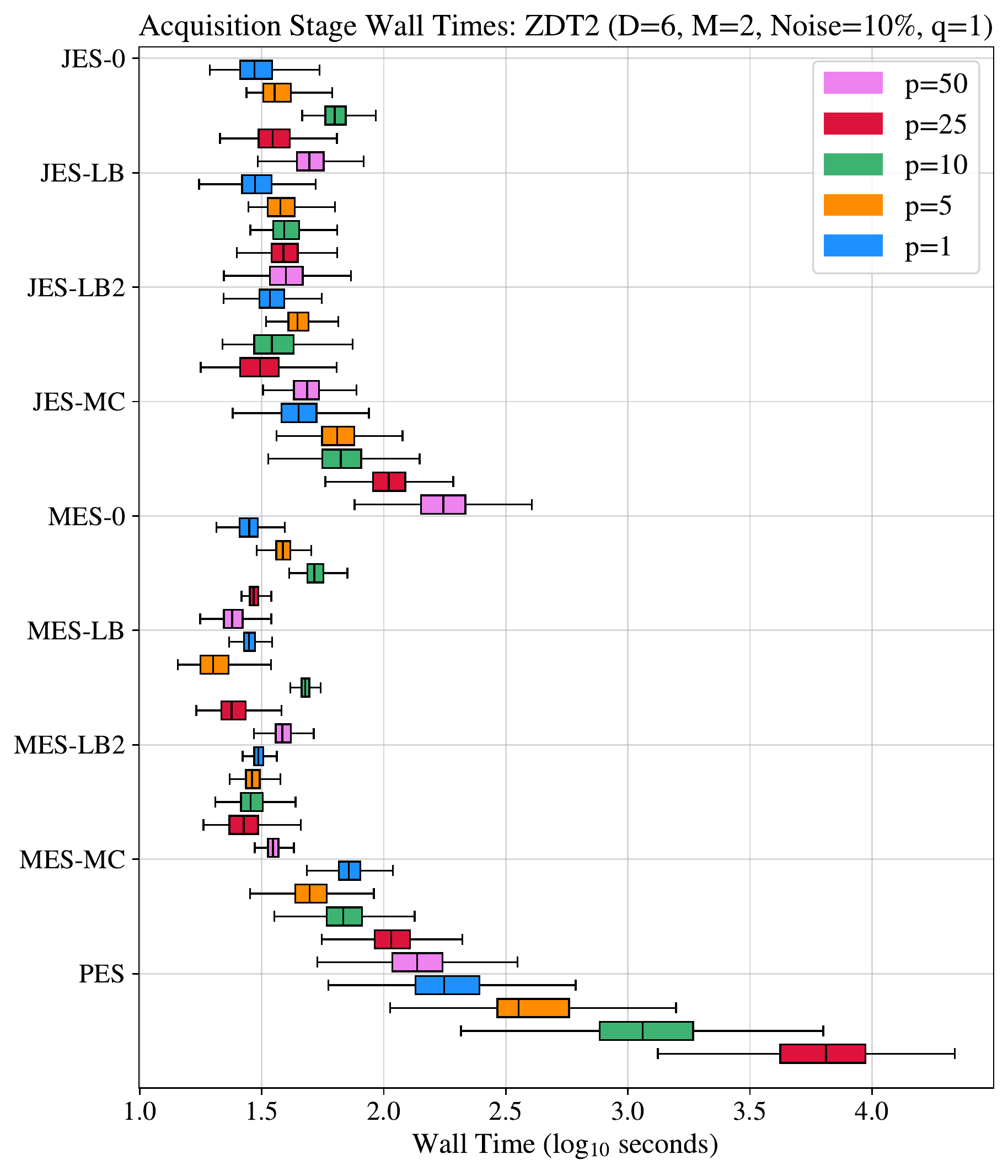}
	\centering
	\caption{A box-plot comparison of the wall times for the acquisition stage over different number of Monte Carlo samples (left) and the number of Pareto optimal samples (right). The box-plots include the median, interquartile range and the extreme values after excluding the outliers. The acquisition stage includes any initialization computations such as the box decompositions and sampling the Pareto optimal points---it does not include initializing the posterior model. All of the runs of each algorithm was performed on a computing cluster, where we restricted the computation to a single CPU core of an AMD EPYC 7742 64-Core Processor @ 2.25GHz.}
	\label{fig:zdt2_sens_walltime}
\end{figure}

In the section we empirically analyse how sensitive the approximations to the information-theoretic acquisition functions are with different choices of Monte Carlo samples $S$ and number of Pareto optimal points $p$. For the experiments in the main section, we set $S=10$ and $p=10$ for all information-theoretic acquisition functions. This selection is comparable to the existing literature \cite{wang2017icml, belakaria2019anips, suzuki2020icml, garrido-merchan2019n}. For the sake of brevity we present results only for one benchmark problem: ZDT2 (D=2, M=2, Noise=10\%, q=1). 
\\ \\
To test the sensitivity with regards to the number of Monte Carlo samples $S$, we fix $p=10$ and ran the benchmark problem $100$ times with $S \in \{1, 5, 10, 25, 50\}$. To test the sensitivity with regards to the number of Pareto optimal points $p$, we fix $S=10$ and ran the benchmark problem $100$ times with $p \in \{1, 5, 10, 25, 50\}$.  We report the mean log HV discrepancy with two standard errors over the runs in \cref{fig:zdt2_sens_mc_hv} and \cref{fig:zdt2_sens_p_hv}. We report the wall times in \cref{fig:zdt2_sens_walltime}. 
\\ \\
Performance-wise there does not appear to be much variation for acquisition function when $S>1$ and $p>1$. Naturally the wall times increase with the number of Monte Carlo samples $S$ because the sampling and optimization of the Gaussian process paths are done in sequence (this step could be parallelized in practice). Overall, for this example problem, there does not appear to be much benefit in using a larger number of Monte Carlo samples $S$ or Pareto optimal points $p$.

\FloatBarrier

\subsection{Noise levels}
\begin{figure}[!htb]
	\includegraphics[width=1\linewidth]{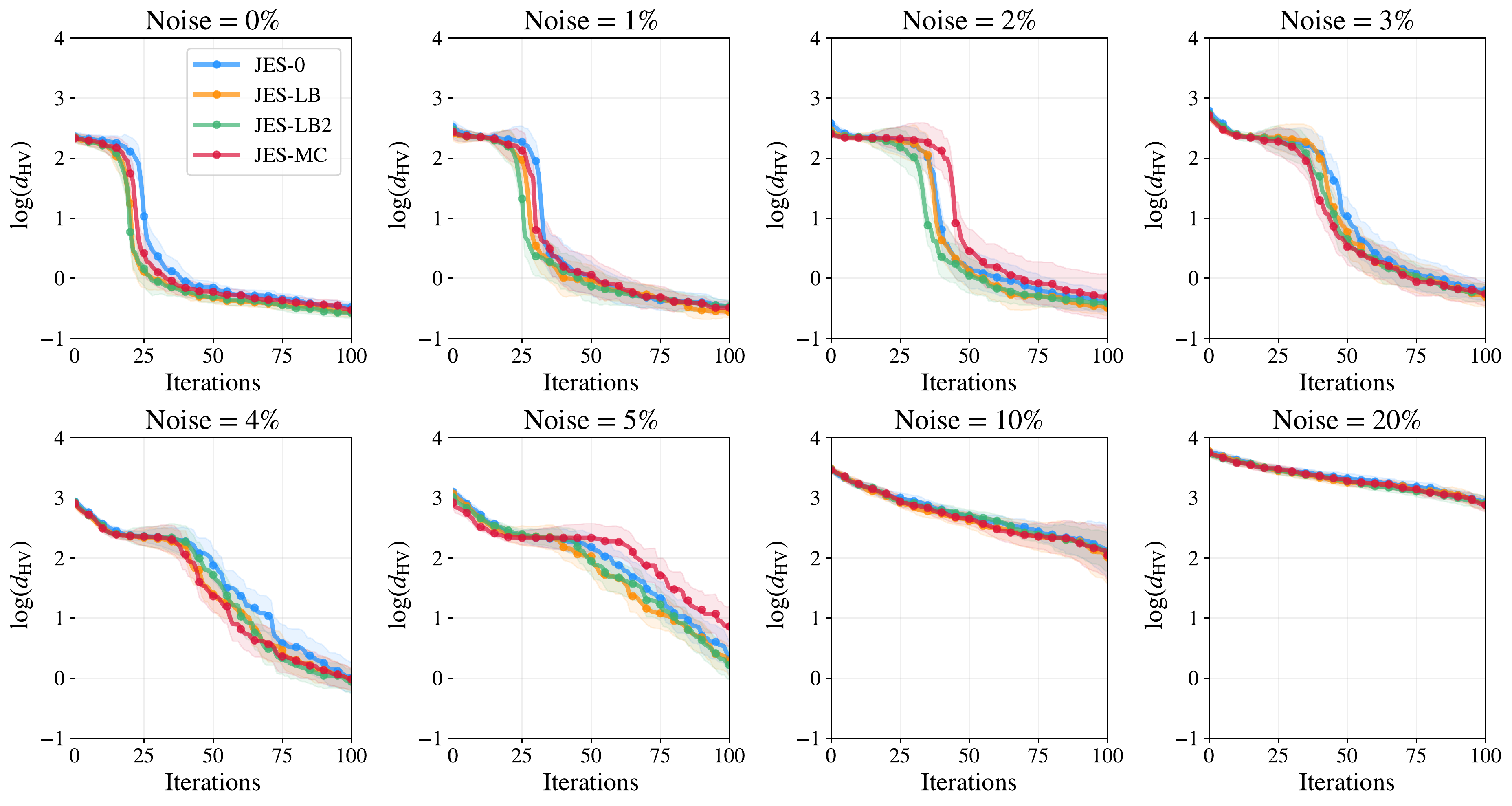}
	\centering
	\caption{A comparison of the mean logarithm HV discrepancy with two standard errors for different noise levels. The recommended set of points were obtained by maximizing the posterior mean.}
	\label{fig:zdt2_noise_hv_jes}
\end{figure}

\begin{figure}[!htb]
	\includegraphics[width=1\linewidth]{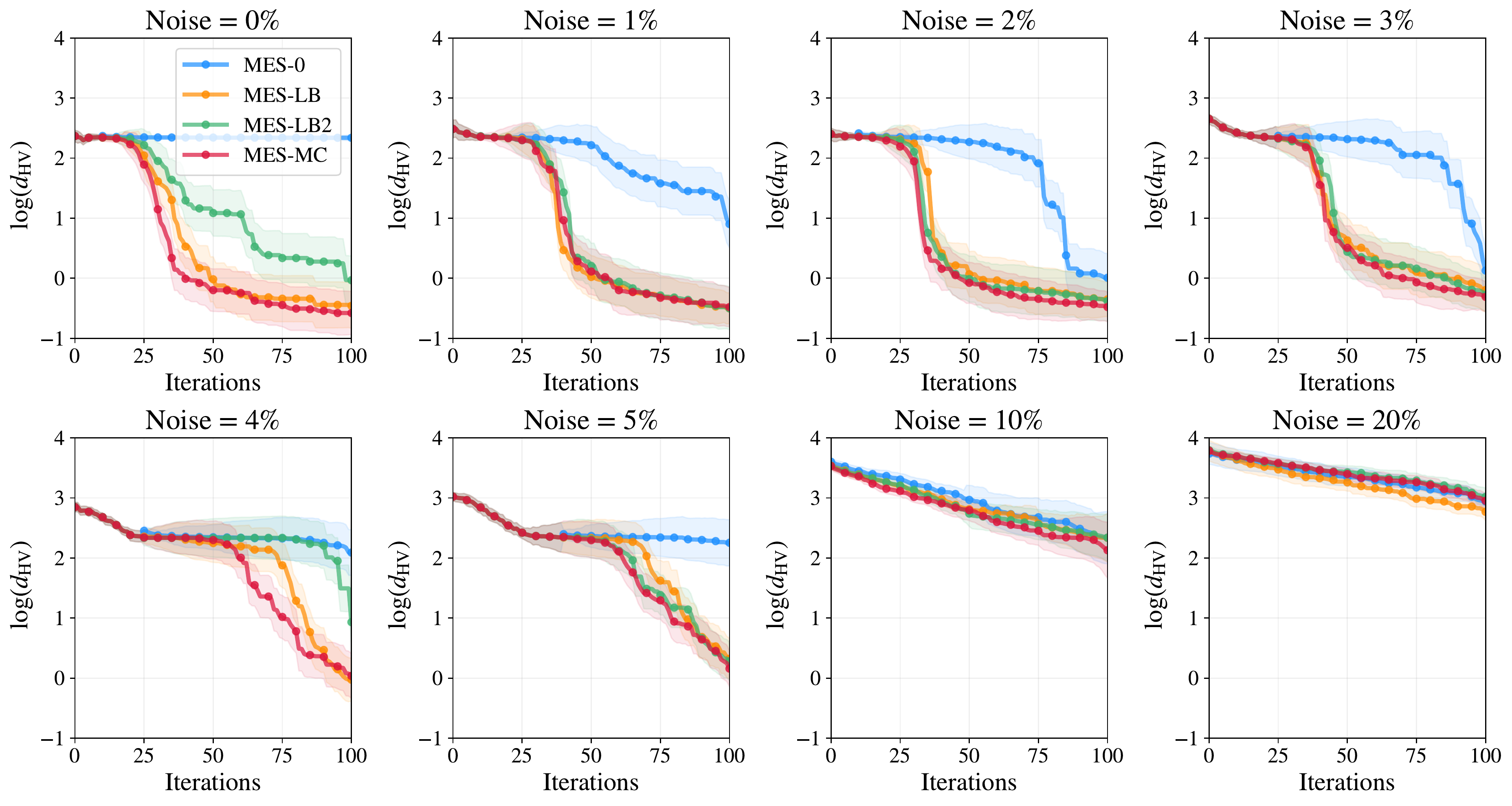}
	\centering
	\caption{A comparison of the mean logarithm HV discrepancy with two standard errors for different noise levels. The recommended set of points were obtained by maximizing the posterior mean.}
	\label{fig:zdt2_noise_hv_mes}
\end{figure}

In the section we empirically analyse how sensitive the approximations to the information-theoretic acquisition functions are when we increase noise levels. In \cref{fig:zdt2_noise_hv_jes} and \cref{fig:zdt2_noise_hv_mes}, we plot the mean logarithm HV discrepancy for the JES and MES estimates on the ZDT2 (D=2, M=2, q=1) benchmark when the recommended points are obtained by maximizing the posterior mean. In both examples, we observe that the performance decreases as the noise levels increases. On the whole the JES estimates perform very similarly across the board. For the MES results, we see that the MES-0 estimate is noticeable weaker than the rest even after the ad hoc correction described in \cref{app:zero_variance}. 

\FloatBarrier
\subsection{In-sample results}
\label{app:in-sample}
\begin{figure}[!htb]
	\includegraphics[width=1\linewidth]{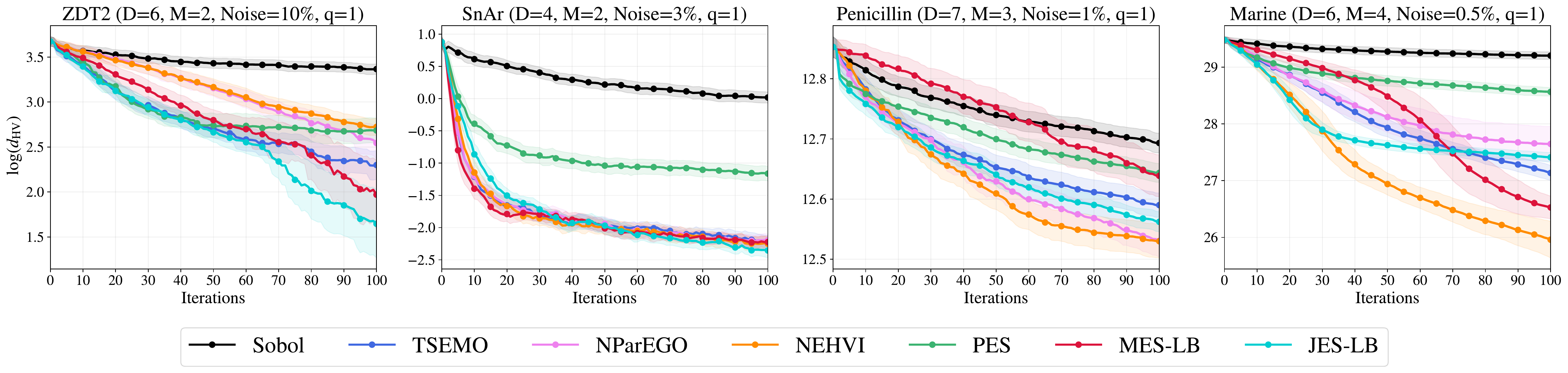}
	\includegraphics[width=1\linewidth]{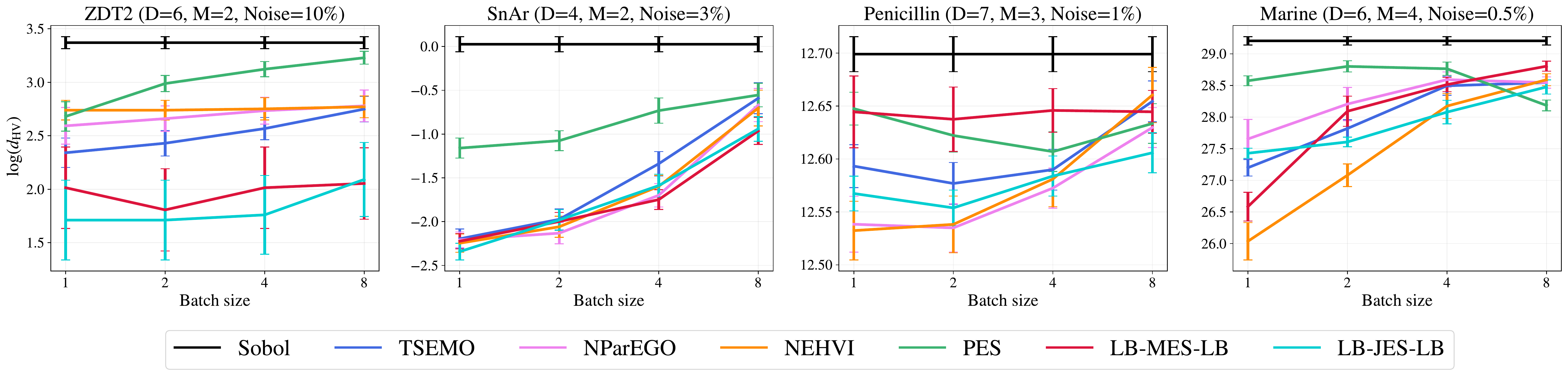}
	\centering
	\caption{A comparison of the mean logarithm HV discrepancy with two standard errors over one hundred runs for the four benchmark problems on a subset of the algorithms when we restrict the approximate Pareto set to the sampled points. On the top, we present the results for the sequential experiment, whilst on the bottom we present the final results on the batch experiments with different batch sizes.}
	\label{fig:experiments_sample_hv}
\end{figure}

In this section we present the results of the standard hypervolume when we restrict the recommended Pareto set $\hat{\mathbb{X}}^*$ to be a subset of the sampled locations: $\hat{\mathbb{X}}^* = \argmax_{\mathbf{x} \in X_N} \boldsymbol{\mu}_N(\mathbf{x})$. The complete results are presented in \cref{fig:experiments_sample_hv}. Our main finding is that information-theoretic strategies have a tendency to not directly query the best performing points but instead opt for more informative points that will reduce the overall model uncertainty over the optimal points. As a result of this behaviour, information-theoretic algorithms tend to perform well when we assess the Pareto set over the whole input space $\mathbb{X}$ and less so when we only assess the performance over the sampled locations $X_N$. If directly querying high-performing points is important, it might be advantageous to use an epsilon greedy strategy, where points are occasionally picked greedily according to the posterior mean. 
\FloatBarrier
\subsection{Comparing the entropy estimates}
\label{app:conditional_entropy_estimates}
\begin{figure}[!htb]
	\includegraphics[width=1\linewidth]{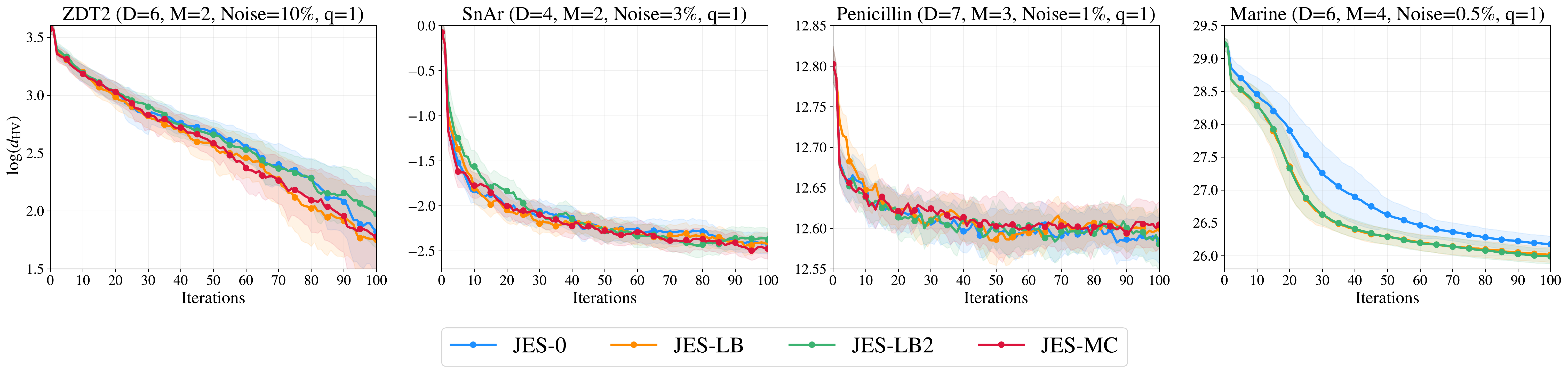}
	\includegraphics[width=1\linewidth]{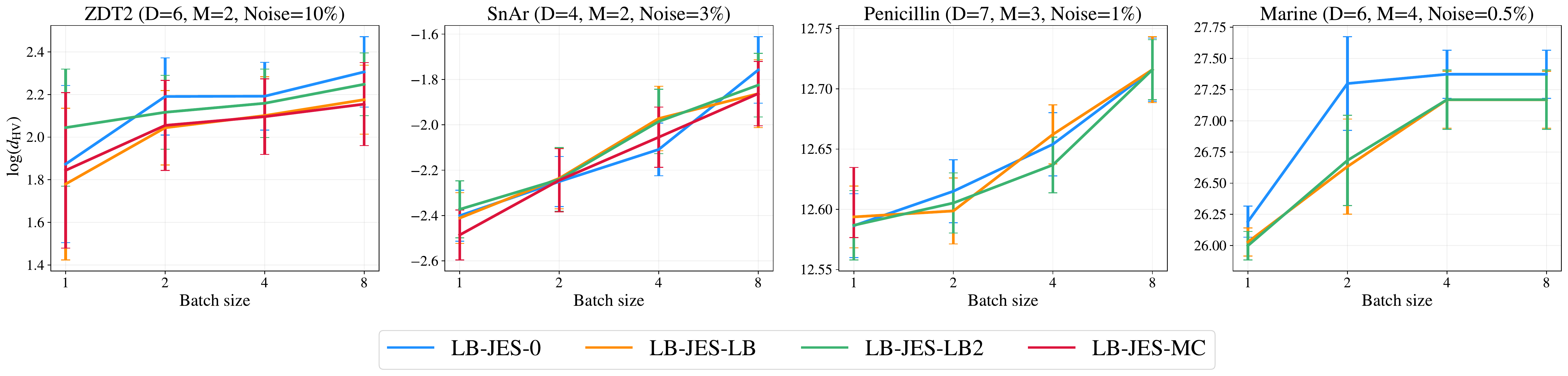}
	\centering
	\caption{A comparison of the mean logarithm HV discrepancy with two standard errors over one hundred runs for the four benchmark problems for the JES algorithms. On the top, we present the results for the sequential experiment, whilst on the bottom we present the final results on the batch experiments with different batch sizes.}
	\label{fig:experiments_hv_jes}
\end{figure}

\begin{figure}[!htb]
	\includegraphics[width=1\linewidth]{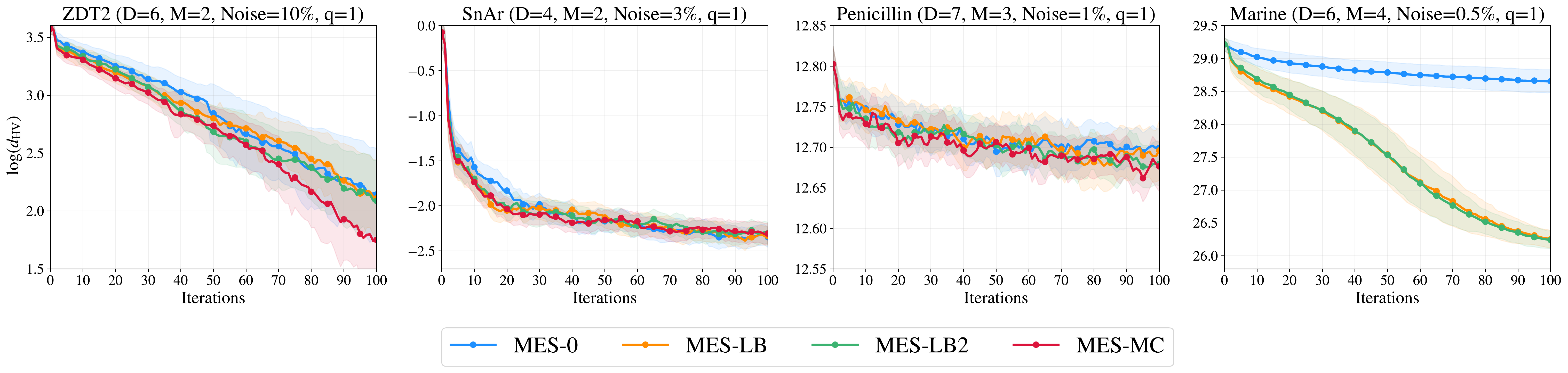}
	\includegraphics[width=1\linewidth]{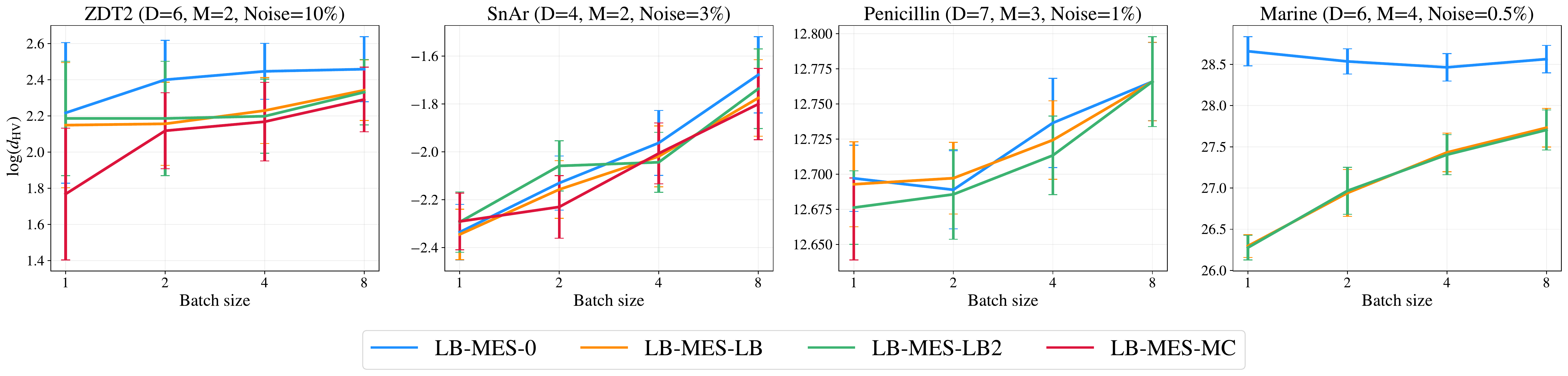}
	\centering
	\caption{A comparison of the mean logarithm HV discrepancy with two standard errors over one hundred runs for the four benchmark problems for the MES algorithms. On the top, we present the results for the sequential experiment, whilst on the bottom we present the final results on the batch experiments with different batch sizes.}
	\label{fig:experiments_hv_mes}
\end{figure}

In this section we compare the results using the different conditional entropy estimates for the different experiments. In \cref{fig:experiments_hv_jes} and \cref{fig:experiments_hv_mes} we present the results for the JES and MES estimates respectively. We observed that the different conditional entropy estimates seem to perform similarly across the board. As a result, we advocate the use of the cheapest estimate, which are typically the lower bound estimates. On the Marine experiment, we observed that the zero-variance estimate was noticeably weaker even after we applied the ad hoc correction described in \cref{app:zero_variance}. Therefore, we generally recommend against using the zero-variance estimate if possible.
\FloatBarrier
\subsection{Comparing the improvement-based algorithms}
In this section we present the results for the improvement-based algorithms: NParEGO, ParEGO, EHVI and NEHVI. On the whole, we observe that the greedier strategy which ignores the noise seems to perform reasonably well compared to the strategy which accounts for the noise. 
\begin{figure}[!htb]
	\includegraphics[width=1\linewidth]{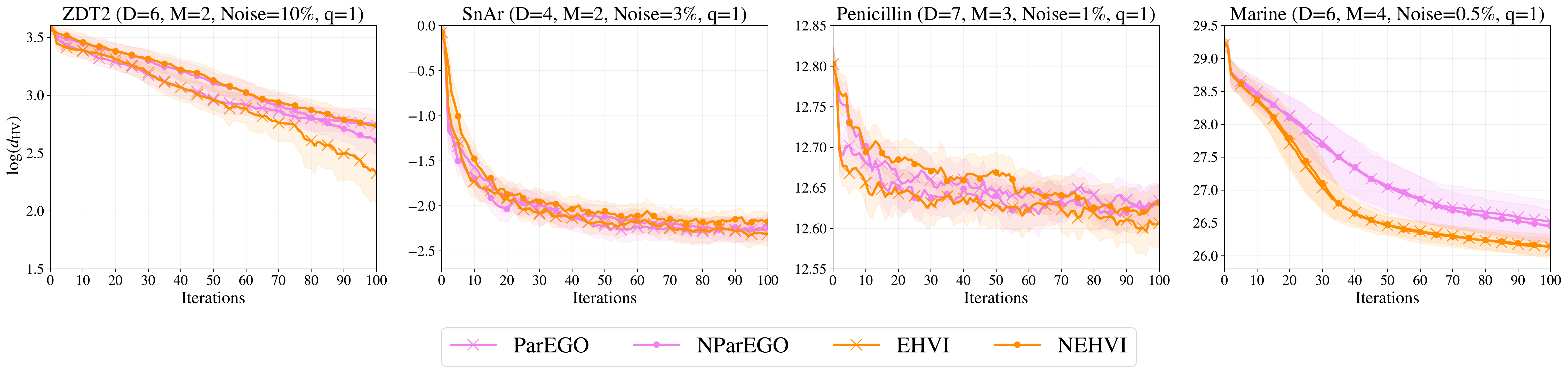}
	\includegraphics[width=1\linewidth]{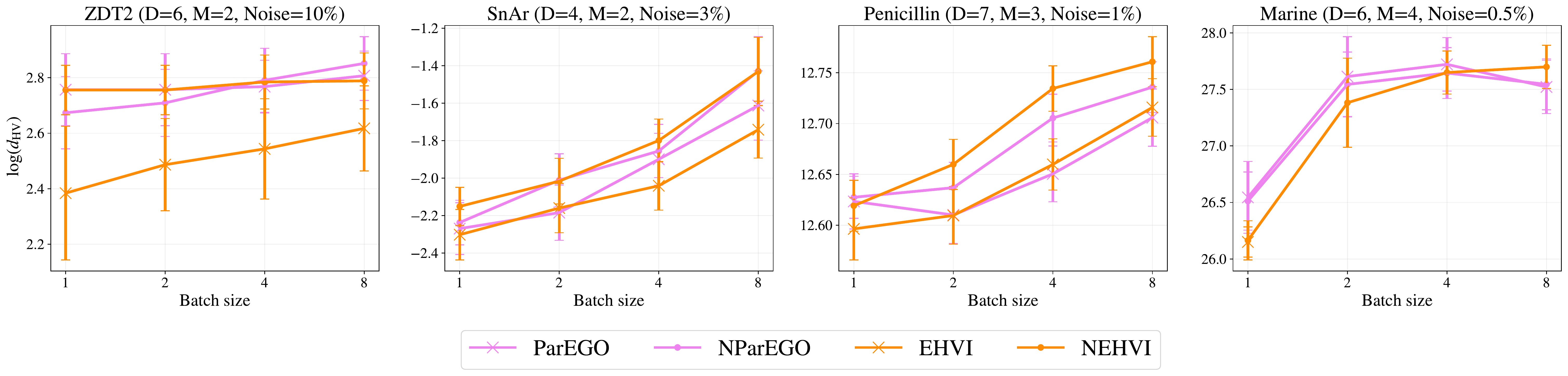}
	\centering
	\caption{A comparison of the mean logarithm HV discrepancy with two standard errors over one hundred runs for the four benchmark problems for the improvement-based algorithms. On the top, we present the results for the sequential experiment, whilst on the bottom we present the final results on the batch experiments with different batch sizes.}
	\label{fig:experiments_hv_ei}
\end{figure}
\FloatBarrier
\subsection{Generalized hypervolume}
\label{app:ghv}
In this section we produce profile plots of the generalized hypervolume at the final time instance. We only consider the median performance from the multiple runs when setting the approximation set to be the maximum of the final posterior mean: $\hat{\mathbb{X}}^* = \argmax_{\mathbf{x} \in \mathbb{X}} \boldsymbol{\mu}_N(\mathbf{x})$ 
\subsubsection{Weight distributions}
\begin{figure}[!htb]
	\includegraphics[width=0.75\linewidth]{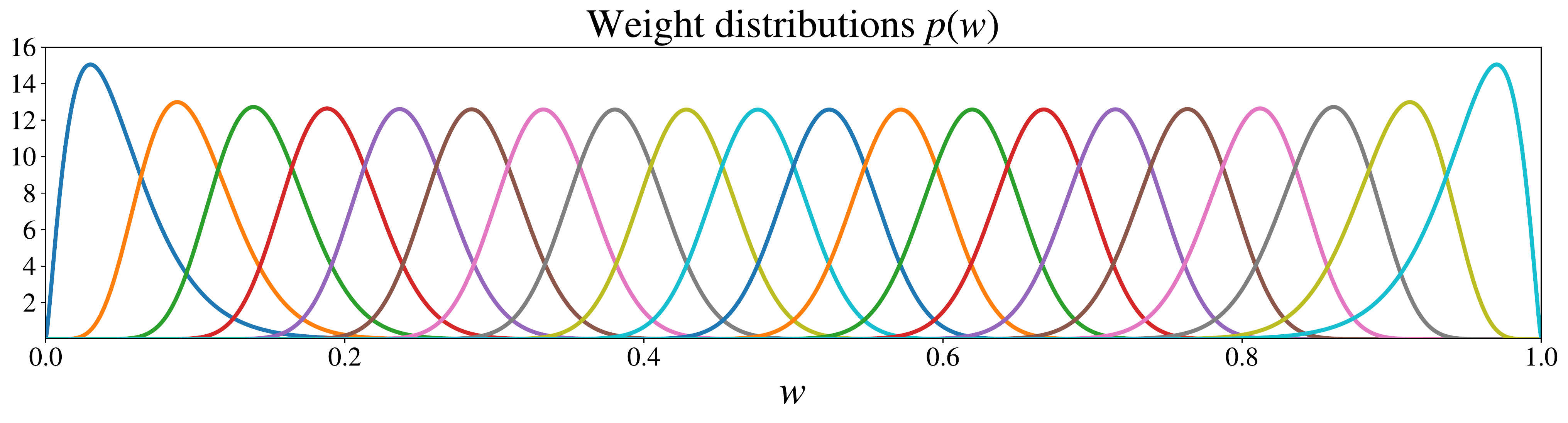}
	\centering
	\caption{The Beta distributions used to generate the weights for the logarithm GHV discrepancy.}
	\label{fig:zdt2_q1_w}
\end{figure}
\begin{figure}[!htb]
	\includegraphics[width=0.75\linewidth]{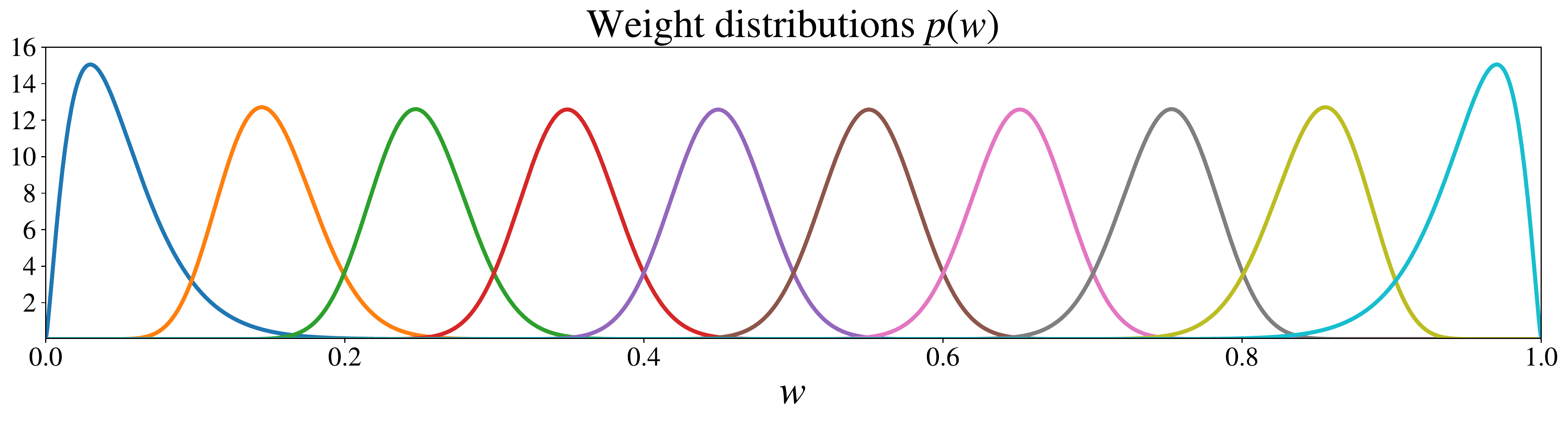}
	\centering
	\caption{The Beta distributions used to generate the weights for the logarithm GHV discrepancy.}
	\label{fig:penicillin_q1_w}
\end{figure}
\begin{figure}
	\includegraphics[width=0.75\linewidth]{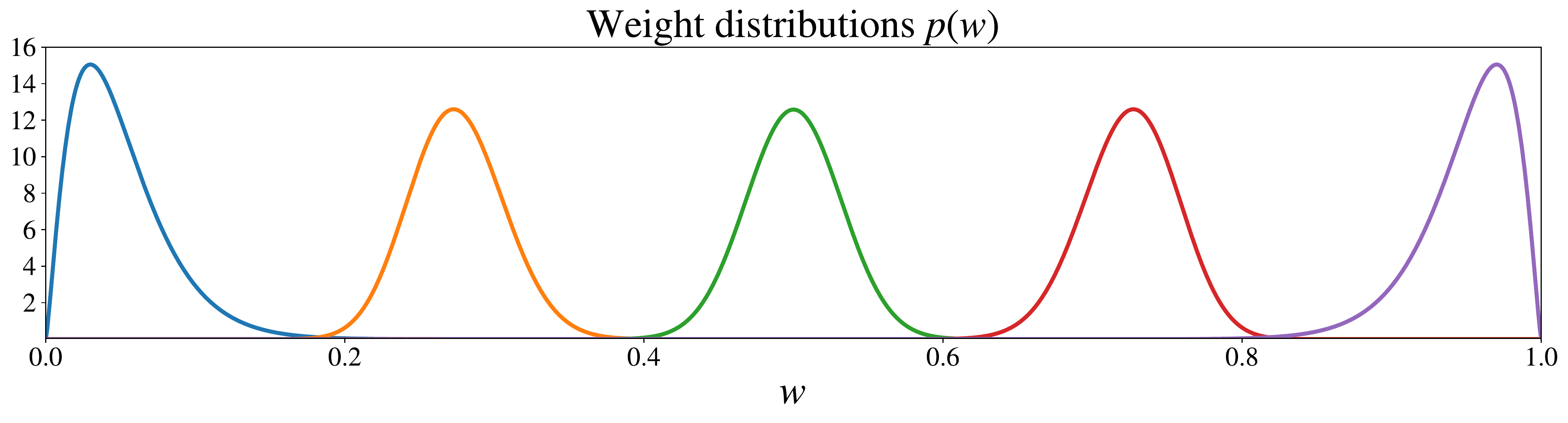}
	\centering
	\caption{The Beta distributions used to generate the weights for the logarithm GHV discrepancy.}
	\label{fig:marine_q1_w}
\end{figure}
\FloatBarrier
\subsubsection{ZDT2(D=2, M=2, Noise=10\%, q=1)}
\begin{figure}[!htb]
	\includegraphics[width=1\linewidth]{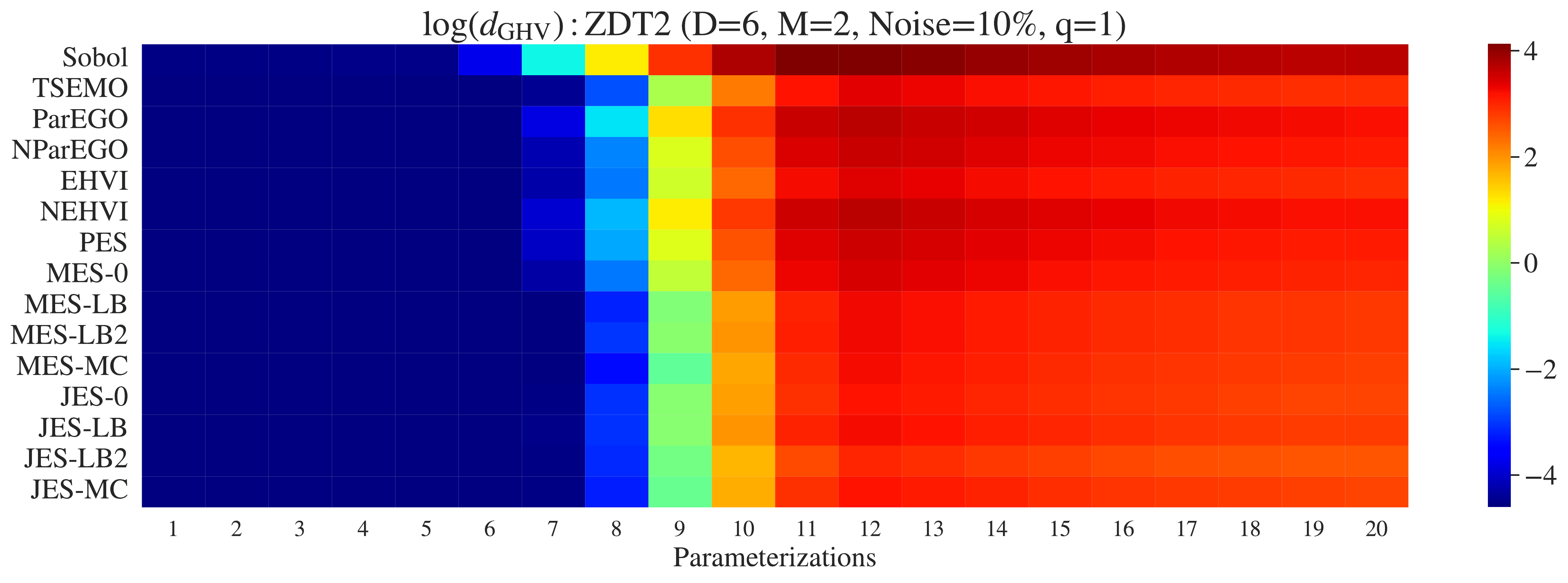}
	\centering
	\caption{A heat map comparison of the final median logarithm GHV discrepancy. The weight distributions are described in \cref{fig:zdt2_q1_w}.}
	\label{fig:zdt2_q1_all_ghv}
\end{figure}
\FloatBarrier
\subsubsection{SnAr (D=4, M=2, Noise=3\%, q=1)}
\begin{figure}[!htb]
	\includegraphics[width=1\linewidth]{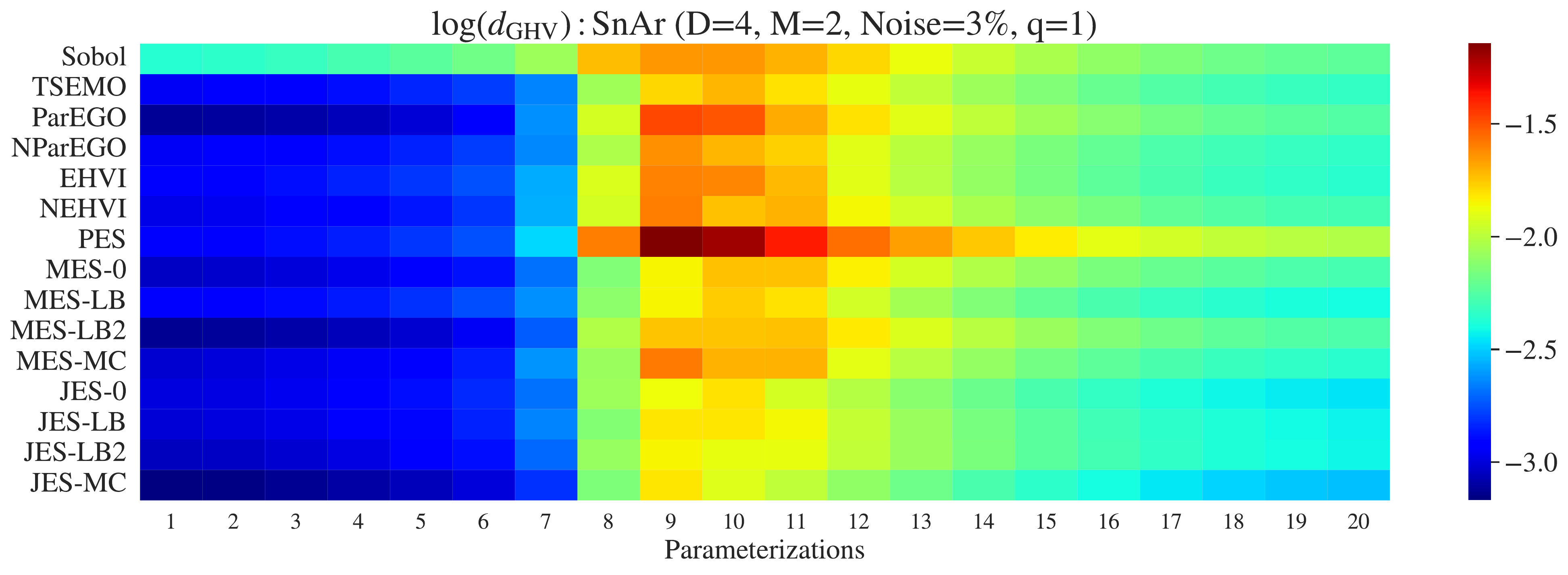}
	\centering
	\caption{A heat map comparison of the final median logarithm GHV discrepancy. The weight distributions are described in \cref{fig:zdt2_q1_w}.}
	\label{fig:snar_q1_all_ghv}
\end{figure}
\FloatBarrier
\subsubsection{Penicillin (D=7, M=3, Noise=1\%, q=1)}
\begin{figure}[!htb]
	\includegraphics[width=1\linewidth]{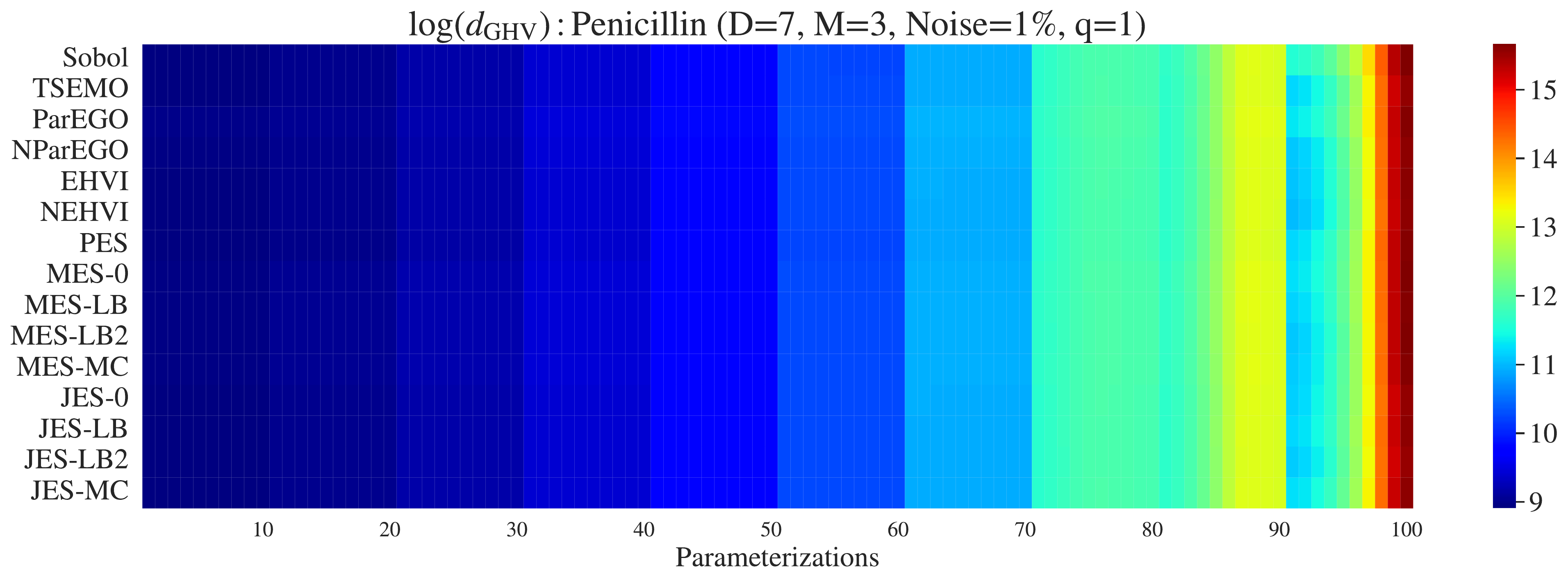}
	\centering
	\caption{A heat map comparison of the final median logarithm GHV discrepancy. The weight distributions are described in \cref{fig:penicillin_q1_w}.} 
	\label{fig:penicillin_q1_all_ghv}
\end{figure}
\FloatBarrier
\subsubsection{Marine (D=6, M=4, Noise=0.5\%, q=1)}
\begin{figure}[!htb]
	\includegraphics[width=1\linewidth]{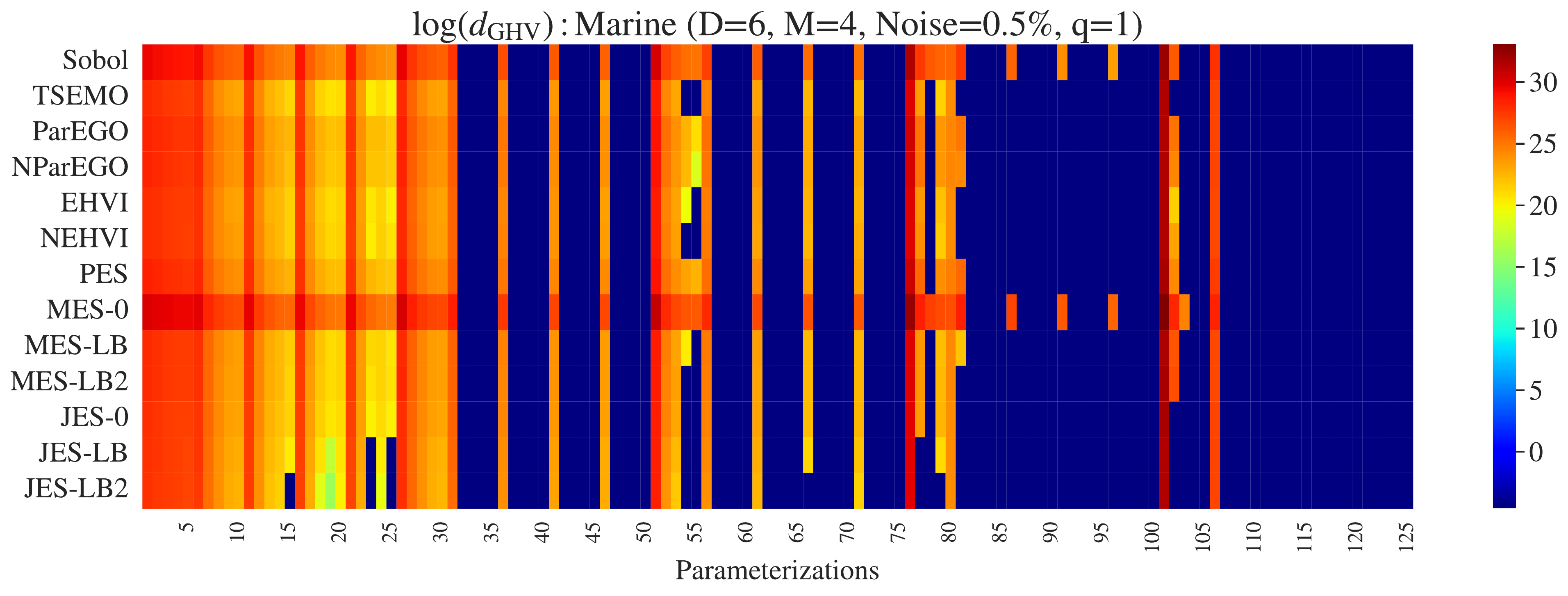}
	\centering
	\caption{A heat map comparison of the final median logarithm GHV discrepancy. The weight distributions are described in \cref{fig:marine_q1_w}.} 
	\label{fig:marine_q1_all_ghv}
\end{figure}
\FloatBarrier
\subsection{Wall times}
\label{app:wall_times}
\begin{figure}[!htb]
	\includegraphics[width=0.48\linewidth]{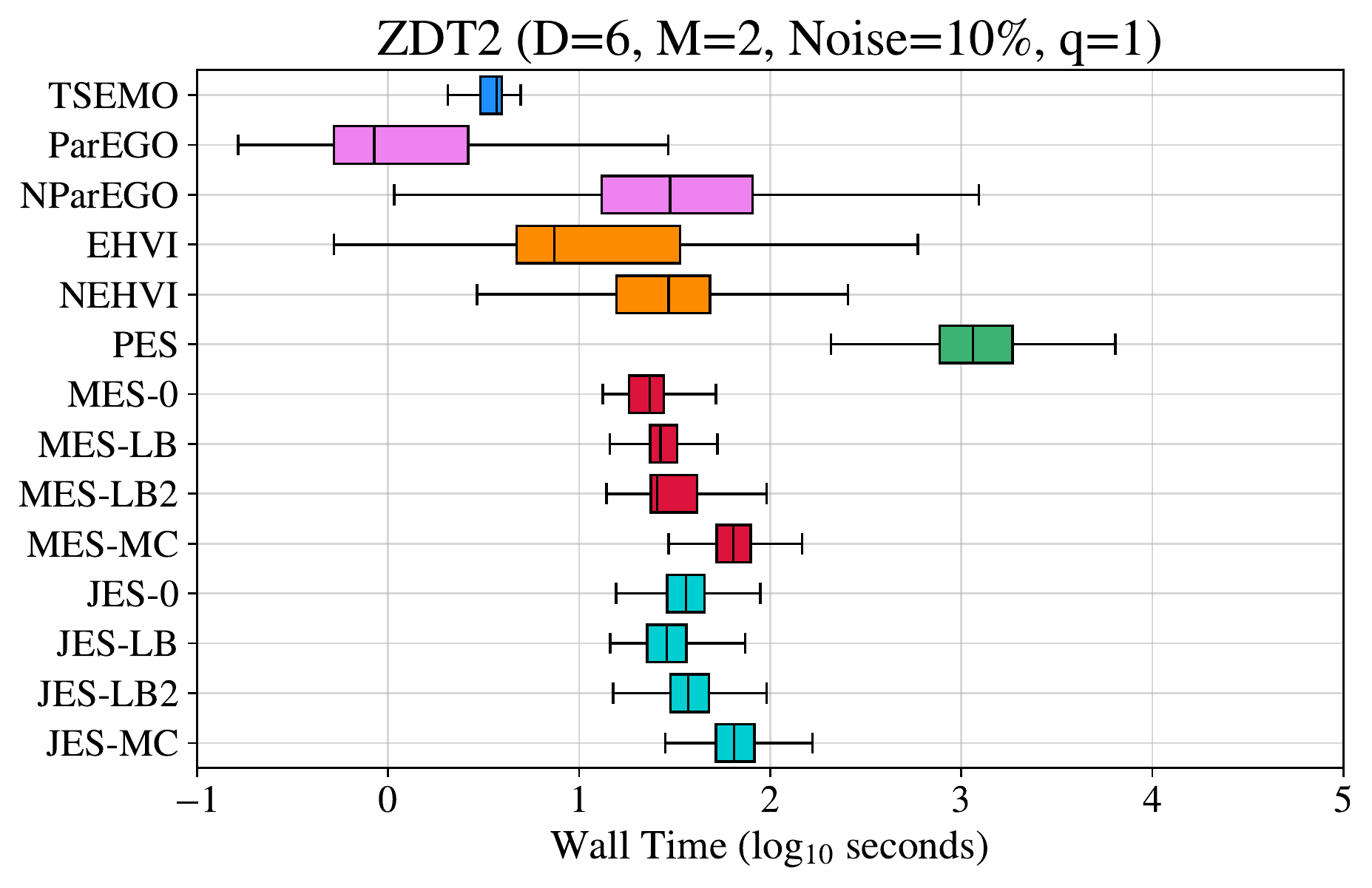}
	\includegraphics[width=0.48\linewidth]{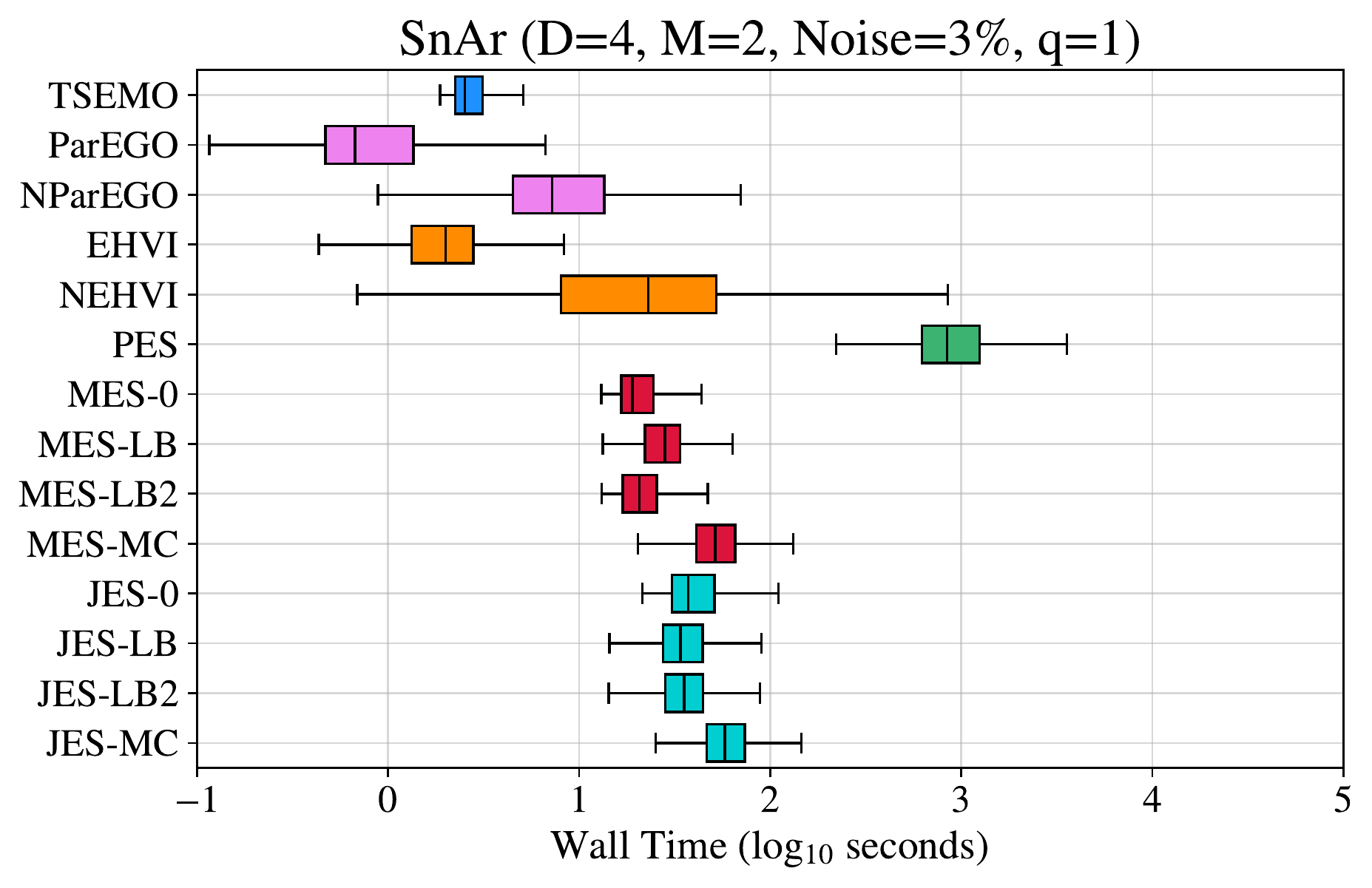}
	\includegraphics[width=0.48\linewidth]{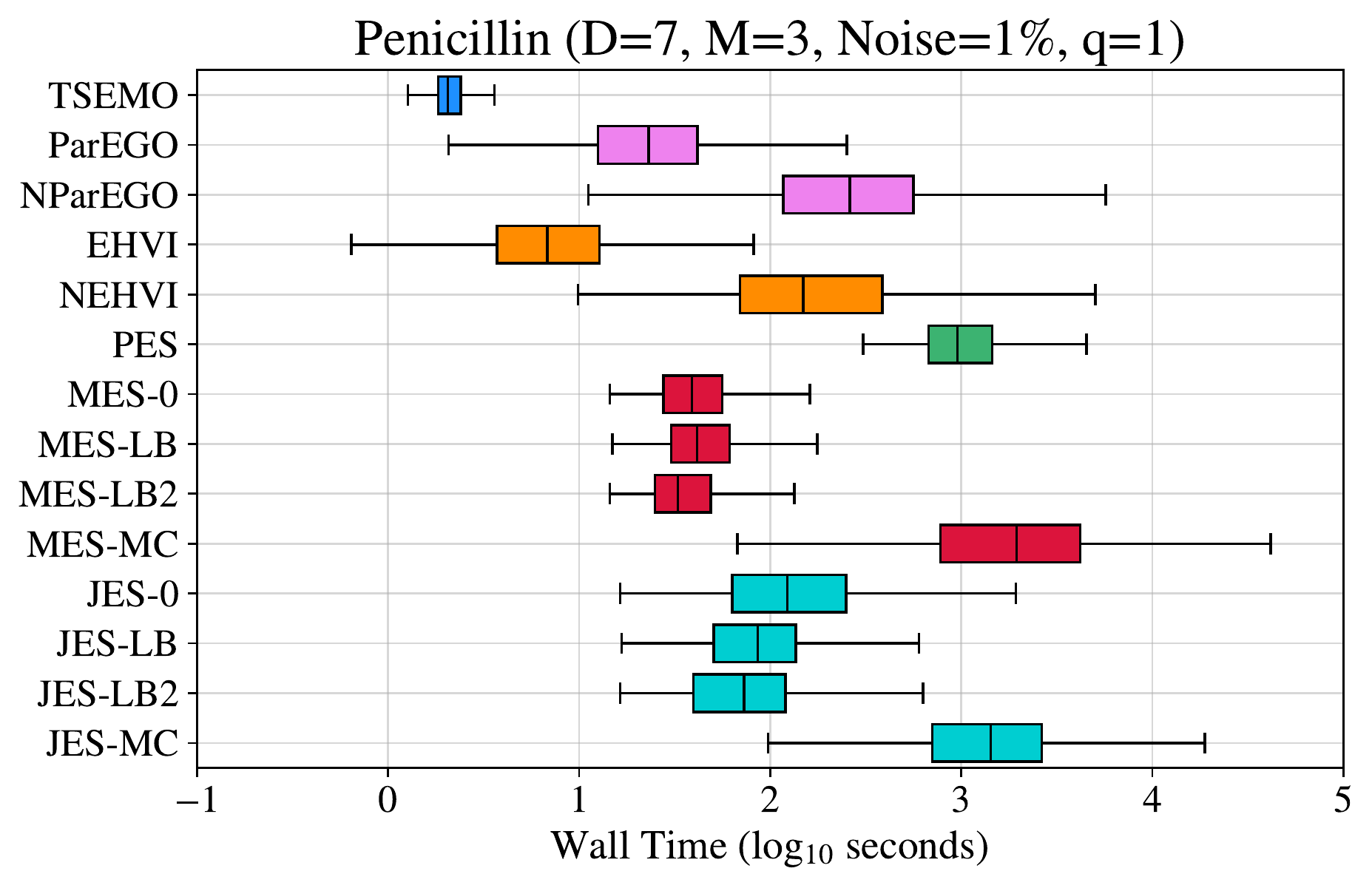}
	\includegraphics[width=0.48\linewidth]{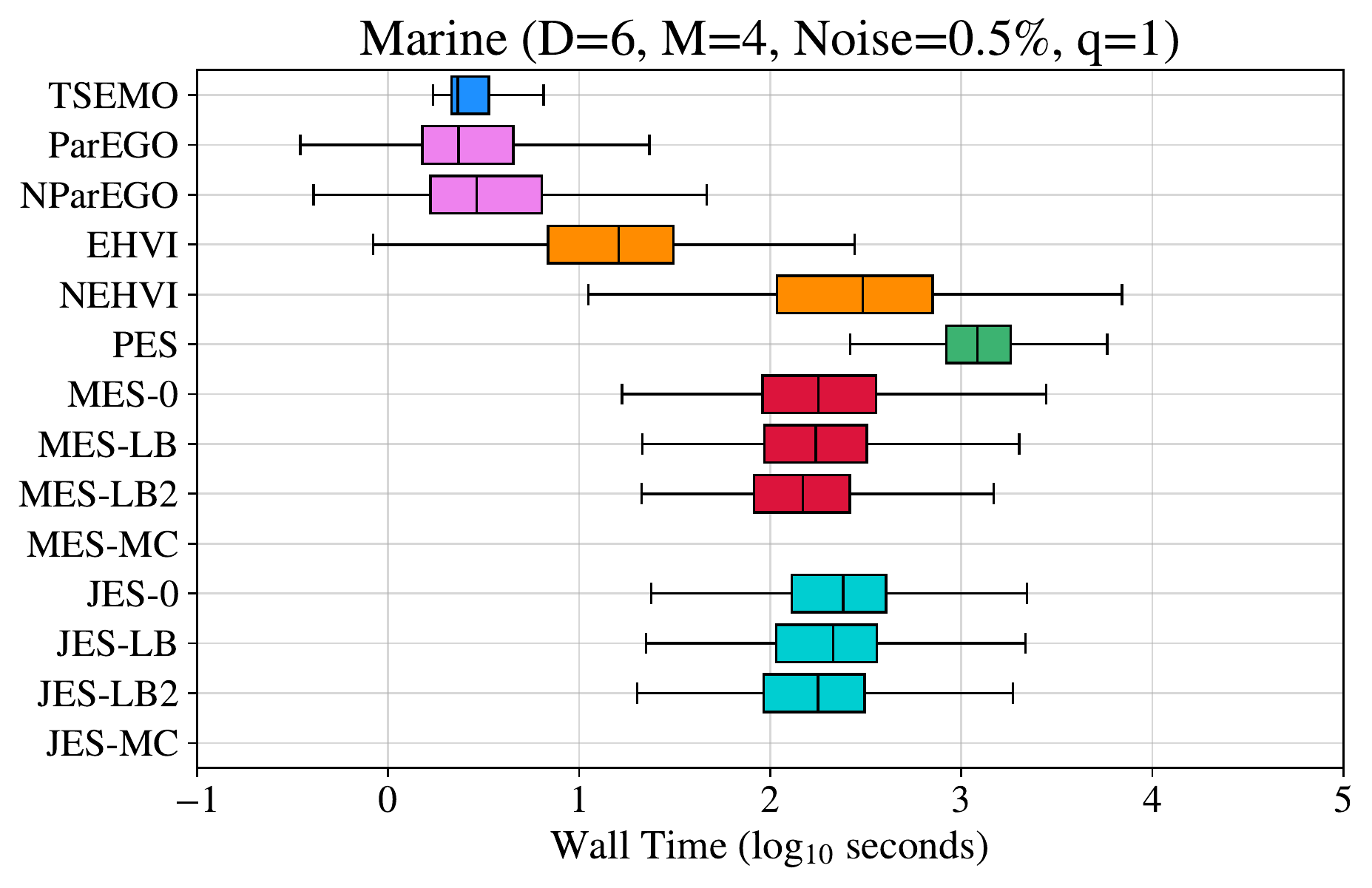}
	\includegraphics[width=0.48\linewidth]{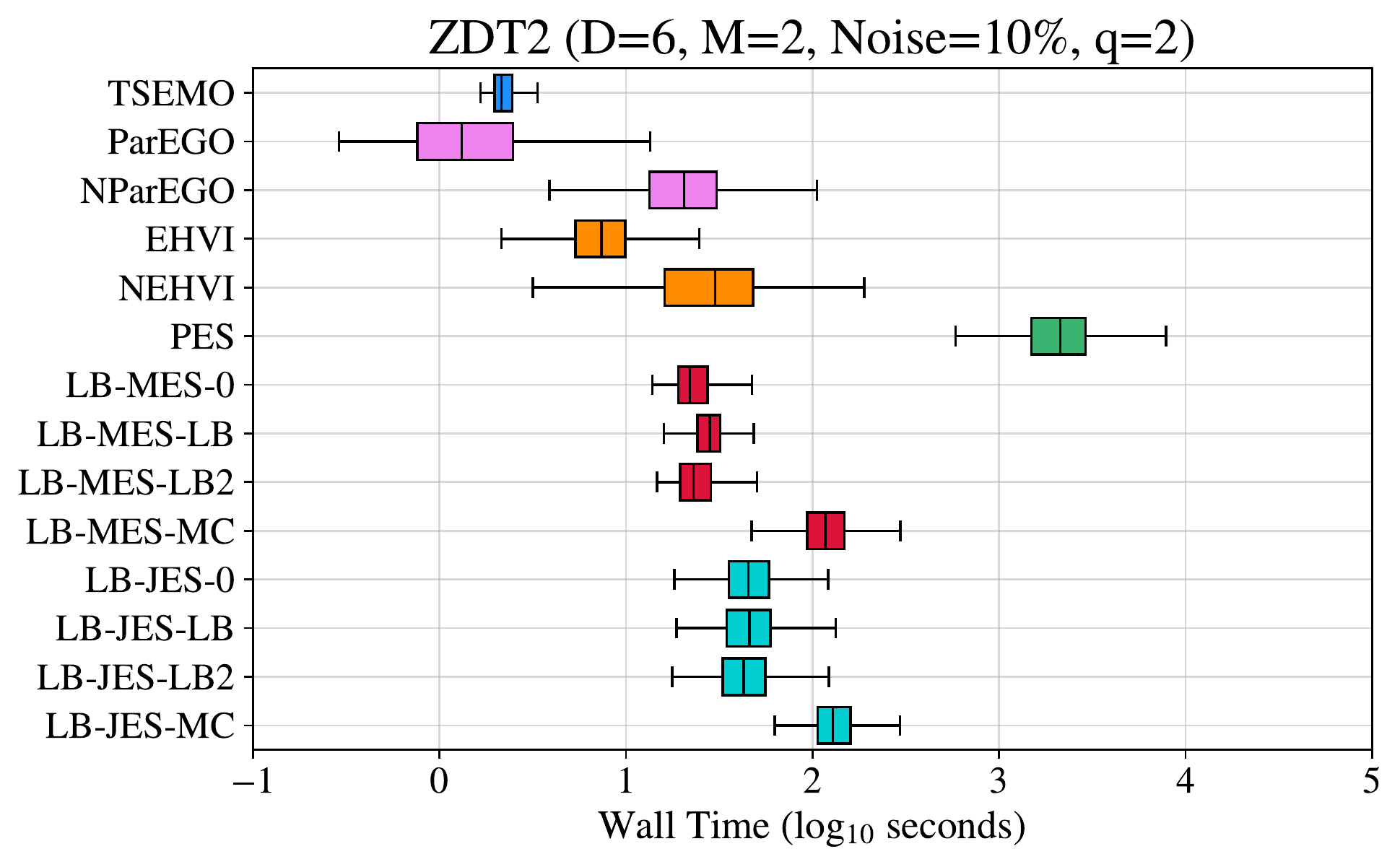}
	\includegraphics[width=0.48\linewidth]{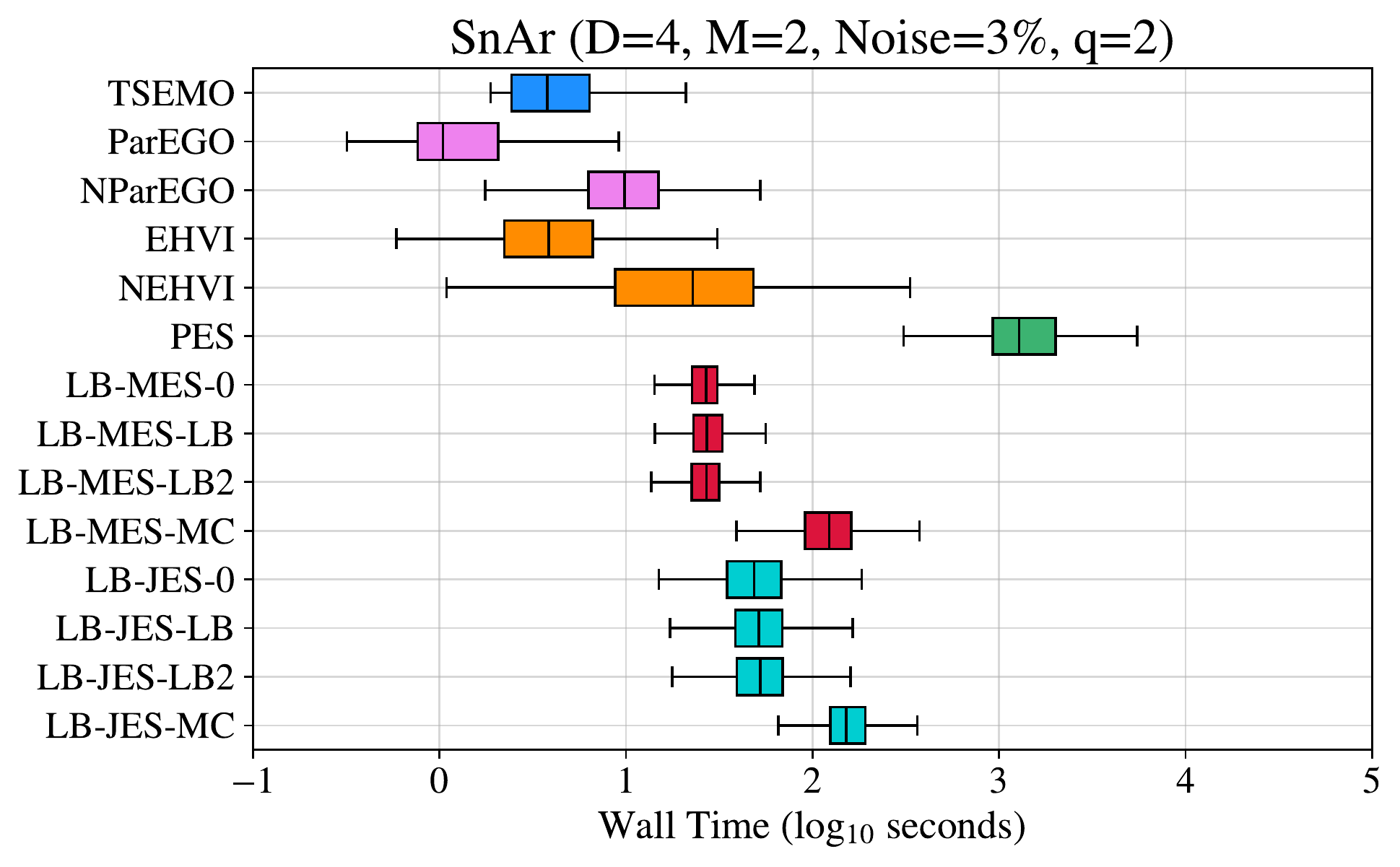}
	\includegraphics[width=0.48\linewidth]{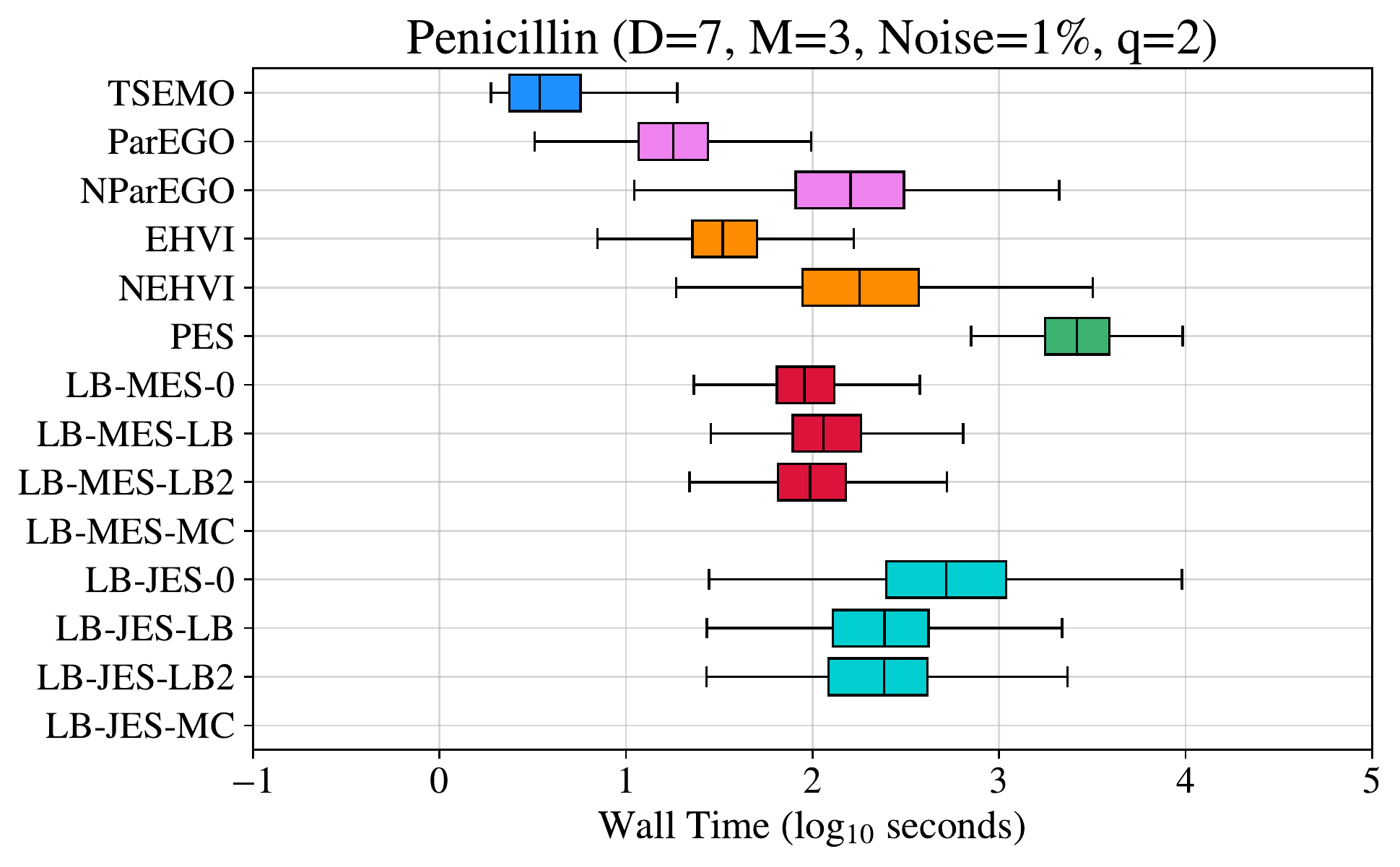}
	\includegraphics[width=0.48\linewidth]{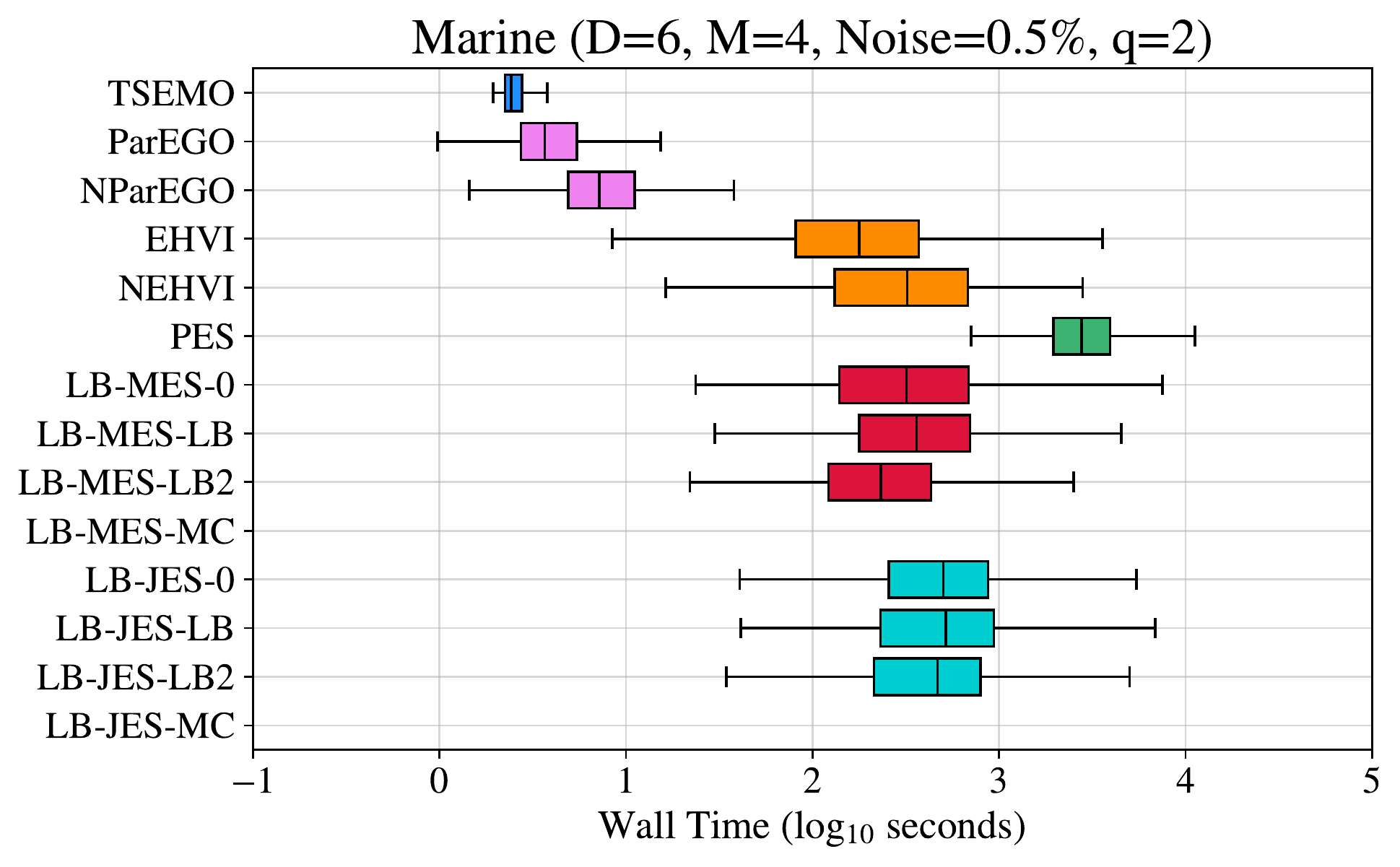}
	\centering
\end{figure}
\begin{figure}	
	\ContinuedFloat
	\includegraphics[width=0.48\linewidth]{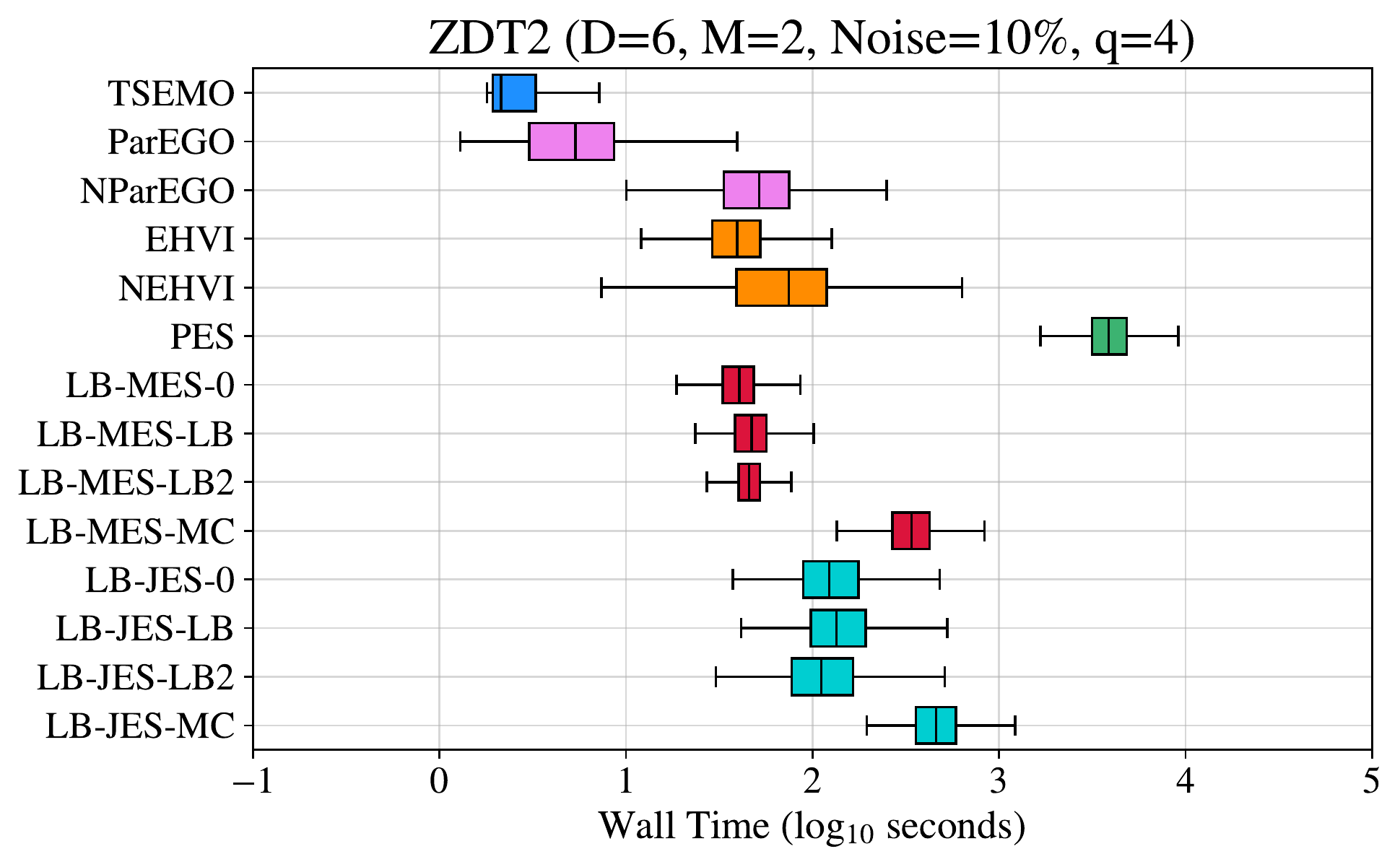}
	\includegraphics[width=0.48\linewidth]{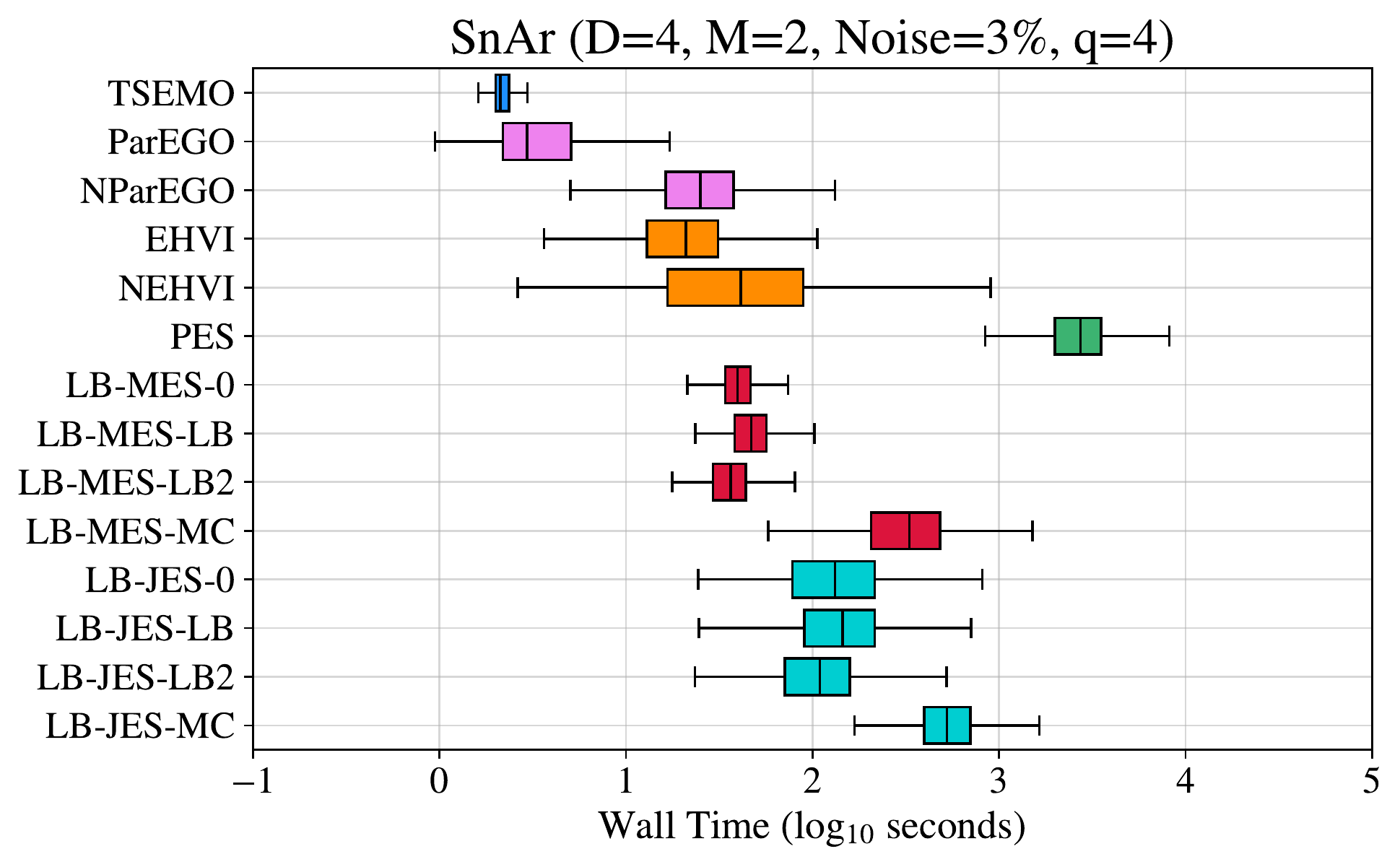}
	\includegraphics[width=0.48\linewidth]{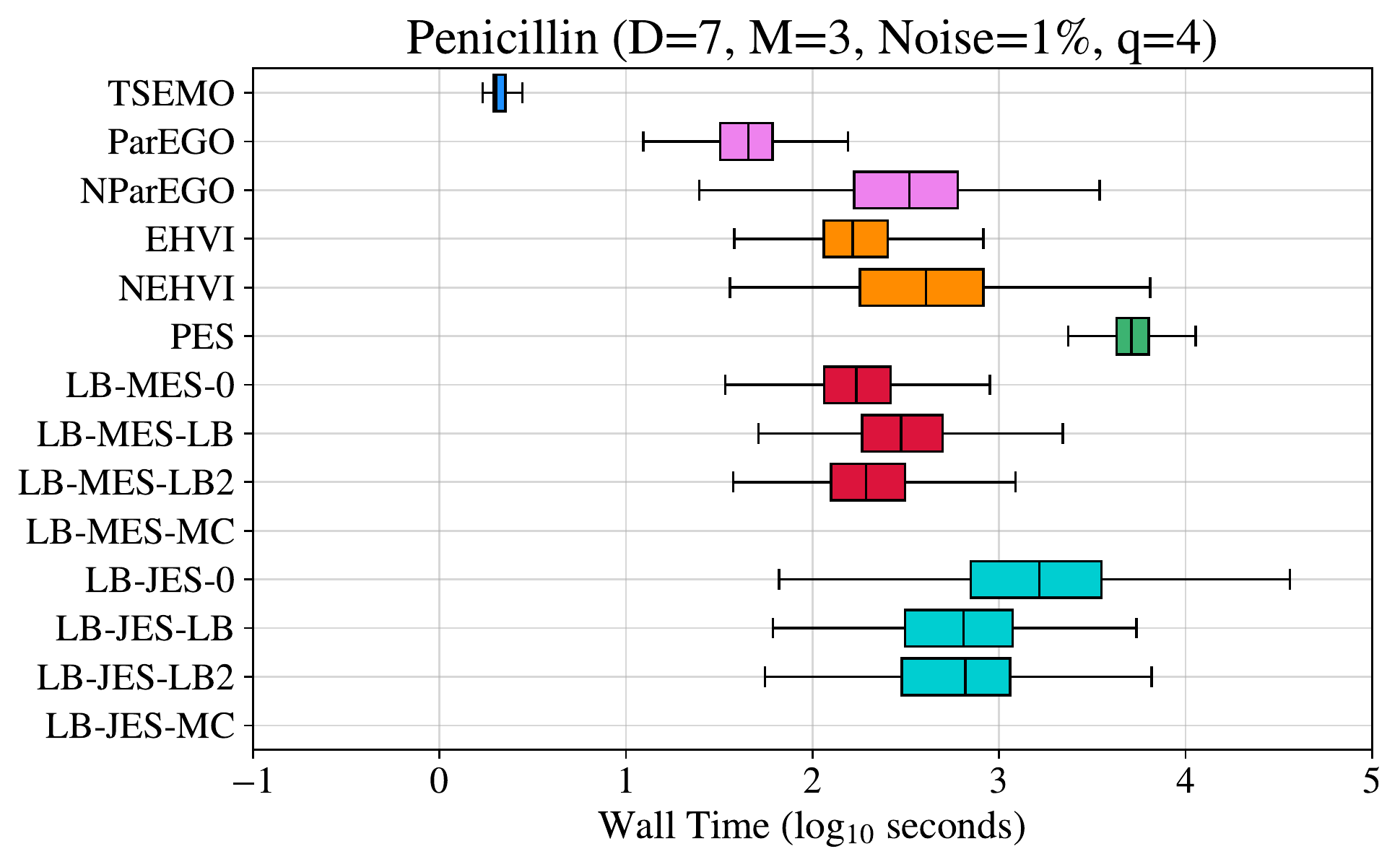}
	\includegraphics[width=0.48\linewidth]{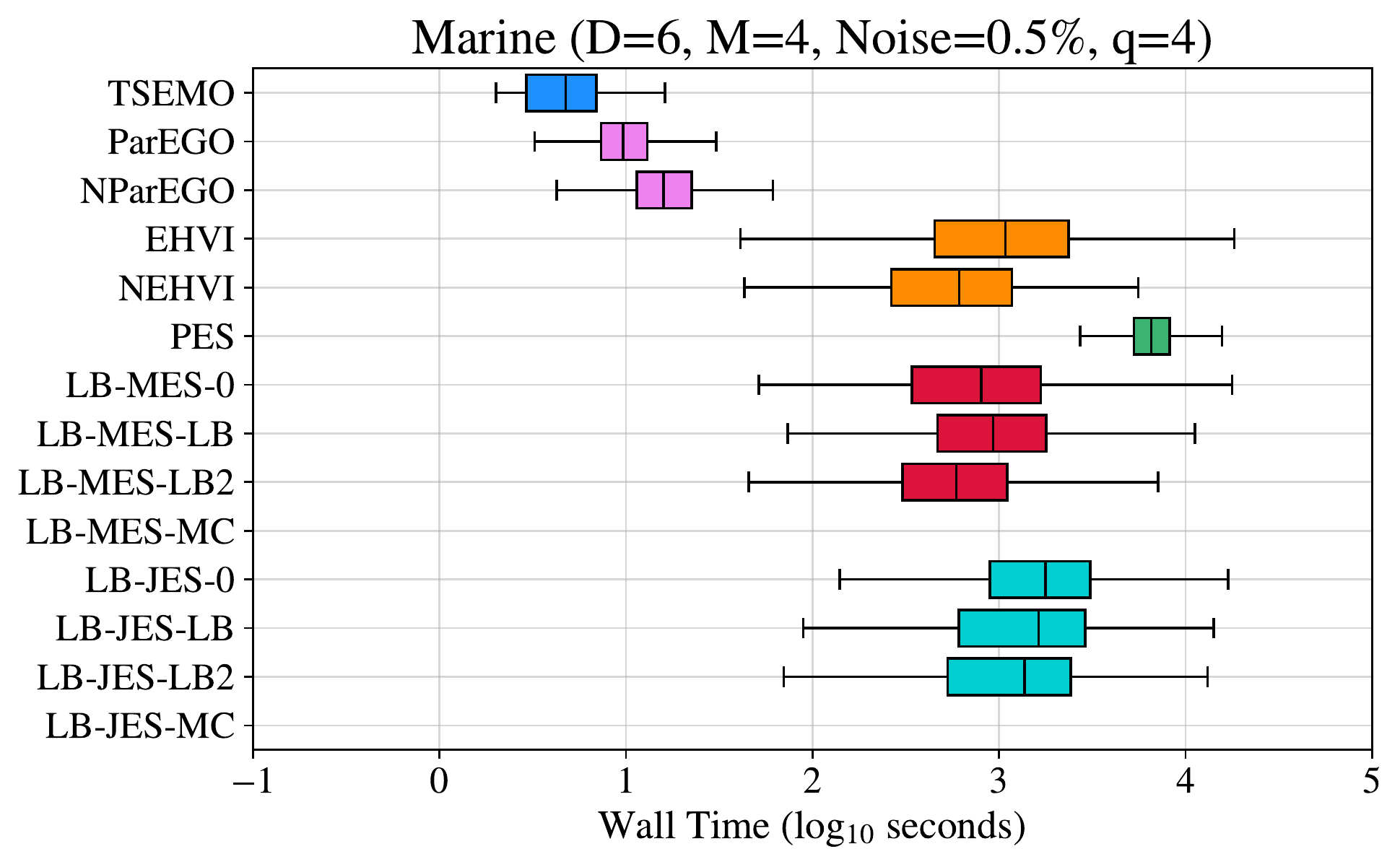}
	\includegraphics[width=0.48\linewidth]{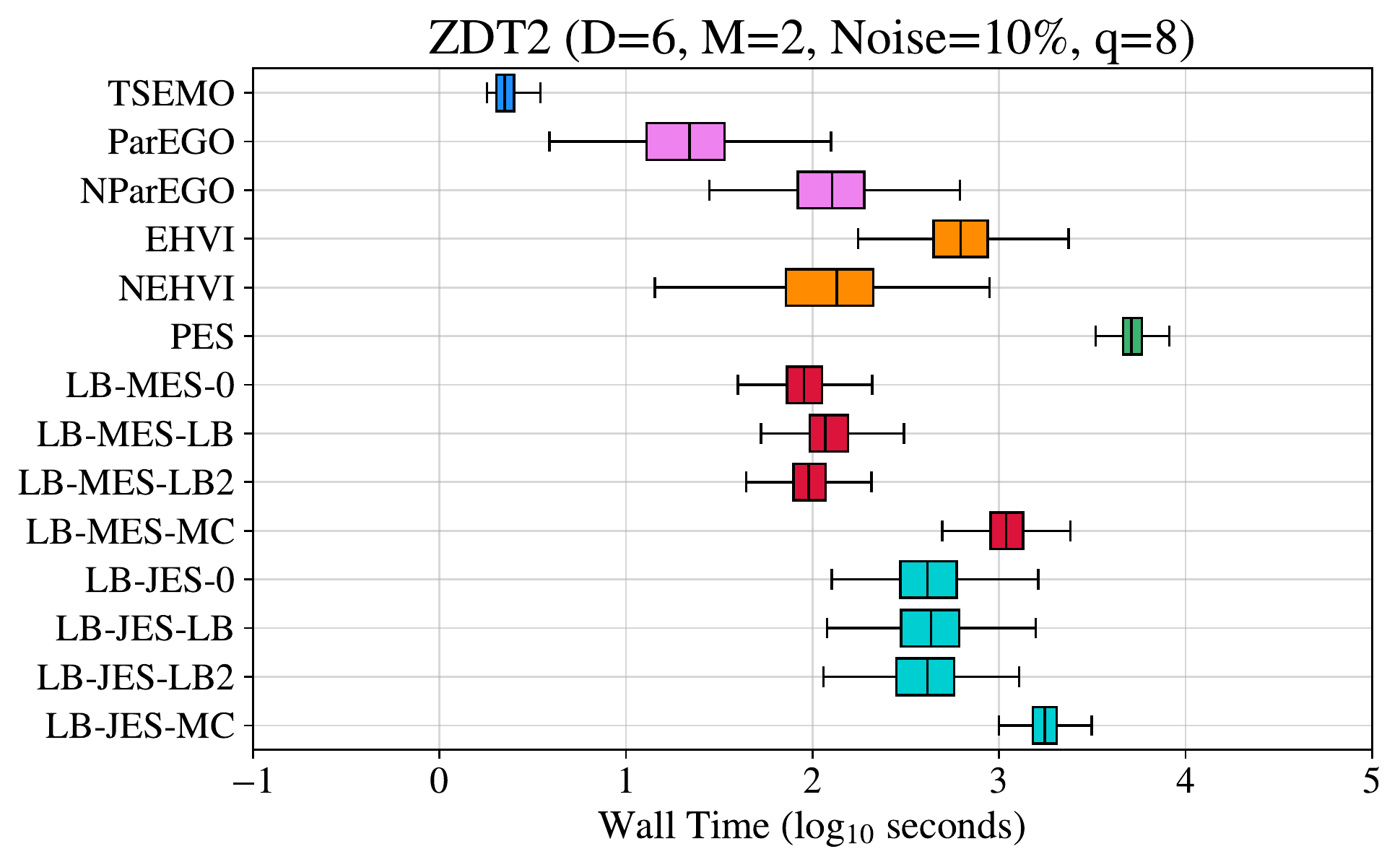}
	\includegraphics[width=0.48\linewidth]{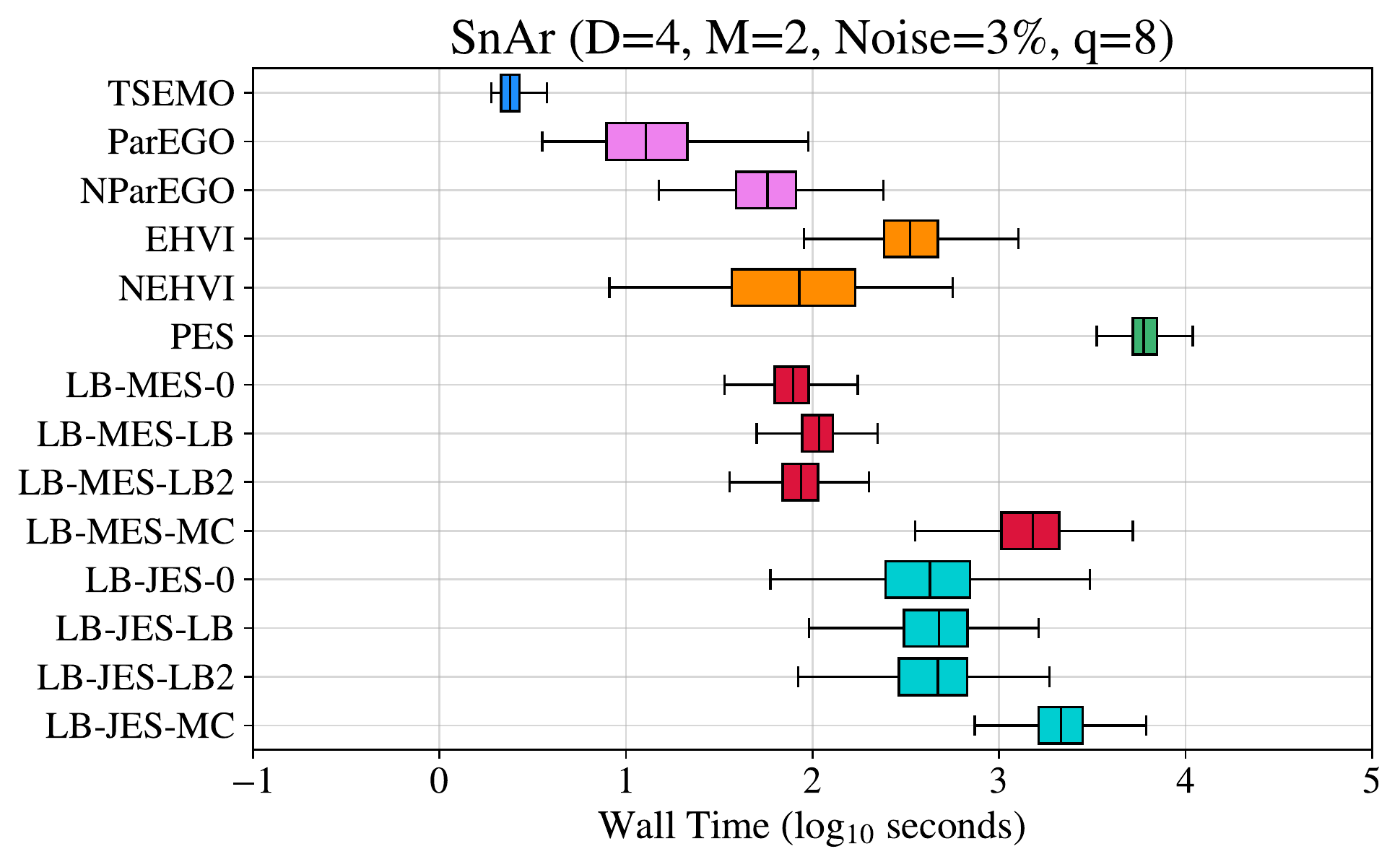}
	\includegraphics[width=0.48\linewidth]{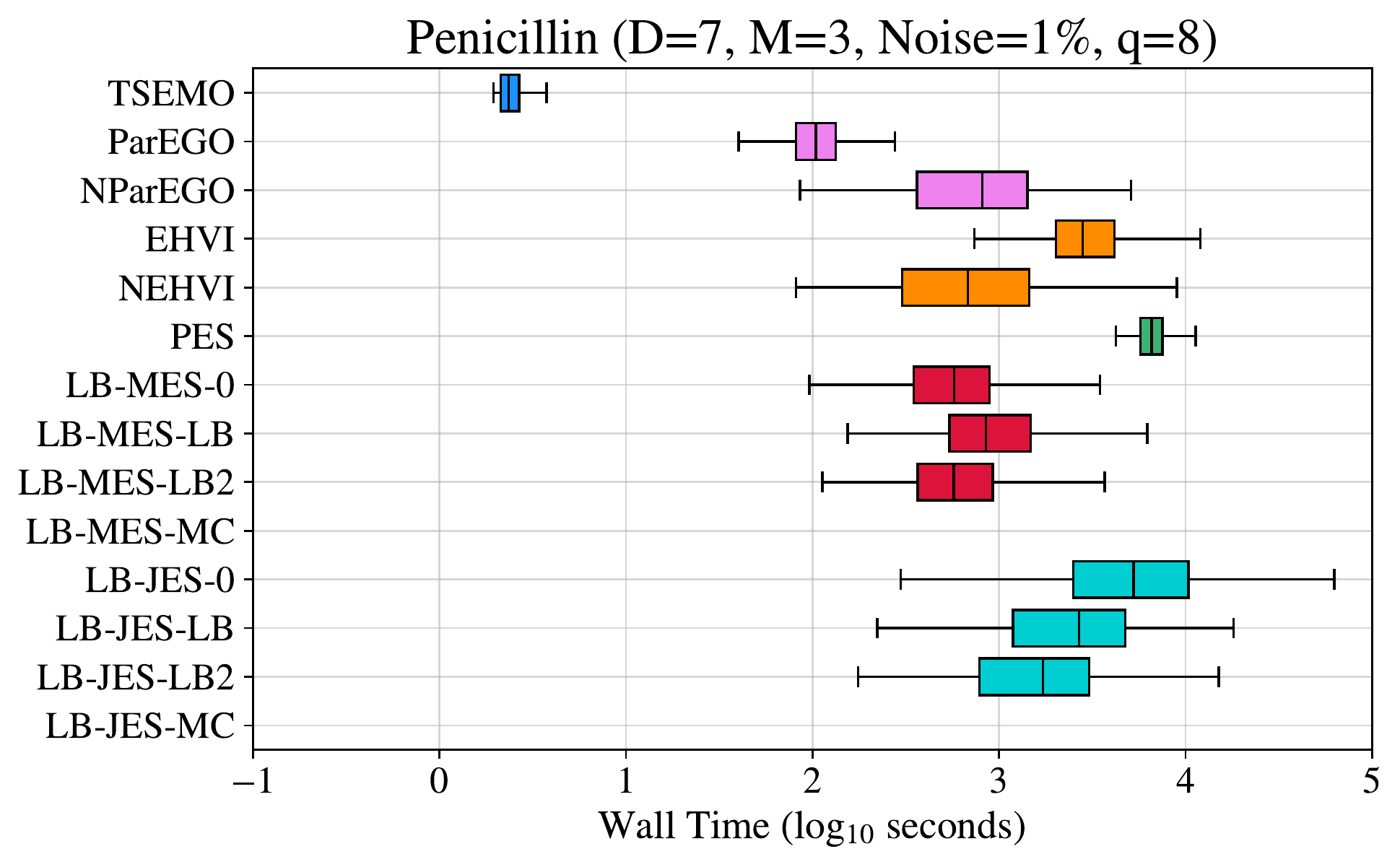}
	\includegraphics[width=0.48\linewidth]{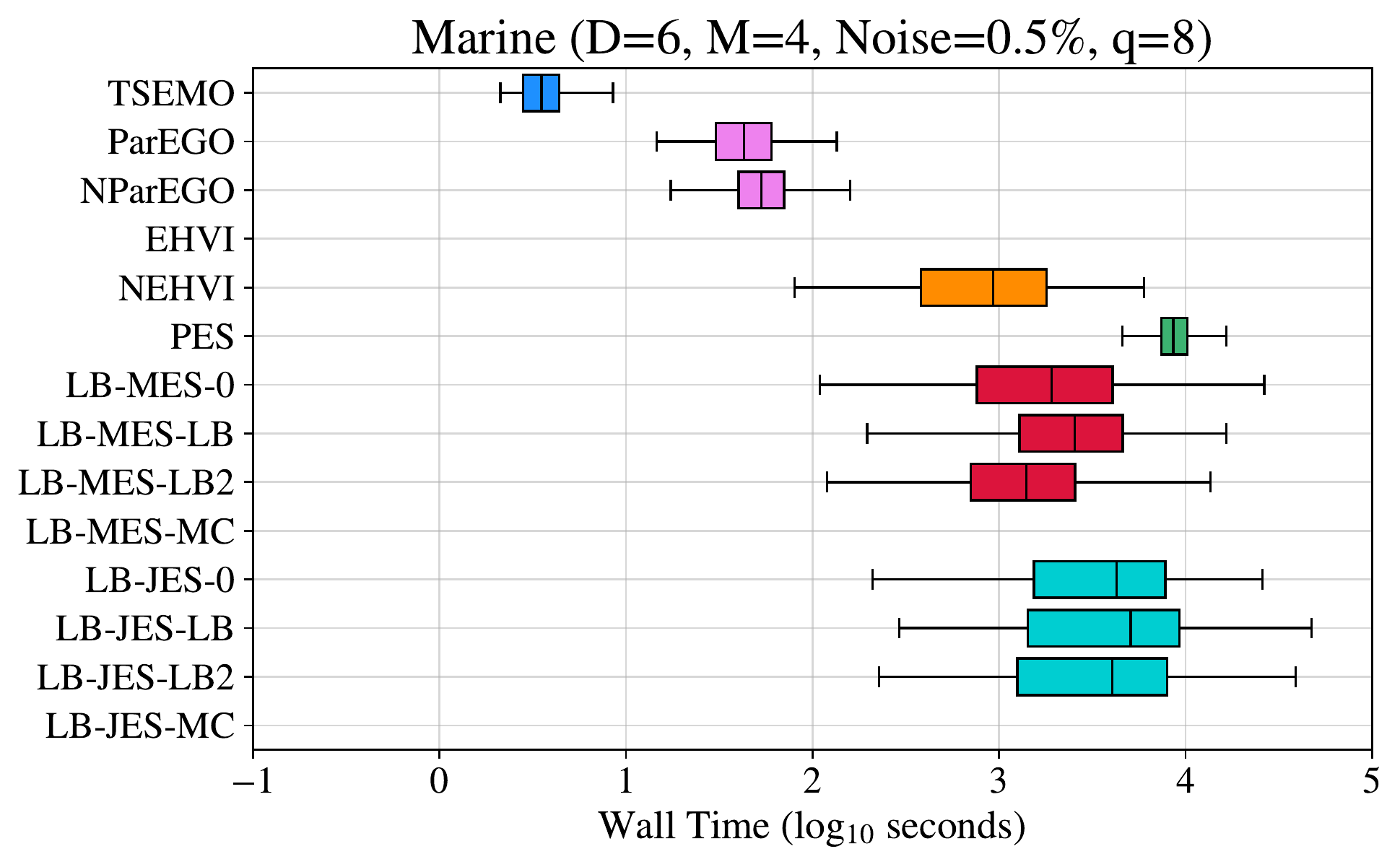}
	\centering
	\caption{A box-plot comparison of the wall times for the acquisition stage, which includes the median, interquartile range and the extreme values after excluding the outliers. The acquisition stage includes any initialization computations such as the box decompositions and sampling the Pareto optimal points---it does not include initializing the posterior model. All of the runs of each algorithm was performed on a computing cluster, where we restricted the computation to a single CPU core of an AMD EPYC 7742 64-Core Processor @ 2.25GHz.}
	\label{fig:q1_time}
\end{figure}
\begin{figure}[!htb]
	\includegraphics[width=0.48\linewidth]{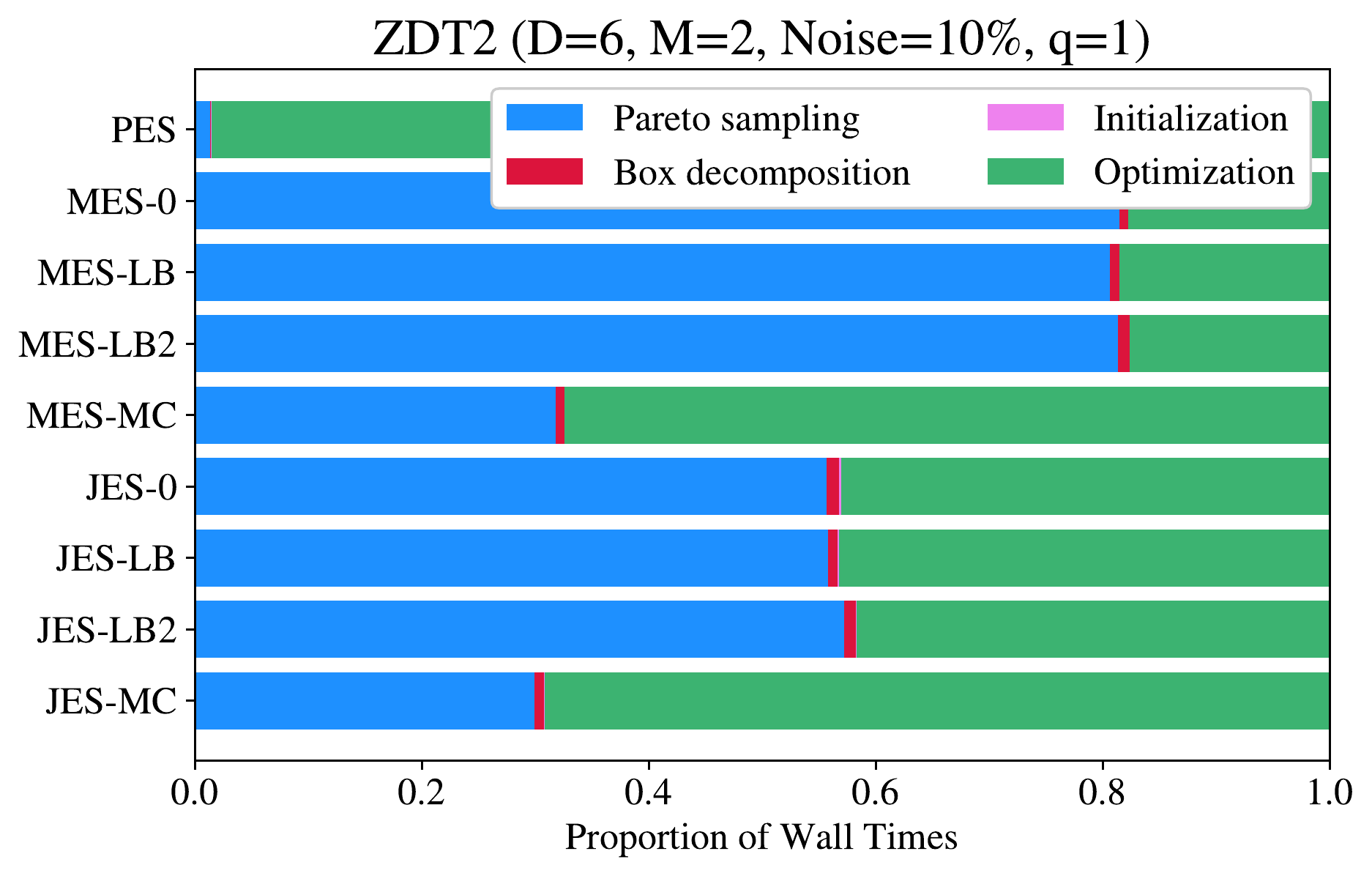}
	\includegraphics[width=0.48\linewidth]{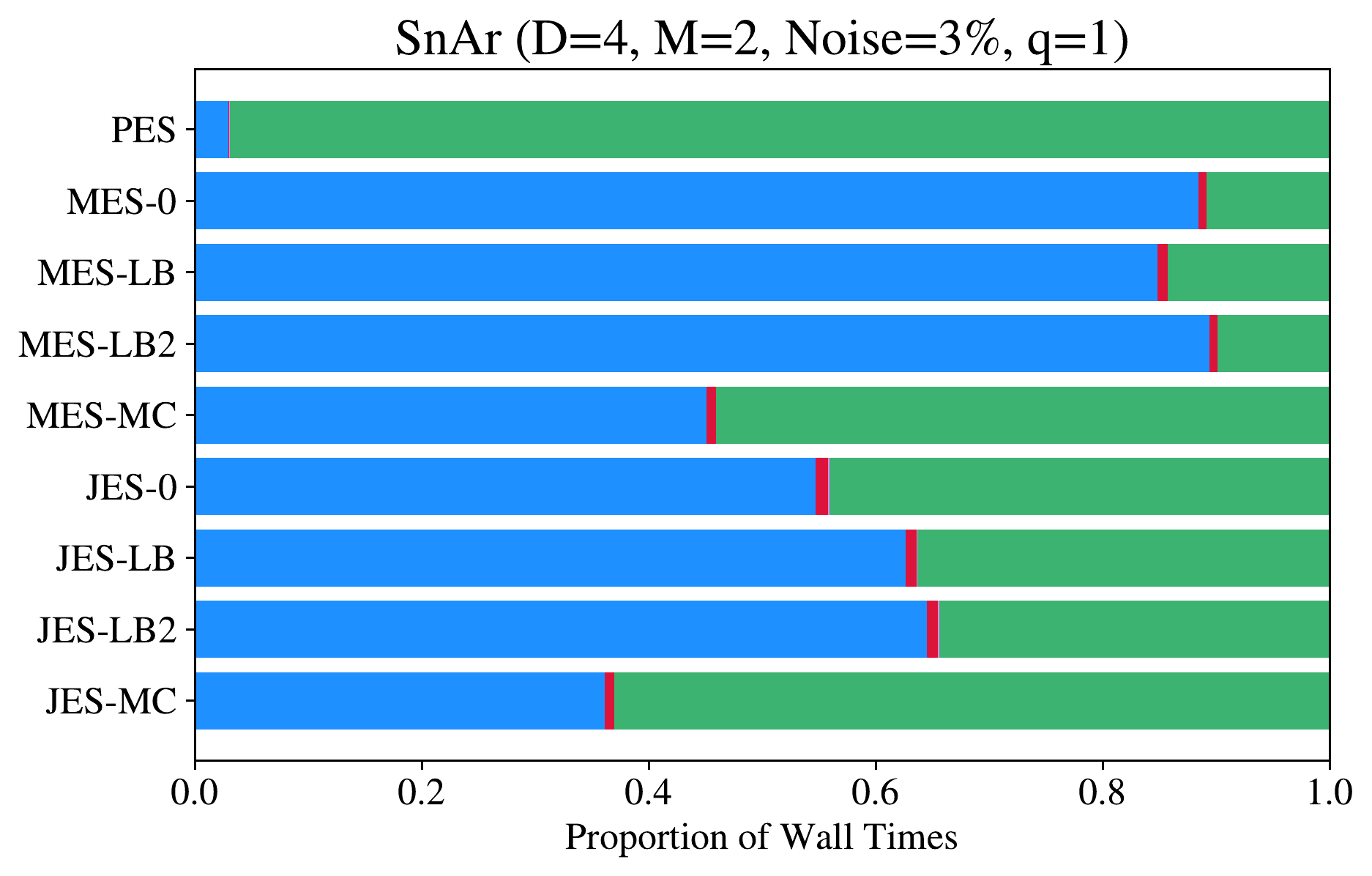}
	\includegraphics[width=0.48\linewidth]{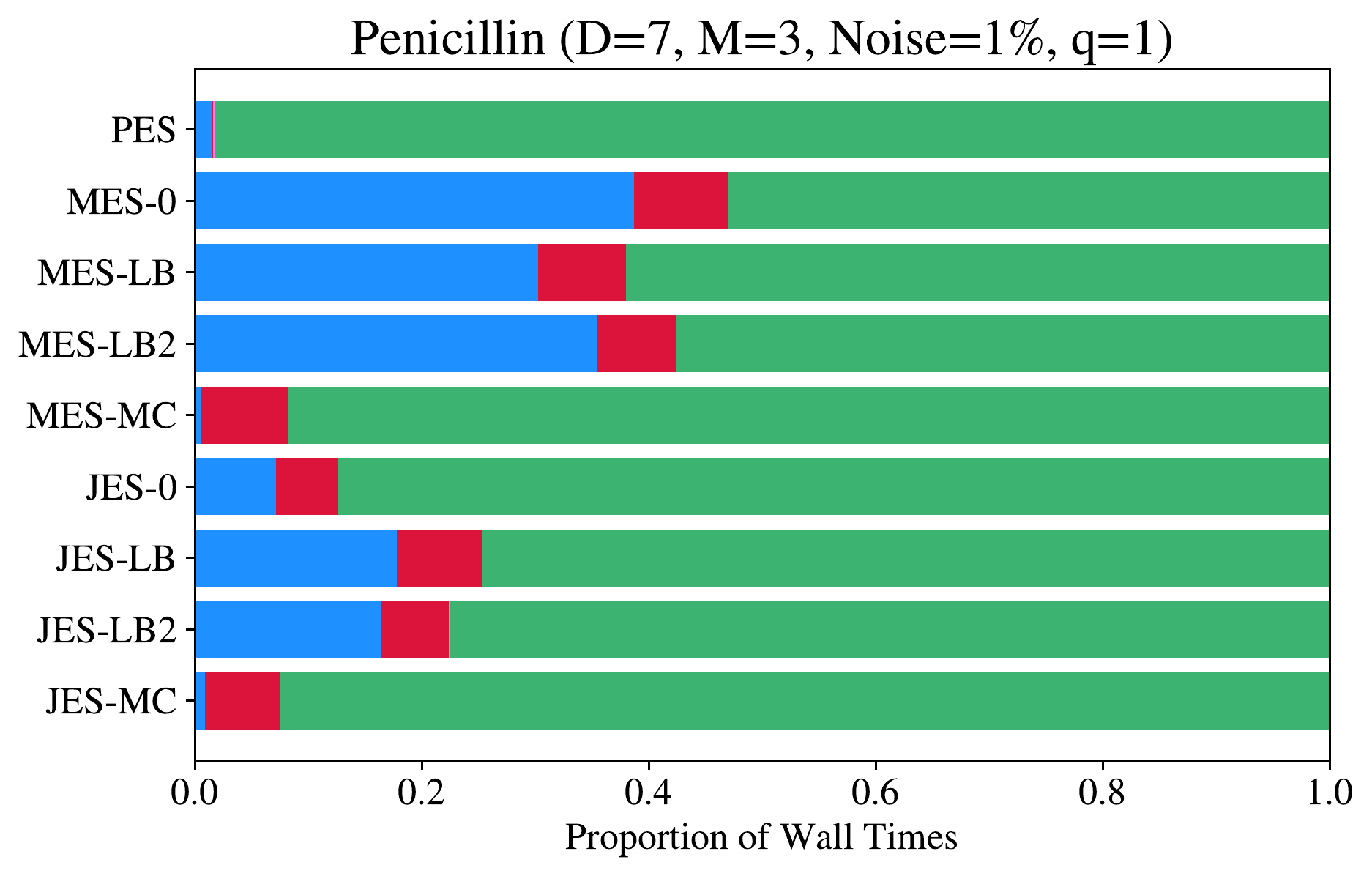}
	\includegraphics[width=0.48\linewidth]{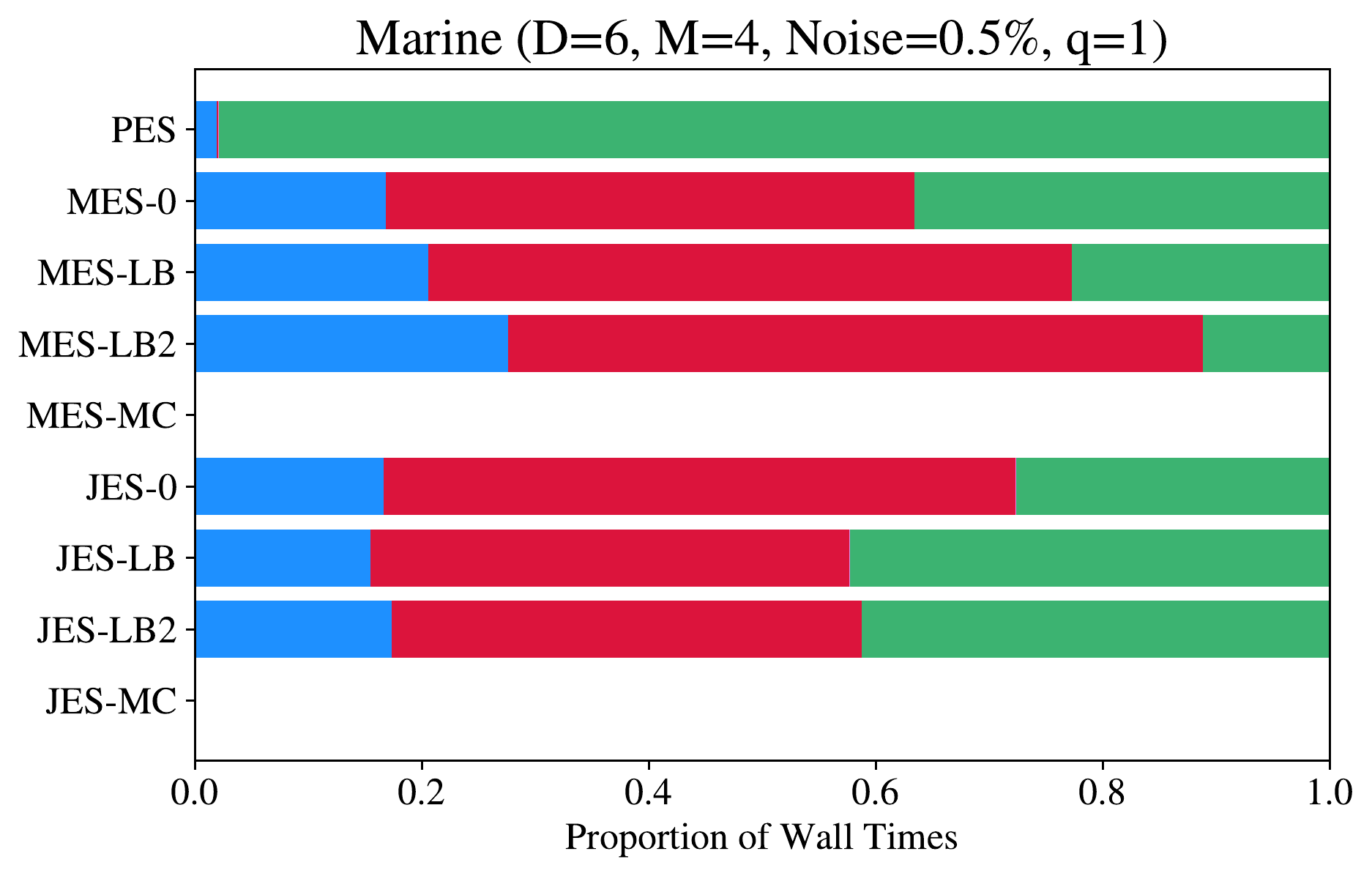}
	\includegraphics[width=0.48\linewidth]{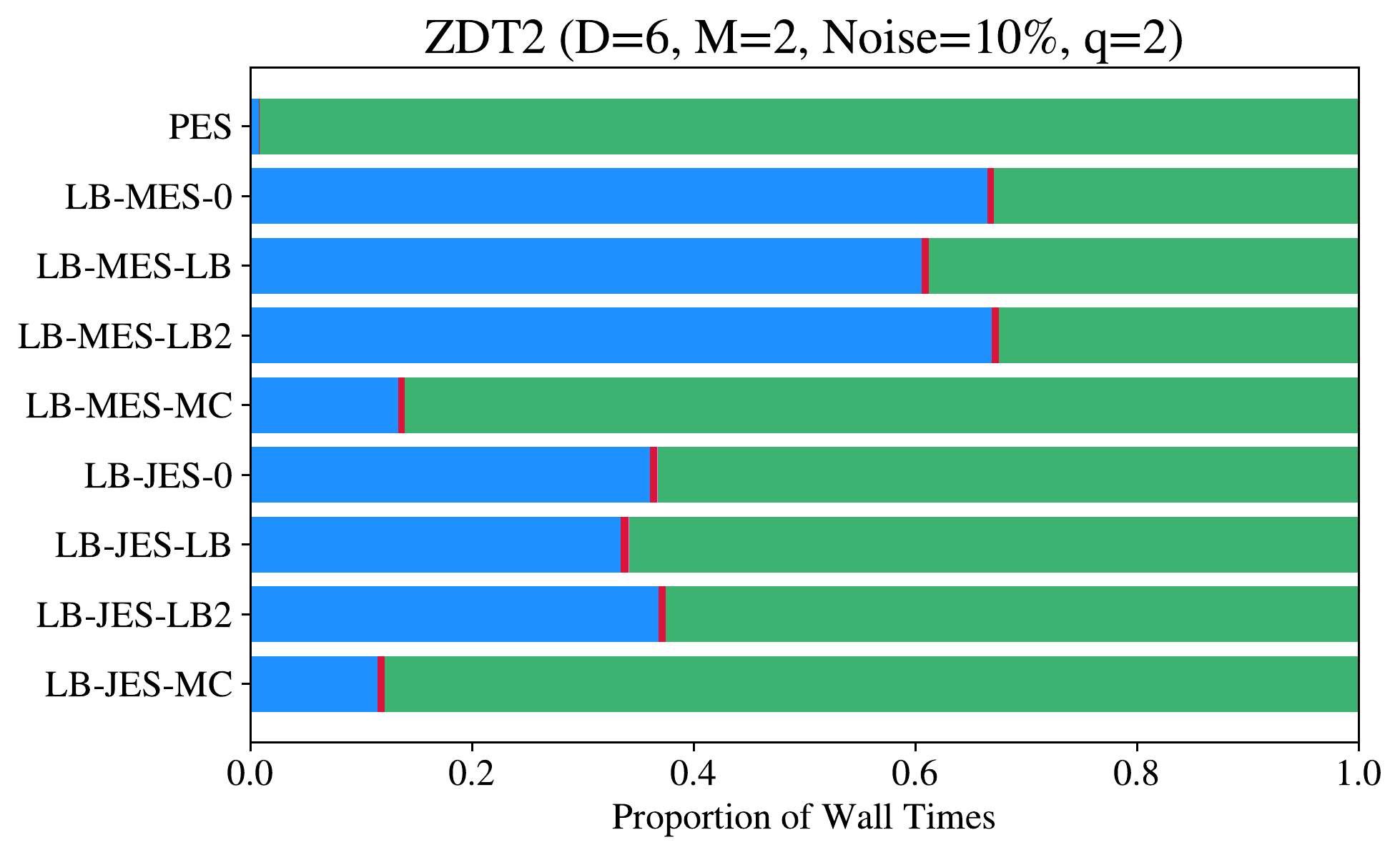}
	\includegraphics[width=0.48\linewidth]{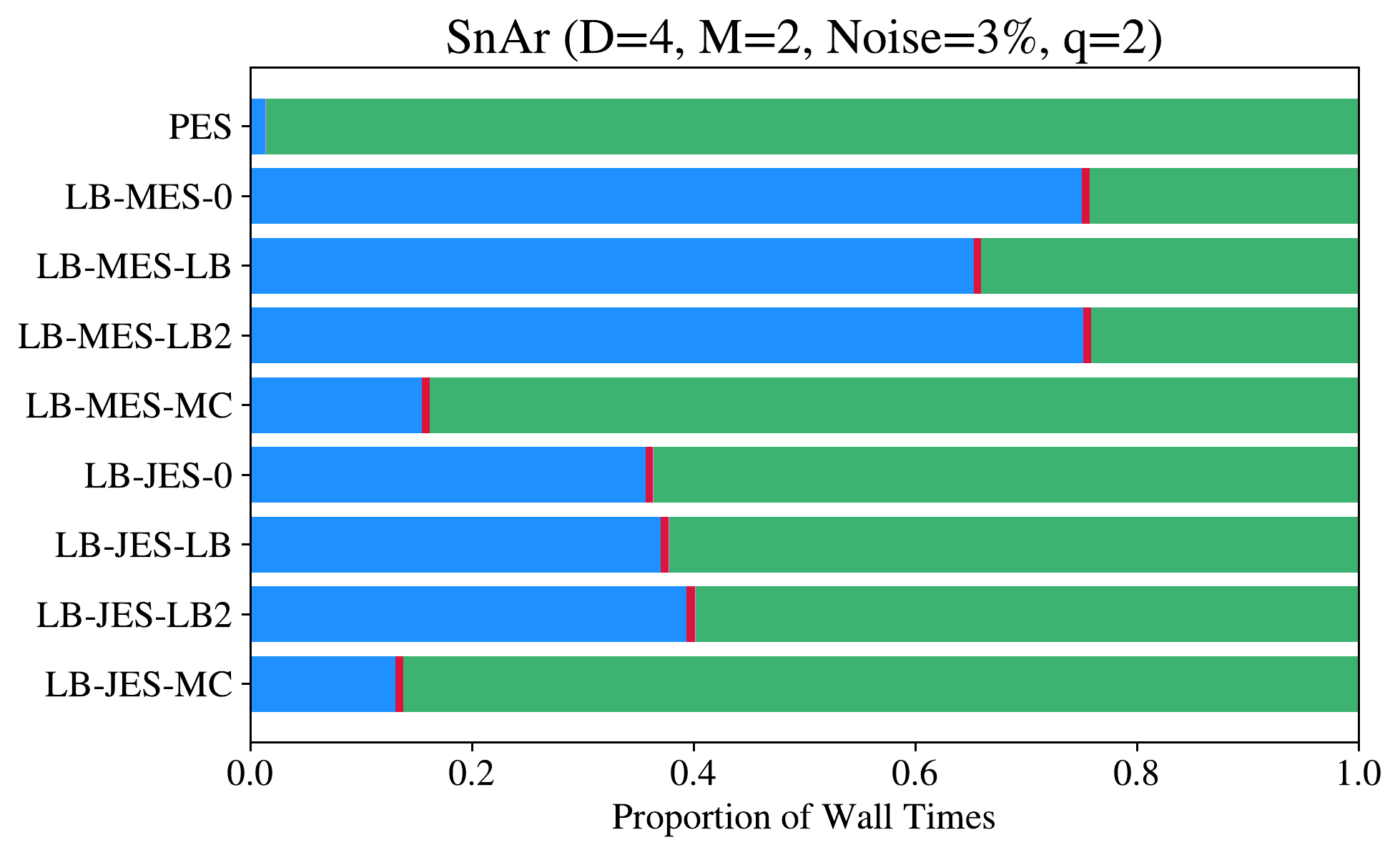}
	\includegraphics[width=0.48\linewidth]{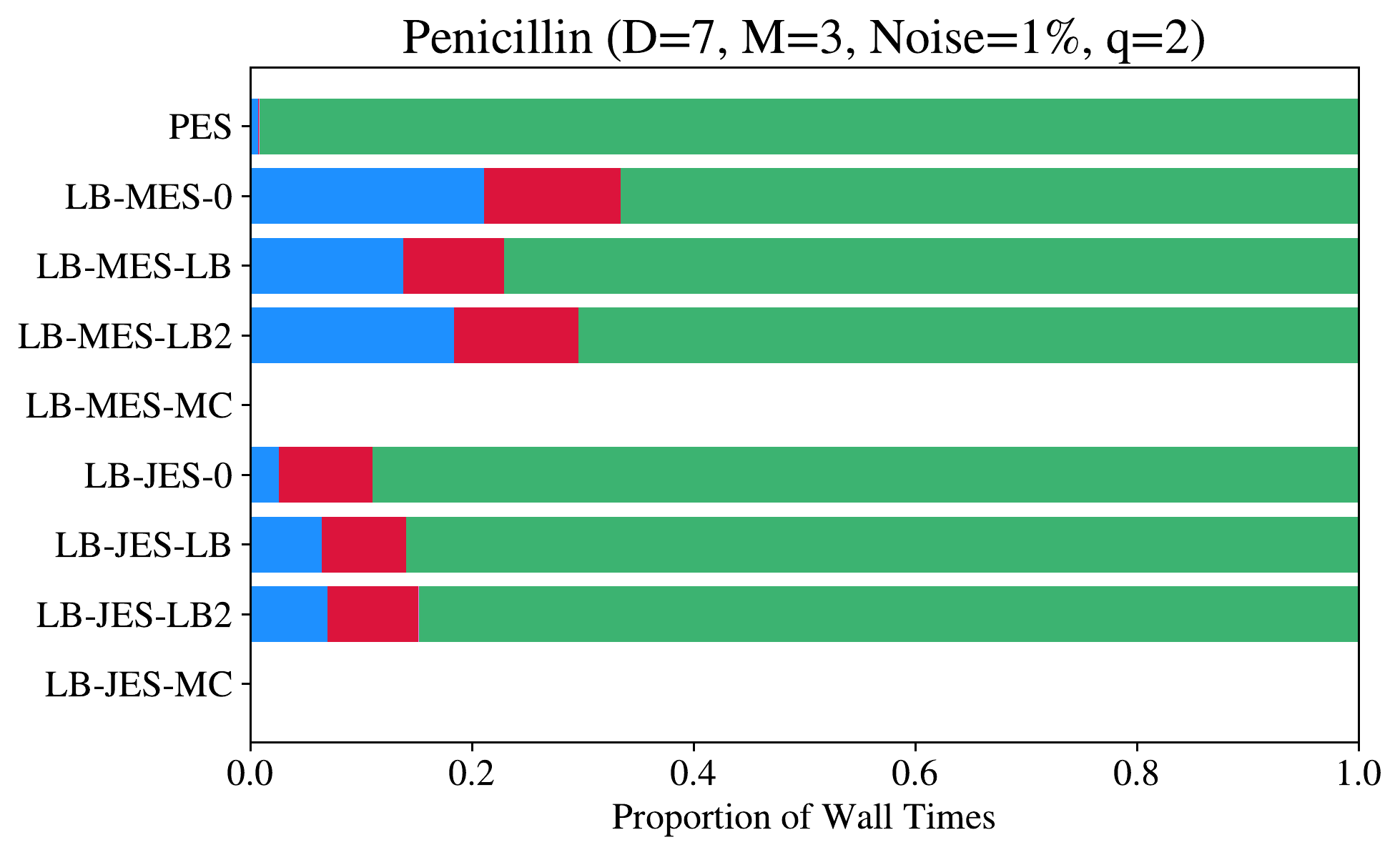}
	\includegraphics[width=0.48\linewidth]{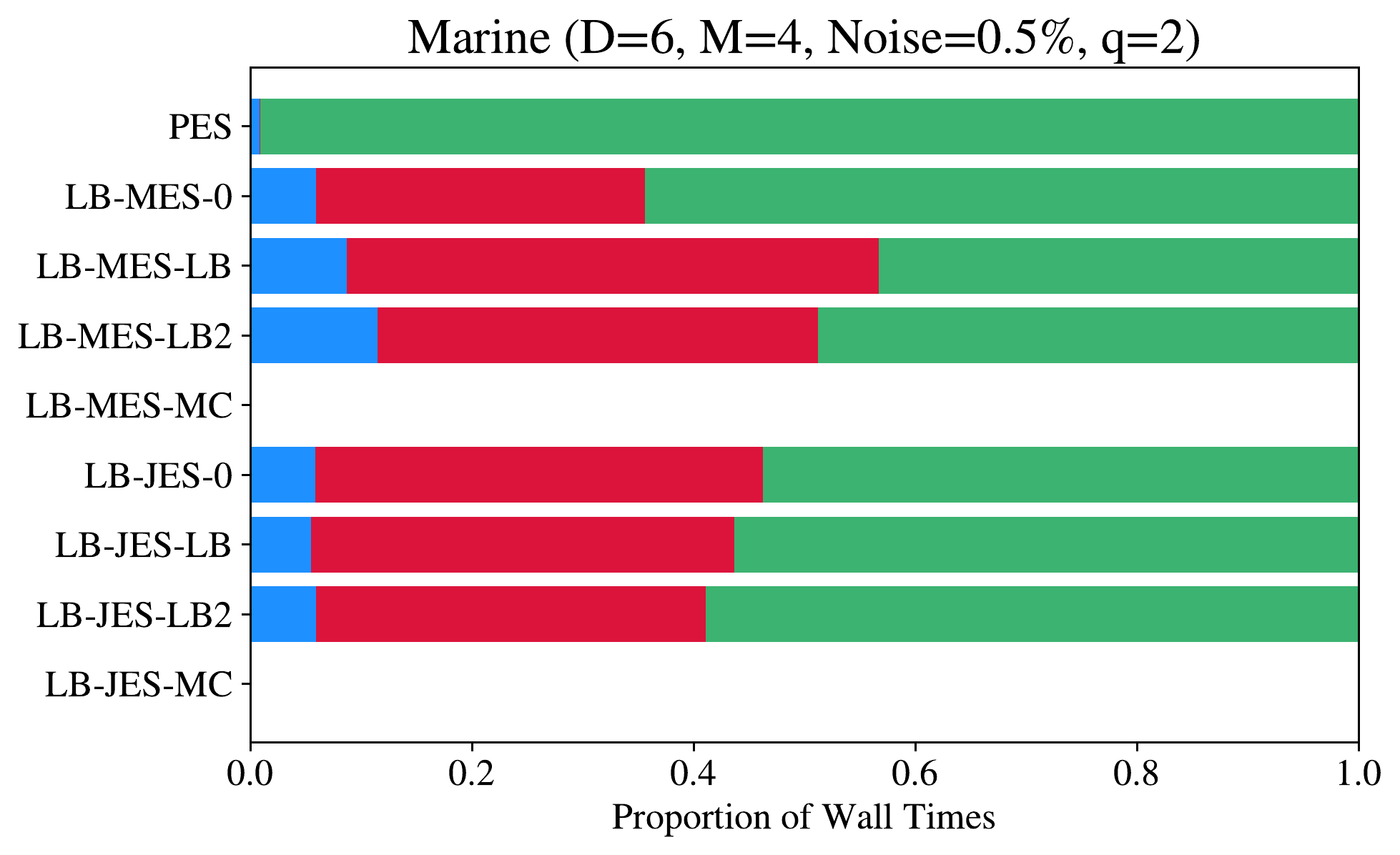}
	\centering
\end{figure}
\begin{figure}	
	\ContinuedFloat
	\includegraphics[width=0.48\linewidth]{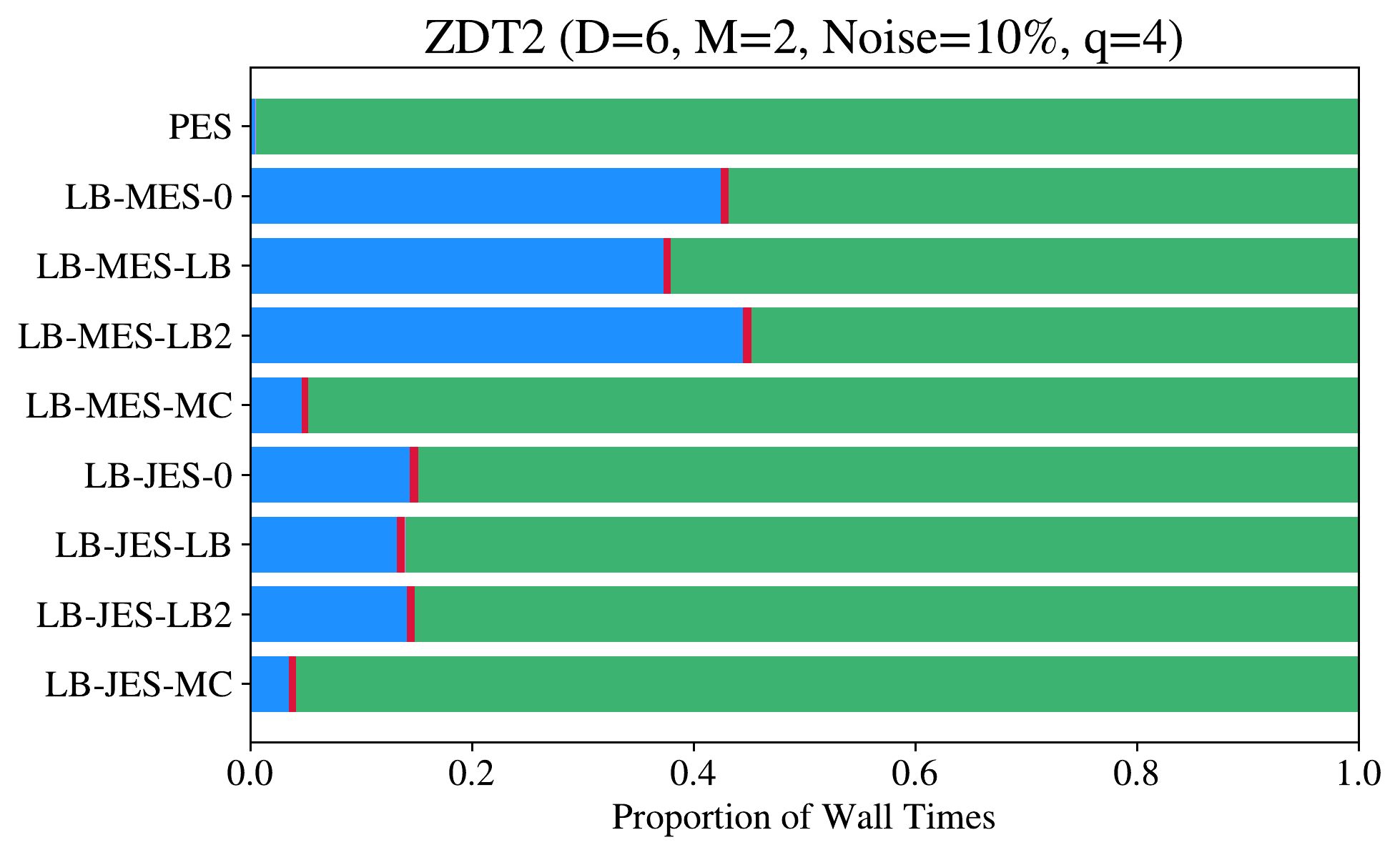}
	\includegraphics[width=0.48\linewidth]{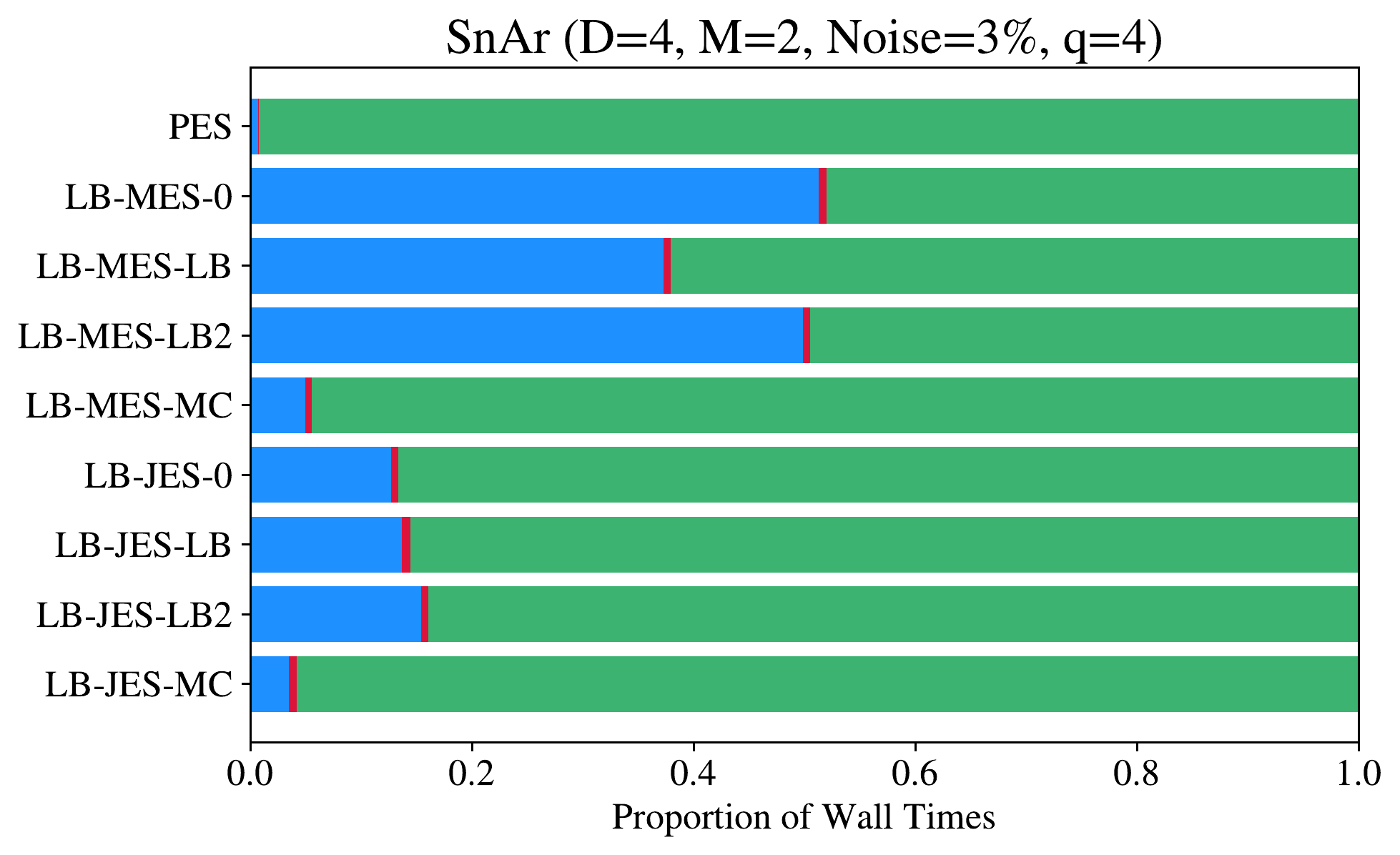}
	\includegraphics[width=0.48\linewidth]{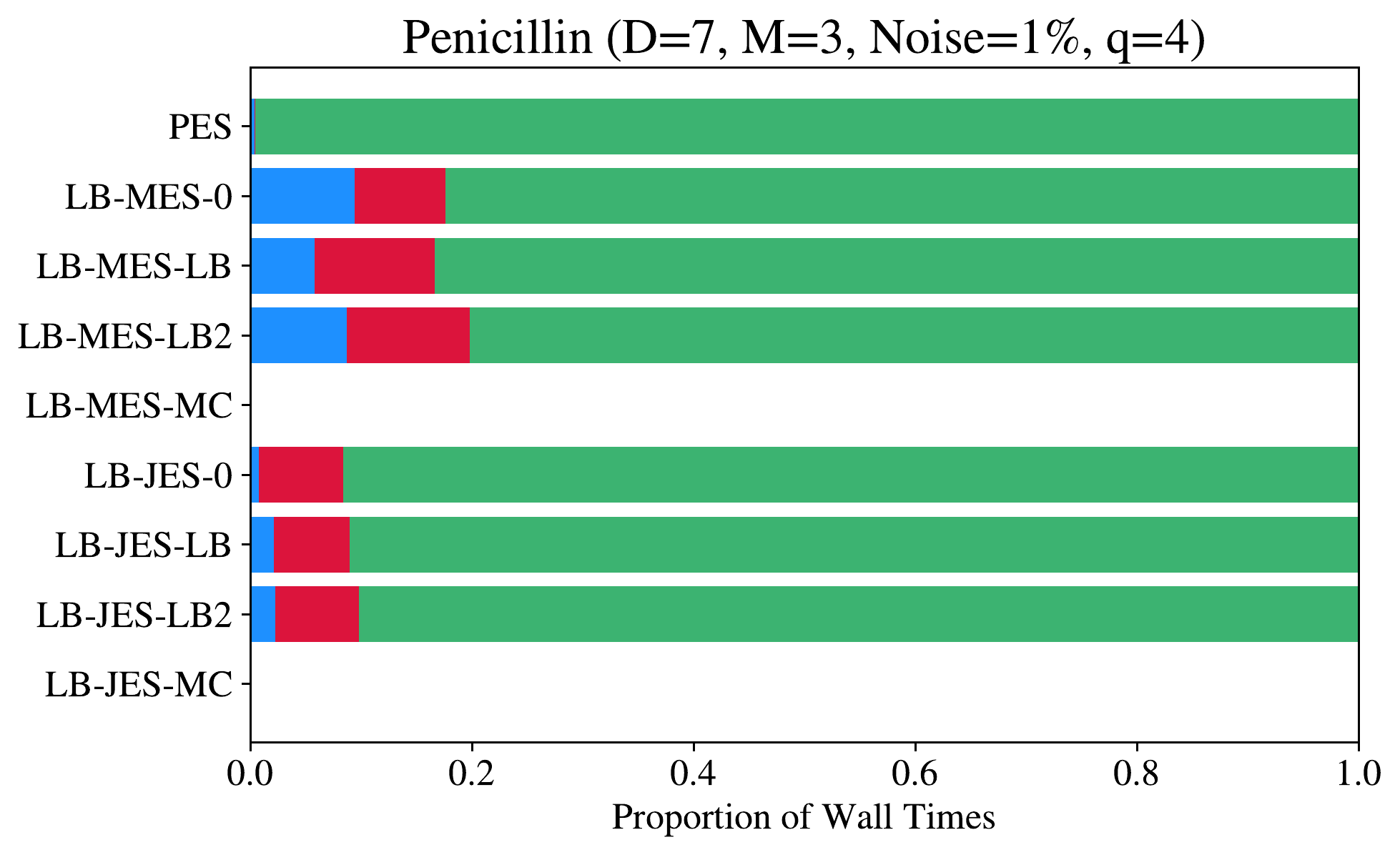}
	\includegraphics[width=0.48\linewidth]{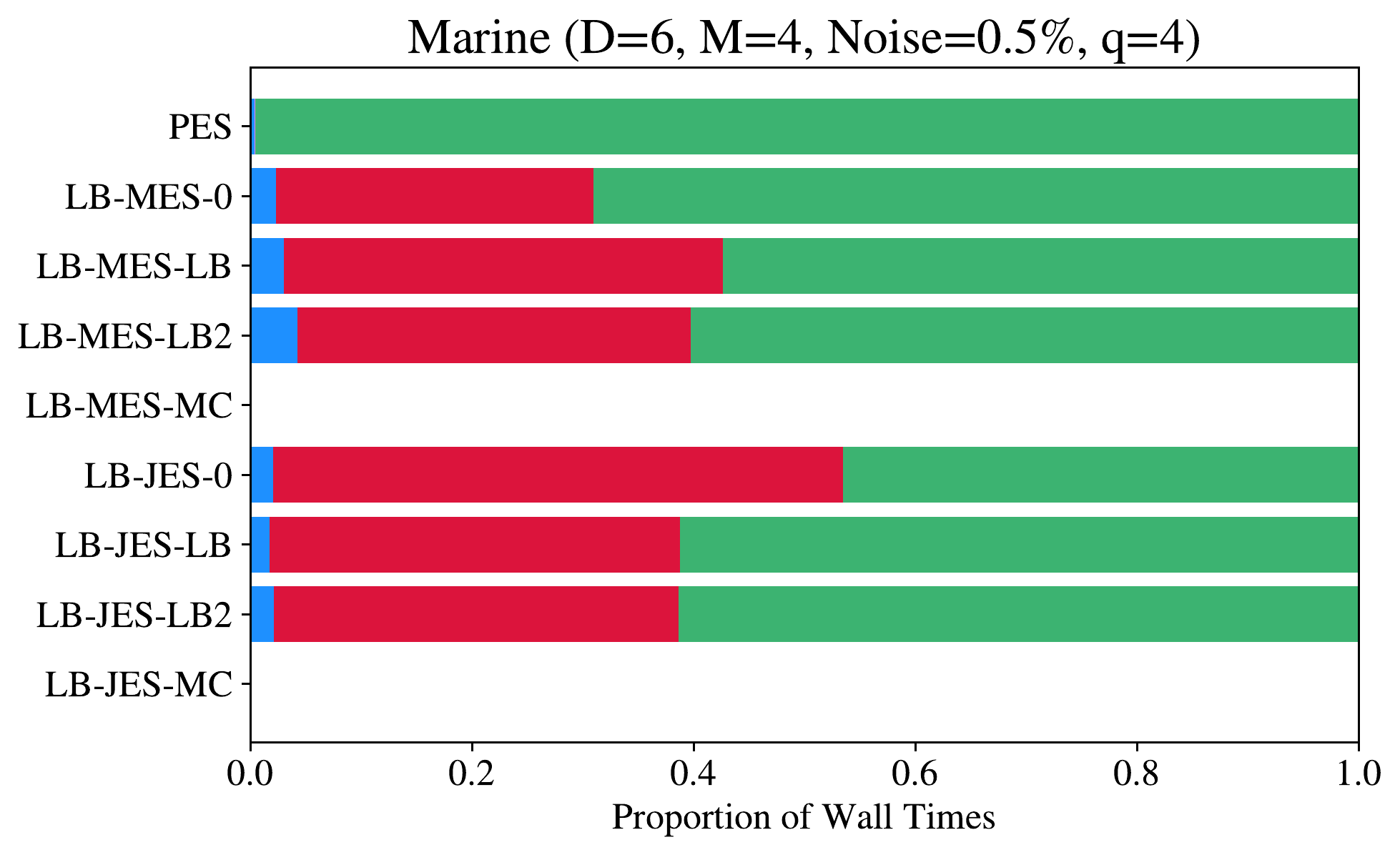}
	\includegraphics[width=0.48\linewidth]{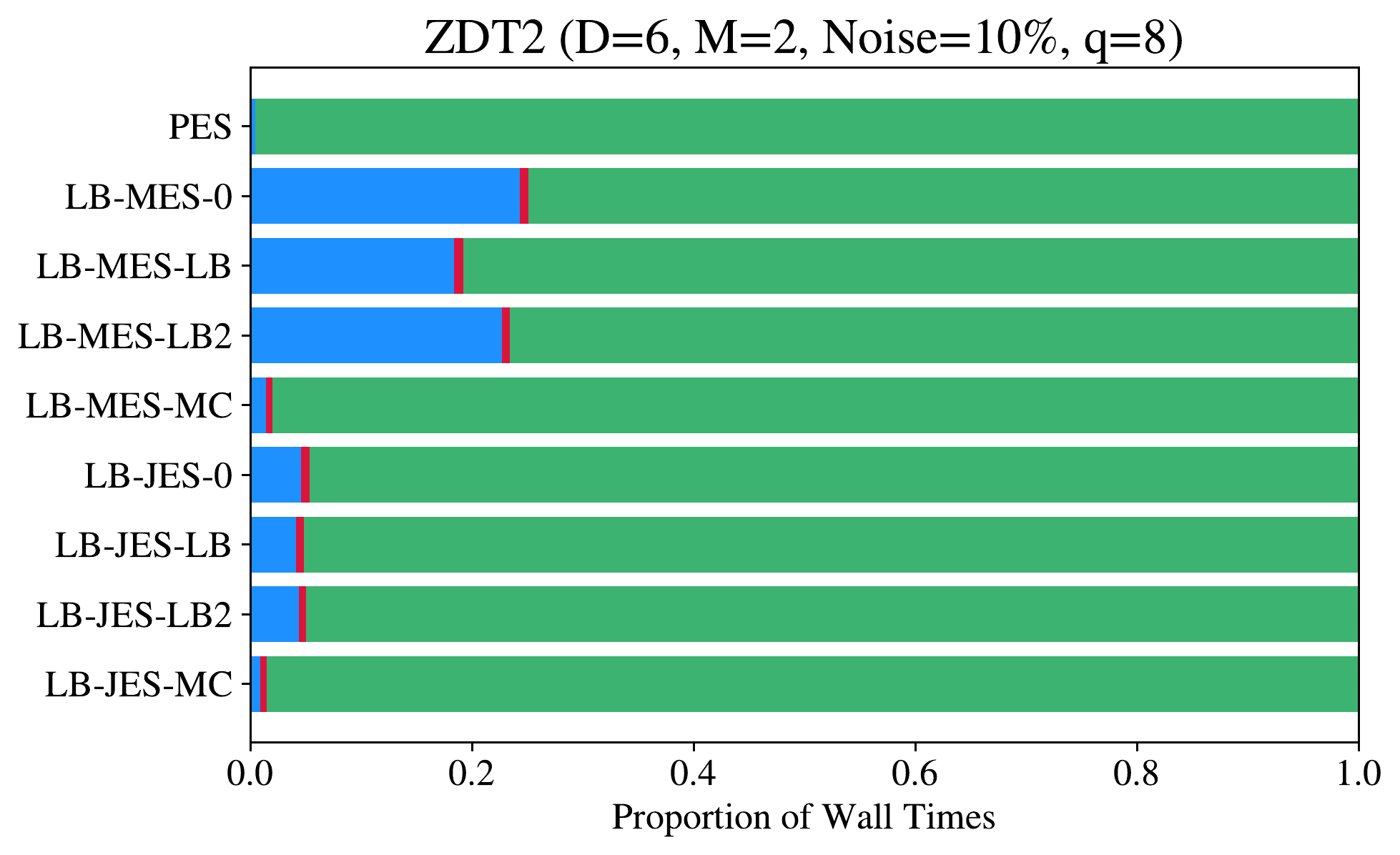}
	\includegraphics[width=0.48\linewidth]{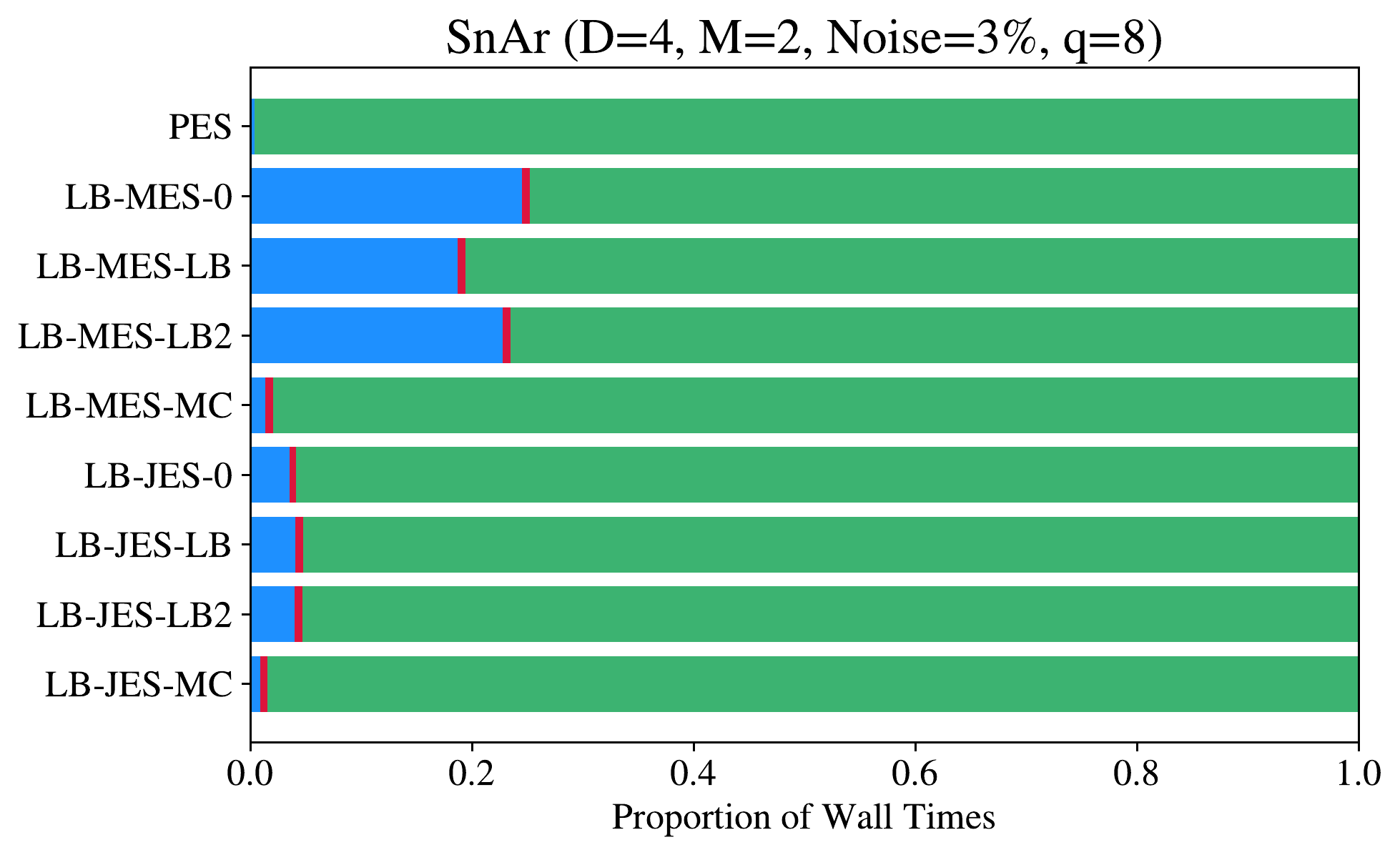}
	\includegraphics[width=0.48\linewidth]{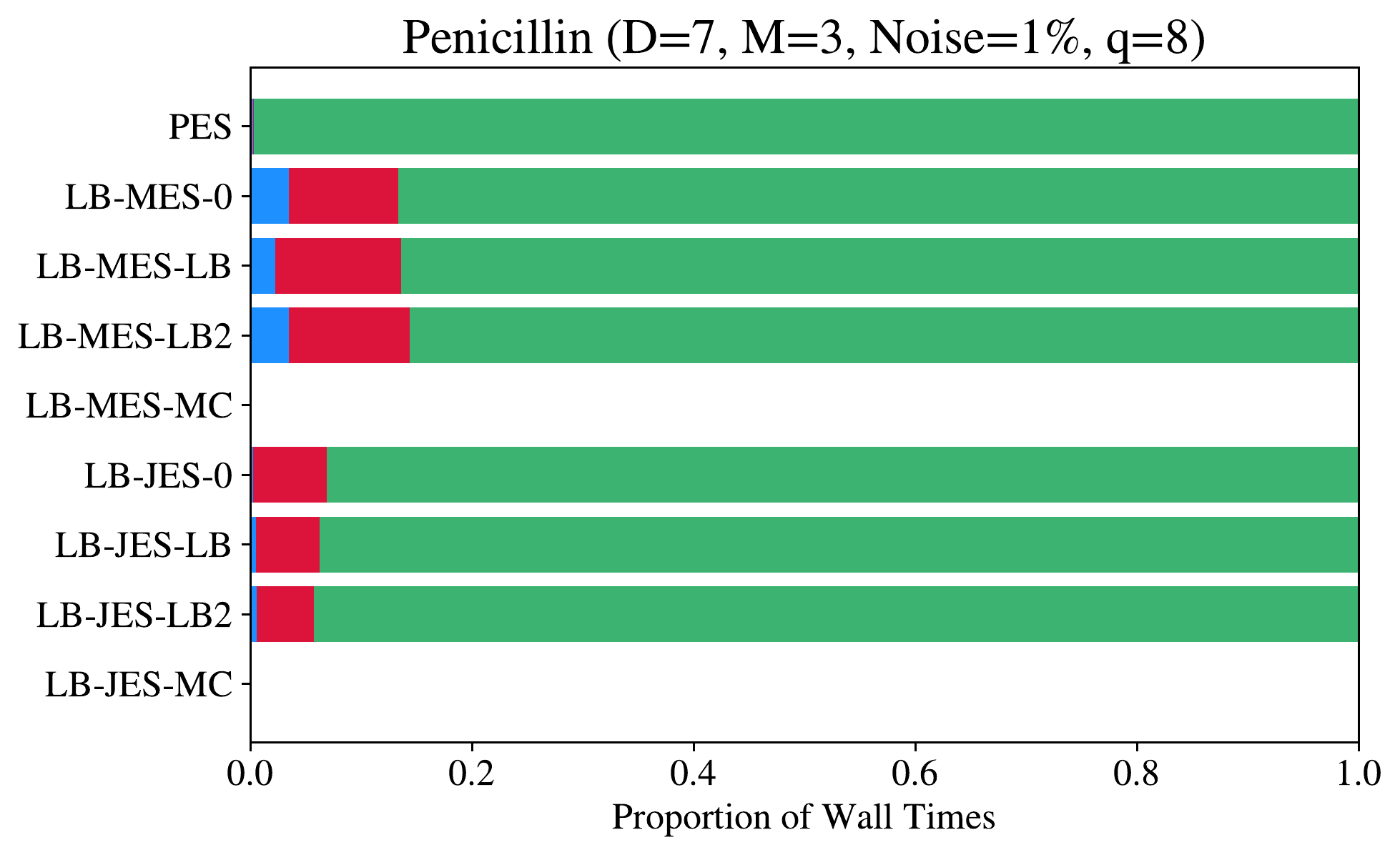}
	\includegraphics[width=0.48\linewidth]{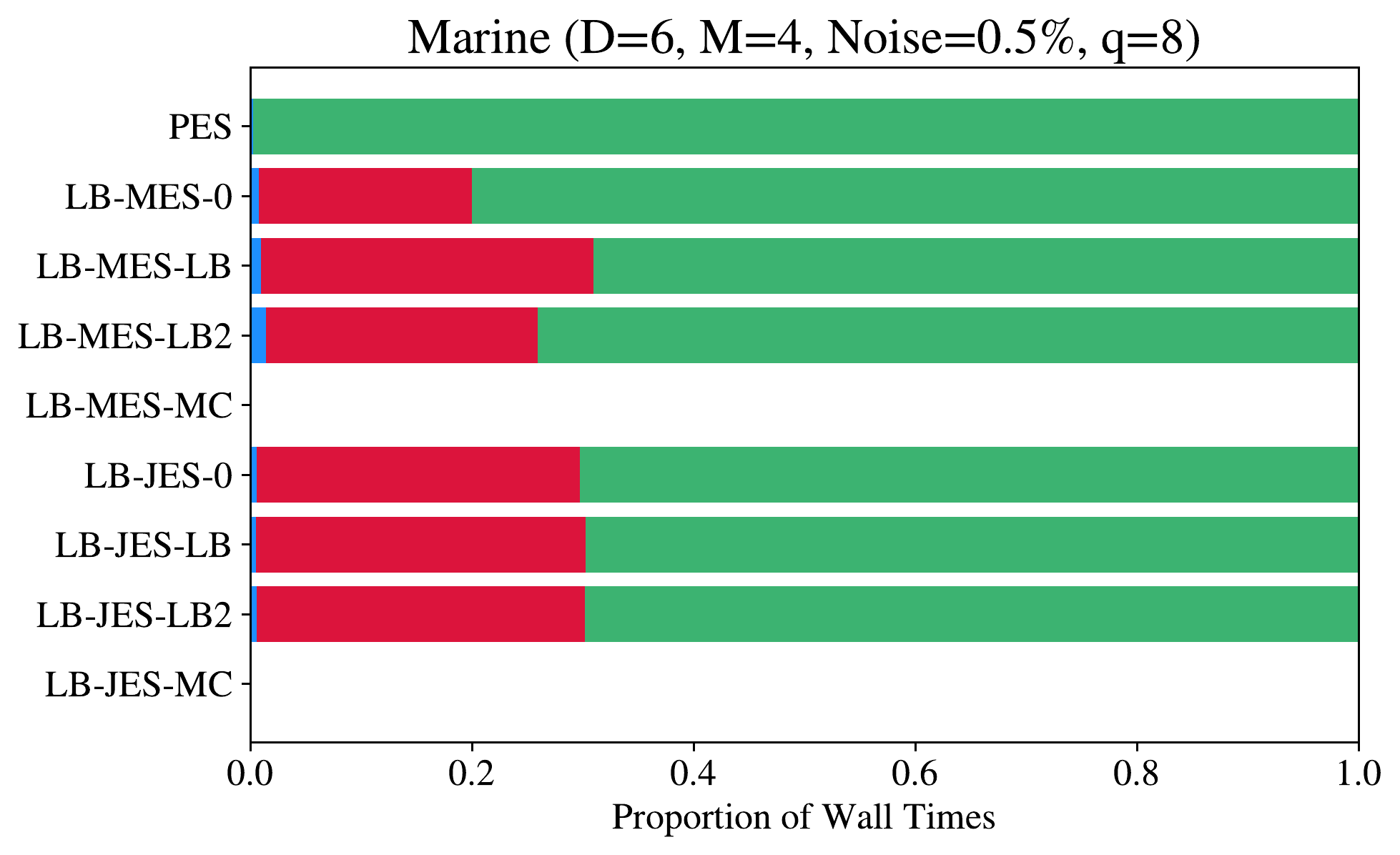}
	\centering
	\caption{A bar chart comparison of the mean wall times proportions for the information-theoretic acquisition stage. The initialization step includes generation of base samples for the Monte Carlo estimates, the conditioning for JES and the initial expectation propagation steps for PES. The Pareto sampling and box decompositions steps are executed in sequence, hence the proportion of these parts could be reduced if these steps were executed in parallel. As the number of objectives increases, the one time box decomposition is the dominant contributor to the wall time for the MES and JES. The dominant cost of the PES is always the gradient optimization step because we estimate the gradients using finite differences, which is both expensive and inefficient.}
	\label{fig:q1_prop_time}
\end{figure}
\FloatBarrier
\section{Extensions}
\label{app:extensions}
The discussion so far focussed on the problem of unconstrained multi-objective optimization where evaluations of the vector-valued function are executed individually or in batch. Not all multi-objective optimization fall into this category, so we propose some additional extensions to handle more general problems.
\paragraph{Constrained optimization.} In the literature there are several examples of how to generalize information-theoretic acquisition functions to constrained optimization problems \cite{perrone2019nwm, garrido-merchan2019n, garrido-merchan2021a, takeno2021a, belakaria2021j, fernandez-sanchez2021a, hernandez-lobato2015icml, hernandez-lobato2016jmlr}. We propose a simple extension for the JES acquisition function, similar to \cite{takeno2021a}, to handle inequality constraints. In particular, suppose we are interested in maximizing an $M$-dimensional black-box function $f^{(1:M)}(\mathbf{x})$ subject to $K$ black-box inequality constraints $f^{(M+k)}(\mathbf{x}) \leq 0$ for $k=1,\dots,K$. We can model the constraints $f^{(M+k:M+K)}(\mathbf{x})$ as additional independent objectives using the same observation model described in \cref{sec:prelim}. The only difference between the constrained setting and the unconstrained setting is the region of integration for the CDF. In the constrained setting, the sampled Pareto front dominates only the vectors satisfying the constraint. Hence the constrained CDF is now of the form $p(\mathbf{z} \in \mathbb{D}^{K}_{\preceq}(\mathbb{Y}^*))$ where
\begin{align*}
	\begin{split}
		\mathbb{D}^{K}_{\preceq}(\mathbb{Y}^*)
		&= \{\mathbf{z} \in \mathbb{R}^{M+K}: 
		(\mathbf{z}^{(1:M)} \preceq \mathbb{Y}^* \text{ and } \mathbf{z}^{(M+1:M+K)} \preceq \mathbf{0}_K)
		\text{ or }  
		(\mathbf{z}^{(M+1:M+K)} \succ \mathbf{0}_K)
		\}
		\\
		&= \{\mathbf{z} \in \mathbb{R}^{M+K}: \mathbf{z}^{(1:M+K)} \preceq (\mathbb{Y}^*, \mathbf{0}_K)\}
		\cup
		\{\mathbf{z} \in \mathbb{R}^{M+K}: \mathbf{z}^{(1:M+K)} \succ (-\boldsymbol{\infty}_M, \mathbf{0}_K)\}
		\\
		&= \mathbb{D}_{\preceq}((\mathbb{Y}^*, \mathbf{0}_K))
		\cup
		\mathbb{D}_{\succeq}((-\boldsymbol{\infty}_M, \mathbf{0}_K)).
	\end{split}
\end{align*}
This region can be decomposed into boxes in the almost same way as before. The only difference is that we now have an additional box arising from region where the constraint is not satisfied. In general, the JES (and MES) acquisition function estimates described here can handle any type of black-box constraint as long as we are able decompose the feasible objective region into boxes. For example, interval constraints of the form $f^{(M+k)}(\mathbf{x}) \in [a, b]$, can also be readily handled in this framework. 

\paragraph{Decoupled evaluations.} 
Evaluating all objectives at each iteration can be costly and perhaps unnecessary for practical problems. To address this problem, researchers in BO have considered the use of decoupled \cite{gelbart2014uai} acquisition functions $\alpha_{\mathcal{M}}$, which considers the quality of querying a subset of objectives $\{f^{(m)}: m \in \mathcal{M}\}$. To the best of our knowledge, all of the existing decoupled acquisition functions in multi-objective BO are based on information-theoretic acquisition functions \cite{garrido-merchan2019n, suzuki2020icml, hernandez-lobato2016icml, hernandez-lobato2016jmlr}. The novel JES-LB2 and MES-LB2 described in this paper possesses this decoupling property because it can be decomposed into a sum of acquisition functions for each objective: $\alpha^{\text{JES-LB2}}(\mathbf{x}| D_n) = \sum_{m=1}^M \alpha_m(\mathbf{x}| D_n)$. By Theorem 4.1 of \cite{suzuki2020icml}, the JES-0 and MES-0 can also be generalized to the decoupled setting via a marginalization argument. Upon reviewing the experimental results of the cited papers, we see that decoupled evaluations typically provide only a marginal improvement over the non-decoupled strategies. This is possibly down to heuristic choice to search the space of subsets. 
A more principled search method based on some emerging ideas about how to optimize over categorical inputs \cite{ru2020icml, nguyen2020potacoai, garrido-merchan2020n} might be useful to obtain further improvements.

\paragraph{Multi-fidelity Bayesian optimization.} For practical optimization problems of interest, it is occasionally possible to evaluate approximations of the true objectives that are much cheaper. This additional degree of freedom has been exploited before in literature under the name of multi-fidelity Bayesian optimization \cite{belakaria2020pacai, belakaria2021j, moss2021mlkdd, moss2021jmlr, zhang2017nwbo, takeno2020icml, kandasamy2017icml, wu2020uai,song2019icais}. These strategies have demonstrated some benefit when optimizing with cost constraints on the function evaluations.  It is possible to adapt JES to the multi-fidelity by combining the ideas introduced here with the ideas before in the papers referred to above such as using cost weights and conditioning arguments on lower fidelities. The main obstacle to extending this work to the multi-fidelity setting will likely arise from some lengthy algebraic exercises relating to the conditional entropy and box decompositions.

\end{document}